\documentclass[runningheads]{llncs}

\usepackage{eccv}

\usepackage{threeparttable}
\usepackage[export]{adjustbox}
\usepackage{bbding}
\usepackage{paralist}
\usepackage{makecell}
\usepackage{multirow}
\usepackage{tabularx}

\usepackage{eccvabbrv}

\usepackage{graphicx}
\usepackage{booktabs}

\usepackage[accsupp]{axessibility}  % Improves PDF readability for those with disabilities.

\usepackage[breaklinks,colorlinks,citecolor=eccvblue]{hyperref}

\usepackage{orcidlink}

\begin{document}

\title{ColorMNet: A Memory-based Deep Spatial-Temporal Feature Propagation Network for Video Colorization}

\titlerunning{ColorMNet: A MSTFP Network for Video Colorization}

\author{
Yixin Yang \and
Jiangxin Dong \and
Jinhui Tang \and
Jinshan Pan}

\authorrunning{Y.~Yang \etal}
% If there are more than two authors, '\etal' is used.

\institute{Nanjing University of Science and Technology}
% \footnotetext{$^*$ Jinshan Pan is the corresponding author.}

\maketitle

% NTIRE testset teaser
\begin{figure}[!h]\tiny
	\vspace{-0.3in} % -2.7in
         % \hspace{-8mm}
	\begin{minipage}{\textwidth}
	\centering
	 \begin{tabular}{cccc}
     {\includegraphics[width=0.230\linewidth, height=0.145\linewidth]{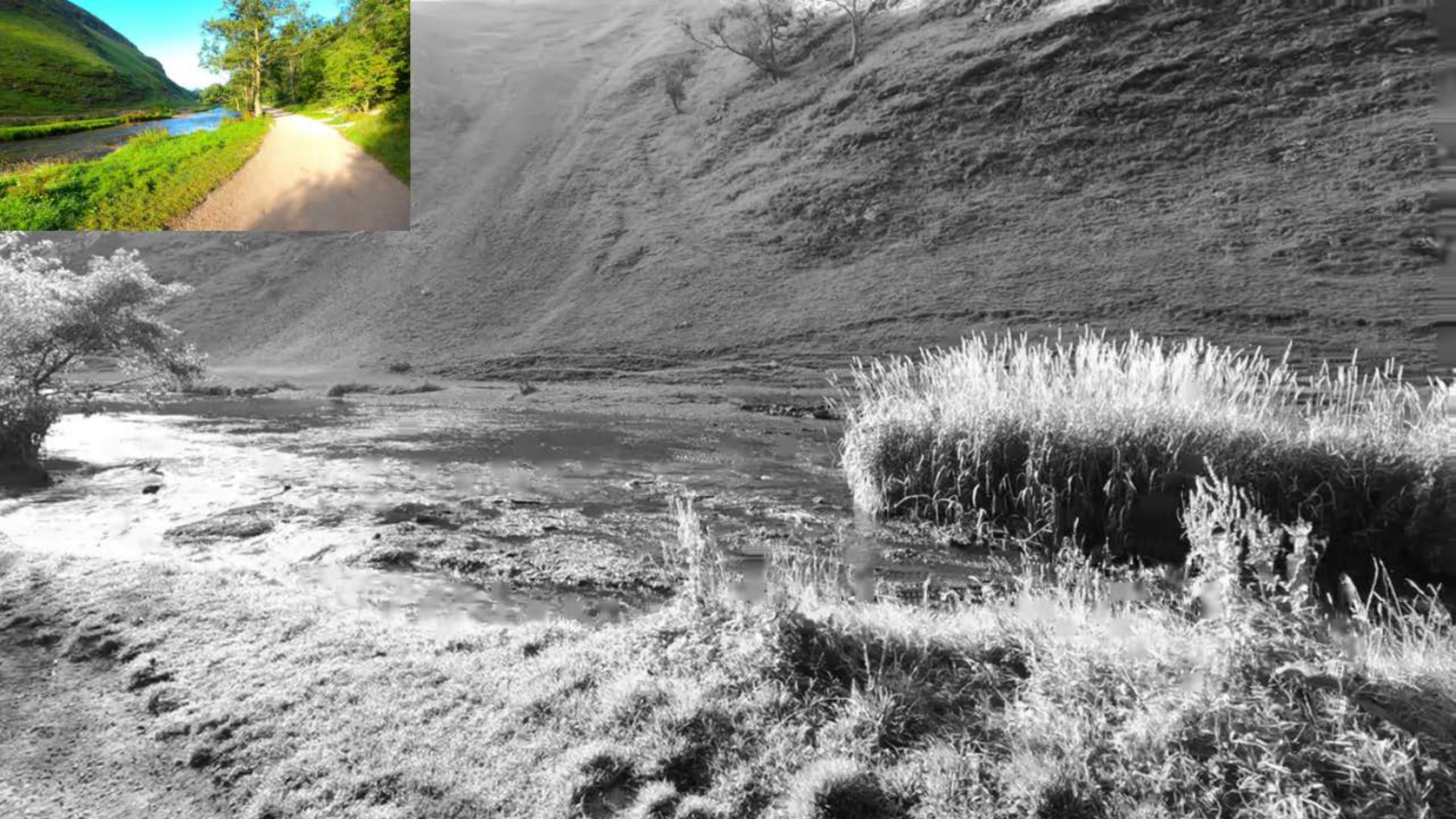}}&
    \includegraphics[width=0.230\linewidth, height = 0.145\linewidth]{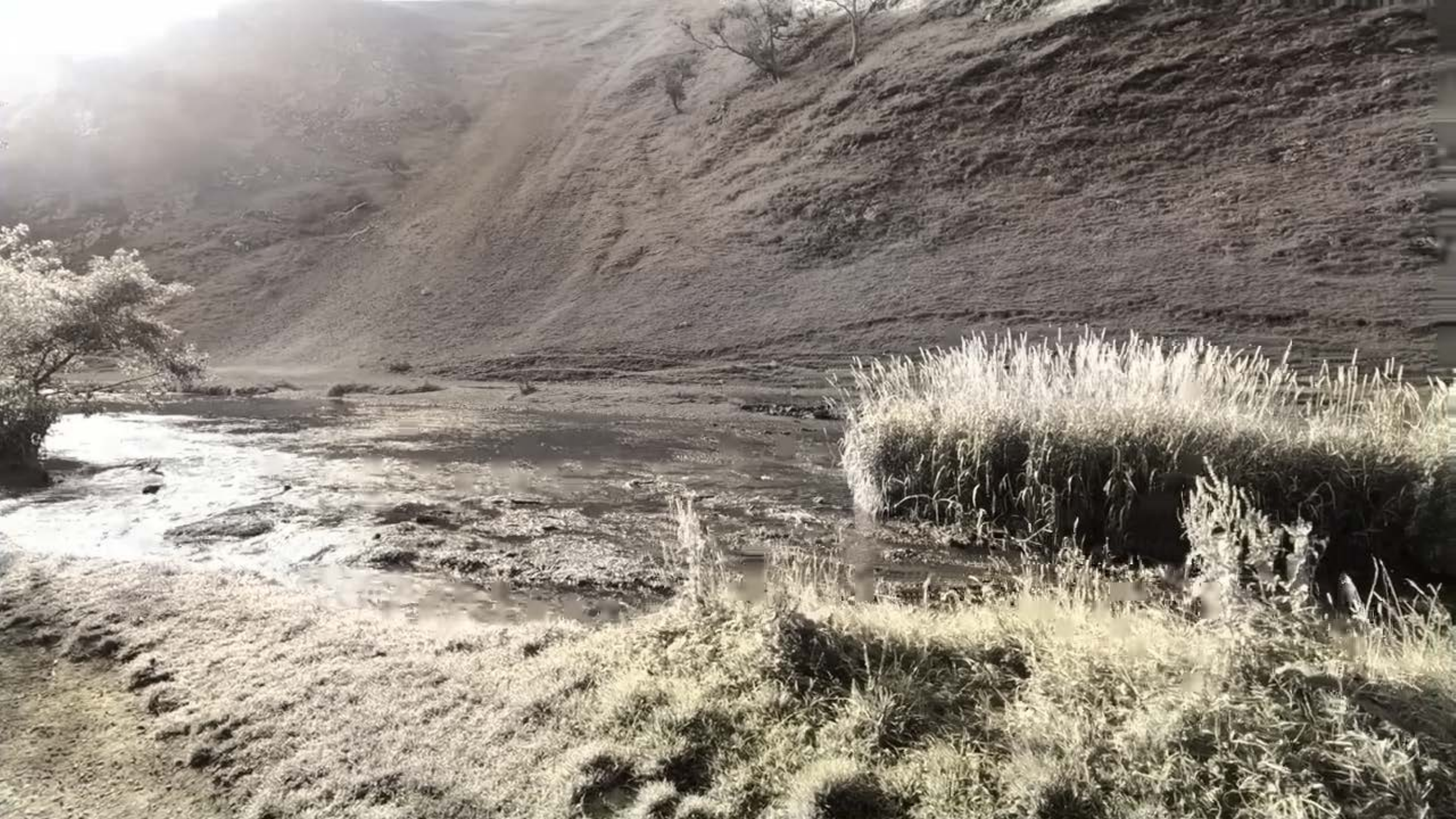} &
     % \hspace{+2mm}
       \multicolumn{2}{c}{\multirow{5}{*}[46pt]{\includegraphics[width=0.46\linewidth, scale=1]{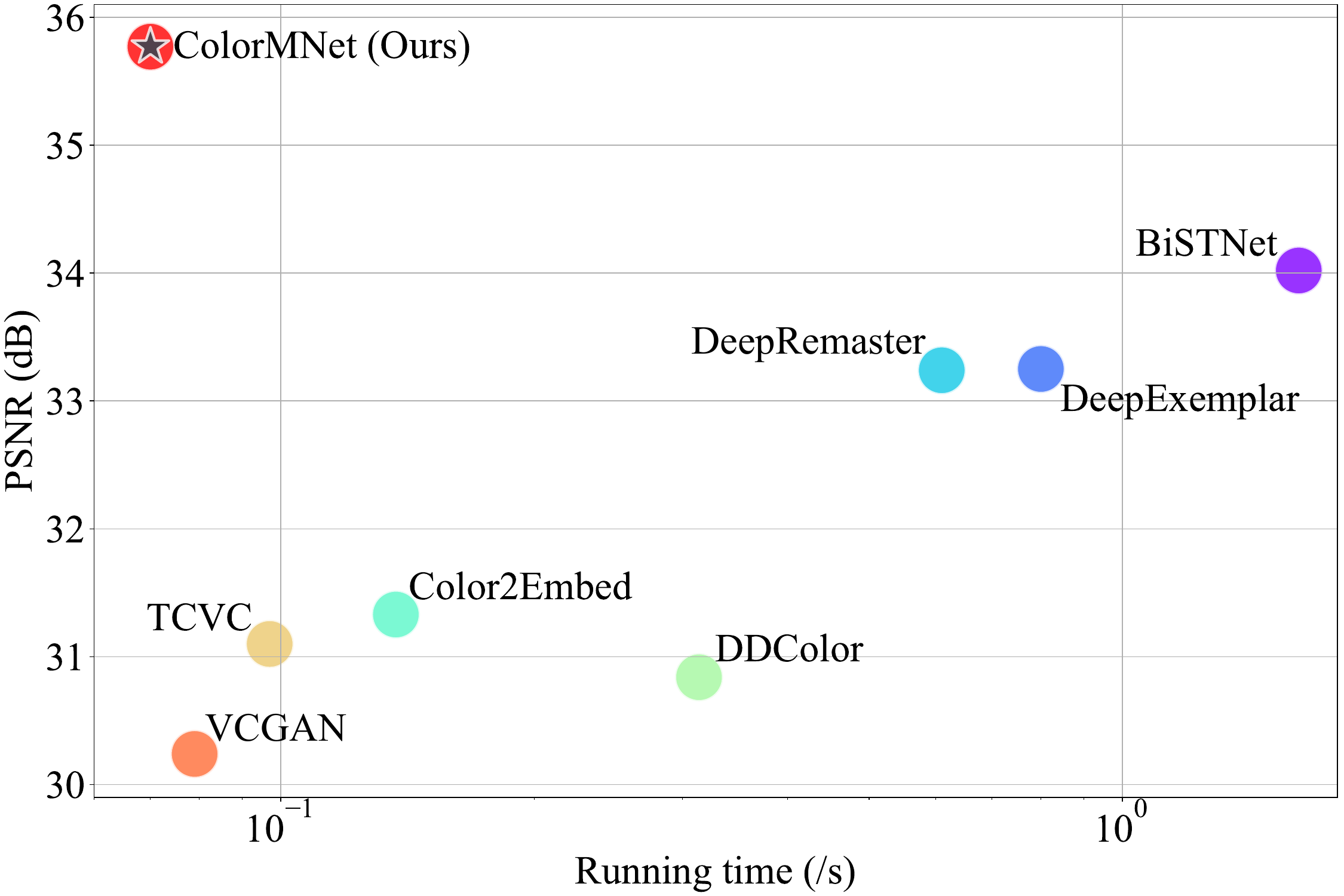}}}
      \\
      (a) Input and exemplar  & (b) DeepRemaster~\cite{IizukaSIGGRAPHASIA2019} & \multicolumn{2}{c}{}
      \\
        \includegraphics[width=0.230\linewidth, height = 0.145\linewidth]{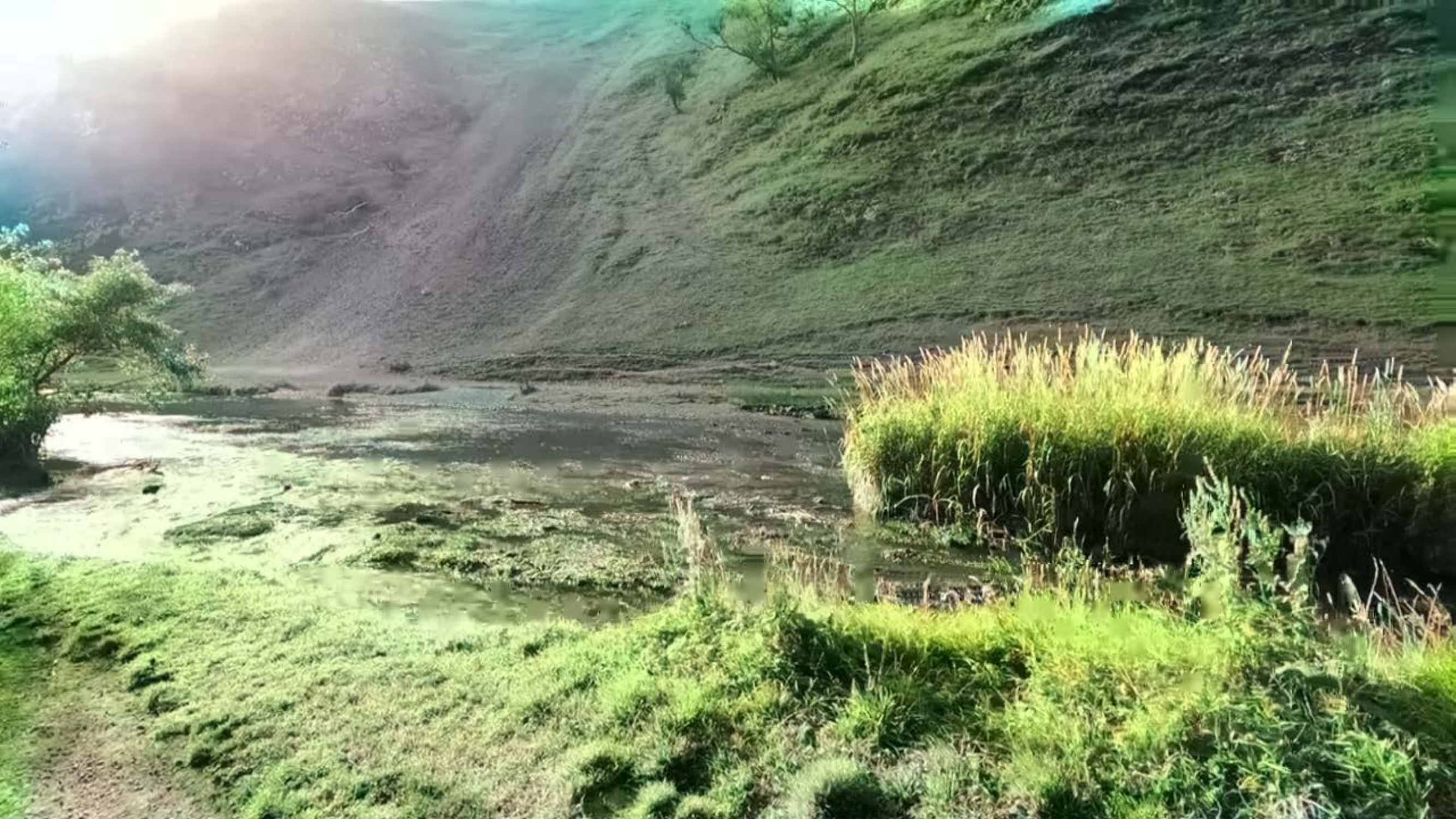} &   \includegraphics[width=0.230\linewidth, height = 0.145\linewidth]{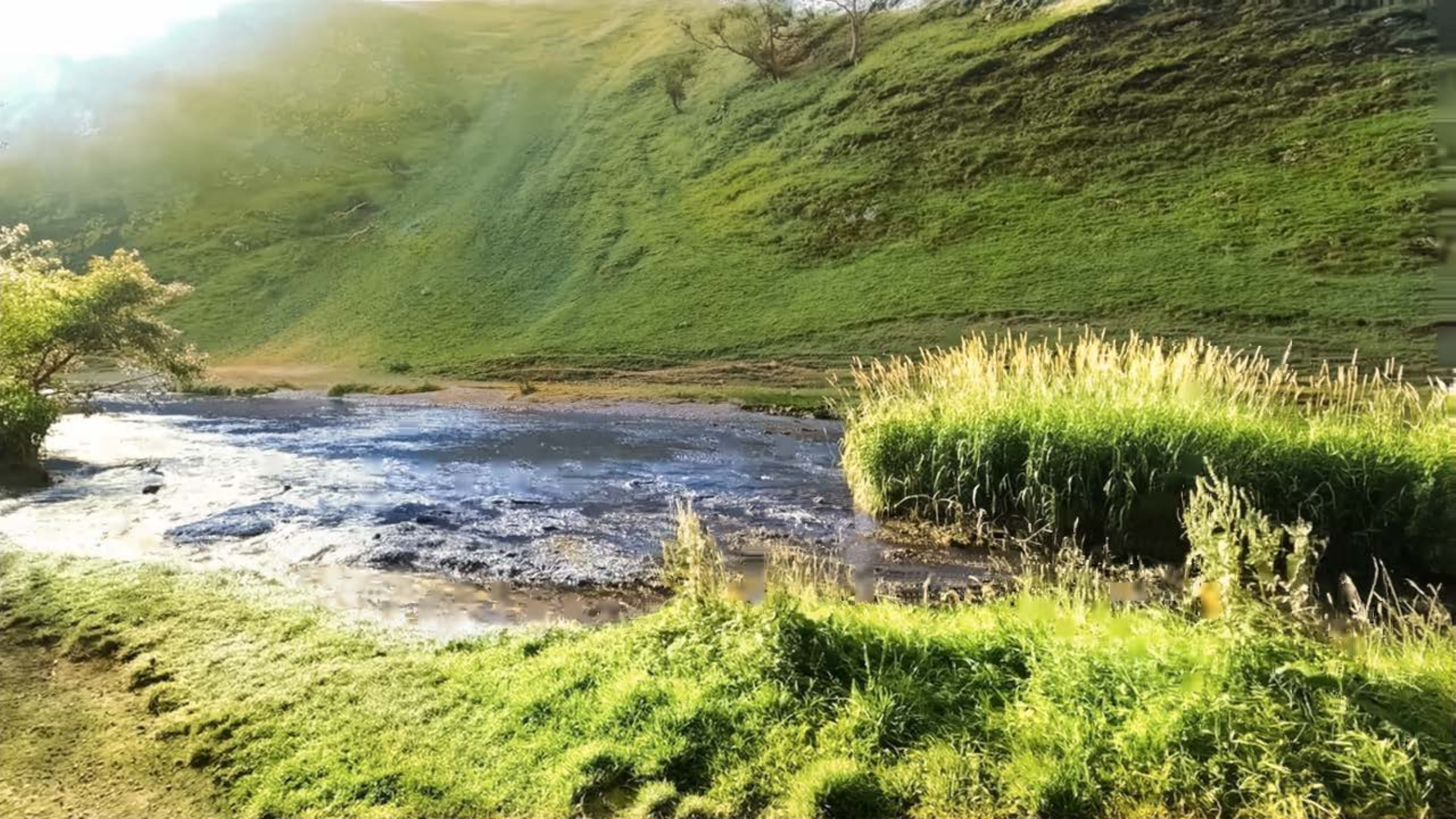} &  \multicolumn{2}{c}{}
    \\
       (c) DeepExemplar~\cite{zhang2019deep} &   (d) ColorMNet (Ours) &  \multicolumn{2}{c}{(e) Performance and running time comparison}
      \\
	\end{tabular}
	\vspace{-2mm}
	\caption{Colorization results on a real-world video and model performance comparisons between our proposed ColorMNet and other methods on the DAVIS~\cite{Perazzi_CVPR_2016} dataset in terms of PSNR and running time. State-of-the-art methods~\cite{zhang2019deep,IizukaSIGGRAPHASIA2019} do not generate well-colorized images in (b) and (c). In contrast, by exploring the features from large-pretrained visual models to estimate robust spatial features for each frame, effectively propagating these features along the temporal dimension based on memory mechanisms for far-apart frames, and exploiting the video property that adjacent frames contain similar contents, our method accurately restores the colors on the grass and generates a realistic image in (d). (e) shows that the proposed ColorMNet performs favorably against state-of-the-art methods in terms of accuracy and running time. The size of the test images for measuring the running time is $960 \times 536$ pixels.}
	\label{fig:teaser}
	\end{minipage}
	\vspace{-8mm}
\end{figure}

\begin{abstract}
How to effectively explore spatial-temporal features is important for video colorization.
Instead of stacking multiple frames along the temporal dimension or recurrently propagating estimated features that will accumulate errors or cannot explore information from far-apart frames, we develop a memory-based feature propagation module that can establish reliable connections with features from far-apart frames and alleviate the influence of inaccurately estimated features.
To extract better features from each frame for the above-mentioned feature propagation, we explore the features from large-pretrained visual models to guide the feature estimation of each frame so that the estimated features can model complex scenarios.
In addition, we note that adjacent frames usually contain similar contents. To explore this property for better spatial and temporal feature utilization, we develop a local attention module to aggregate the features from adjacent frames in a spatial-temporal neighborhood.
We formulate our memory-based feature propagation module, large-pretrained visual model guided feature estimation module, and local attention module into an end-to-end trainable network (named ColorMNet) and show that it performs favorably against state-of-the-art methods on both the benchmark datasets and real-world scenarios.
The source code and pre-trained models will be available at \href{https://github.com/yyang181/colormnet}{https://github.com/yyang181/colormnet}.
  \keywords{Exemplar-based video colorization \and Deep convolutional neural network \and Feature propagation}
  % \vspace{-3mm}
\end{abstract}

% \vspace{-3mm}
\section{Introduction}
\label{sec:intro}
\vspace{-2mm}

Due to the technical limitations of old imaging devices, lots of videos captured in the last century are in black and white, making them less visually appealing on modern display devices.
As most of these videos have historical values and are difficult to reproduce, it is of great need to colorize them.

Restoring high-quality colorized videos is challenging as it not only needs to handle the colorization of each frame but also requires exploring temporal information from the video sequences.
Therefore, directly applying existing image colorization methods~\cite{Levin2004, Cheng2015, cic, Larsson2016, let, tog17, inscolor, kang2022ddcolor, Xu_2020_CVPR} does not generate satisfactory colorized videos as minor perturbations in consecutive input video frames may lead to substantial differences in colorized video results.
To overcome this, numerous methods model the temporal information from inter-frames by stacking multiple frames along the temporal dimension~\cite{IizukaSIGGRAPHASIA2019, chen2023exemplar} or recurrently propagating features~\cite{atcvc, favc, wan2022oldfilm, zhang2019deep}.
Although these approaches show better performance than the ones based on single image colorization, stacking multiple frames along the temporal dimension cannot effectively leverage spatial-temporal prior from adjacent frames and requires a large amount of GPU memory. In addition, recurrent-based feature propagation is not able to effectively explore long-range information, leading to unsatisfactory results for frames far apart.

To better explore long-range temporal information, several approaches~\cite{bistnet, liu2021temporally} develop bidirectional recurrent-based feature propagation methods for video colorization.
As the recurrent-based feature propagation treats the features of each frame equally, if the features are not estimated accurately, the errors will accumulate, thus affecting the final video colorization.
Therefore, it is still challenging to effectively model temporal information from long-range frames.

In addition to the temporal information exploration, how to extract good features from each frame plays a significant role in video colorization. Existing methods~\cite{favc, atcvc, zhang2019deep, bistnet,chen2023exemplar, vcgan} usually utilize a pretrained VGG~\cite{vgg2014} or ResNet-101~\cite{He_2016_CVPR} to extract features from each frame. These methods are able to model local structures but are less effective for exploiting non-local and semantic structures, \eg, complex scenes with multiple objects.
To restore high-quality videos, it is of great interest to develop a better feature representation method that is able to characterize the non-local and semantic properties of each frame.

In this paper, we present a memory-based deep spatial-temporal feature propagation network for video colorization.
Note that the robustness of the spatial features extracted from each frame is important, we first develop a large-pretrained visual model guided feature estimation (PVGFE) module, which is motivated by the success of large-pretrained visual models~\cite{oquab2023dinov2} in generating robust visual features to facilitate the spatial feature estimation.
However, simply recurrently propagating the estimated spatial features or directly stacking them along the temporal dimension does not effectively explore temporal information for video colorization.
Moreover, it requires a large amount of GPU memory capacity to store past frame representations when the spatial resolution is large and videos are long.
To overcome these problems, we then propose a memory-based feature propagation (MFP) module that can not only adaptively explore and propagate useful features from far-apart frames but also reduce memory consumption.
In addition, we note that adjacent frames of a video usually contain similar contents and thus develop a local attention (LA) module to better utilize the spatial and temporal features.
Taken together, the memory-based deep spatial-temporal feature propagation network, called ColorMNet, is able to generate high-quality video colorization results (see Figure~\ref{fig:teaser}(e)).

The main contributions are summarized as follows:
\begin{compactitem}
     \item We propose a large-pretrained visual model guided feature estimation module to model non-local and semantic structures of each frame for colorization.
    \item We develop a memory-based feature propagation module to adaptively explore temporal features from far-apart frames and reduce memory usage.
    \item We develop a local attention module to explore similar contents of adjacent frames for better video colorization.
    \item We formulate the proposed network into an end-to-end trainable framework and show that it performs favorably against state-of-the-art methods on both the benchmark datasets and real-world scenarios.
\end{compactitem}

\vspace{-3mm}
\section{Related Work}
\vspace{-2mm}

\noindent{\bf User-guided image colorization.}
Since the colorization problem is ill-posed, conventional image colorization methods usually adopt local user hints~\cite{Levin2004,Qu2006,Luan2007,Chen2012,Yatziv2006, zhao2021color2embed, Sangkloy_2017_CVPR, tog17, he2018deep} to make this problem well-posed.
However, these methods do not fully exploit the property of video sequences and usually need to solve temporal consistency problems.
In addition, their colorization performance for individual frame is usually far from satisfactory as estimating global and semantic features is challenging.

\noindent{\bf Automatic video colorization.}
Instead of using user-guided image methods, several approaches explore deep learning to solve video colorization.
In~\cite{favc,vcgan}, both the colorization performance and the temporal consistency are enhanced by recurrently propagating features from adjacent frames for video colorization.
In~\cite{liu2021temporally}, Liu \etal use bidirectional propagation to explore temporal features and introduces a self-regularization learning scheme to minimize the prediction difference obtained with different time steps.
Although using feature propagation improves the temporal consistency, it is not a trivial task to estimate the spatial features from each frame due to the inherent complexity of scenes in videos.
Moreover, these automatic video colorization methods~\cite{paul2016spatiotemporal,sheng2013video, Xu_2020_CVPR} may work well on synthetic datasets but lack generality on diverse real-world scenarios.

\noindent{\bf Exemplar-based video colorization.}
Exemplar-based methods aim to generate videos that are faithful to the exemplar with enhanced temporal consistency.
In~\cite{zhang2019deep}, Zhang \etal employ a recurrent-based feature propagation module to explore temporal features for latent frame restoration.
In~\cite{IizukaSIGGRAPHASIA2019}, Iizuka \etal propose stacking multiple frames along the temporal dimension to obtain better performance.
To better explore spatial and temporal features, Chen \etal~\cite{chen2023exemplar} adopt ResNets~\cite{He_2016_CVPR} instead of the commonly-used VGG~\cite{vgg2014} model for better spatial feature estimation and split video sequences into frame blocks for long-term spatiotemporal dependency.
In~\cite{bistnet}, Yang \etal use bidirectional propagation to gradually propagate features and use optical flow models~\cite{teed2020raft} to align adjacent frames.
Recurrently propagating information from long-range frames or stacking multiple frames improves the colorization performance.
However, if there are inaccurately estimated features of long-range frames, the errors will be accumulated, which thus affects video colorization.
Additionally, when dealing with long videos, these methods often consume significant GPU memory and suffer from slow inference speeds, posing substantial challenges to video colorization.

\vspace{-3mm}
\section{ColorMNet}
\label{sec:method}
\vspace{-2mm}

\begin{figure*}[!t]
    \centering
     \includegraphics[width=1.0\textwidth]{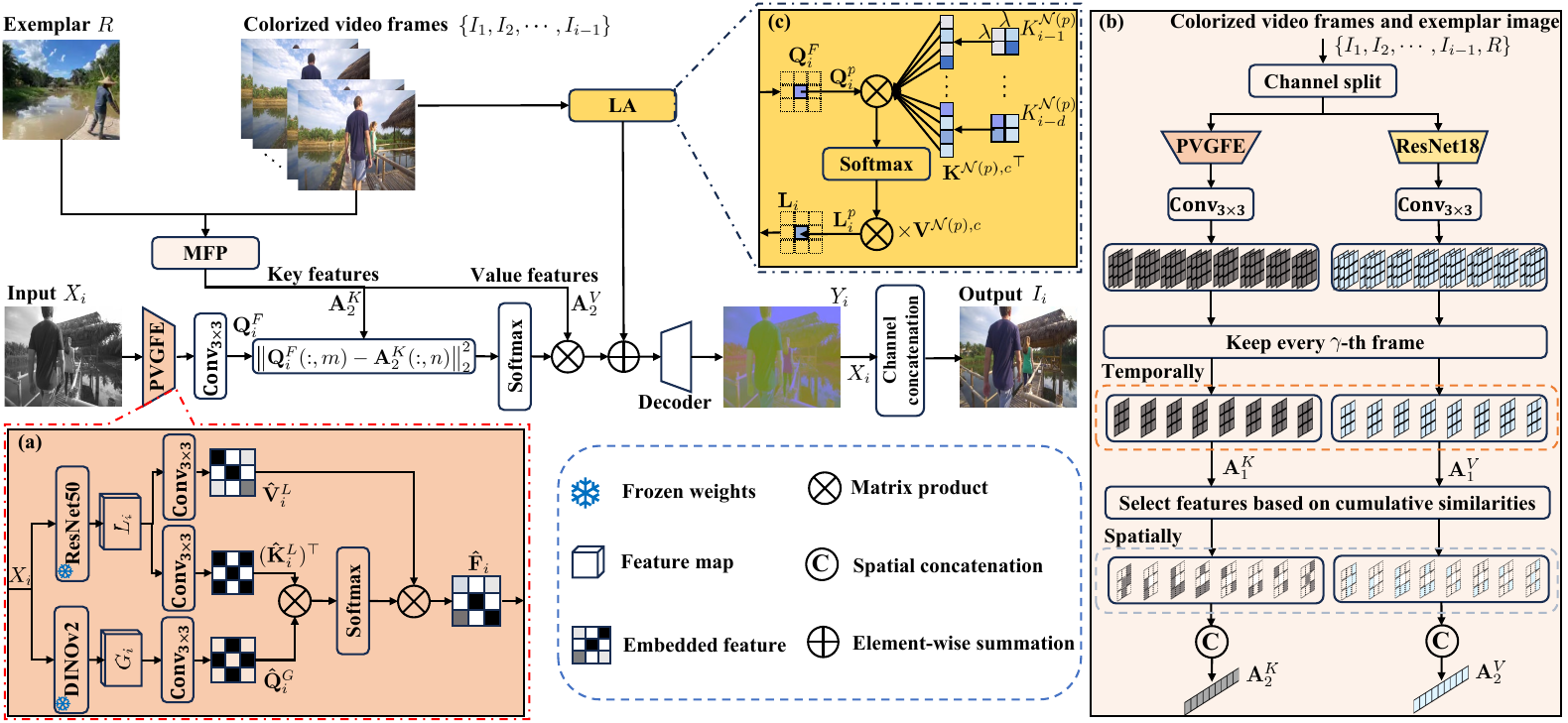}
    \vspace{-7mm}
    \caption{An overview of the proposed ColorMNet. The core components of our method include: \textbf{(a)} large-pretrained visual model guided feature estimation (PVGFE) module, \textbf{(b)} memory-based feature propagation (MFP) module and \textbf{(c)} local attention (LA).
    }
    \label{fig:framework}
    \vspace{-7mm}
\end{figure*}

Our goal is to develop an effective and efficient video colorization method to restore high-quality videos with low GPU memory requirements.
The proposed ColorMNet contains a large-pretrained visual model guided feature estimation (PVGFE) module to extract spatial features from each frame, a memory-based feature propagation (MFP) module that is able to adaptively explore the temporal features from far-apart frames, and a local attention (LA) module that is used to explore the similar contents from adjacent frames for better spatial and temporal feature utilization.
Figure~\ref{fig:framework} illustrates the overview of the proposed method.
In the following, we explain each module in detail.

\vspace{-3mm}
\subsection{PVGFE module}\label{sec:31}
\vspace{-2mm}

To estimate robust spatial features, we explore the features learned from large-pretrained visual models as they are able to model the non-local and semantic information and are robust to numerous scenarios.
Note that one of the large-pretrained visual models, \ie, DINOv2~\cite{oquab2023dinov2}, adopts ViT~\cite{vit} as the main feature extractor and generates all-purpose visual features facilitating both image-level (classification) and pixel level (segmentation) tasks due to its robust feature representation ability. In this paper, we utilize the global features learned from DINOv2 to guide the local features learned from CNNs for better feature estimation of each frame.

Given the input grayscale frames $\{{X}_i\}_{i=1}^N$ (where $X_i \in \mathbb{R}^{H \times W \times 1}$, ${H}\times {W}$ denotes the spatial resolution, $N$ is the number of frames of the input video),
we first extract features $\{{G}_i\}_{i=1}^N$ and $\{{L}_i\}_{i=1}^N$ by applying the pretrained DINOv2 and ResNet50~\cite{He_2016_CVPR} to $\{{X}_i\}_{i=1}^N$, respectively.
Then we use the cross attention~\cite{zamir2022restormer} to fuse $\{{G}_i\}_{i=1}^N$ and $\{{L}_i\}_{i=1}^N$ for obtaining robust spatial features.

Specifically, we first extract the query feature $Q_i^G$ from ${G}_i$, the key feature $K_i^L$ and the value feature $V_i^L$ from ${L}_i$ by:
\setlength{\abovedisplayskip}{3pt}
\setlength{\belowdisplayskip}{3pt}
\setlength{\abovedisplayshortskip}{0pt}
\setlength{\belowdisplayshortskip}{0pt}
\begin{subequations}
% \vspace{-3mm}
    \begin{align}
        &Q_i^G  = \mathrm{Conv}_{3\times 3}({G}_i),\\
        &K_i^L  = \mathrm{Conv}_{3\times 3}({L}_i),\\
        &V_i^L  = \mathrm{Conv}_{3\times 3}({L}_i),
    \end{align}
\end{subequations}
where $\{Q_i^G, K_i^L , V_i^L \} \in \mathbb{R}^{ \hat{H}\times \hat{W}\times \hat{C}}$, $\hat{H}\times \hat{W}$ and $\hat{C}$ denote the spatial and channel dimensions, respectively; $\mathrm{Conv}_{3\times3}(\cdot)$ denotes a convolution with the filter size of $3\times 3$ pixels.
We denote the matrix forms of $Q_i^G$, $K_i^L$, $V_i^L$ as $\mathbf{\hat{Q}}_i^G$, $\mathbf{\hat{K}}_i^L$, $\mathbf{\hat{V}}_i^L$, and obtain the fused feature by:
\begin{equation}
% \vspace{-2mm}
\mathbf{\hat{F}}_i  = \mathrm{softmax}\left(\frac{\mathbf{\hat{Q}}_i^G(\mathbf{\hat{K}}_i^L)^{\top}}{\alpha}\right) \mathbf{\hat{V}}_i^L,
\label{eq: cross-attention}
\end{equation}
where $\mathbf{\hat{Q}}_i^G \in \mathbb{R}^{\hat{C} \times \hat{H}\hat{W}}$, $\mathbf{\hat{K}}_i^L \in \mathbb{R}^{\hat{C} \times \hat{H}\hat{W}}$ and $\mathbf{\hat{V}}_i^L\in \mathbb{R}^{\hat{C} \times \hat{H}\hat{W}}$ are obtained by reshaping tensors~\cite{zamir2022restormer} from the original size $\mathbb{R}^{ \hat{H}\times \hat{W}\times \hat{C}}$; $\mathrm{softmax}(\cdot)$ denotes the softmax operation that is applied to each row of the matrix; $\alpha$ is a scaling factor.

In our implementation, we take the features of the last 4 layers of ViT-S/14 from DINOv2 and concatenate them together in channel dimension as the feature ${G}_i$.
For the feature ${L}_i$, we take stage-4 features with stride 16 from the base ResNet50~\cite{He_2016_CVPR}.
Finally, the feature  $\mathbf{\hat{F}}_i~\in~\mathbb{R}^{\hat{C} \times \hat{H}\hat{W}}$ is reshaped into $F_i~\in~\mathbb{R}^{\hat{H}\times \hat{W}\times \hat{C}}$ for the following processing.

\vspace{-3mm}
\subsection{MFP module}\label{sec:32}
\vspace{-2mm}

Inspired by the efficient mechanisms of human brains in memorizing long-term information, \ie, paying more attention to frequently used information, we propose a memory-based feature propagation module to effectively and efficiently explore temporal features by establishing reliable connections with features from far-apart frames and alleviating the influence of inaccurately estimated features.

Assuming that we have predicted $i-1$ color frames $\{{I}_1,\cdots,{I}_{i-1}\}$ with their chrominance channels $\{{Y}_1,\cdots,{Y}_{i-1}\}$ and luminance channels $\{{X}_1,\cdots,{X}_{i-1}\}$
 (\ie, the input grayscale frames), we aim to estimate the chrominance channel ${Y}_{i}$ of the $i$-th color frame ${I}_i$ based on $\{{Y}_1,\cdots,{Y}_{i-1}\}$ and the given exemplar $R$.

First, we extract features $\{F_1, F_2, \cdots, F_{i}\}$ and $F_r$ from $\{X_1, X_2, \cdots, X_i\}$ and the luminance channel ${X}_r$ of $R$ using the proposed PVGFE module for global and semantic features in the spatial dimension.
Meanwhile, we extract features $\{E_1,\cdots,E_{i-1}\}$ and $E_r$ from $\{Y_1,\cdots,Y_{i-1}\}$ and the chrominance channel ${Y}_r$ of $R$ using a lightweight pretrained ResNet18~\cite{He_2016_CVPR}.
Then, we generate the embedded query, key, and value features by:
\begin{subequations}
\begin{align}
&Q_i^F  = \mathrm{Conv}_{3\times 3}({F}_i), \label{eq:qif}\\
&\{K_j^F\}_{j=1}^{i-1}  = \{\mathrm{Conv}_{3\times 3}({F}_j)\}_{j=1}^{i-1},\label{eq:kjf} \\
&K_r^F  = \mathrm{Conv}_{3\times 3} (F_r), \\
&\{V_j^E\}_{j=1}^{i-1}  = \{\mathrm{Conv}_{3\times 3}({E}_j)\}_{j=1}^{i-1}, \label{eq:vje}\\
&V_r^E  = \mathrm{Conv}_{3\times 3} (E_r).
\label{eq:query-key}
\end{align}
\end{subequations}

As $\{K_1^F,\cdots,K_{i-1}^F\}$ and $\{V_1^E,\cdots,V_{i-1}^E\}$ contain valuable historical information of previously colorized video frames $\{{I}_1,\cdots,{I}_{i-1}\}$, they facilitate the establishment of long-range temporal correspondences between the contents of the current frame and those of the previously colorized video frames.
However, memorizing all of the historical frames results in significant GPU memory consumption, especially as the number of colorized frames increases.
As a trade-off between long-range correspondence and memory consumption, we keep every $\gamma$ frame and discard the remaining ones that contain similar contents to obtain more temporally compact and representative features.

In particular, given that adjacent frames are mutually redundant, we first merge the temporal features by concatenating every $\gamma$ frame, thereby establishing reliable connections with far-apart frames under constrained memory consumption, and obtain the aggregated key features $\mathbf{A}_1^K$ by:
\begin{equation}
        \mathbf{A}_1^K = \mathrm{Concat}(\mathbf{K}_{\gamma}^F, \mathbf{K}_{2\gamma}^F, \cdots, \mathbf{K}_{z\gamma}^F),
    \label{eq:c2}
\end{equation}
where $\{ \mathbf{K}_{\gamma}^F, \mathbf{K}_{2\gamma}^F, \cdots, \mathbf{K}_{z\gamma}^F\}$ are the matrix forms of $\{K_{\gamma}^F,K_{2\gamma}^F,\cdots,K_{z\gamma}^F \}$, $z = \lfloor \frac{i-1}{\gamma} \rfloor$, $\lfloor \cdot \rfloor$ denotes the round down operation, and $\mathrm{Concat}(\cdot)$ denotes the spatial dimension concatenation operation.

Note that errors are likely to accumulate when the features of some frames are not estimated accurately.
In addition, the high dimension ($\hat{H}\hat{W}$) of $\mathbf{A}_1^K$ makes it impossible to handle lots of video frames with limited GPU memory capacity.
To overcome these problems, we then aggregate $\mathbf{A}_1^K$ into a more spatially compact and less error-prone form by selecting better features with higher usage.
However, the computation of the feature usage frequency relies on a sufficient number of colorized frames.
Therefore, when the number of colorized frames is small (\ie, $z<N_s$), we directly use $\mathbf{A}_2^K$ as the output of our proposed MFP module, which is defined as:
\begin{equation}
        \mathbf{A}_2^K = \mathrm{Concat}\left(\mathbf{K}_{r}^F,\mathbf{A}_1^K \right),
    \label{eq:a2k_small}
\end{equation}
% \useshortskip
where $\mathbf{K}_{r}^F$ is the matrix form of $K_{r}^F$.
When a sufficient number of frames is colorized (\ie, $z=N_s$), we aggregate $\mathbf{A}_1^K$ by compressing the earlier $N_e$ frames to reduce GPU memory consumption and alleviate the influence of inaccurately estimated features for better video colorization.
After the aggregation, the total number of aggregated frames reduced from $z=N_s$ to $z=N_s - N_e$ and we use $\mathbf{A}_2^K$ in \eqref{eq:a2k_small} as the output of the MFP module until $z$ reaches $N_s$ again.
In the following, we explain the situation when $z=N_s$ in detail.

We first define the cumulative similarities $\{\mathbf{S}_{\gamma},\cdots,\mathbf{S}_{z\gamma} \}$ for the key features $\{\mathbf{K}_{\gamma}^F,\cdots,\mathbf{K}_{z\gamma}^F \}$, which are aggregated in \eqref{eq:c2}, as:
\begin{subequations}
{  \setlength{\abovedisplayskip}{0pt}
  \setlength{\belowdisplayskip}{\abovedisplayskip}
  \setlength{\abovedisplayshortskip}{0pt}
  \setlength{\belowdisplayshortskip}{0pt}
\begin{align}
     &\mathbf{S}_{\gamma} = \sum_{j={\gamma}+1}^{i-1} \sum_{m=1}^{\hat{H}\hat{W}} \left(\mathrm{softmax}\left( -\mathbf{C}^j  \right) \right), \mathbf{C}^j=(\mathbf{C}_{m,n}^j),\\
     &\mathbf{C}_{m,n}^j = {\left \Vert \mathbf{K}_{j}^F(:,m) - \mathbf{K}_{\gamma}^F(:,n) \right\Vert_2^2},
    \label{eq:sz}
\end{align}
}%
\end{subequations}
where $\mathbf{S}_{\gamma} \in \mathbb{R}^{1\times \hat{H}\hat{W}}$ and $\mathbf{C}^{j} \in \mathbb{R}^{\hat{H}\hat{W} \times \hat{H}\hat{W}}$,
$\mathbf{K}_{j}^F(:,m)$ denotes the $m$-th feature vector in $\mathbf{K}_j^F$ and $\mathbf{K}_{\gamma}^F(:,n)$ denotes the $n$-th feature vector in $\mathbf{K}_{\gamma}^F$, $\{\mathbf{K}_{j}^F, \mathbf{K}_{\gamma}^F\}\in \mathbb{R}^{\hat{C}^k \times \hat{H}\hat{W}}$, $\left \Vert \cdot \right\Vert_2$ denotes the Euclidean distance calculation.
Then, we normalize $\mathbf{S}_{{\gamma}}$ by dividing it by the number of frames in $\{\mathbf{K}_{j}^F\}_{j=\gamma+1}^{i-1}$ for fairness consideration and obtain $\mathbf{{S}}^{'}_{{\gamma}} = \frac{\mathbf{{S}}_{{\gamma}}}{i-1-{\gamma}}$, which can be utilized to estimate the probability that the feature $\mathbf{K}_{\gamma}^F$ is accurately predicted as $\mathbf{{S}}^{'}_{{\gamma}}$ describes how frequently the feature is used.
We further define a top-$M$ operation $\mathcal{T}_M(\cdot)$ to select the best $M$ pixels of all features from the earlier $N_e$ frames $\{\mathbf{K}_{\gamma}^F , \cdots , \mathbf{K}_{N_e\gamma}^F\}$, based on the highest $M$ values in $\{\mathbf{{S}}^{'}_{\gamma} , \cdots , \mathbf{{S}}^{'}_{N_e\gamma}\}$ as:
% in both the spatial and temporal dimensions, \eg, $N_e\hat{H}\hat{W}$,  by:
\begin{subequations}
\begin{align}
            &\mathbf{K}^{c} = \mathrm{Concat}\left(\mathbf{K}_{\gamma}^F , \cdots , \mathbf{K}_{N_e\gamma}^F \right),
            \\
            &\mathbf{{S}}^{c} = \mathrm{Concat}\left(\mathbf{{S}}^{'}_{\gamma} , \cdots , \mathbf{{S}}^{'}_{N_e\gamma} \right),
            \\  &\mathcal{T}_M\left(\mathbf{K}^{c} \right) = \mathrm{Concat}\left(\mathbf{K}^{c}(:, 1),\cdots,\mathbf{K}^{c}(:, M)  \right),
    \label{eq:topl}
\end{align}
\end{subequations}
where $\mathbf{K}^{c}\in \mathbb{R}^{\hat{C}^k \times N_e\hat{H}\hat{W}}$, $\mathbf{{S}}^{c}\in \mathbb{R}^{1\times N_e\hat{H}\hat{W}}$, and $\{\mathbf{K}^{c}(:,1),\cdots,\mathbf{K}^{c}(:,M)\}$ denote the $M$ feature vectors in $\mathbf{K}^{c}$ that satisfy $\{\mathbf{{S}}^{c}(:,1),\cdots,\mathbf{{S}}^{c}(:,M)\}$ are the $\text{top-}M$ values in $\mathbf{{S}}^{c}$.
Then we obtain the aggregated key features $\mathbf{A}_2^K$ by:
% \vspace{-1mm}
\begin{equation}
% \vspace{-1mm}
% \footnotesize
        \mathbf{A}_2^K = \mathrm{Concat}\left(
           \mathcal{T}_M \left( \mathbf{K}^{c}\right)  ,   \mathbf{K}_{r}^F , \mathbf{K}_{(N_e+1)\gamma}^F , \cdots , \mathbf{K}_{z\gamma}^F \right),
    \label{eq:c3}
\end{equation}
where $\mathbf{A}_2^K \in \mathbb{R}^{\hat{C}^k \times T}, T=M+(1+z-N_e)\hat{H}\hat{W}$.
Similarly, we obtain the aggregated value features $\mathbf{A}_2^V \in \mathbb{R}^{\hat{C}^v \times T}$ by aggregating $\{V_r^E,\{V_j^E\}_{j=1}^{i-1}\}$.

To fully exploit the temporal information contained in $\mathbf{A}_2^K$ and $\mathbf{A}_2^V$, we use a commonly used $L_2$ similarity to find the features in $\mathbf{A}_2^K$ that are most similar to $\mathbf{Q}_i^F$, where $\mathbf{Q}_i^F \in \mathbb{R}^{\hat{C}^k \times \hat{H}\hat{W}}$ is the matrix form of $Q_i^F$, by:
\begin{subequations}
\begin{align}
% \vspace{-5mm}
    &\mathbf{W}_i = \mathrm{softmax}\left( -\mathbf{D}^i\right), \mathbf{D}^i=(\mathbf{D}_{m,n}^i),\\
    & \mathbf{D}^i_{m,n} =  {\left\Vert \mathbf{Q}_i^F(:,m) - {\mathbf{A}_2^{K}(:,n)} \right\Vert_2^2},
    \label{eq:si}
\end{align}
\end{subequations}
where $\{\mathbf{W}_i,\mathbf{D}^i\}$~$\in$~$\mathbb{R}^{\hat{H}\hat{W}\times  T}$, $\mathbf{Q}_{i}^F(:,m)$ denotes the $m$-th feature vector in $\mathbf{Q}_i^F$ and $\mathbf{A}_{2}^{K}(:,n)$ denotes the $n$-th feature vector in $\mathbf{A}_{2}^{K}$.
Then, we reconstruct the $i$-th estimated feature $\mathbf{V}_i$~$\in$~$\mathbb{R}^{\hat{C}^v \times \hat{H}\hat{W}}$ of the chrominance channel by mapping the aggregated value features $\mathbf{A}_2^V$ based on $\mathbf{W}_i$ as:
% \vspace{-2mm}
\begin{equation}
% \vspace{-2mm}
      \mathbf{V}_i = \mathbf{A}_2^V {(\mathbf{W}_i)}^{\top}.
    \label{eq:cross-channel-attention}
\end{equation}
Finally, we obtain the estimated feature ${V}_i$ by reshaping $\mathbf{V}_i$ to its original size $\mathbb{R}^{\hat{H} \times \hat{W} \times \hat{C}^v}$.
In the following, we enhance $V_i$ by a local attention (LA) module.

\vspace{-3mm}
\subsection{LA module}\label{sec:33}
\vspace{-2mm}
As adjacent frames contain similar contents that may be useful to complement the long-range information captured by MFP, we develop a local attention (LA) module to explore better spatial-temporal features.

We first use $Q_{i}^p$~$\in$~$\mathbb{R}^{1\times 1 \times \hat{C}^k}$ to represent the feature $Q_{i}^F$ in \eqref{eq:qif} at the spatial location $p$~$\in$~$\mathbb{R}^{\hat{H}\times \hat{W}}$.
Next, we formulate the past $d$ key features in \eqref{eq:kjf} and past $d$ value features in \eqref{eq:vje} of the spatial-temporal neighborhood corresponding to $Q_i^p$ as:
% \vspace{-2mm}
\begin{subequations}
\begin{align}
% \vspace{-2mm}
        &{K}^{\mathcal{N}(p),c} = \mathrm{Concat}(K_{i-d}^{\mathcal{N}(p)},\cdots,K_{i-1}^{\mathcal{N}(p)} ),\\
        &{V}^{\mathcal{N}(p),c} = \mathrm{Concat}(V_{i-d}^{\mathcal{N}(p)},\cdots,V_{i-1}^{\mathcal{N}(p)} ),
    \label{eq:kd}
\end{align}
\end{subequations}
where $K_{j}^{\mathcal{N}(p)}$ and $V_{j}^{\mathcal{N}(p)}$ denote the $j$-th key feature and value feature corresponding to a $\lambda\times\lambda$ patch $\mathcal{N}(p)$ centered at $p$, ${K}^{\mathcal{N}(p),c}$ $\in$~$\mathbb{R}^{d \times{\lambda} \times \lambda  \times \hat{C}^k}$ and ${V}^{\mathcal{N}(p),c}$ $\in$~$\mathbb{R}^{d \times{\lambda} \times \lambda  \times \hat{C}^v}$.
Then we apply the local attention to $Q_{i}^p$ with ${K}^{\mathcal{N}(p),c}$ and ${V}^{\mathcal{N}(p),c}$ and obtain the output of LA module at location $p$ as:
% \vspace{-2mm}
\begin{equation}
% \vspace{-2mm}
\mathbf{L}_i^p  = \mathrm{softmax}\left(\frac{\mathbf{Q}_{i}^p{(\mathbf{K}^{\mathcal{N}(p),c})}^{\top}}{\beta}\right) \mathbf{V}^{\mathcal{N}(p),c},\\
\label{eq:lam}
\end{equation}
where $\mathbf{Q}_{i}^p$~$\in$~$\mathbb{R}^{1\times \hat{C}^k }$, ${\mathbf{K}^{\mathcal{N}(p),c}}$~$\in$~$\mathbb{R}^{d{\lambda}^2 \times \hat{C}^k}$ and ${\mathbf{V}^{\mathcal{N}(p),c}}$~$\in$~$\mathbb{R}^{d{\lambda}^2 \times \hat{C}^v}$ are matrices obtained by reshaping $Q_{i}^p$, ${K}^{\mathcal{N}(p),c}$ and ${V}^{\mathcal{N}(p),c}$ from their original size, $\beta$ is the scaling factor.
Finally, we obtain the feature $L_i$~$\in$~$\mathbb{R}^{\hat{H} \times \hat{W} \times \hat{C}^v}$ by reshaping $\mathbf{L}_i$ to its original size.

To restore the colors of the input frame $X_i$, we further adopt a simple but effective decoder.
Specifically, we use a decoder $\mathcal{D}(\cdot)$ consisting of ResBlocks~\cite{He_2016_CVPR} followed by up-sampling interpolation layers to gradually refine the enhanced feature ${V}_i+L_i$ and obtain the predicted $i$-th chrominance channel $Y_i$ as:
\begin{equation}
% \vspace{-2mm}
    Y_i  =  \mathcal{D}({V}_i + L_i).
\label{eq:yi_2}
\end{equation}

\begin{table*}[!t]
	\centering
	\small
 \caption{Quantitative comparisons of the proposed method against state-of-the-art ones on the DAVIS~\cite{Perazzi_CVPR_2016} validation set (short frame length), the Videvo~\cite{Lai2018videvo} validation set (medium frame length) and the NVCC2023~\cite{Kang_2023_CVPR} validation set (long frame length). Our method achieves favorable performance in most of the metrics. Top $1_{st}$ and $2_{nd}$ results are marked in \textcolor{red}{\textbf{bold red}} and \textcolor{blue}{blue} respectively. $^*$ denotes that we apply DVP~\cite{lei2020dvp} to the results of Color2Embed and DDColor as post-processing method. Note that for the LPIPS matrix on Videvo, we examine the performance of our method and BiSTNet in additional decimal places to determine the best one and the second best one as the performance of these two methods appears to be identical when displayed in the table with a limited number of decimal places. $^\dagger$ denotes that two exemplars are used.}
 \vspace{-2mm}
	\begin{adjustbox}{max width=\linewidth}
            \begin{threeparttable}
    		\begin{tabular}{l|cccc|cc|cccc}
    			\noalign{\hrule height 0.3mm}
    			Methods       & \multicolumn{1}{c}{\makecell{DDColor\\~\cite{kang2022ddcolor}} } & \multicolumn{1}{c}{\makecell{Color2Embed\\~\cite{zhao2021color2embed}}}  &  \multicolumn{1}{c}{\makecell{$\text{DDColor}^*$\\~\cite{kang2022ddcolor}} } & \multicolumn{1}{c|}{\makecell{$\text{Color2Embed}^*$\\~\cite{zhao2021color2embed}} } & \multicolumn{1}{c}{\makecell{VCGAN\\~\cite{vcgan}}} & \multicolumn{1}{c|}{\makecell{TCVC\\~\cite{liu2021temporally}}}  & \makecell{DeepExemplar\\~\cite{zhang2019deep}} & \multicolumn{1}{c}{\makecell{DeepRemaster\\~\cite{IizukaSIGGRAPHASIA2019}}} & \multicolumn{1}{c}{\makecell{BiSTNet$^\dagger$\\~\cite{bistnet}}} &  \multicolumn{1}{c}{\makecell{{ColorMNet}\\ {(Ours)}} } \\ \hline

                Categories & \multicolumn{4}{c}{Image-based} \vline & \multicolumn{2}{c}{Fully-automatic} \vline & \multicolumn{4}{c}{Exemplar-based} \\ \hline
                % DAVIS
                \textbf{DAVIS} & \multicolumn{4}{c|}{} & \multicolumn{2}{c|}{} & \multicolumn{4}{c}{} \\
                PSNR (dB)${\uparrow}$ & 30.84 & 31.33 & 30.67 & 31.05 & 30.24 & 31.10 & 33.24 & 33.25 & \textcolor{blue}{34.02} &  \textcolor{red}{\textbf{35.77}}\\
                FID${\downarrow}$ & 65.13 & 101.08 & 86.96 & 118.11 & 128.48 & 116.41 & 69.56 & 92.28 & \textcolor{blue}{44.69} &  \textcolor{red}{\textbf{38.39}} \\
                SSIM${\uparrow}$ & 0.926 & 0.951 & 0.936 & 0.943 & 0.924 & 0.955 & 0.950 & 0.961 & \textcolor{blue}{0.964} &  \textcolor{red}{\textbf{0.970}}\\
                LPIPS${\downarrow}$ & 0.085 & 0.076 & 0.086 & 0.086 & 0.100 & 0.080 & 0.062 & 0.060 & \textcolor{blue}{0.043} &  \textcolor{red}{\textbf{0.035}}

                \\ \hline
                % Videvo
                \textbf{Videvo} & \multicolumn{4}{c|}{} & \multicolumn{2}{c|}{} & \multicolumn{4}{c}{} \\
                PSNR (dB)${\uparrow}$ & 30.76 & 31.65 & 30.60 & 31.62 & 30.62 & 31.29 & 33.11 & 32.95 & \textcolor{blue}{34.12} &  \textcolor{red}{\textbf{34.35}}\\
                FID${\downarrow}$ & 45.08 & 66.73 & 50.80 & 73.02 & 97.86 & 80.74 & 54.93 & 68.15 & \textcolor{blue}{32.25} &  \textcolor{red}{\textbf{30.76}} \\
                SSIM${\uparrow}$ & 0.925 & 0.958 & 0.940 & 0.960 & 0.934 & 0.956 & 0.956 & 0.964 & \textcolor{blue}{0.968} &  \textcolor{red}{\textbf{0.972}}\\
                LPIPS${\downarrow}$ & 0.079 & 0.060 & 0.077 & 0.061 & 0.085 & 0.068 & 0.053 & 0.053 & \textcolor{blue}{0.036} &  \textcolor{red}{\textbf{0.036}}

                \\ \hline
                % NTIRE 2023
                \textbf{NVCC2023} & \multicolumn{4}{c|}{} & \multicolumn{2}{c|}{} & \multicolumn{4}{c}{} \\
                PSNR (dB)${\uparrow}$ & 29.95 & 30.90 & 30.16 & 30.66 & 29.77 & 30.45 & 32.03 & 32.25 & \textcolor{blue}{33.18} &  \textcolor{red}{\textbf{33.26}}\\
                FID${\downarrow}$ & 48.50 & 65.92 & 95.83 & 70.35 & 86.59 & 72.76 & 35.39 & 53.03 & \textcolor{blue}{25.55} &  \textcolor{red}{\textbf{20.16}} \\
                SSIM${\uparrow}$ & 0.888 & 0.933 & 0.911 & 0.927 & 0.866 & 0.935 & 0.930 & \textcolor{blue}{0.951} & 0.949 &  \textcolor{red}{\textbf{0.959}}\\
                LPIPS${\downarrow}$ & 0.114 & 0.091 & 0.099 & 0.098 & 0.130 & 0.089 & 0.073 & 0.071 & \textcolor{blue}{0.054} &  \textcolor{red}{\textbf{0.039}}
                \\ \hline
    			\noalign{\hrule height 0.3mm}

    		\end{tabular}

            \end{threeparttable}
	\end{adjustbox}
	\label{tab:psnr}
	\vspace{-6mm}
\end{table*}

\vspace{-3mm}
\section{Experimental Results}
\vspace{-2mm}
\label{sec:experiments}
In this section, we first describe the experimental settings of the proposed ColorMNet.
Then we evaluate the effectiveness of our approach against state-of-the-art methods.
More experimental results are included in the supplemental material. Code and models are available at \href{https://github.com/yyang181/colormnet}{https://github.com/yyang181/colormnet}.

\vspace{-3mm}
\subsection{Experimental settings}
\label{ssec:datasets}
\vspace{-1mm}

\noindent\textbf{Datasets.}
Following previous works~\cite{liu2021temporally, chen2023exemplar, bistnet}, we use the datasets of DAVIS~\cite{Perazzi_CVPR_2016} and Videvo~\cite{Lai2018videvo} for training and generate grayscale video frames using OpenCV library.
For testing, we use three popular benchmark test datasets including the DAVIS validation set, the Videvo validation set and the official validation set of NTIRE 2023 Video Colorization Challenge~\cite{Kang_2023_CVPR} (NVCC2023 for short).

\noindent\textbf{Exemplars.}
Following~\cite{bistnet,IizukaSIGGRAPHASIA2019, zhang2019deep, chen2023exemplar}, we adopt a similar strategy to utilize the first frame of each video clip as an exemplar to colorize the video clip.

\noindent\textbf{Evaluation metrics.}
Following the experimental protocol of most existing colorization methods, we use peak signal-to-noise ratio (PSNR), structural similarity index measurement (SSIM)~\cite{wang2004image}, the Fr\'echet Inception Distance (FID)~\cite{heusel2017gans}, and the learned perceptual image patch similarity (LPIPS)~\cite{zhang2018unreasonable} as evaluation metrics.
These assessment matrices cover a spectrum of pixel-wise considerations, the distribution similarity between generated images as well as ground truth images, and the perception similarity.

\noindent\textbf{Implementation details.}
We train our model on a machine with one RTX A6000 GPU. We adopt the Adam optimizer~\cite{kingma2014adam} with default parameters using PyTorch~\cite{paszke2019pytorch} for 160,000 iterations. The batch size is set to 4.
We adopt the CIE LAB color space for each frame in our experiments.
The learning rate is set to a constant $2 \times {10}^{-5}$.
We empirically set $\gamma=5$, $N_e=5$, $N_s=10$, $M=128$ and $d=1$.
We employ an $L_1$ loss, computed as the mean absolute errors between the predicted images and the ground truths.

\vspace{-3mm}
\subsection{Comparisons with state-of-the-art methods}
\vspace{-1mm}
\noindent\textbf{Quantitative comparison.}
We benchmark our method against state-of-the-art ones on three datasets and report quantitative results.
The competing methods include the automatic colorization techniques~\cite{vcgan, liu2021temporally}, single exemplar-based approaches~\cite{IizukaSIGGRAPHASIA2019, zhang2019deep}, and a double exemplar-based method~\cite{bistnet}.
Furthermore, for enhanced self-containment, we incorporate comparisons with leading image colorization methods~\cite{zhao2021color2embed, kang2022ddcolor} and enhance the temporal consistency of these methods by applying DVP~\cite{lei2020dvp} as post-processing method.
For~\cite{liu2021temporally, bistnet, IizukaSIGGRAPHASIA2019, vcgan}, whose weights are trained on the same datasets as our method, we conduct tests using their official codes and weights provided by the authors.
However, for~\cite{kang2022ddcolor, zhao2021color2embed, zhang2019deep}, we train from scratch using the official codes on identical datasets as our method to ensure fair comparisons.
In addition, we adhere to the commonly adopted protocol of using the same first frame ground truth image of each video clip as the exemplar for testing all exemplar-based methods~\cite{zhang2019deep, IizukaSIGGRAPHASIA2019, zhao2021color2embed, bistnet}, with the exception that we also include the last frame as an extra exemplar for testing the double exemplar-based method, BiSTNet~\cite{bistnet}.

Table~\ref{tab:psnr} shows that the ColorMNet consistently generates competitive colorization results on the DAVIS~\cite{Perazzi_CVPR_2016} validation set, the Videvo~\cite{Lai2018videvo} validation set, and the NVCC2023~\cite{Kang_2023_CVPR} validation set, where our method performs better than the evaluated methods in terms of PSNR, SSIM, FID, and LPIPS, indicating that our method not only can generate both high-quality and high-fidelity colorization results, but also demonstrates the good generalization based on the favorable performance on the validation set in NVCC2023 (the training set in NVCC2023 is not included in the training phase).

\noindent\textbf{Qualitative evaluation.}
\begin{figure*}[!t]\tiny
 \begin{center}
  \begin{tabular}{cccccccc}
 \multicolumn{3}{c}{\multirow{5}*[42pt]
 {\includegraphics[width=0.368\linewidth, height=0.290\linewidth]{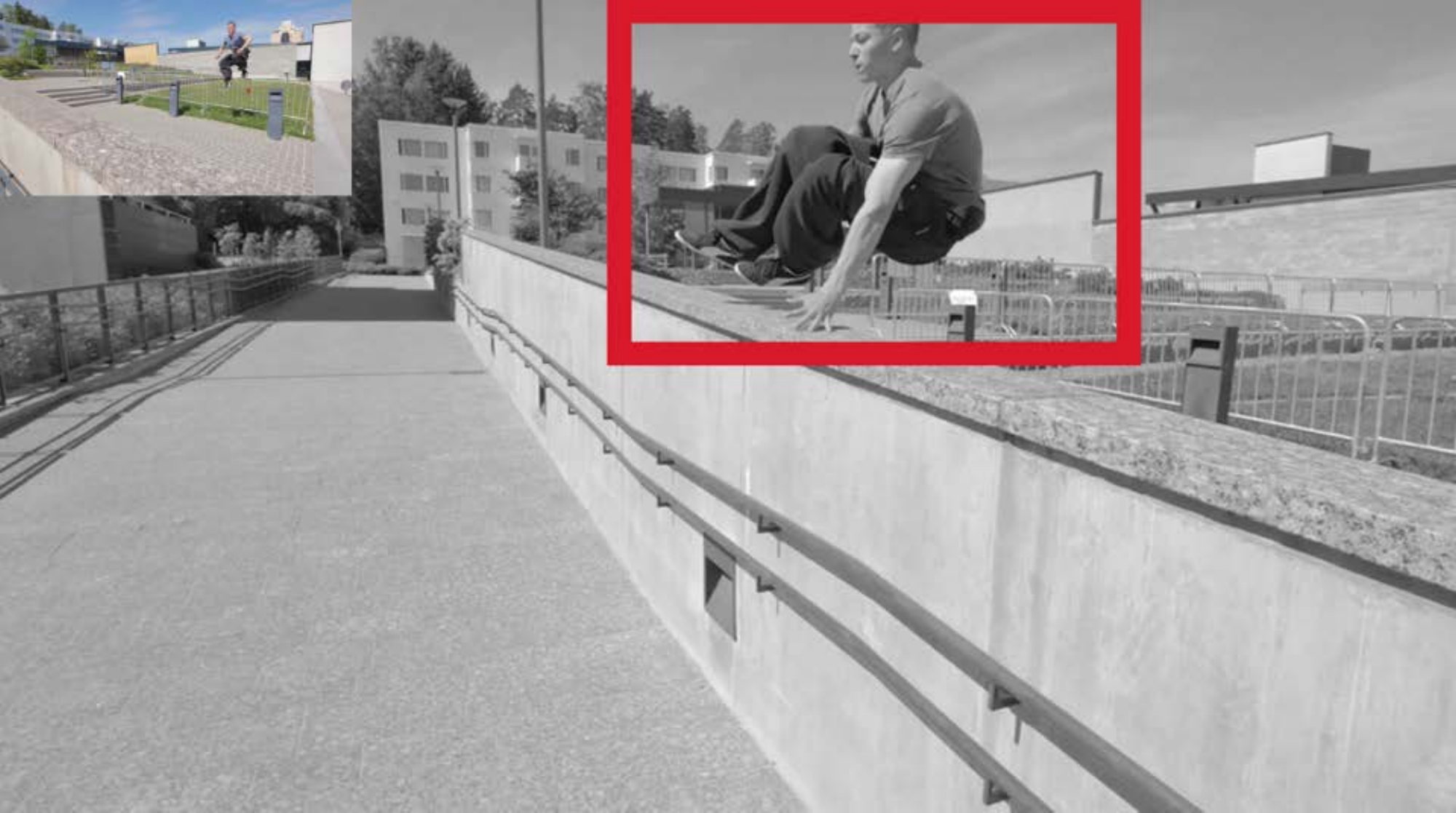}}}&\hspace{0.0mm}
  \includegraphics[width=0.142\linewidth, height = 0.132\linewidth]{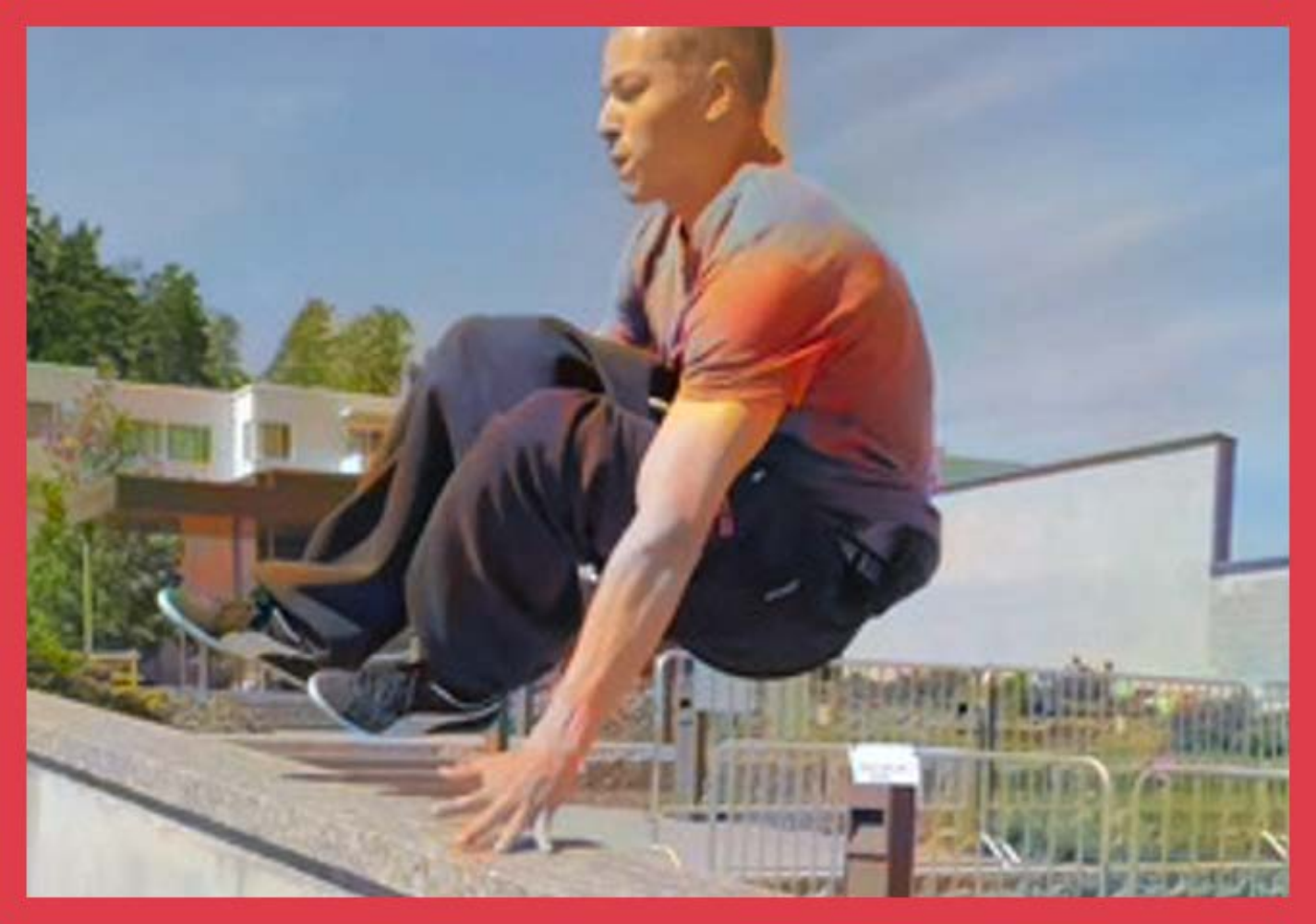} &\hspace{0.0mm}
  \includegraphics[width=0.142\linewidth, height = 0.132\linewidth]{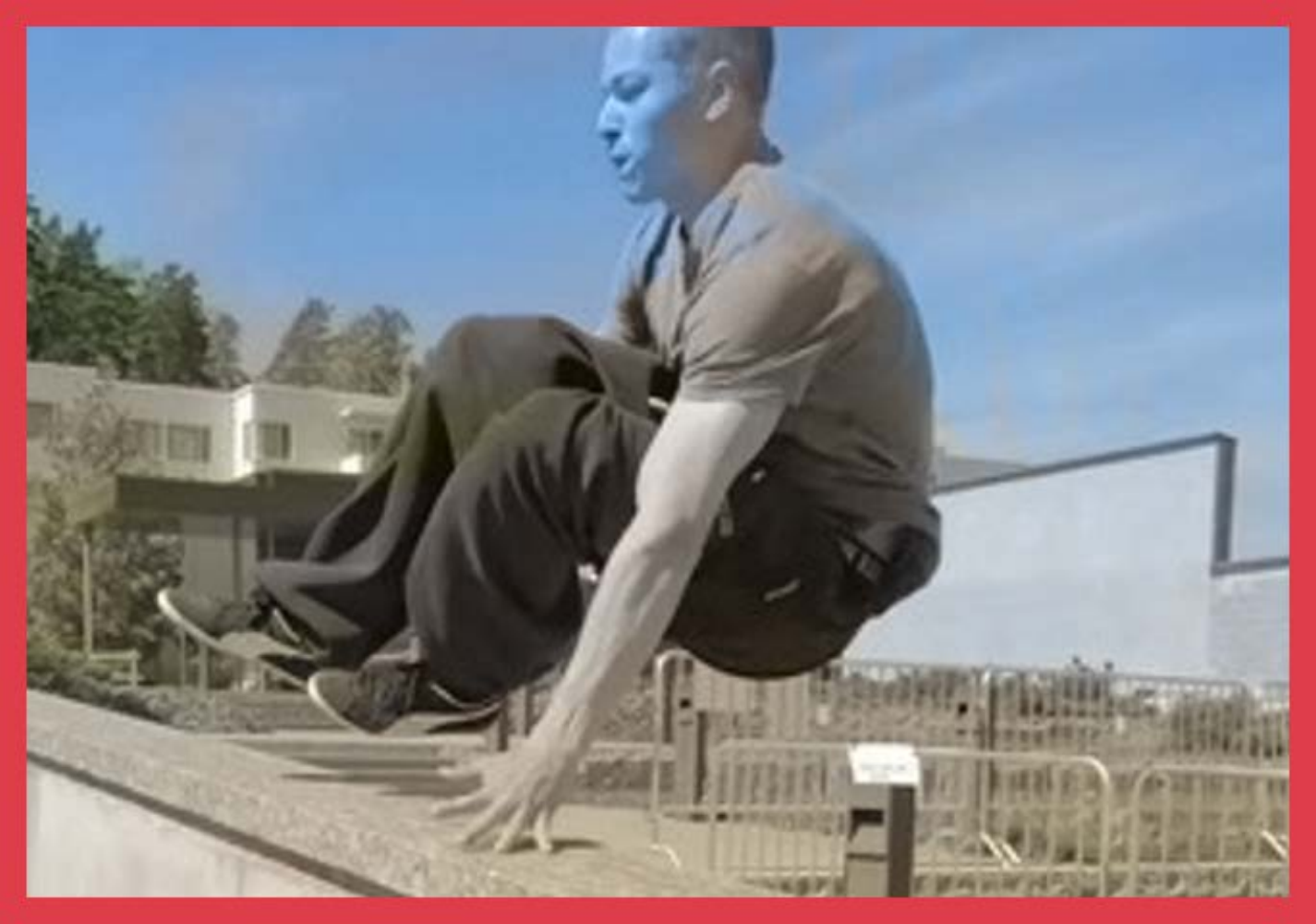} &\hspace{0.0mm}
  \includegraphics[width=0.142\linewidth, height = 0.132\linewidth]{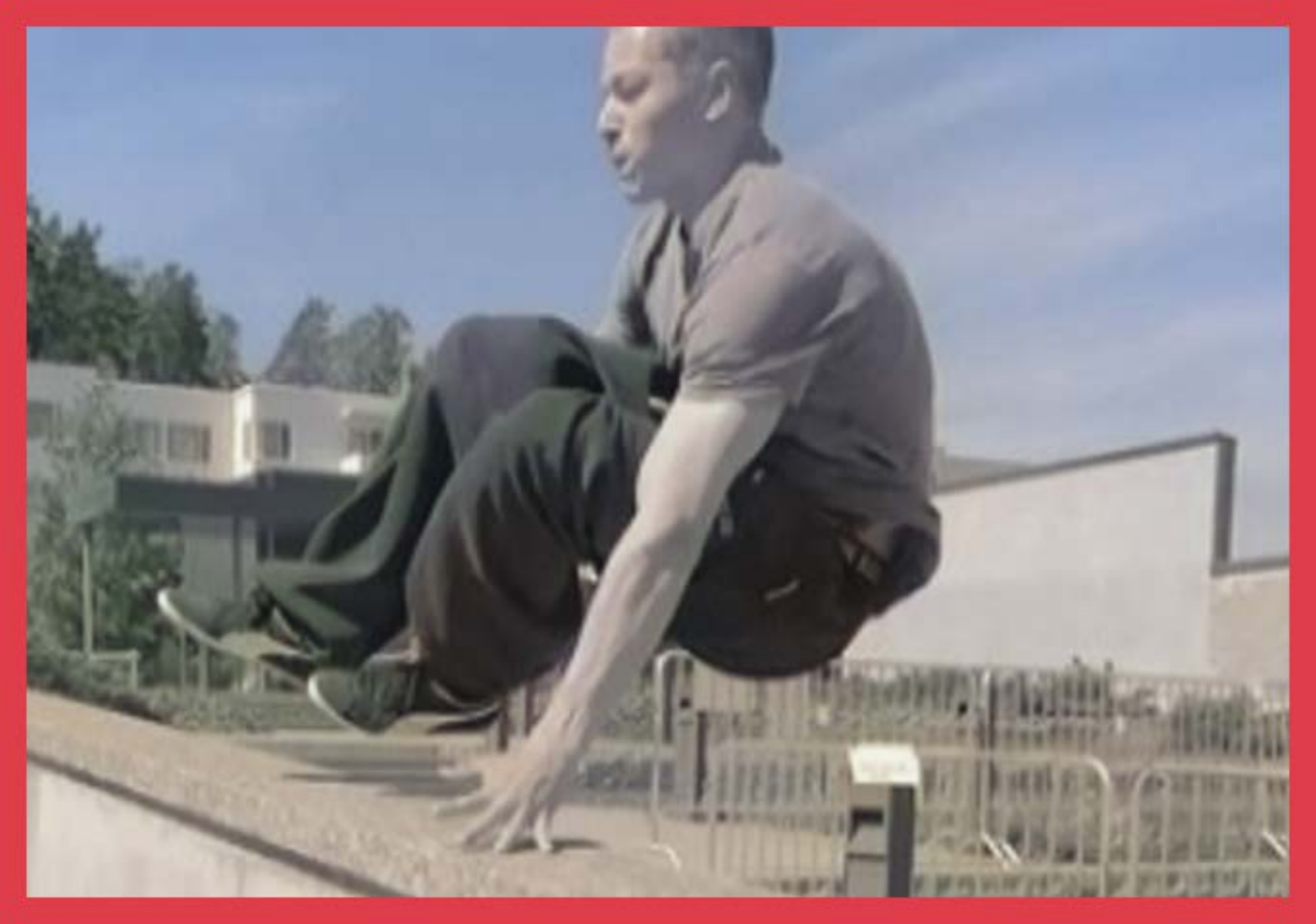} &\hspace{0.0mm}
  \includegraphics[width=0.142\linewidth, height = 0.132\linewidth]{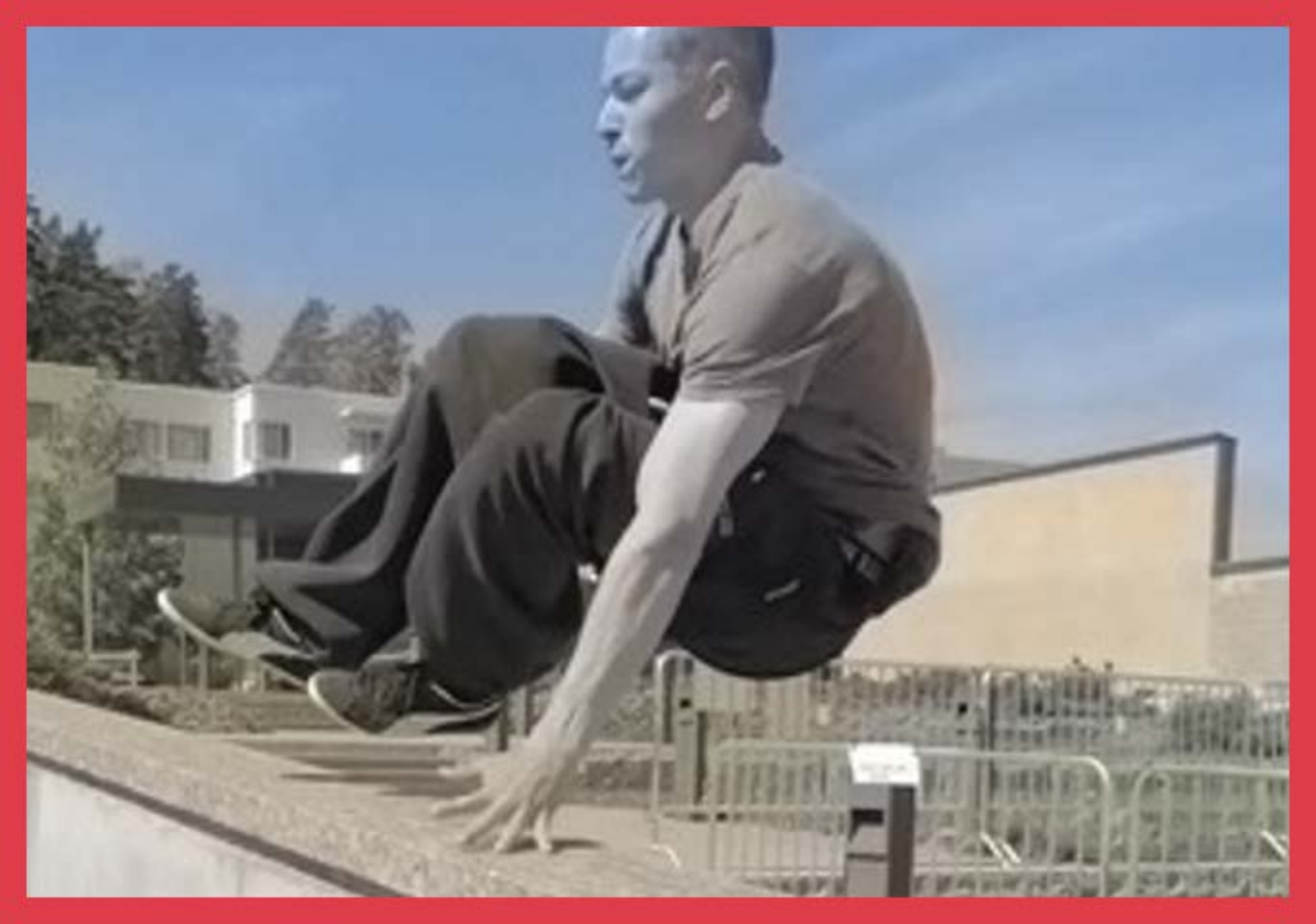}&\hspace{0.0mm}\\
  \multicolumn{3}{c}{~} &\hspace{0.0mm}   (a)  &\hspace{0.0mm}  (b)   &\hspace{0.0mm}  (c)
  &\hspace{0.0mm} (d) \\

  \multicolumn{3}{c}{~} & \hspace{0.0mm}
    \includegraphics[width=0.142\linewidth, height = 0.132\linewidth]{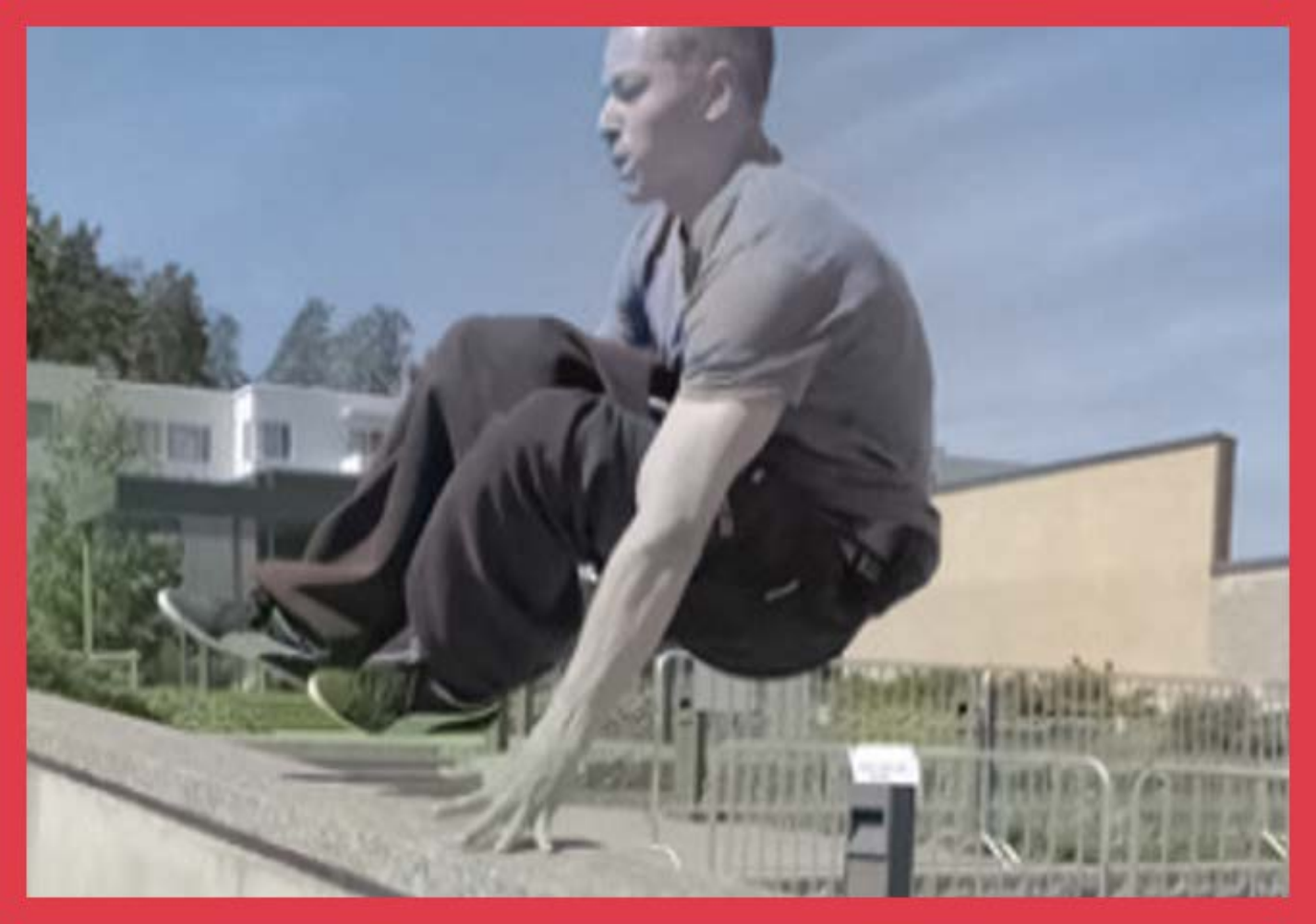} & \hspace{0.0mm}
  \includegraphics[width=0.142\linewidth, height = 0.132\linewidth]{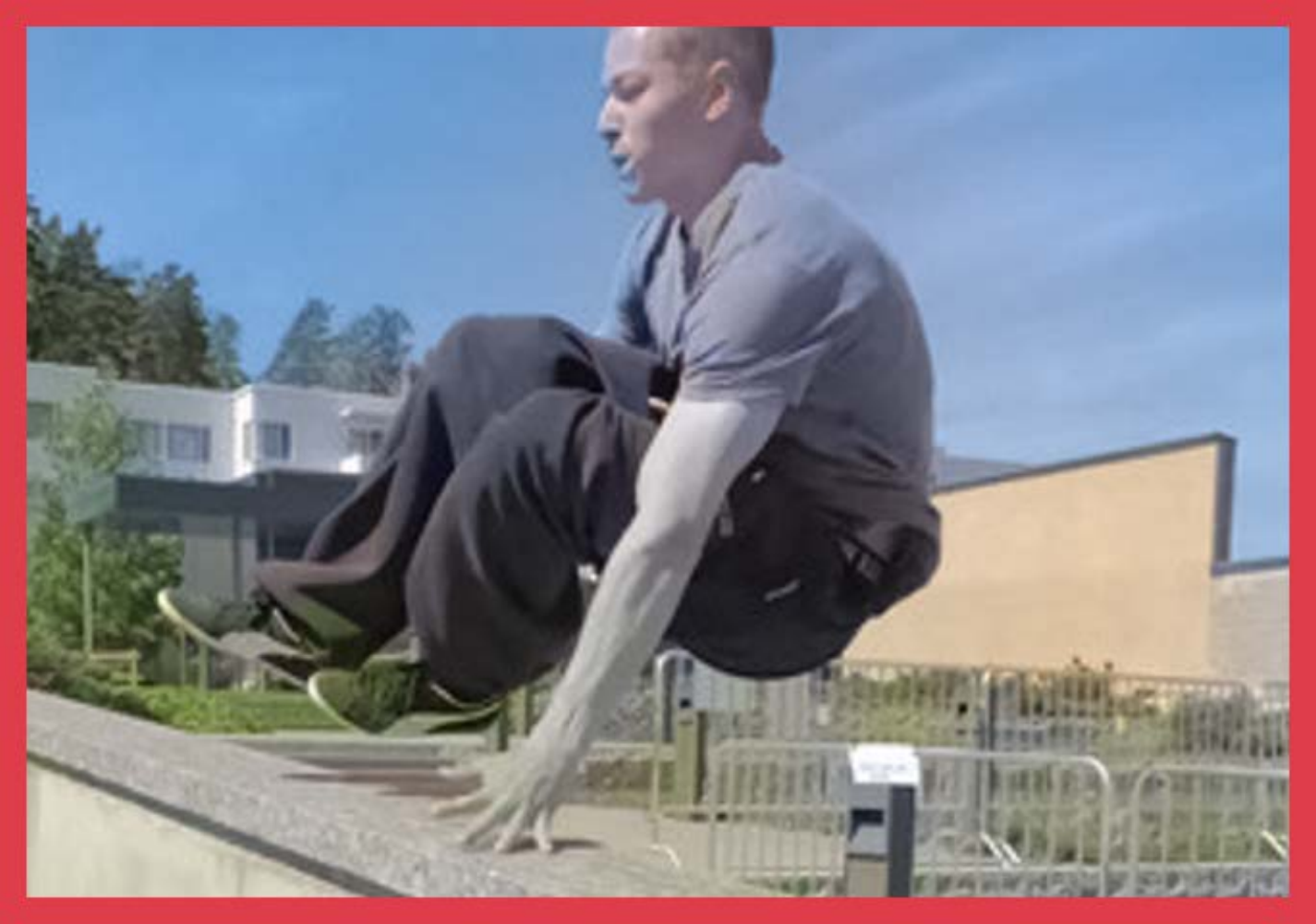} & \hspace{0.0mm}
\includegraphics[width=0.142\linewidth, height = 0.132\linewidth]{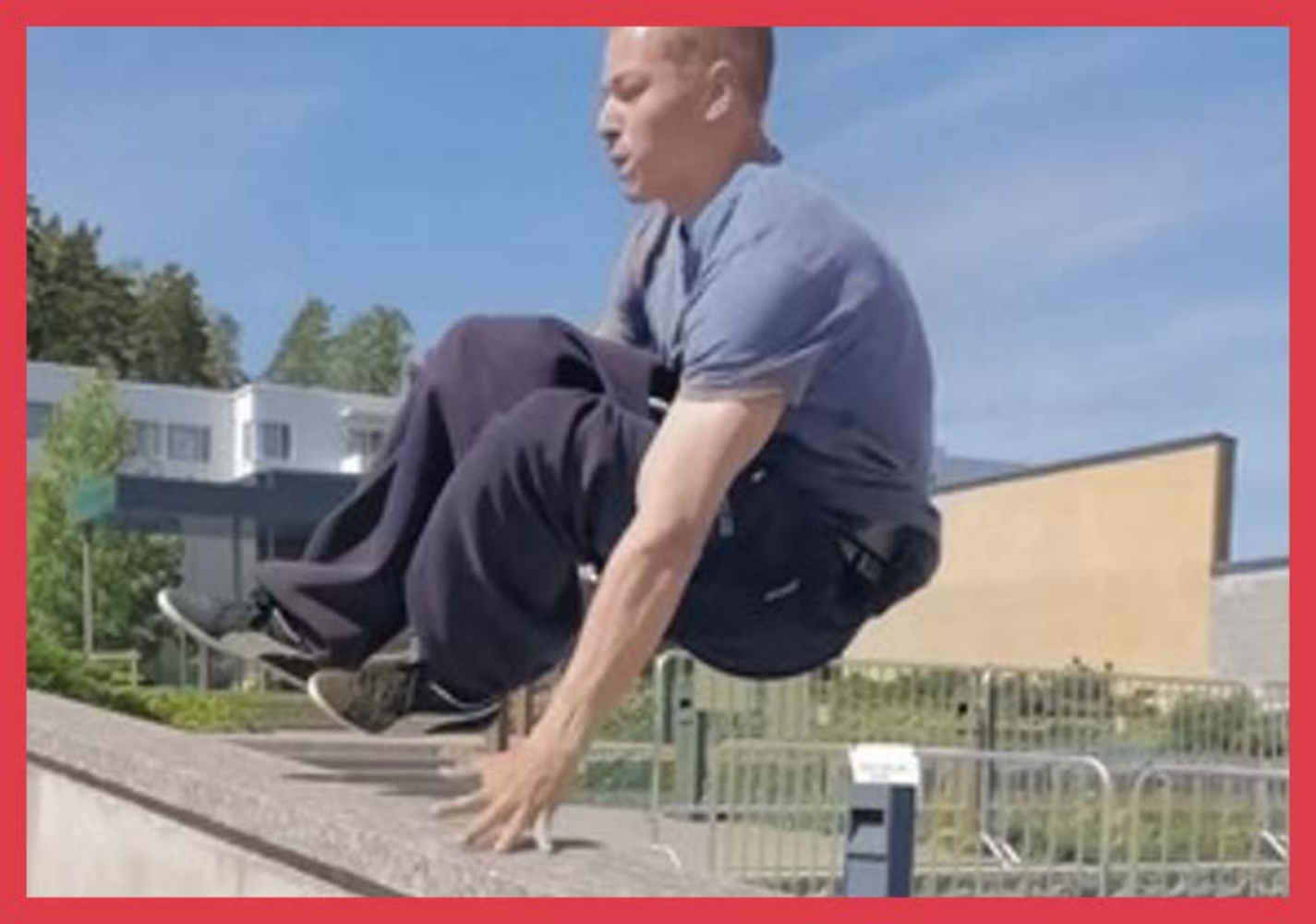} & \hspace{0.0mm}
  \includegraphics[width=0.142\linewidth, height = 0.132\linewidth]{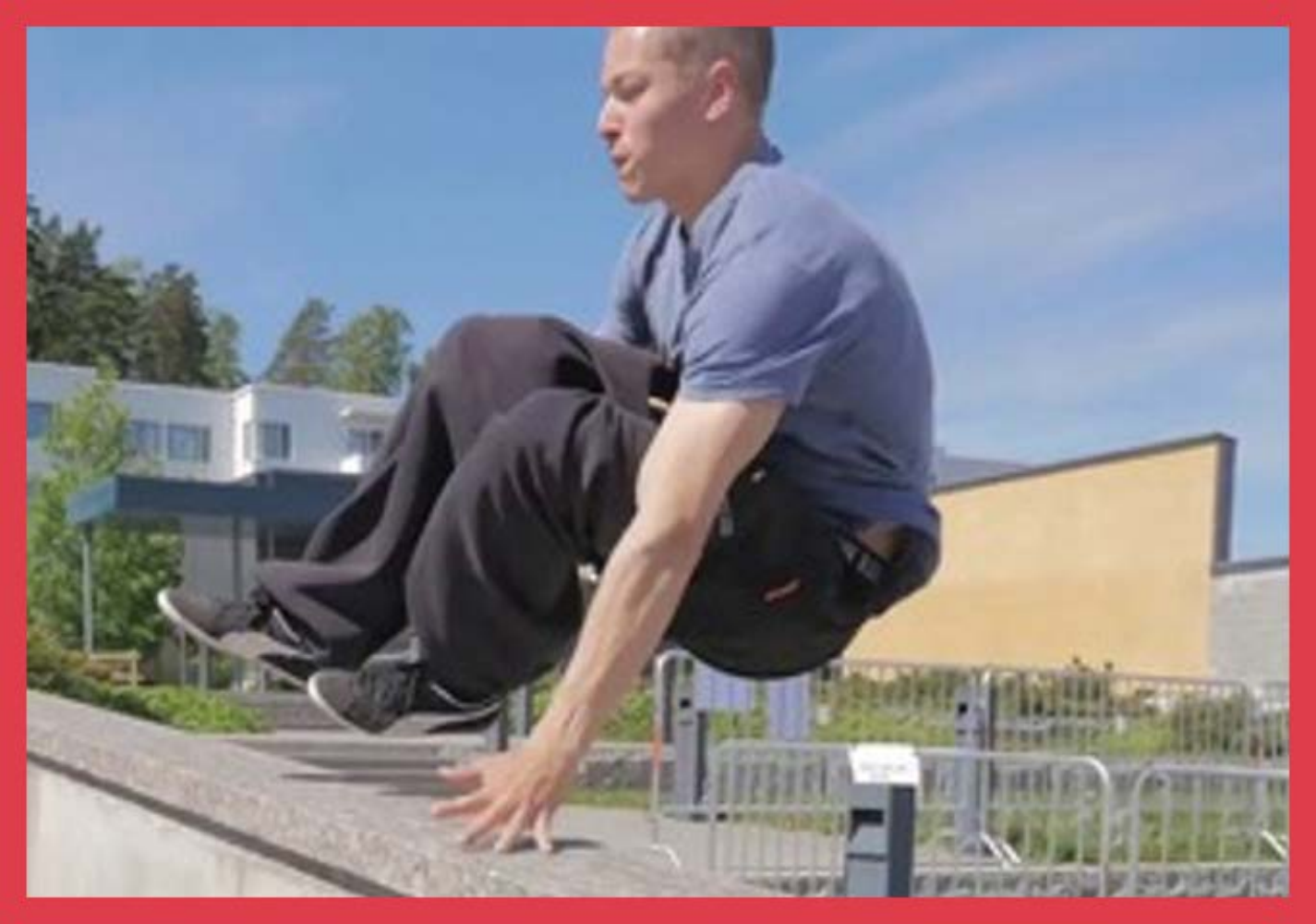} \\

  \multicolumn{3}{c}{\hspace{0.0mm} Input frame and exemplar image} &\hspace{0.0mm}  (e) &  \hspace{0.0mm} (f)  &  \hspace{0.0mm}  (g)   & \hspace{0.0mm}(h)
 \end{tabular}
 \end{center}
 \vspace{-6mm}
 \caption{Qualitative comparisons on clip \textit{parkour} from the validation set of DAVIS~\cite{Perazzi_CVPR_2016} dataset. (a)-(g) are the colorization results by DDColor~\cite{kang2022ddcolor}, TCVC~\cite{liu2021temporally}, VCGAN~\cite{vcgan}, DeepRemaster~\cite{IizukaSIGGRAPHASIA2019}, DeepExemplar~\cite{zhang2019deep}, BiSTNet$^\dagger$~\cite{bistnet} and ColorMNet (Ours). (h) Ground truth. The evaluated methods do not generate realistic colorful images in (a)-(f). In contrast, our approach generates a well-colorized image in (g).}
 \label{fig:002}
 \vspace{-8mm}
\end{figure*}
Figure~\ref{fig:002}(a) shows that the image-based method~\cite{kang2022ddcolor} generates results with non-uniform colors on the man's face and cloth.
Automatic video colorization techniques~\cite{liu2021temporally, vcgan} can not generate vivid colors (Figure~\ref{fig:002}(b) and (c)).
Exemplar-based methods~\cite{IizukaSIGGRAPHASIA2019, zhang2019deep, bistnet} can not establish long-range correspondence and thus fail to restore the colors on the man's arms and cloth (Figure~\ref{fig:002}(d)-(f)).
In contrast, the proposed ColorMNet generates a vivid colorized image (Figure~\ref{fig:002}(g)) by modeling both the spatial information from each frame and the temporal information from far-apart frames.

\begin{table}
\vspace{-5mm}
    \begin{minipage}{0.48\textwidth}
    \vspace{+4mm}
    \setlength\tabcolsep{1.0pt}
	\centering
	\tiny
	\begin{tabularx}{1.0\textwidth}{ccccc}
        \includegraphics[width=0.187\textwidth, height = 0.15\textwidth] {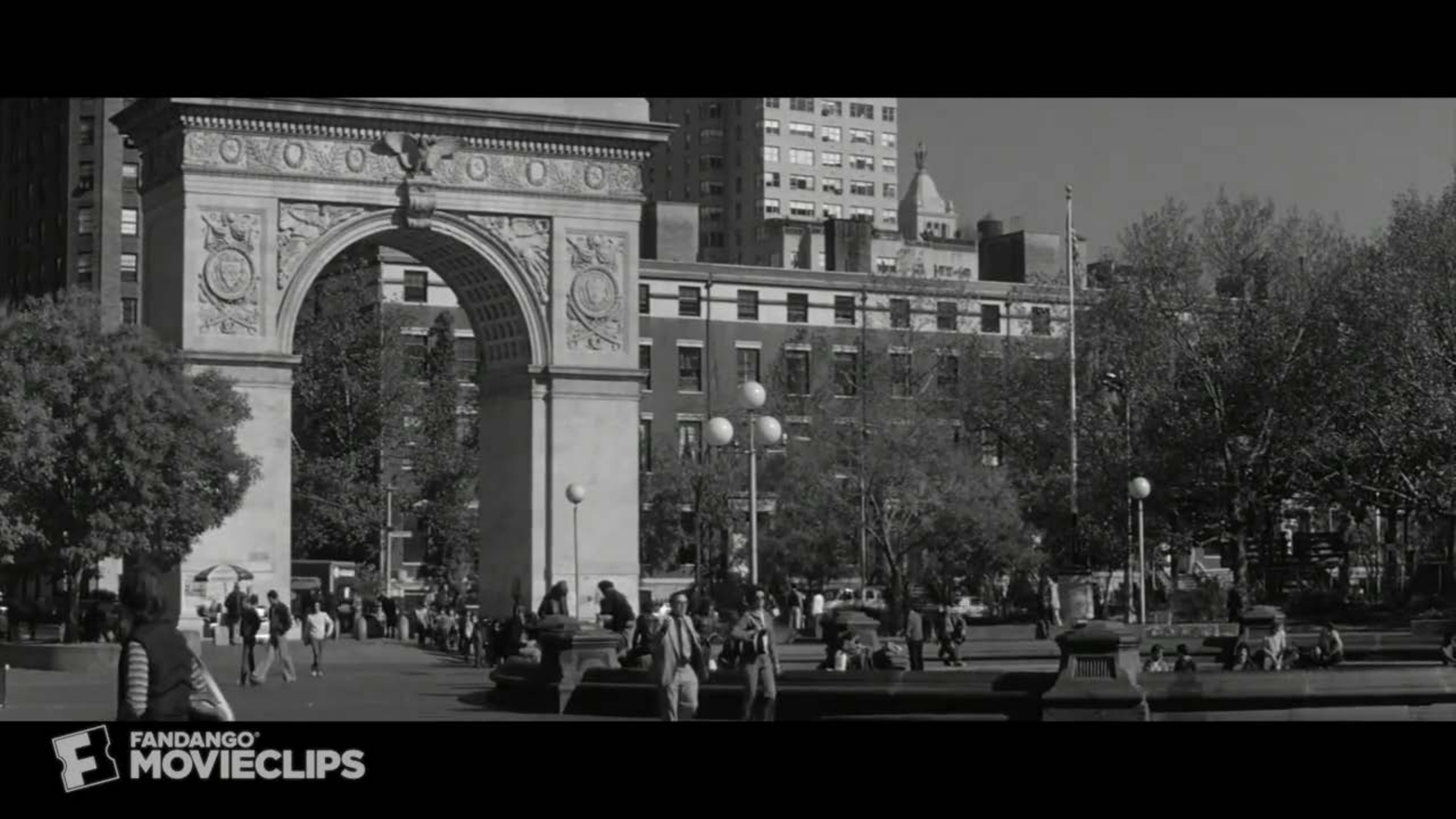} &
        \includegraphics[width=0.187\textwidth, height = 0.15\textwidth] {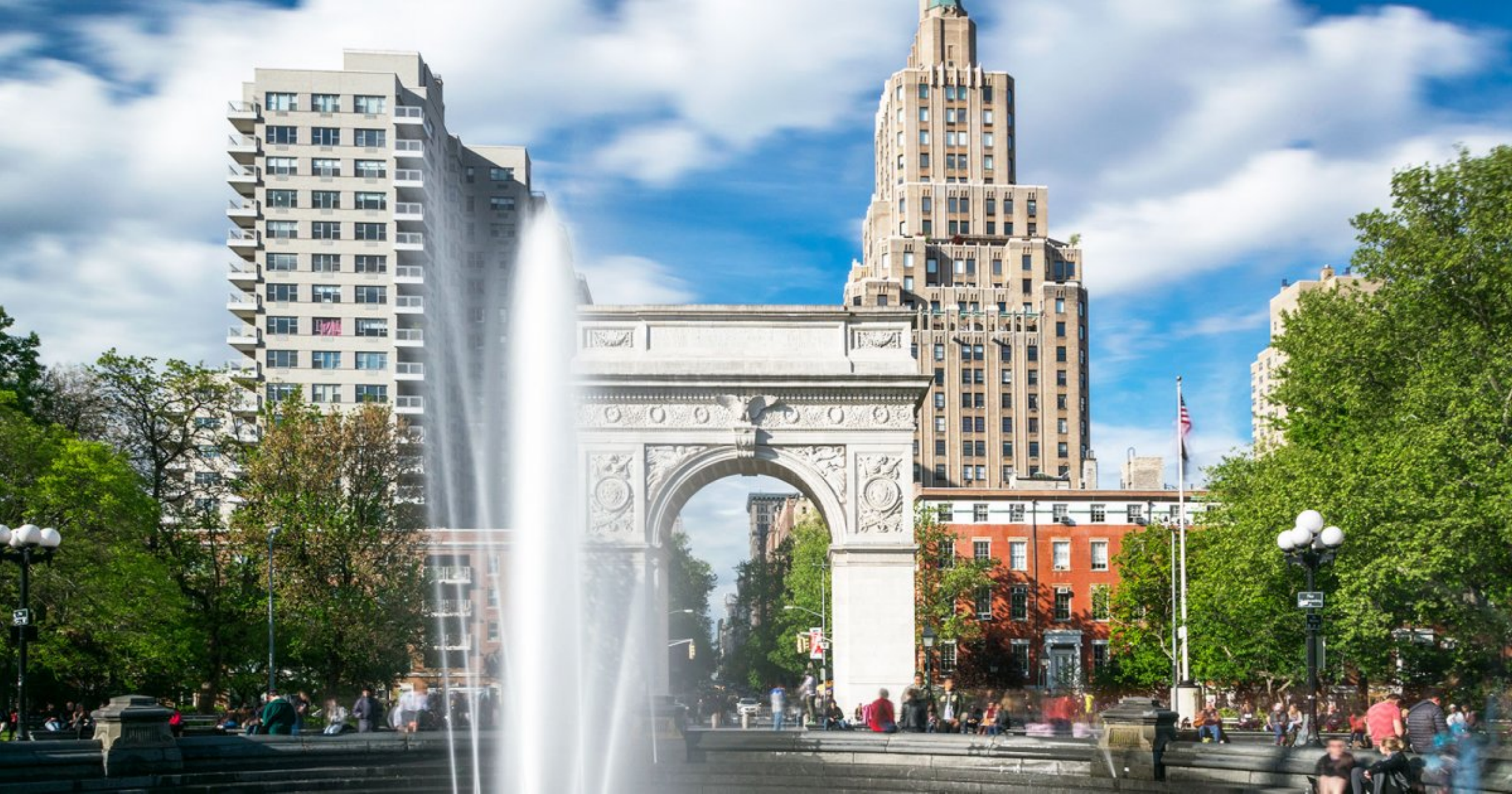} &
        \includegraphics[width=0.187\textwidth, height = 0.15\textwidth] {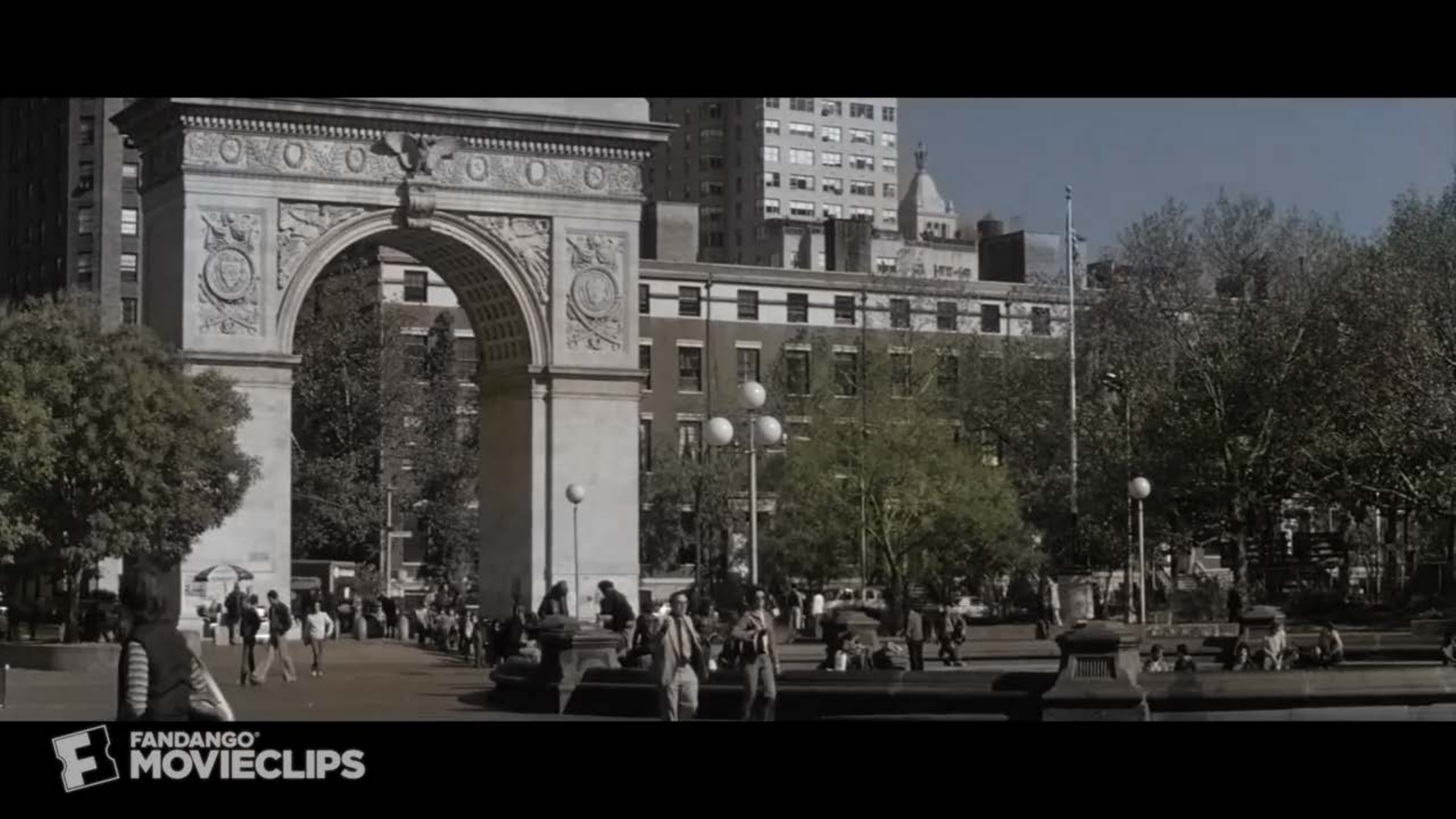} &
        \includegraphics[width=0.187\textwidth, height = 0.15\textwidth]{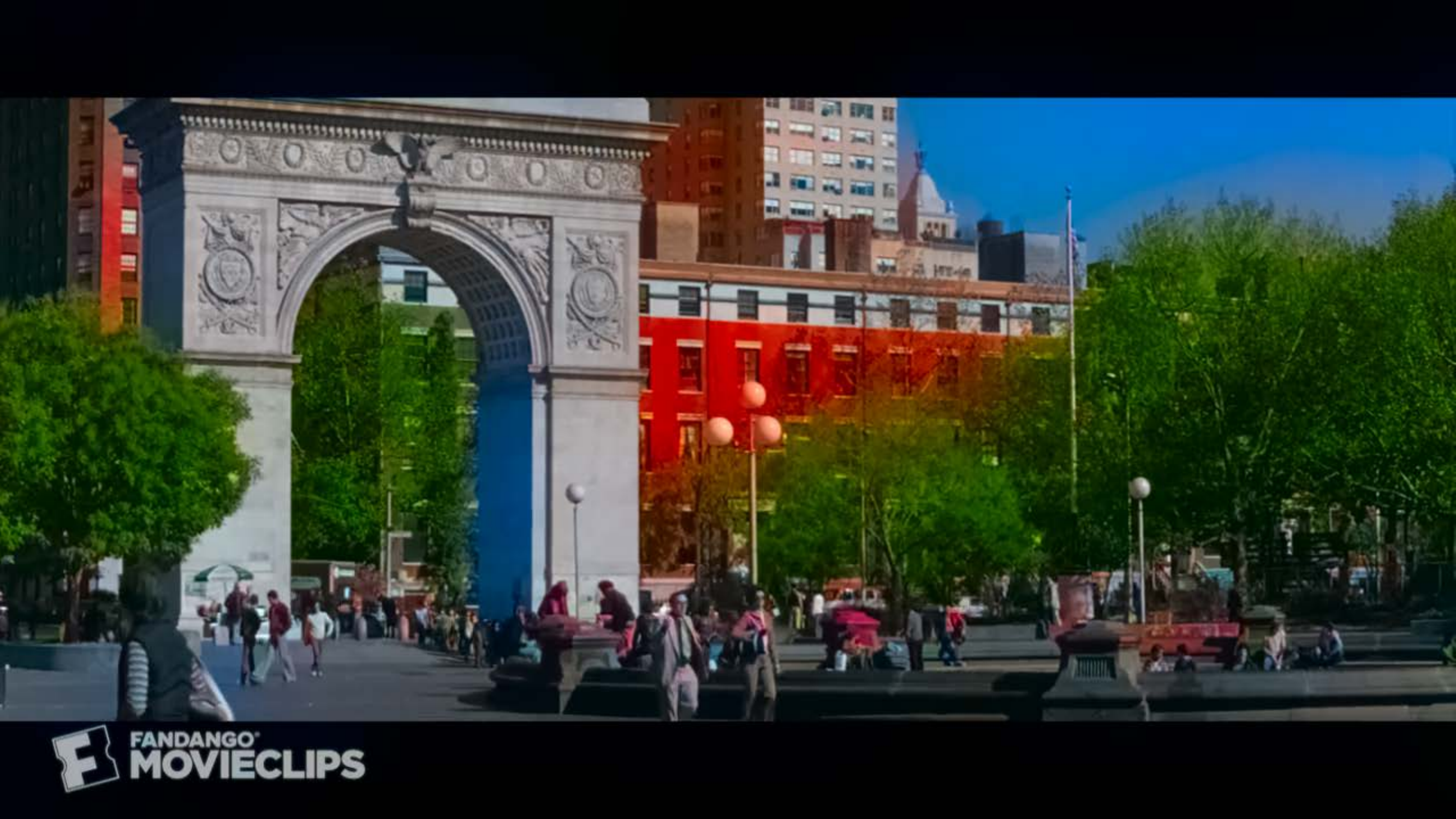}  &
        \includegraphics[width=0.187\textwidth, height = 0.15\textwidth]{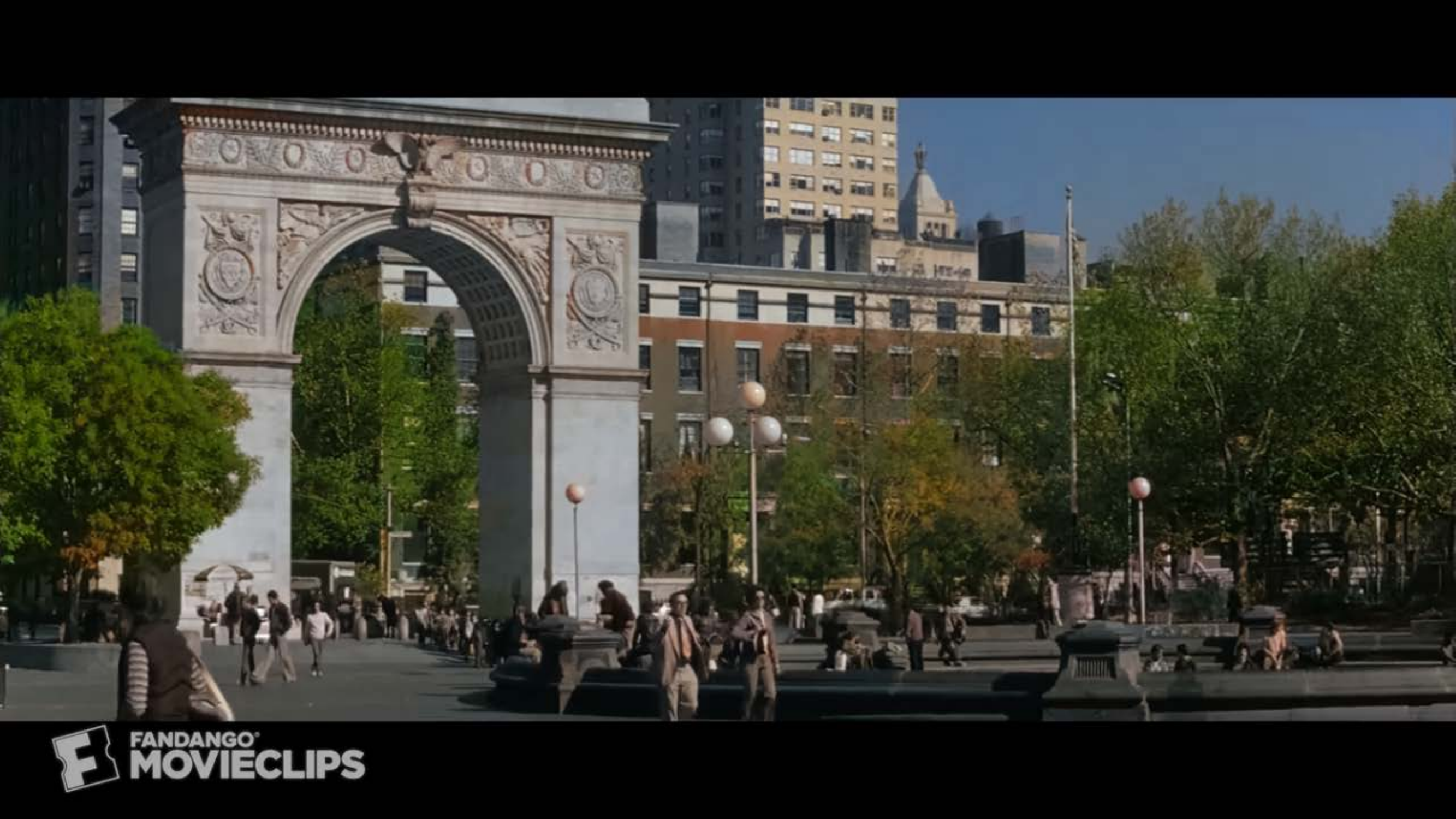}
        \\
        \includegraphics[width=0.187\textwidth, height = 0.15\textwidth] {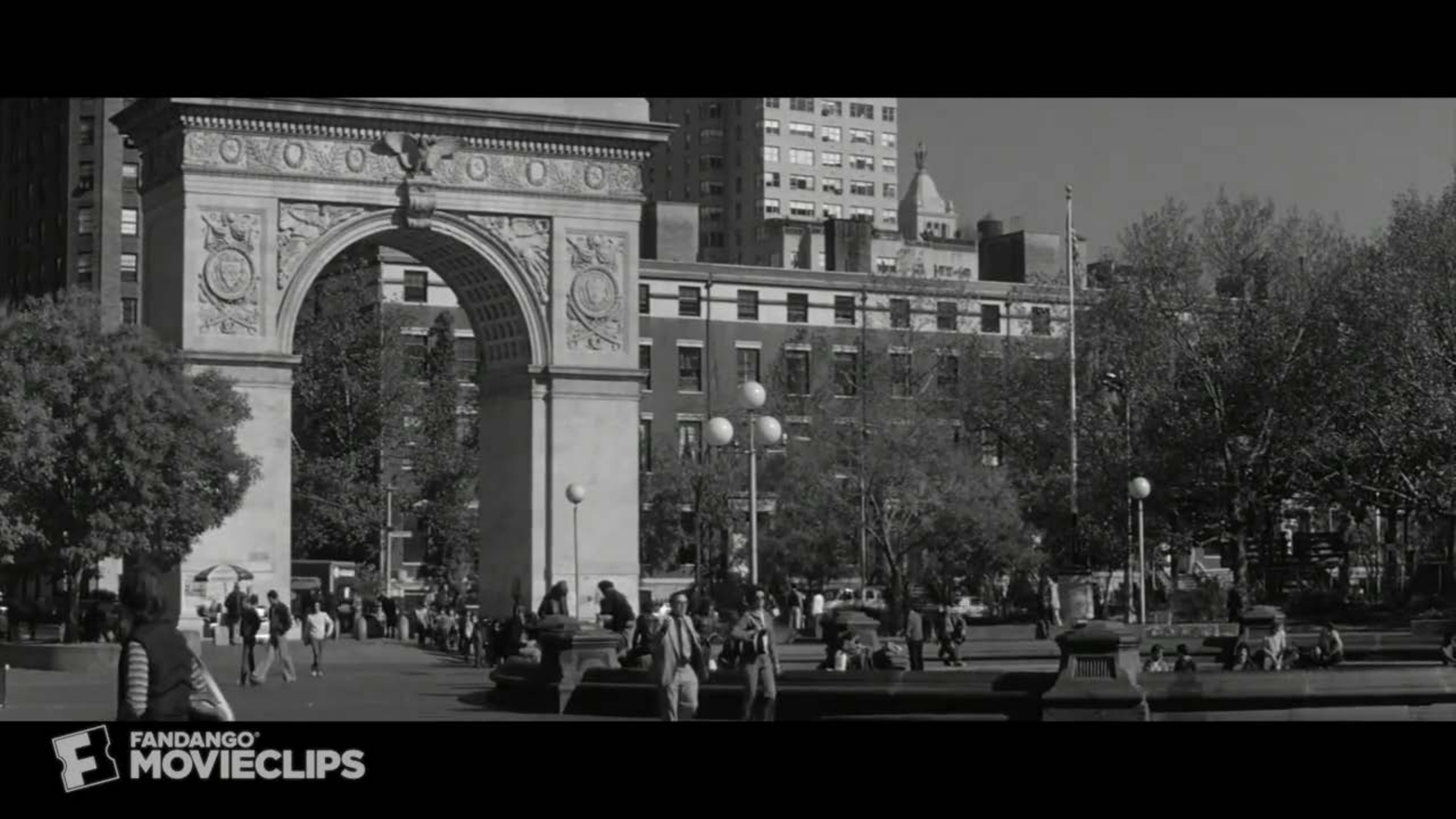} &
        \includegraphics[width=0.187\textwidth, height = 0.15\textwidth] {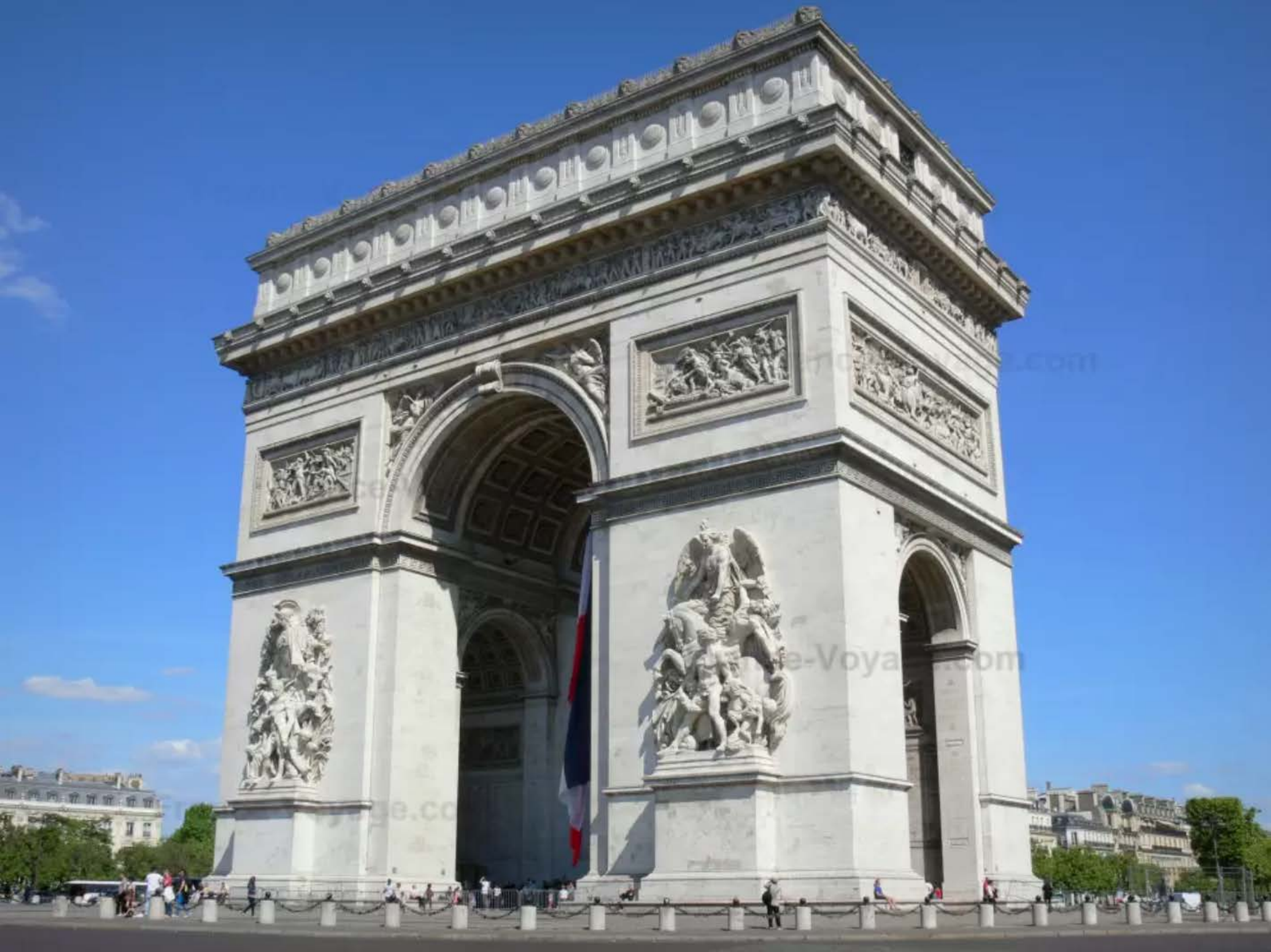} &
        \includegraphics[width=0.187\textwidth, height = 0.15\textwidth] {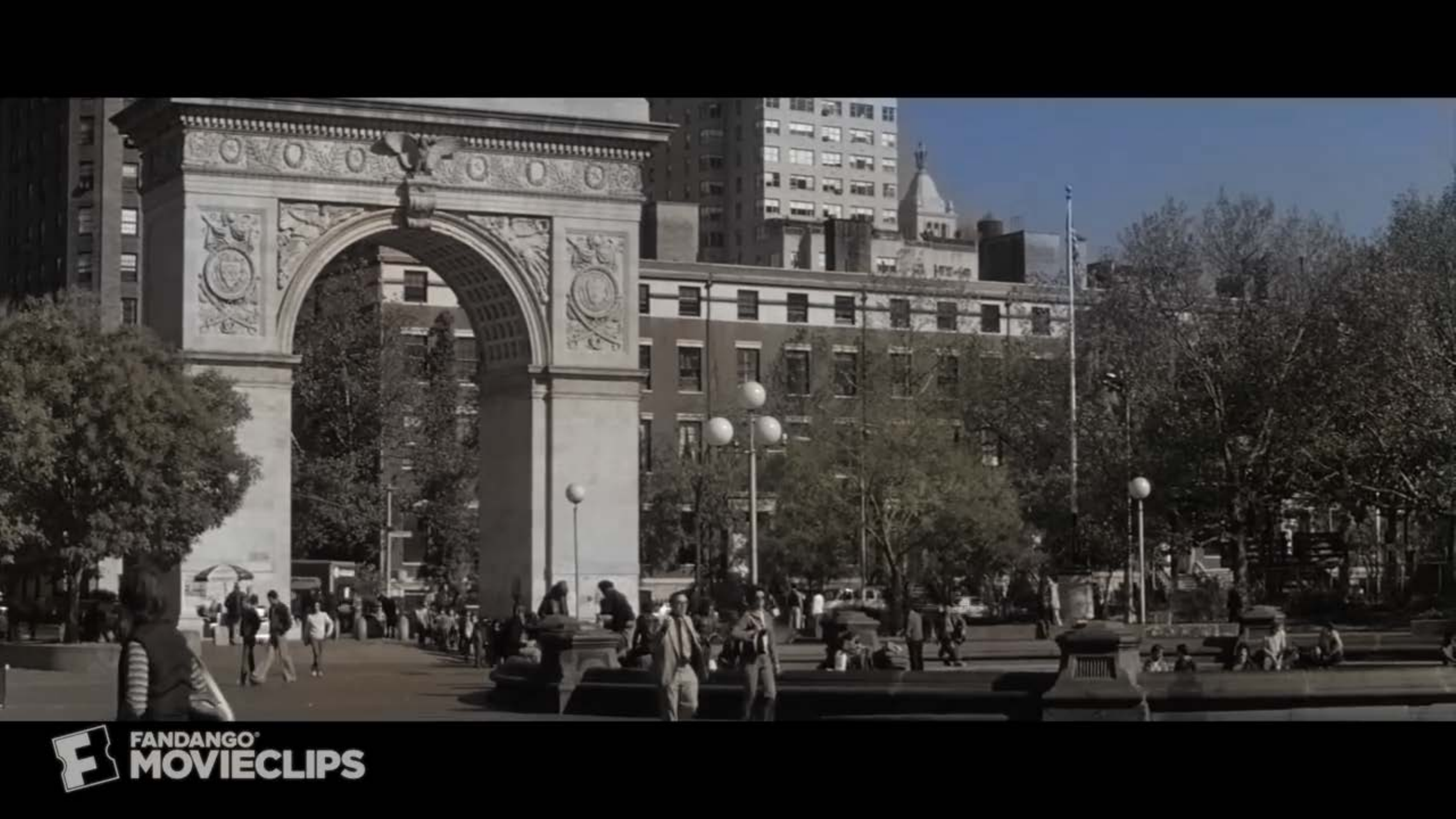} &
        \includegraphics[width=0.187\textwidth, height = 0.15\textwidth]{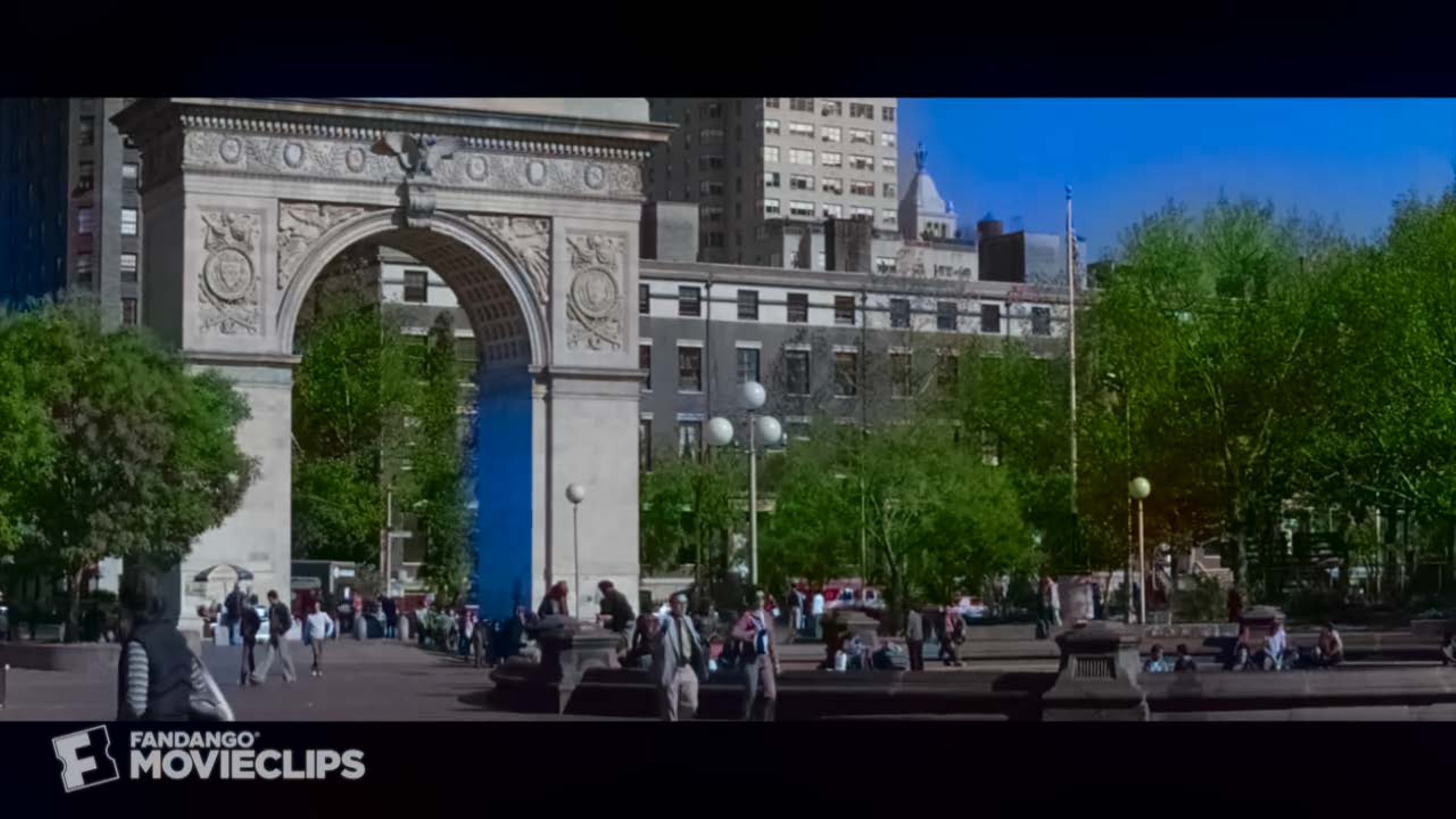}  &
        \includegraphics[width=0.187\textwidth, height = 0.15\textwidth]{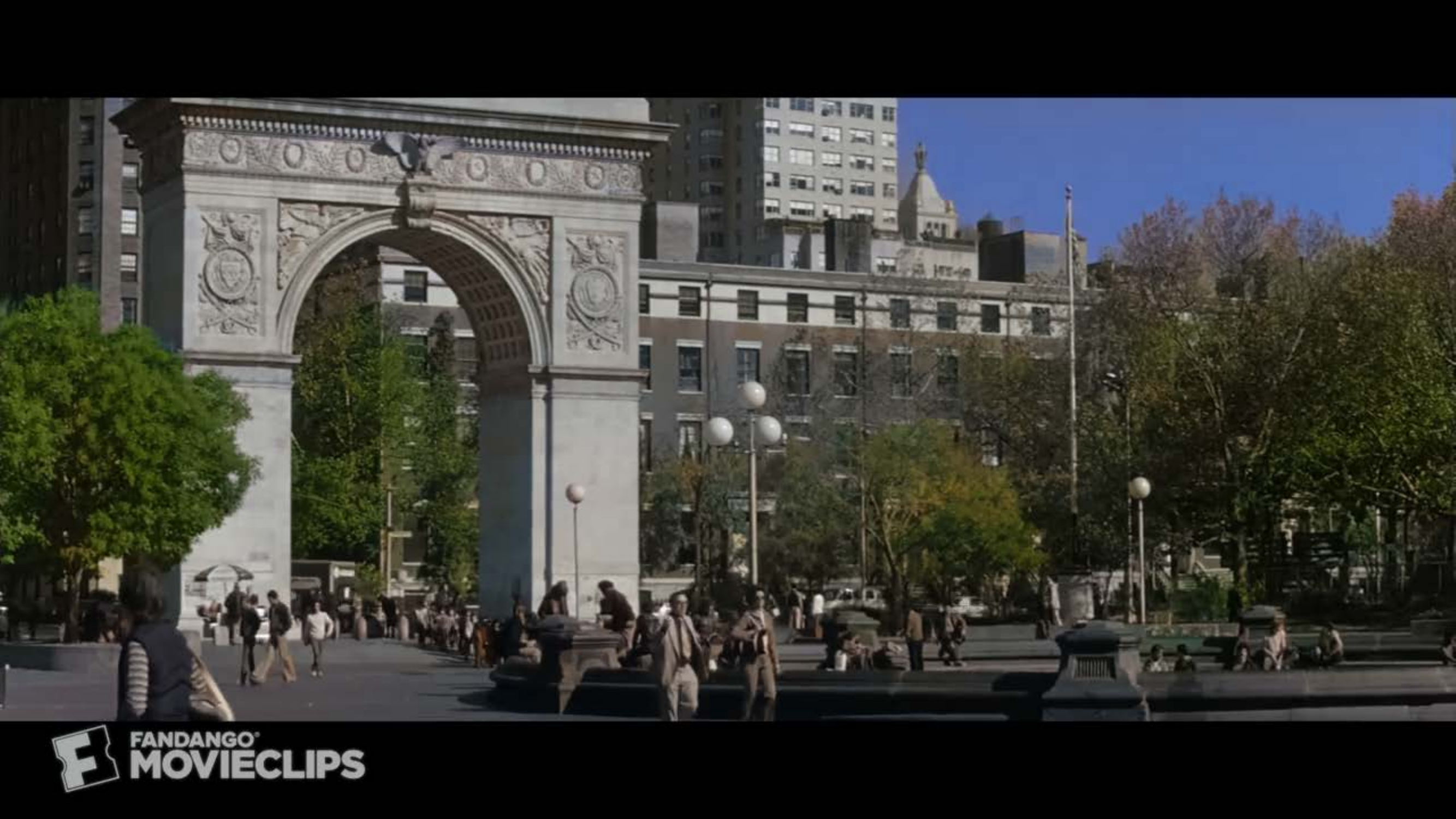}
        \\
        \makebox[0.187\textwidth]{\tiny (a)} &
        \makebox[0.187\textwidth]{\tiny (b)} &
        \makebox[0.187\textwidth]{\tiny (c)} &
        \makebox[0.187\textwidth]{\tiny (d)} &
        \makebox[0.187\textwidth]{\tiny (e)}
         %& \vspace{-0.7em}
	% \vspace{-0.8em}
  	\end{tabularx}
	\captionof{figure}{{Qualitative colorization comparisons on real-world video \textit{Manhattan (1979)}. (a) Input frame. (b) Exemplar images obtained by Google Image Search. (c)-(e) are the colorization results by DeepRemaster~\cite{IizukaSIGGRAPHASIA2019}, DeepExemplar~\cite{zhang2019deep} and ColorMNet (Ours), respectively. The methods~\cite{zhang2019deep,IizukaSIGGRAPHASIA2019} do not colorize the wall of the building, the trees, and the sky well in (c) and (d). Our ColorMNet generates error-free and realistic colors in (e).}}
	\label{fig:real_world}
     \end{minipage}
     \hspace{+2mm}
    \begin{minipage}{0.48\textwidth}
        \scriptsize
        \centering
	\caption{Quantitative evaluations of the video colorization methods with better accuracy performance on the DAVIS~\cite{Perazzi_CVPR_2016} dataset in terms of maximum GPU memory consumption, average running time and temporal consistency index CDC.}
	\vspace{-3mm}
 \resizebox{1.\textwidth}{!}{
	\centering
	\small
        \setlength{\tabcolsep}{2.8mm}
	\begin{tabular}{lccc}
            \noalign{\hrule height 0.3mm}
		Methods & Memory (G)   &Running time (/s) & CDC${\downarrow}$\\
		\hline
  		DeepExemplar~\cite{zhang2019deep}     & 19.0  & 0.80  & 0.003876 \\
            DeepRemaster~\cite{IizukaSIGGRAPHASIA2019}     & 16.6  & 0.61 & 0.004285  \\
            BiSTNet~\cite{bistnet} & 34.9  & 1.62  & 0.003870 \\
            ColorMNet (Ours)     & \textbf{1.9}  & \textbf{0.07} & \textbf{0.003763}  \\
		\noalign{\hrule height 0.3mm}
	\end{tabular}
 }
	\label{tab:runtime}
    \vspace{2pt}
	\small
	\centering
 \caption{{Effectiveness of the proposed PVGFE, MFP, and LA modules in our ColorMNet. Evaluated on the DAVIS~\cite{Perazzi_CVPR_2016}.}}
 \vspace{-3mm}
	\begin{adjustbox}{max width=\linewidth}
		\begin{tabular}{l|ccc|ccc|c|cc}
			\noalign{\hrule height 0.3mm}
   		Components & \multicolumn{3}{c|}{Feature extractor}   &\multicolumn{3}{c|}{Feature propagation} & Locality & \multicolumn{2}{c}{Metrics}  \\ \hline
		Methods & ResNet50& DINOv2  &PVGFE   & Stacking & Recurrent & MFP & LA & PSNR${\uparrow}$ & SSIM${\uparrow}$  \\ \hline
           ColorMNet$_{\text{w/ ResNet50}}$ & \Checkmark & & & & & \Checkmark   & \Checkmark & 35.01  & 0.962
           \\
            ColorMNet$_{\text{w/ DINOv2}}$  & & \Checkmark &	 & & &\Checkmark & \Checkmark & 35.38 & 0.963
            \\
            ColorMNet$_{\text{w/ Concatenation}}$  & \Checkmark & \Checkmark & & & &\Checkmark & \Checkmark & 35.26  & 0.965
            \\
            ColorMNet$_{\text{w/ Stacking}}$ & \Checkmark & \Checkmark & \Checkmark & \Checkmark &&  &\Checkmark    & {33.94}  & {0.961}
            \\
            ColorMNet$_{\text{w/ Recurrent}}$  & \Checkmark & \Checkmark & \Checkmark & &\Checkmark &  &\Checkmark    & {35.26} & {0.966}
            \\
            ColorMNet$_{\text{w/o LA}}$ & \Checkmark & \Checkmark & \Checkmark & & &\Checkmark &  & 35.44 & 0.967
            \\ \hline
            ColorMNet (Ours) & \Checkmark & \Checkmark & \Checkmark &  & &\Checkmark   & \Checkmark & \textbf{35.77} & \textbf{0.970}
            \\

			\noalign{\hrule height 0.3mm}
		\end{tabular}
	\end{adjustbox}
	%\vspace{-0.3em}
	\label{tab:pvgfe}
    \end{minipage}
    \vspace{-10mm}
\end{table}

\noindent \textbf{Evaluations on real-world videos.}
We further evaluate the proposed method on a real-world grayscale video, \textit{Manhattan (1979)}.
We obtain the exemplars (Figure~\ref{fig:real_world}(b)) by searching the internet to find the most visually similar images to the input video frames.
Figure~\ref{fig:real_world}(c) and (d) show that state-of-the-art methods~\cite{IizukaSIGGRAPHASIA2019, zhang2019deep} do not colorize the objects (\eg, the wall of the building, the trees and the sky) well.
In contrast, our method generates better-colorized frames, where the colors look natural and realistic (Figure~\ref{fig:real_world}(e)).
In addition, our method demonstrates its robustness by consistently generating similar results even when provided with exemplar images that possess diverse colors and contents.

\noindent \textbf{Efficiency evaluation.}
Given that practical applications of video colorization often involve processing longer videos, where maximum GPU memory usage and inference speed are critical metrics, we further evaluate our method against three representative state-of-the-art exemplar-based video colorization approaches~\cite{bistnet, zhang2019deep,IizukaSIGGRAPHASIA2019}.
Specifically, we record the maximum GPU memory consumption during the inference on a machine with an NVIDIA RTX A6000 GPU.
The average running time is obtained using 300 test images with a $960 \times 536$ resolution.

Table~\ref{tab:runtime} shows that the maximum GPU consumption of ColorMNet (ours) is only 11.2\% of DeepRemaster~\cite{IizukaSIGGRAPHASIA2019}, 10.0\% of DeepExemplar~\cite{zhang2019deep} and 5.4\% of BiSTNet~\cite{bistnet}; the running time is at least 8$\times$ faster than the evaluated methods.

\noindent \textbf{Temporal consistency evaluations.}
To examine whether the colorized videos generated by our method have a better temporal consistency property, we use the color distribution consistency index (CDC)~\cite{liu2021temporally} as the metric.
Table~\ref{tab:runtime} shows that our method has a lower CDC value when compared to exemplar-based methods~\cite{bistnet, zhang2019deep, IizukaSIGGRAPHASIA2019} on the DAVIS~\cite{Perazzi_CVPR_2016} validation set, which indicates that our method is capable of generating videos with improved temporal consistency by exploring better temporal information.

\vspace{-3mm}
\section{Analysis and Discussion}
\vspace{-2mm}
\label{sec:analysis}
To better understand how our method solves video colorization and demonstrate the effectiveness of its main components, we conduct a deeper analysis of the proposed approach. For the ablation studies in this section, we train our method and all alternative baselines on the training set of the DAVIS~\cite{Perazzi_CVPR_2016} dataset and the Videvo~\cite{Lai2018videvo} dataset with 160,000 iterations for fair comparisons.

\begin{figure}[!t]
% \hspace{-6mm}
\begin{minipage}[!t]{1.0\textwidth}
% \hspace{-0mm}
% \begin{figure}[!tb]
	\setlength\tabcolsep{1.0pt}
	\centering
	\scriptsize
	\begin{tabularx}{1.0\textwidth}{ccccc}
  \multicolumn{2}{c}{\multirow{2}*[29pt]
  {\includegraphics[width=0.420\linewidth, height=0.230\linewidth]{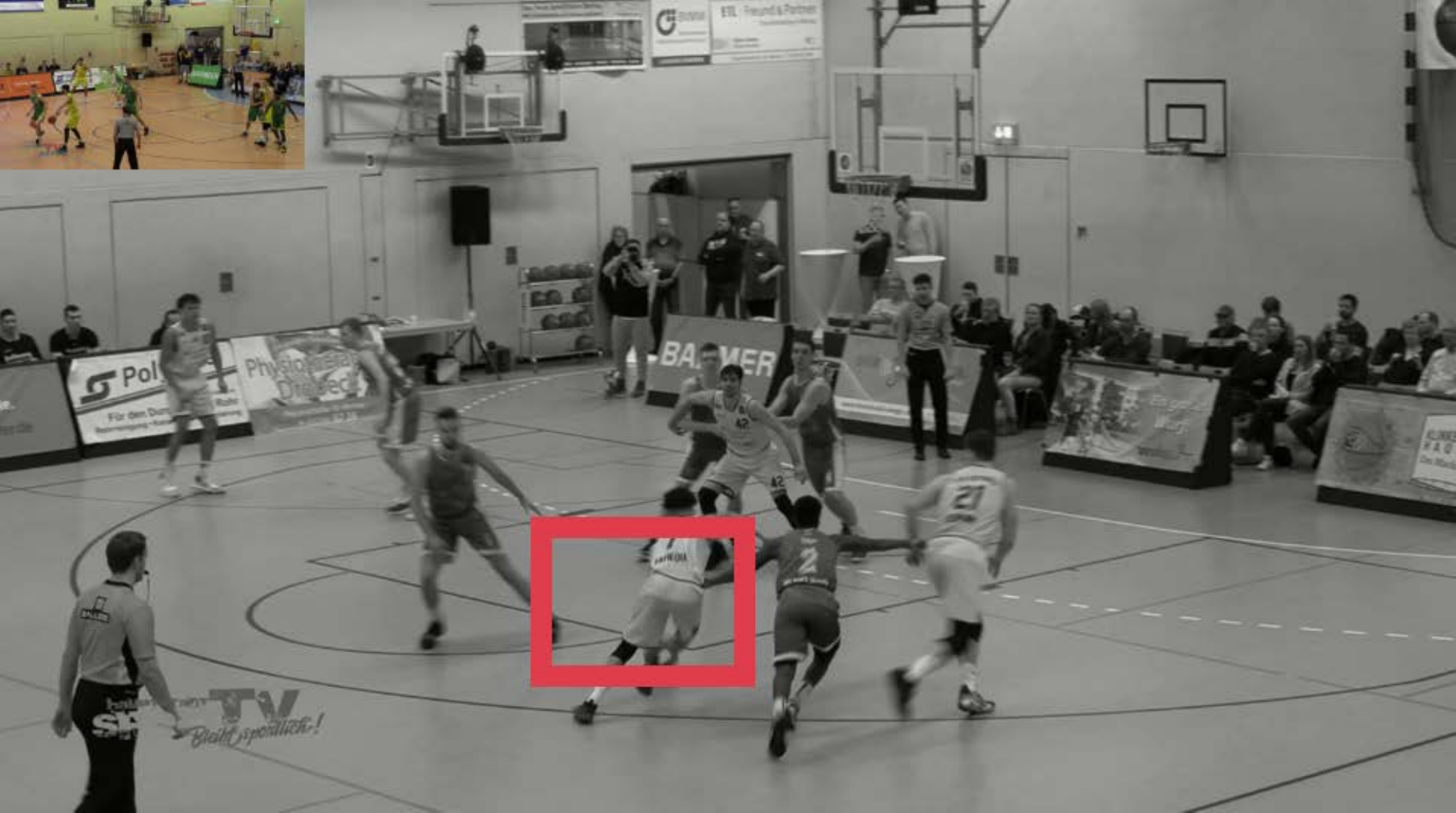}}}&
  \includegraphics[width=0.187\linewidth, height = 0.10\linewidth]{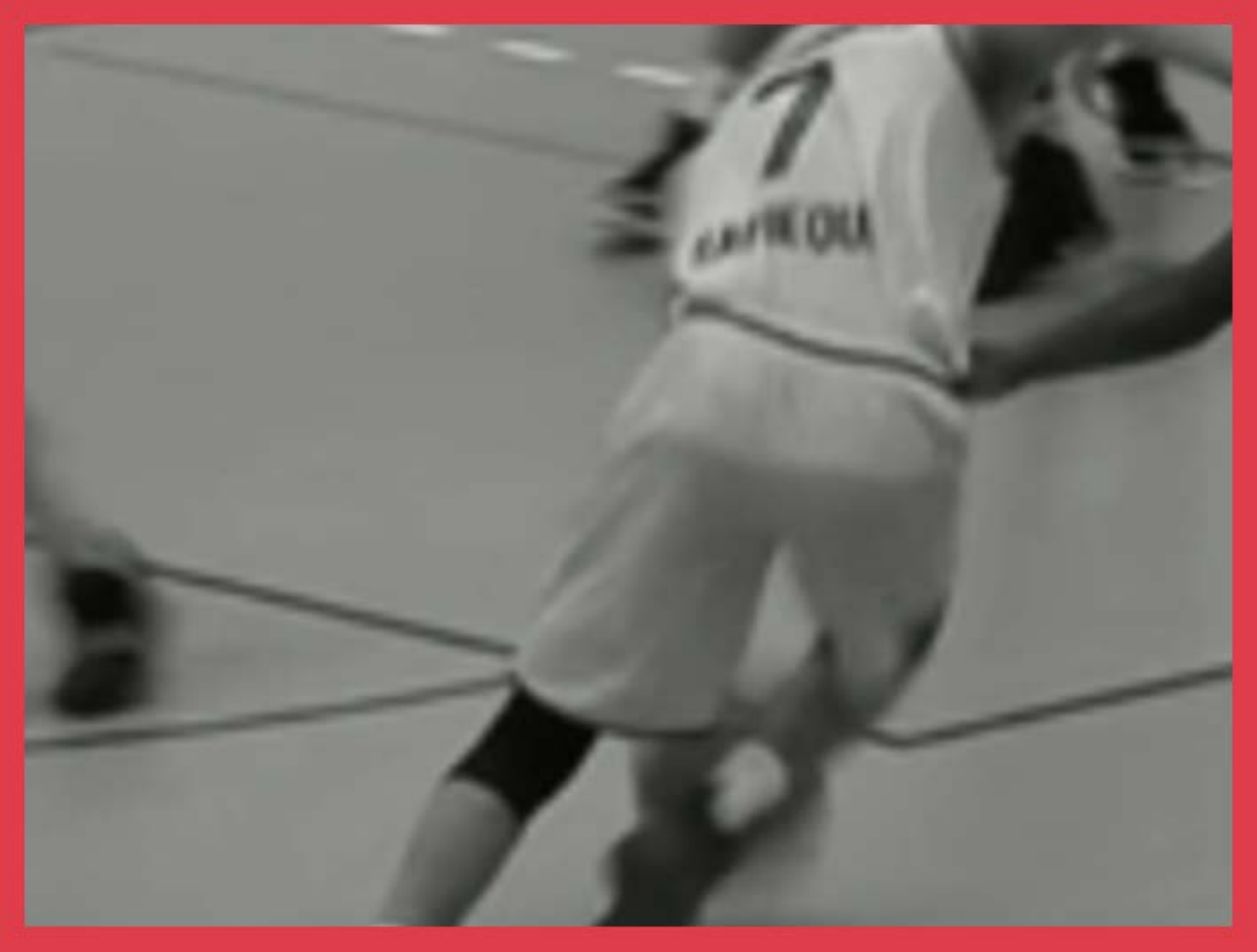} &
    \includegraphics[width=0.187\linewidth, height = 0.10\linewidth]{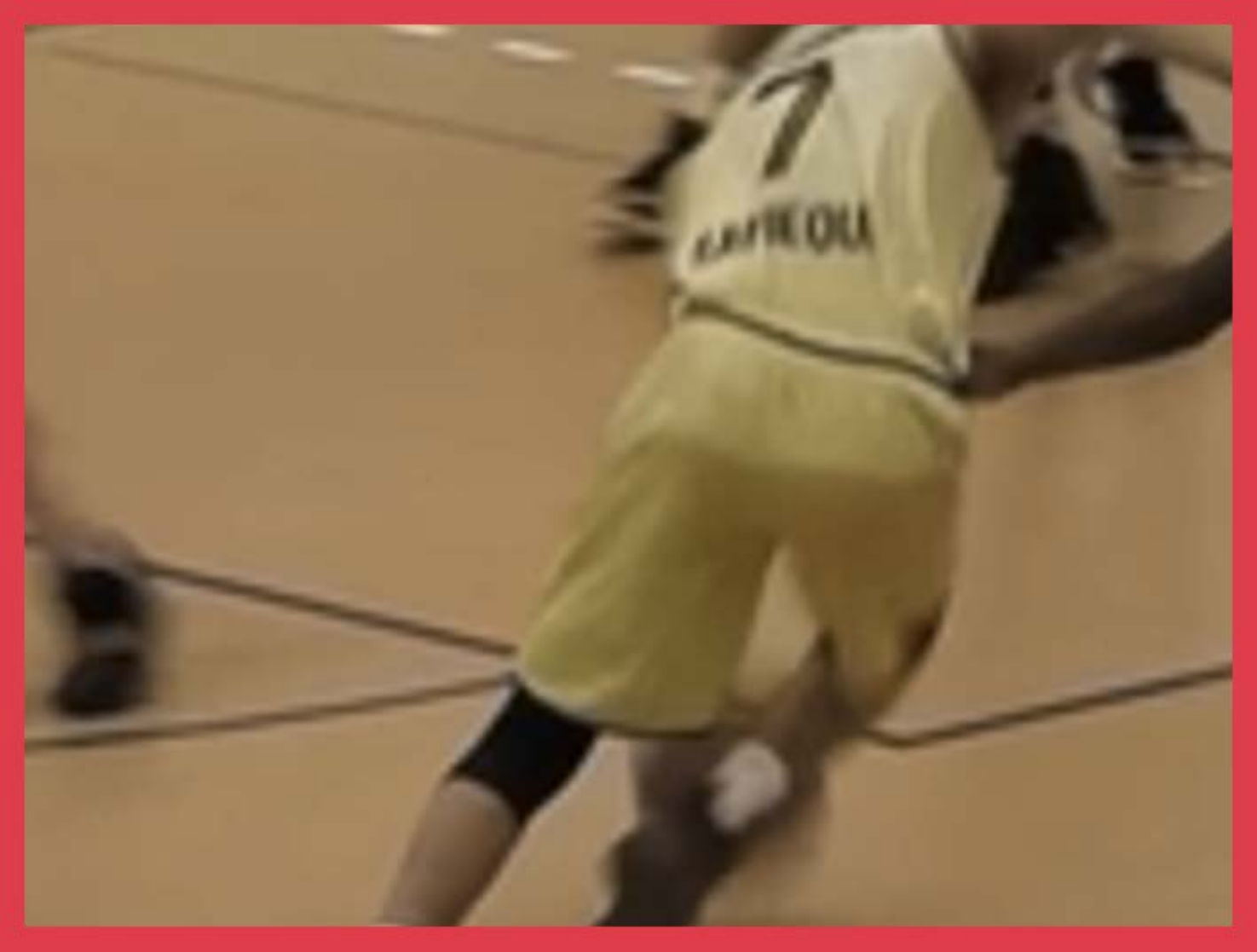} &
  \includegraphics[width=0.187\linewidth, height = 0.10\linewidth]{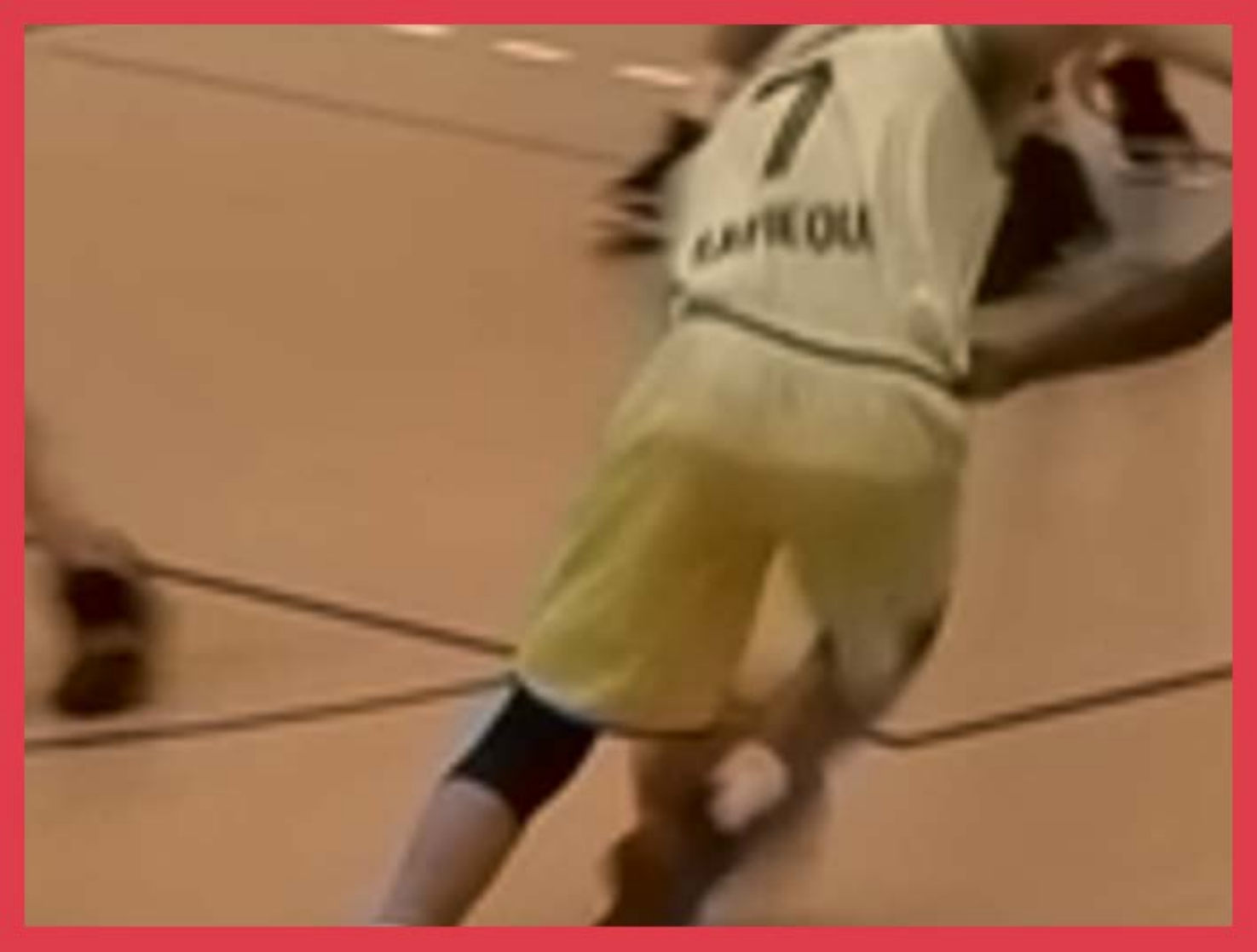}\\
  \multicolumn{2}{c}{~}  &  (a)  &  (b)   &  (c)
   \\
  \multicolumn{2}{c}{~} &
  \includegraphics[width=0.187\linewidth, height = 0.10\linewidth]{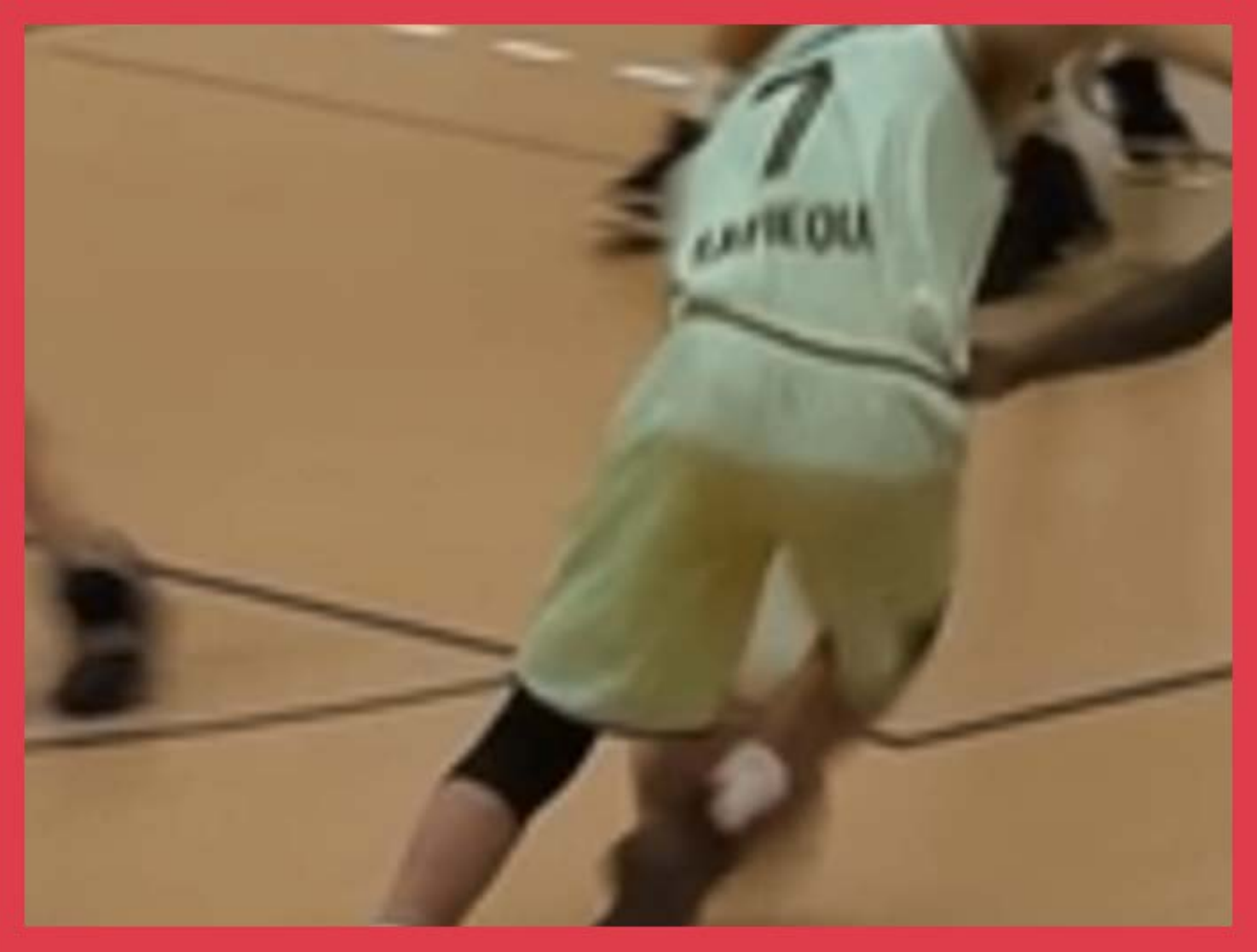} &
  \includegraphics[width=0.187\linewidth, height = 0.10\linewidth]{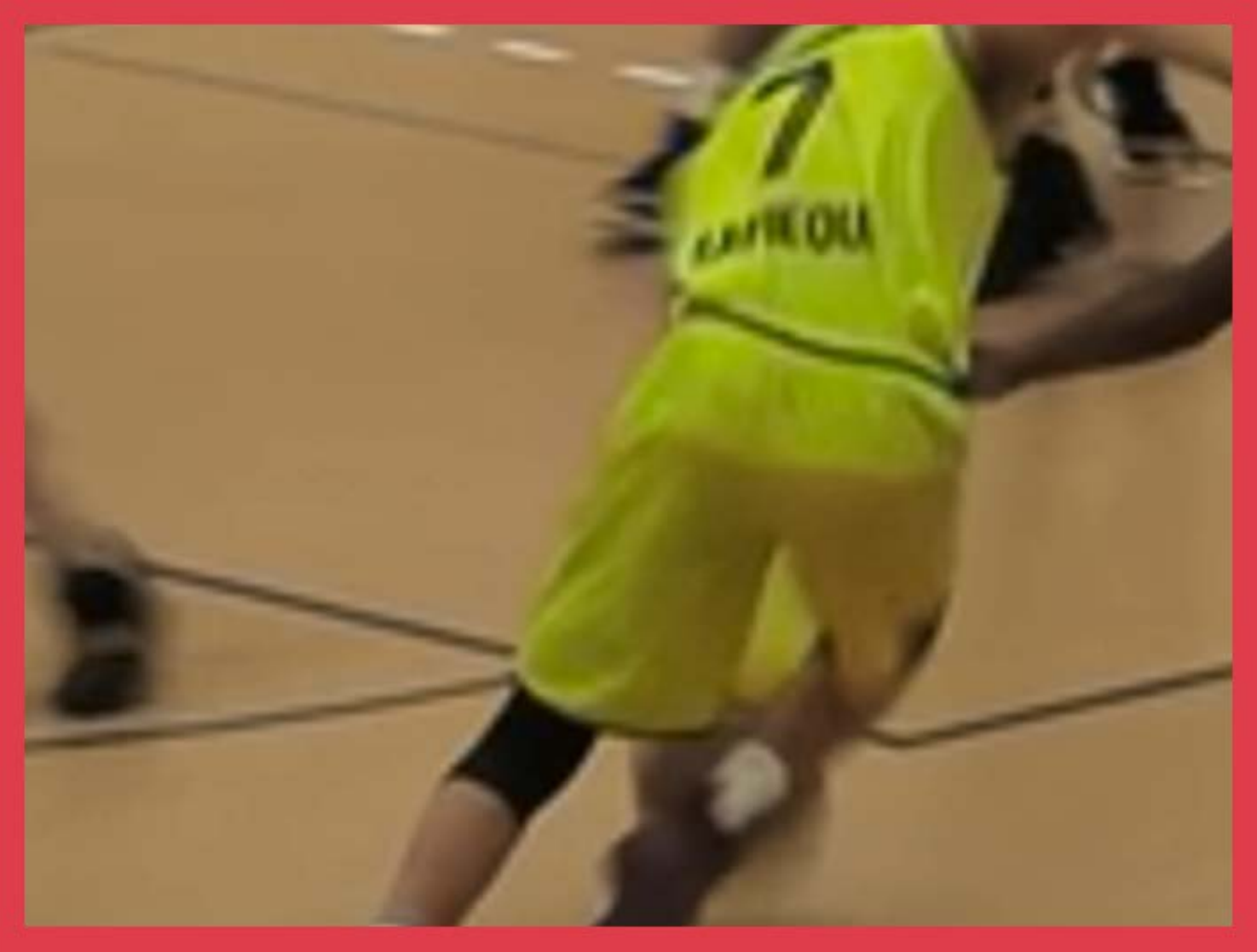} &
  \includegraphics[width=0.187\linewidth, height = 0.10\linewidth]{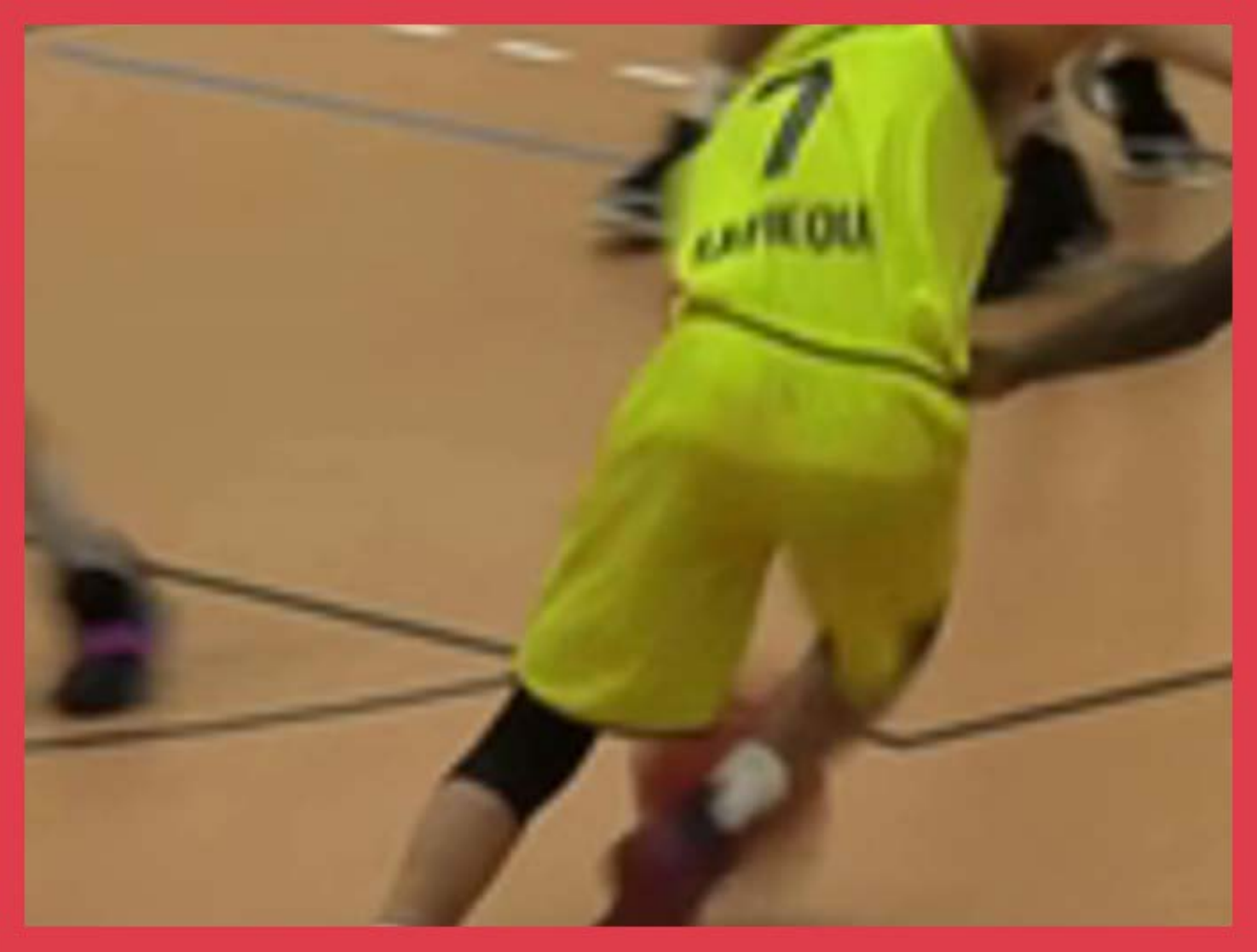} \\

  \multicolumn{2}{c}{Input frame and exemplar image} &   (d)  &    (e)   & (f)  \\
	\end{tabularx}
	\vspace{-3mm}
	\captionof{figure}{{Effectiveness of PVGFE for video colorization. (a) Input patch. (b)-(e) are the colorization results by ColorMNet$_{\text{w/ ResNet50}}$, ColorMNet$_{\text{w/ DINOv2}}$, ColorMNet$_{\text{w/ Concatenation}}$ and ColorMNet (Ours), respectively. (f) Ground truth.} Compared to the baselines, our approach yields a more natural colorized result in (e).}
	\label{fig:pvgfe}

 \vspace{+1mm}

% \begin{figure}
    % \vspace{-3mm}
    \setlength\tabcolsep{1.0pt}
	\centering
	\small
	\begin{tabularx}{1.0\textwidth}{cccccccccc}
         \includegraphics[width=0.094\textwidth, height = 0.08\textwidth] {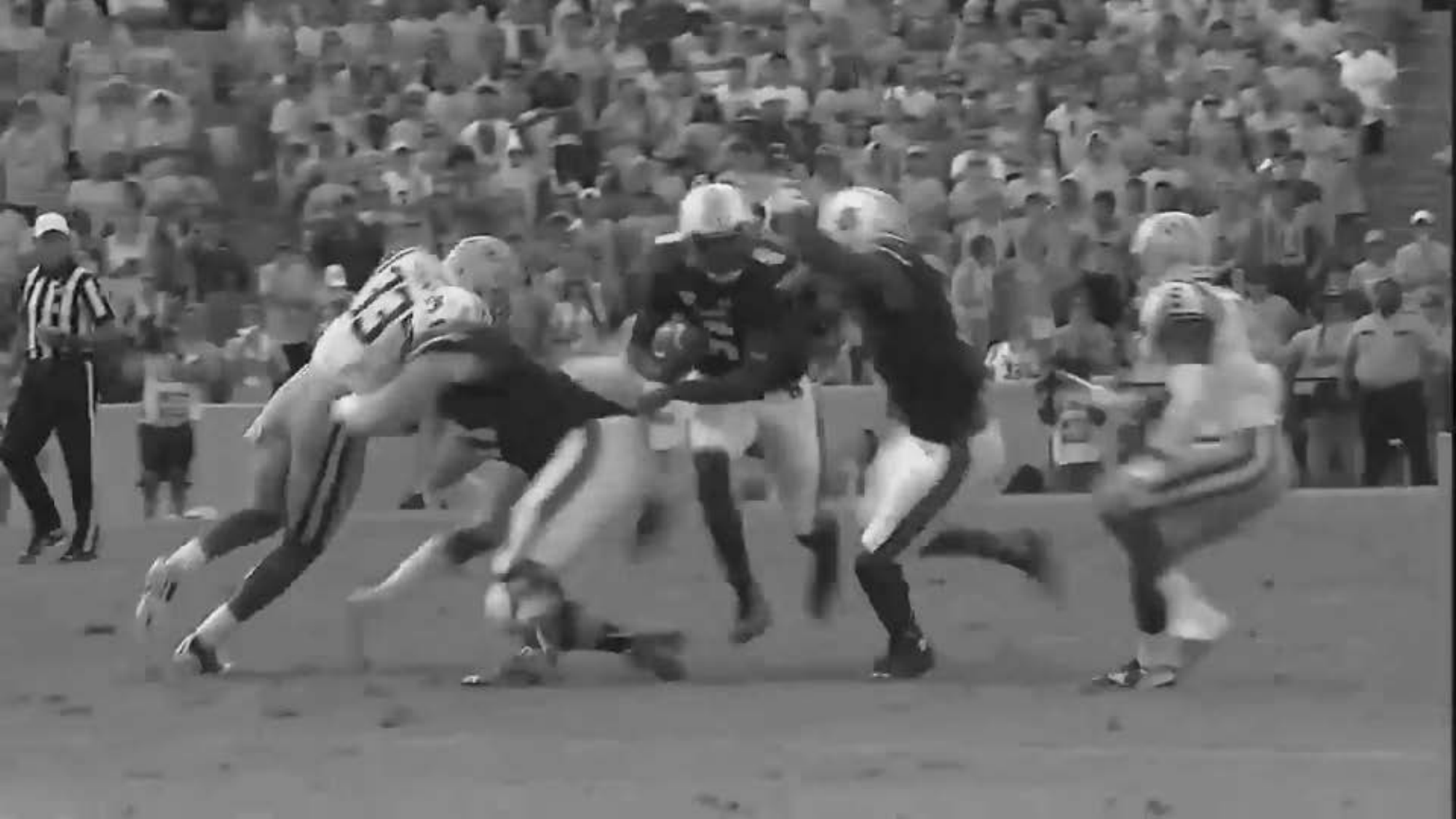} &
        \includegraphics[width=0.094\textwidth, height = 0.08\textwidth] {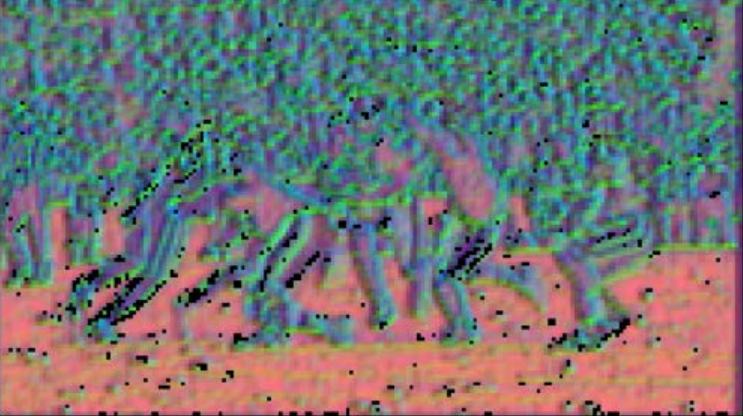} &
        \includegraphics[width=0.094\textwidth, height = 0.08\textwidth] {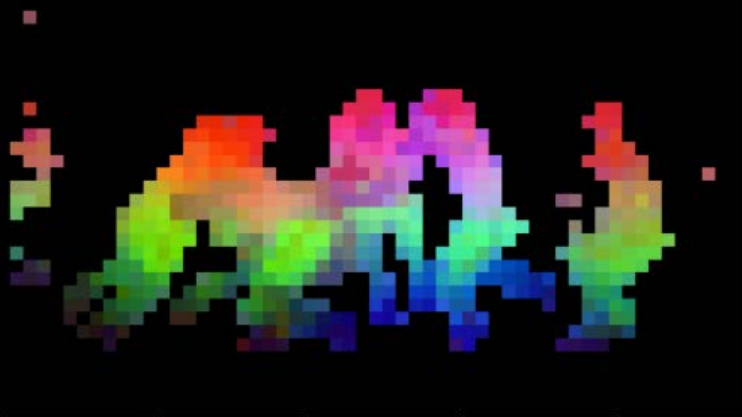} &
        \includegraphics[width=0.094\textwidth, height = 0.08\textwidth]{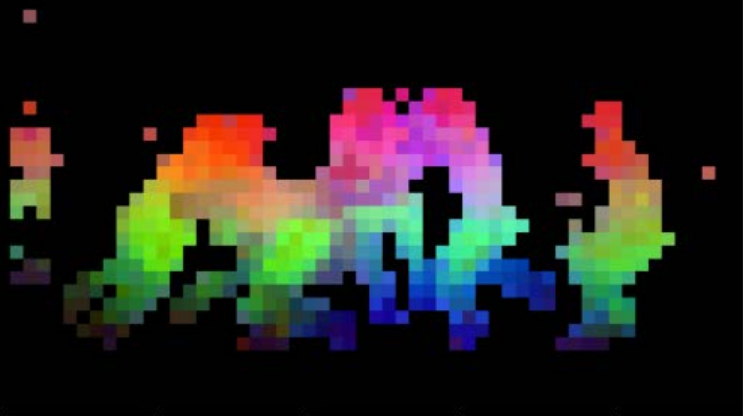}  &
        \includegraphics[width=0.094\textwidth, height = 0.08\textwidth]{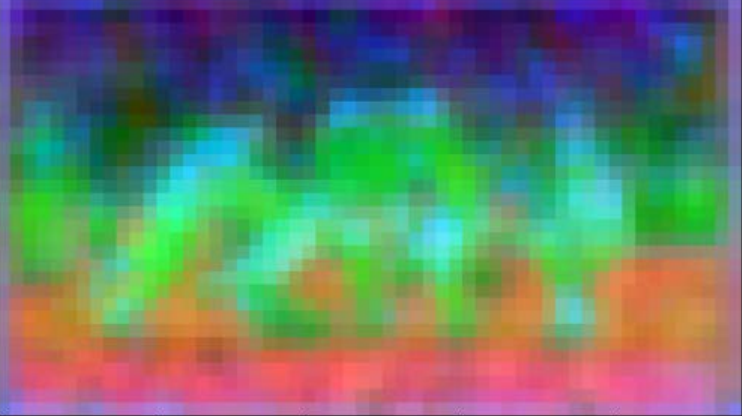} & \hspace{0.0mm}
        \includegraphics[width=0.094\textwidth, height = 0.08\textwidth] {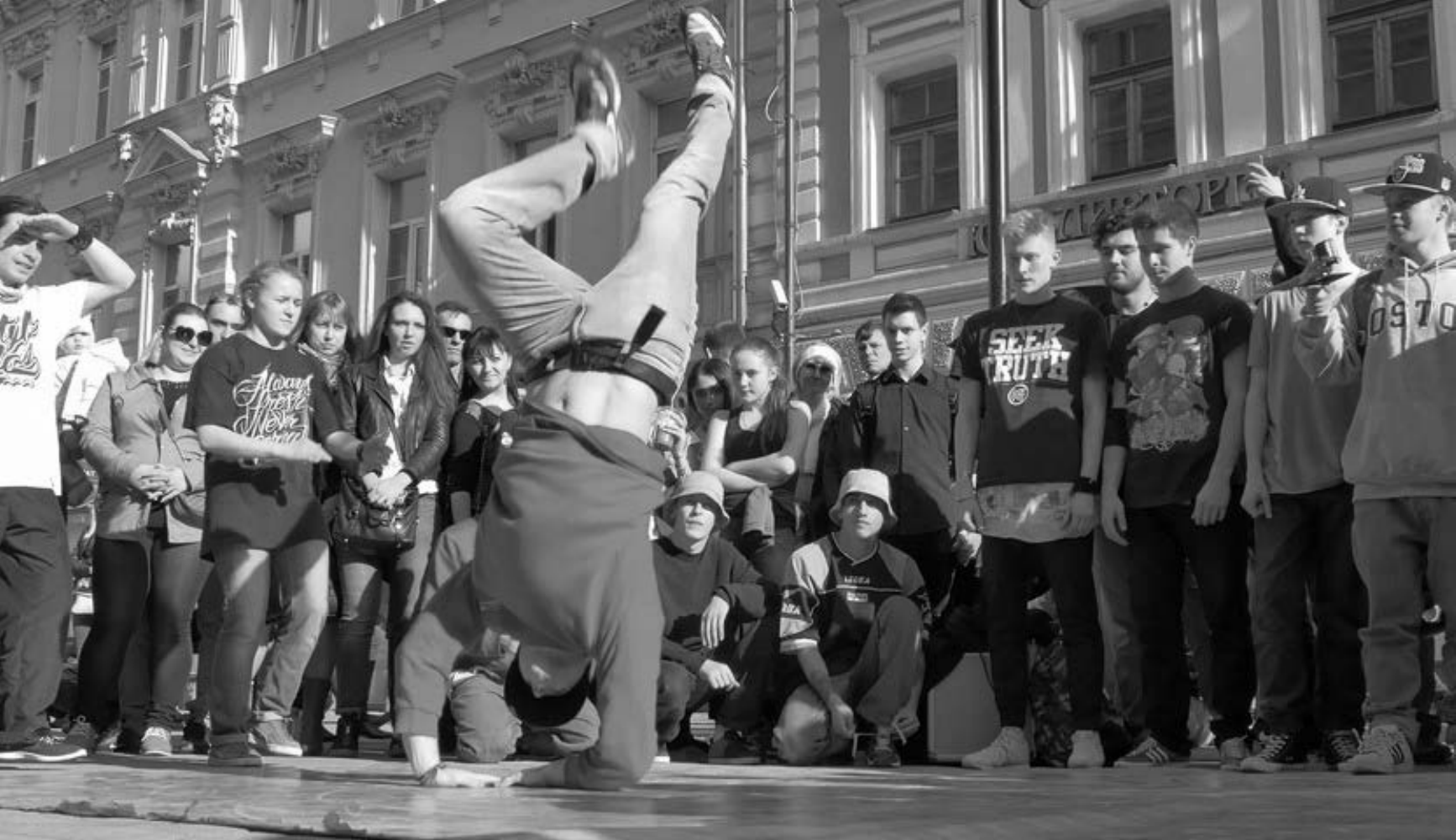} &
        \includegraphics[width=0.094\textwidth, height = 0.08\textwidth] {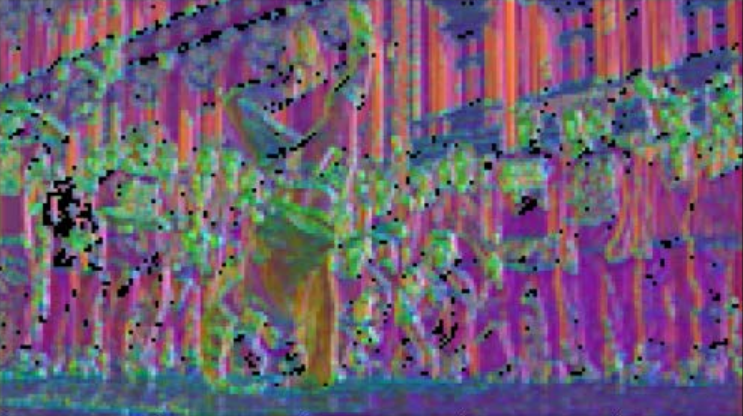} &
        \includegraphics[width=0.094\textwidth, height = 0.08\textwidth] {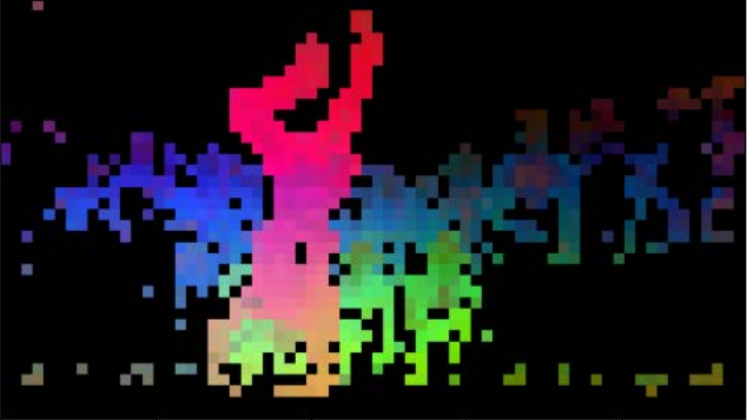} &
        \includegraphics[width=0.094\textwidth, height = 0.08\textwidth]{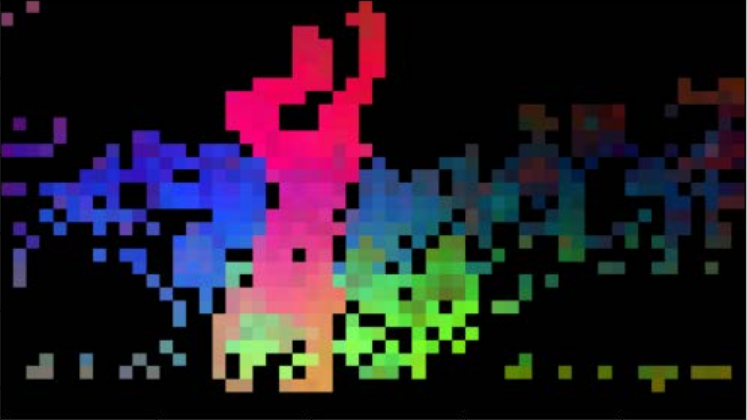}  &
        \includegraphics[width=0.094\textwidth, height = 0.08\textwidth]{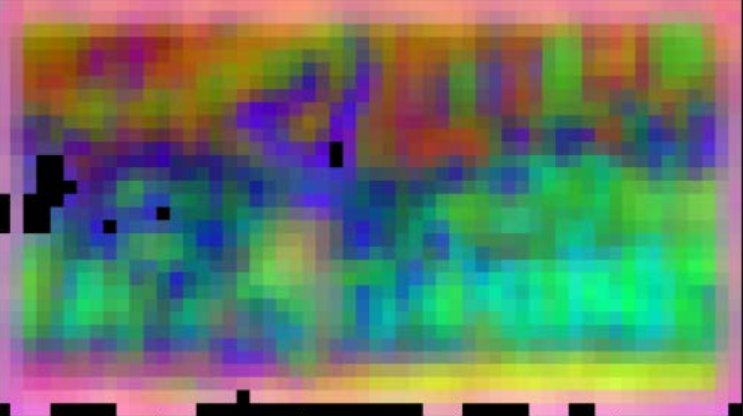}
        \vspace{-1mm}
        \\
        \vspace{-1mm}
        \makebox[0.092\textwidth]{\scriptsize (a) } &
        \makebox[0.092\textwidth]{\scriptsize (b) } &
        \makebox[0.092\textwidth]{\scriptsize (c) } &
        \makebox[0.092\textwidth]{\scriptsize (d) } &
        \makebox[0.092\textwidth]{\scriptsize (e) } & \hspace{0.0mm}
        \makebox[0.092\textwidth]{\scriptsize (a) } &
        \makebox[0.092\textwidth]{\scriptsize (b) } &
        \makebox[0.092\textwidth]{\scriptsize (c) } &
        \makebox[0.092\textwidth]{\scriptsize (d) } &
        \makebox[0.092\textwidth]{\scriptsize (e) }
        \\ %& \vspace{-0.7em}
	\end{tabularx}
	\vspace{-3mm}
	\captionof{figure}{{Visualization of features. We use the PCA tools by~\cite{oquab2023dinov2}. (a) Input frame. (b)-(e) are the features generated by the feature extractors of ColorMNet$_{\text{w/ ResNet50}}$, ColorMNet$_{\text{w/ DINOv2}}$, ColorMNet$_{\text{w/ Concatenation}}$ and ColorMNet (Ours), respectively. Compared with (b), (c) and (d), our proposed PVGFE can generate features that are not only semantic-aware (\ie, the players and the dancer in the foreground) but also sensitive to local details (\ie,  a crowd of spectators in the background) in (e).}}
	\label{fig:visfeature1}

    \vspace{+1mm}

	\setlength\tabcolsep{4.8pt}
	\centering
	\small
        \hspace{-4mm}
	\begin{tabularx}{1.0\textwidth}{ccc}
        \includegraphics[width=0.312\textwidth, height = 0.16\textwidth] {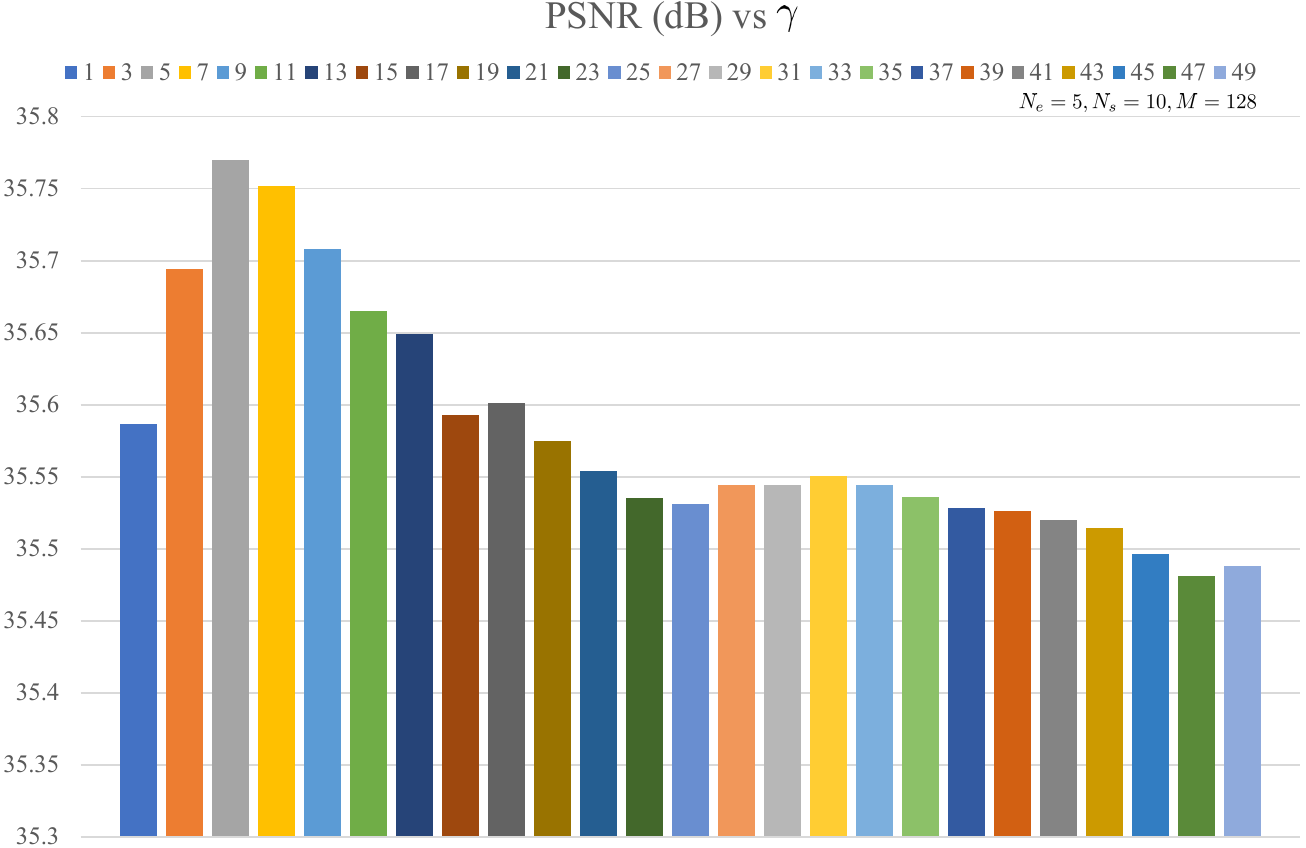} &
        \includegraphics[width=0.312\textwidth, height = 0.16\textwidth] {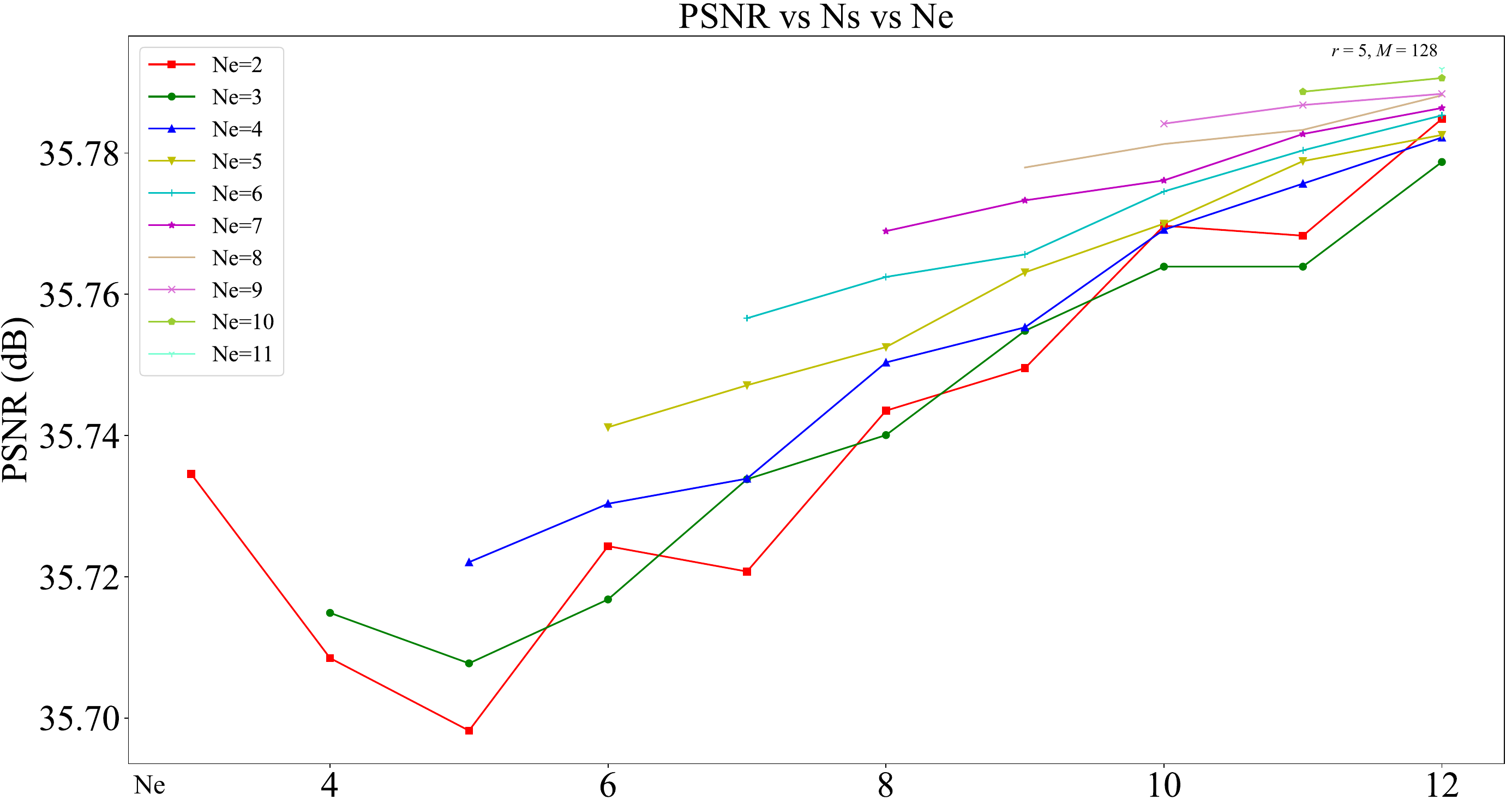} &
        \includegraphics[width=0.312\textwidth, height = 0.16\textwidth] {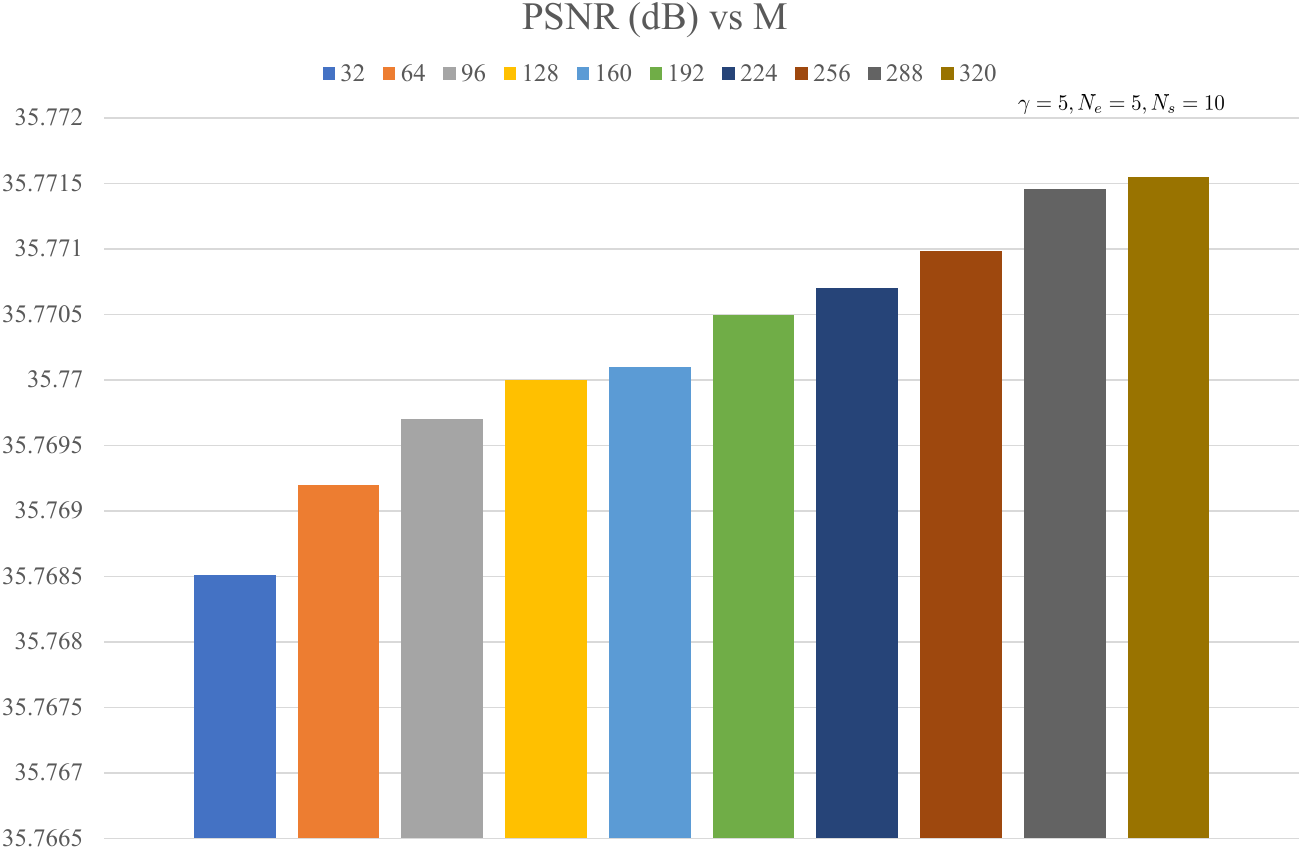}
        \vspace{-1mm}
        \\
        \makebox[0.312\textwidth]{\tiny (a) Ablation on different $\gamma$} &
        \makebox[0.312\textwidth]{\tiny (b) Ablation on $N_e$ and $N_s$} &
        \makebox[0.312\textwidth]{\tiny (c) Ablation on different $M$}
	\end{tabularx}
	\vspace{-3mm}
	\captionof{figure}{Extensive ablation study on the detailed design of the proposed MFP module.}
	\label{fig:ablation}
	% \vspace{-5mm}
% From left to right: $\gamma$, $N_e$ \& $N_s$, $M$.
    \end{minipage}
    \vspace{-8mm}
    \end{figure}

\noindent \textbf{Effectiveness of PVGFE.}
The proposed PVGFE explores robust spatial features that can model both global semantic structures and local details for better video colorization.
To demonstrate its effectiveness, we compare with baseline methods that respectively replace the PVGFE with the pretrained ResNet50~\cite{He_2016_CVPR} (ColorMNet$_{\text{w/ ResNet50}}$ for short), the pretrained DINOv2~\cite{oquab2023dinov2} (ColorMNet$_{\text{w/ DINOv2}}$ for short), and the concatenation of both pretrained ResNet50 and DINOv2 (ColorMNet$_{\text{w/ Concatenation}}$ for short) in our implementation.
Table~\ref{tab:pvgfe} shows that our ColorMNet with the PVGFE outperforms all baseline methods.
The qualitative comparisons in Figure~\ref{fig:pvgfe} show that the results obtained by baseline methods exhibit severe color distortions on the cloth of the player (Figure~\ref{fig:pvgfe}(b)-(d)). In contrast, our proposed ColorMNet with the PVGFE generates a better-colorized frame in Figure~\ref{fig:pvgfe}(e).
Note that the PVGFE can adaptively enhance useful features while reducing the influence of useless information based on the similarities computed on input features by employing cross-attention (\ie, \eqref{eq: cross-attention}).
However, a direct concatenation of features evenly without discrimination, \ie, ColorMNet$_{\text{w/ Concatenation}}$, is less effective for reducing the impact of useless features, thus degrading performance in Table~\ref{tab:pvgfe} and Figure~\ref{fig:pvgfe}(d).
% ,

To better understand the feature estimators mentioned above, we use the PCA tools by~\cite{oquab2023dinov2} to visualize the features generated by them.
Figure~\ref{fig:visfeature1}(b) shows that ResNet50 cannot generate features that are aware of semantic structures.
Although DINOv2 generates more semantic features in Figure~\ref{fig:visfeature1}(c) and (d), it lacks the local details vital for colorization tasks, which explains why DINOv2 performs favorably in high-level vision tasks, \ie, classification and segmentation (see~\cite{oquab2023dinov2} for details), but fails in colorization as the exact colors for pixels of objects are crucial considerations in colorization, unlike in segmentation where the primary decision is whether or not a pixel belongs to a human.
Figure~\ref{fig:visfeature1}(e) shows that our proposed PVGFE module is capable of generating better features optimized for colorization, retaining both semantic relevance and local details.

\noindent \textbf{Effectiveness of MFP.}
The proposed MFP propagates temporal features for better long-range correspondences.
To investigate whether directly stacking multiple frames along the temporal dimension or recurrently propagating features can already generate competitive results, we compare with baseline methods that respectively replace the MFP with direct stacking of the features from all previous colorized frames and the exemplar image along the temporal dimension (ColorMNet$_{\text{w/ Stacking}}$ for short) and the recurrent-based feature propagation~\cite{atcvc, favc, wan2022oldfilm, zhang2019deep} (ColorMNet$_{\text{w/ Recurrent}}$ for short) in our implementation.

Table~\ref{tab:pvgfe} shows that the PSNR value of our ColorMNet is at least $0.51$dB higher than each baseline method, which illustrates the effectiveness of the proposed MFP in propagating features for video colorization.
Figure~\ref{fig:mfp}(b) shows that the baseline that directly stacks frames is not able to generate a realistic image as spatial-temporal priors are not well-explored.
The result obtained by the baseline with recurrent-based feature propagation contains significant color distortions on the boy (Figure~\ref{fig:mfp}(c)), as the errors accumulate in the recurrent-based propagation steps.
In contrast, the proposed ColorMNet using the MFP generates a vivid and error-free image in Figure~\ref{fig:mfp}(d).

We further conduct an extensive ablation study on the parameters of the proposed MFP module.
Figure~\ref{fig:ablation}(a) shows that our method achieves its peak PSNR when $\gamma$ equals 5, as a higher $\gamma$ risks potential information loss, while a lower $\gamma$ could contribute redundant data.
Figure~\ref{fig:ablation}(b) and (c) show that our method generally achieves slightly higher PSNR values with $N_e$, $N_s$ and $M$ increasing, respectively.
However, note that the GPU memory usage escalates correspondingly with larger values of $N_e$, $N_s$ and $M$.

\noindent \textbf{Efficiency of MFP.}
To examine the efficiency of the proposed MFP, we further evaluate the proposed ColorMNet against the baseline method with direct stacking (\ie, ColorMNet$_{\text{w/ Stacking}}$) on the validation set of NVCC2023 \cite{Kang_2023_CVPR} in terms of the maximum GPU memory consumption and the average running time.
Table~\ref{tab:abl_runtime} shows that our ColorMNet only requires $7.8\%$ of the maximum GPU consumption of the baseline method, but the average running time of our approach is nearly 14$\times$ faster than the baseline method, which indicates the efficiency of the proposed MFP.

Tables~\ref{tab:pvgfe} and~\ref{tab:abl_runtime} show that our approach using the MFP achieves a favorable performance in terms of faster inference speed, lower GPU memory consumption, and better colorization results, which demonstrates the effectiveness and efficiency of the proposed MFP in video colorization.

\begin{table}[!t]
\vspace{-6mm}
    \begin{minipage}{0.48\textwidth}
    \vspace{+4mm}
	\caption{Efficiency of the proposed MFP module for video colorization.}
	\vspace{-3mm}
 \resizebox{1.0\textwidth}{!}{
	\centering
	\begin{tabular}{lcc}
	\noalign{\hrule height 0.3mm}
	Methods &Memory consumption (G)   &Running time (/s)\\
	\hline
	ColorMNet$_{\text{w/ Stacking}}$     & 24.1  & 1.04   \\
    ColorMNet (Ours)     & \textbf{1.9}  & \textbf{0.07   }\\
	\noalign{\hrule height 0.3mm}
	\end{tabular}
    }
	\label{tab:abl_runtime}

    \vspace{1mm}

    \setlength\tabcolsep{1.0pt}
	\centering
	\small
	\begin{tabularx}{1.0\textwidth}{ccccc}
        \includegraphics[width=0.187\textwidth, height = 0.15\textwidth] {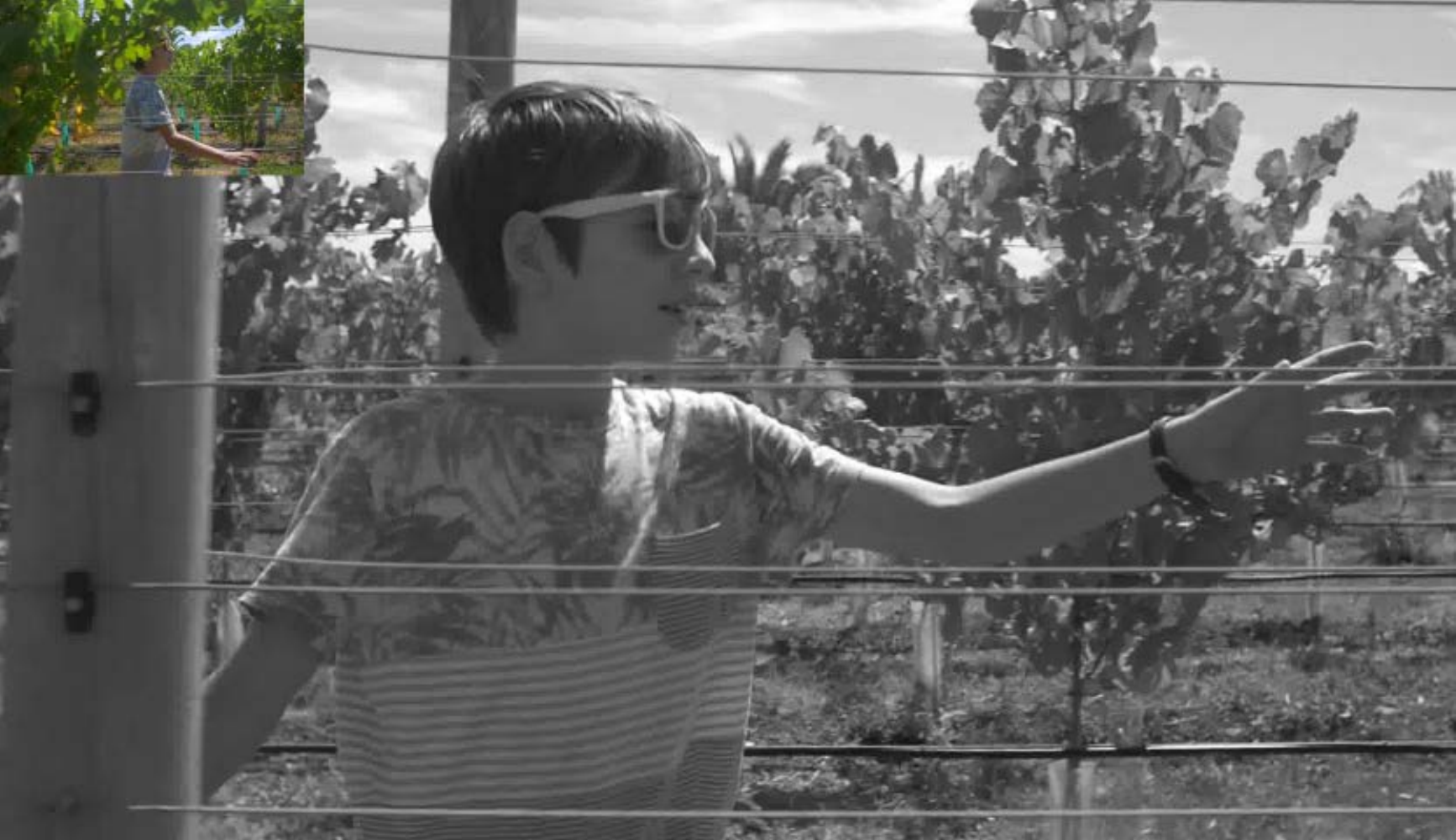} &
        \includegraphics[width=0.187\textwidth, height = 0.15\textwidth] {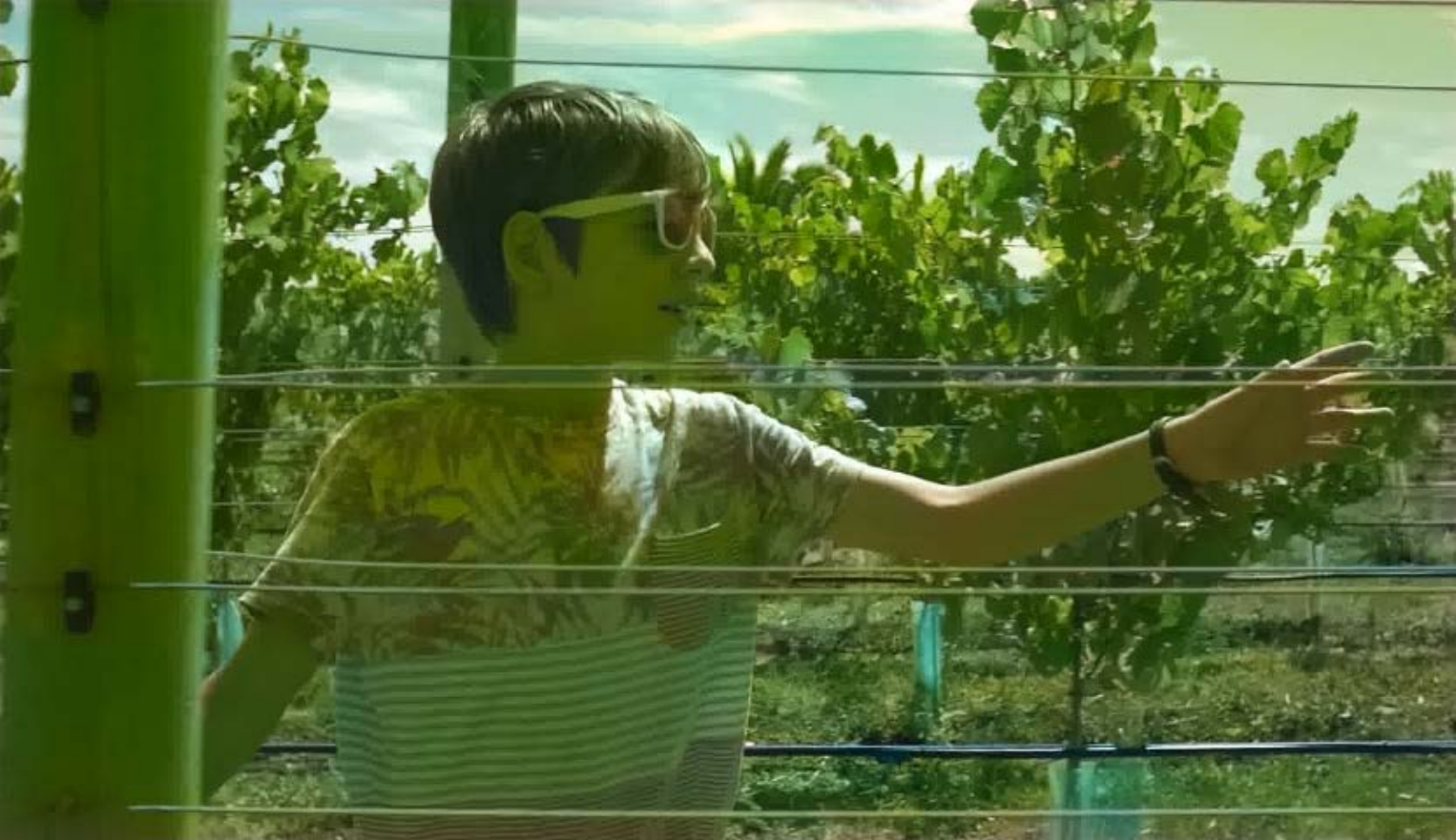} &
        \includegraphics[width=0.187\textwidth, height = 0.15\textwidth] {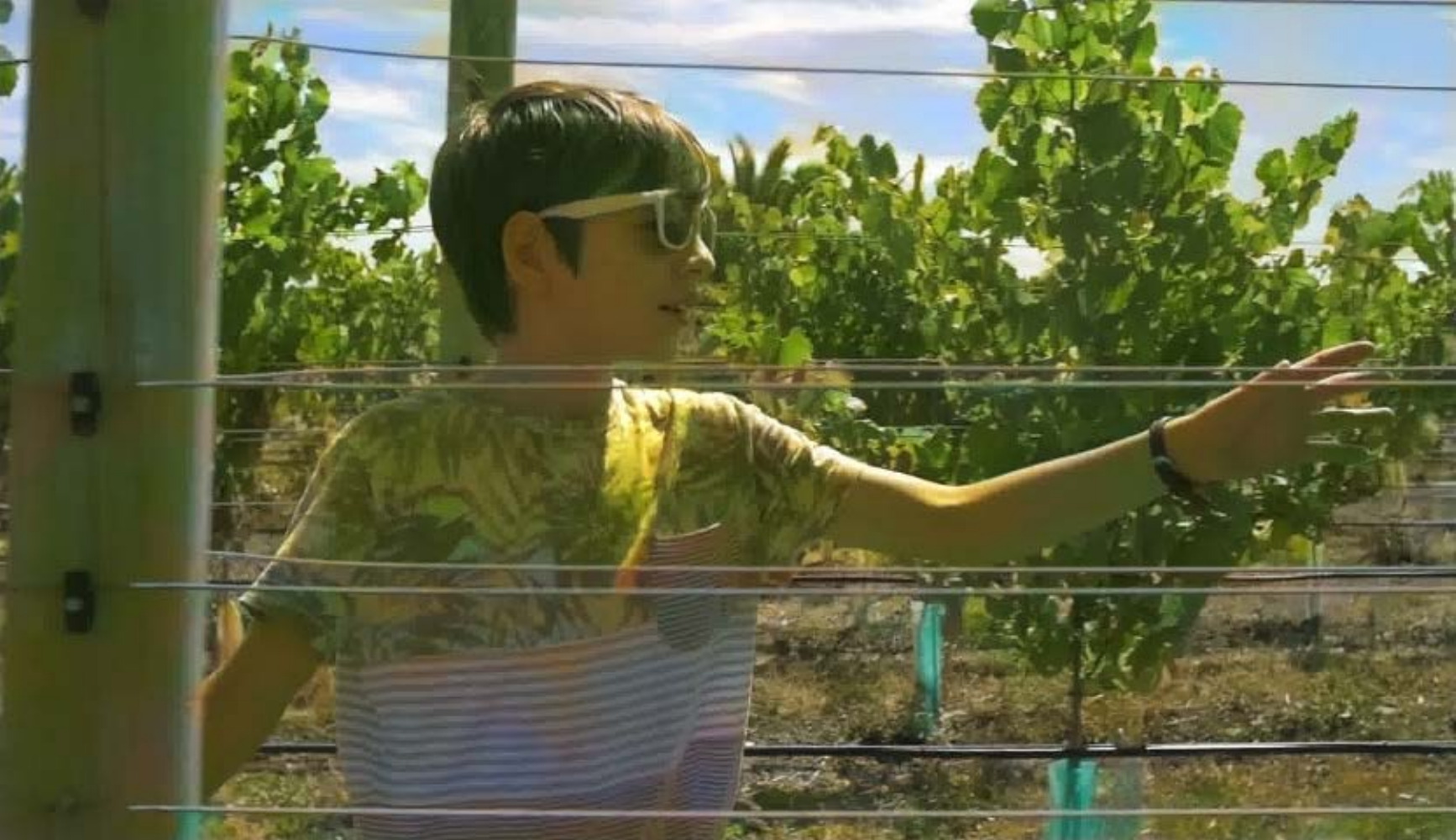} &
        \includegraphics[width=0.187\textwidth, height = 0.15\textwidth]{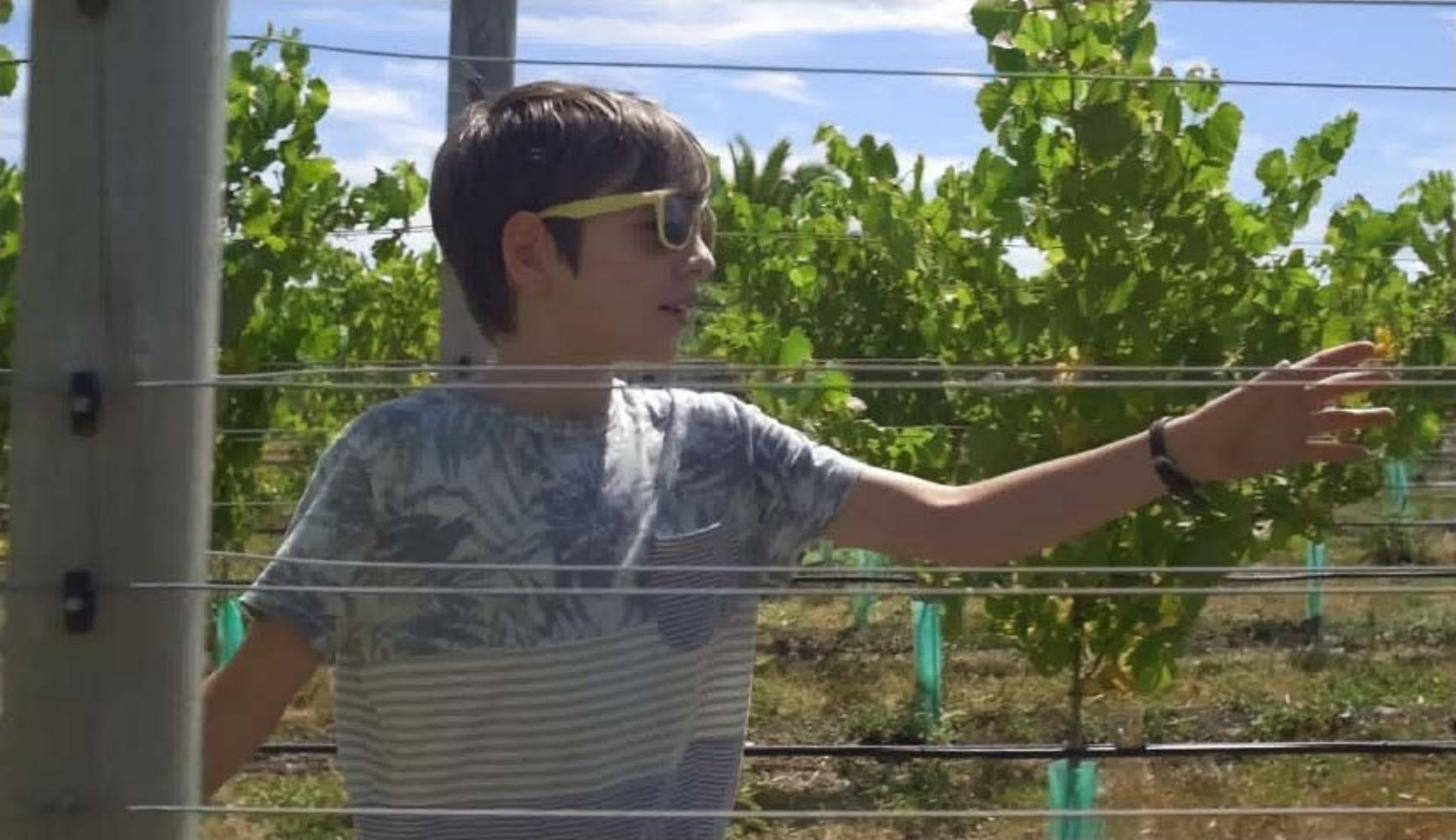}  &
        \includegraphics[width=0.187\textwidth, height = 0.15\textwidth]{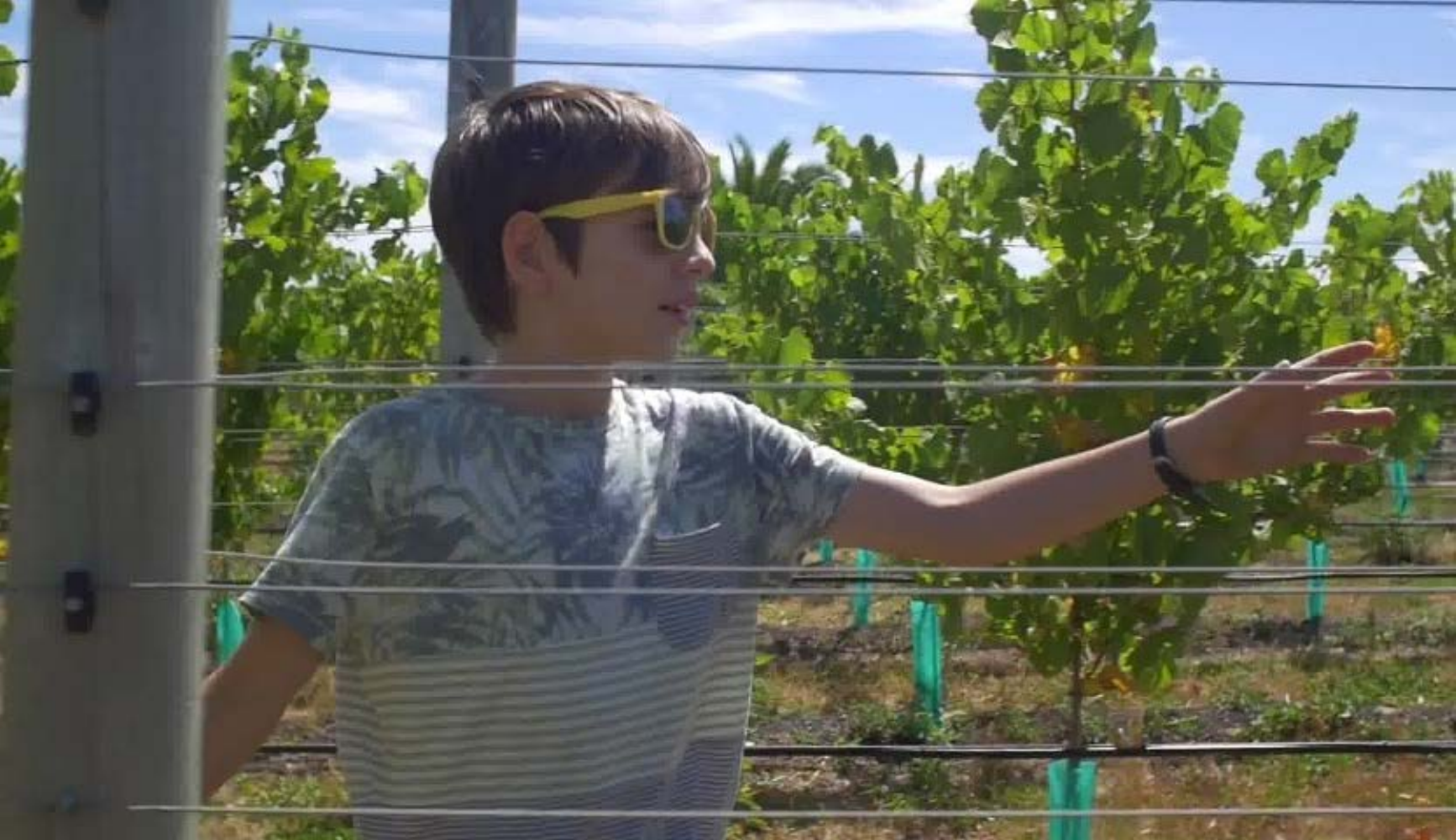}
        \vspace{-1mm}
        \\
        \vspace{-1mm}
        \makebox[0.092\textwidth]{\scriptsize (a) } &
        \makebox[0.092\textwidth]{\scriptsize (b) } &
        \makebox[0.092\textwidth]{\scriptsize (c) } &
        \makebox[0.092\textwidth]{\scriptsize (d) } & \hspace{-2mm}
        \makebox[0.092\textwidth]{\scriptsize (e) }
        \\ %& \vspace{-0.7em}
	\end{tabularx}
	\vspace{0mm}
	\captionof{figure}{{Effectiveness of the MFP module for video colorization. (a) Input frame and exemplar image. (b)-(d) are the colorization results by ColorMNet$_{\text{w/ Stacking}}$, ColorMNet$_{\text{w/ Recurrent}}$ and ColorMNet (Ours), respectively. (e) Ground truth. Compared with (b) and (c), the colorized result (d) by our method contains fewer color distortions and more vivid details. }}
	\label{fig:mfp}

     \end{minipage}
     \hspace{+2mm}
    \begin{minipage}{0.48\textwidth}
     \vspace{4.7mm}
	\setlength\tabcolsep{1.0pt}
	\centering
	\small
	\begin{tabularx}{1.0\textwidth}{cccc}
        \includegraphics[width=0.238\textwidth, height = 0.165\textwidth] {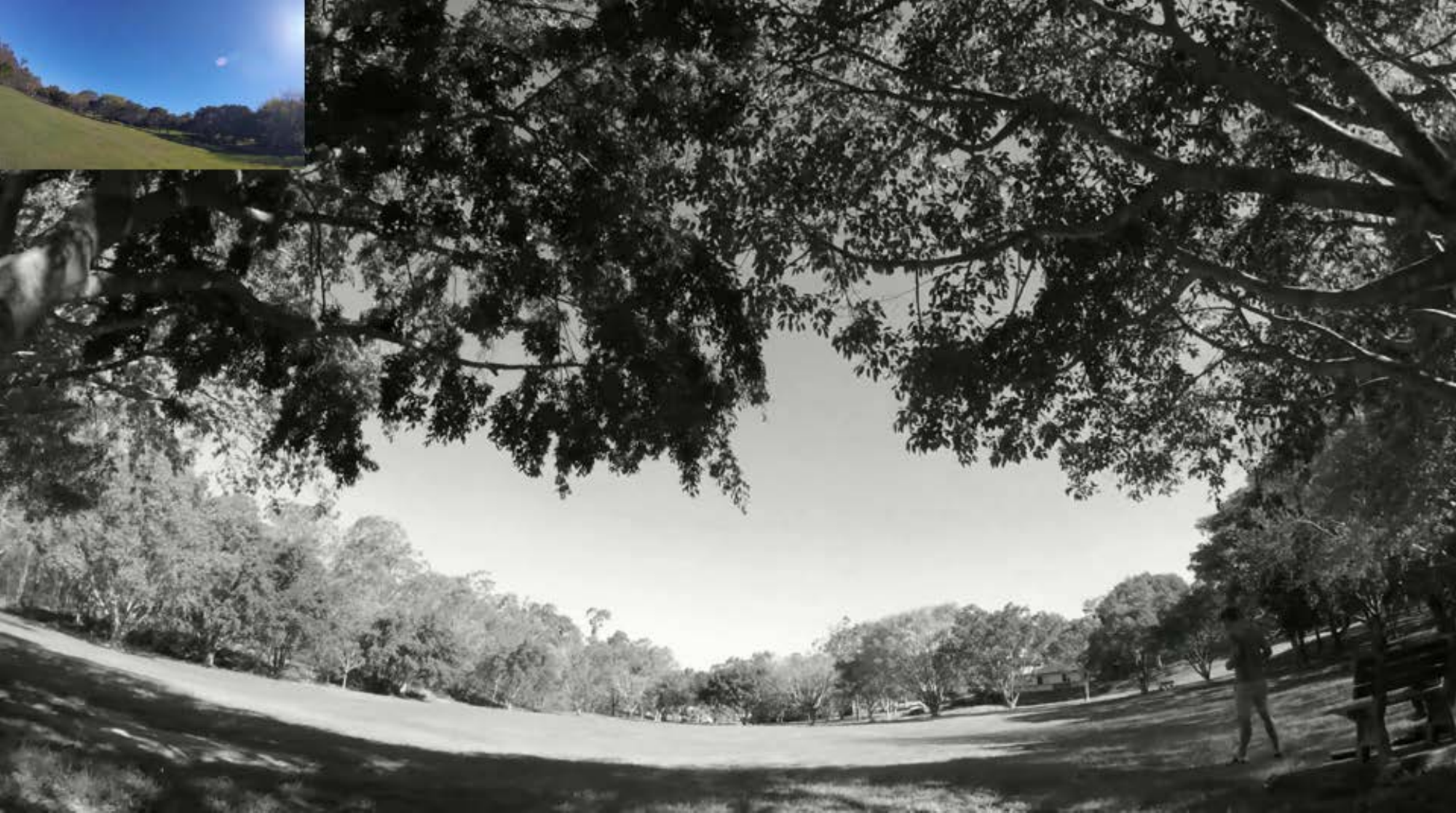} &
        \includegraphics[width=0.238\textwidth, height = 0.165\textwidth] {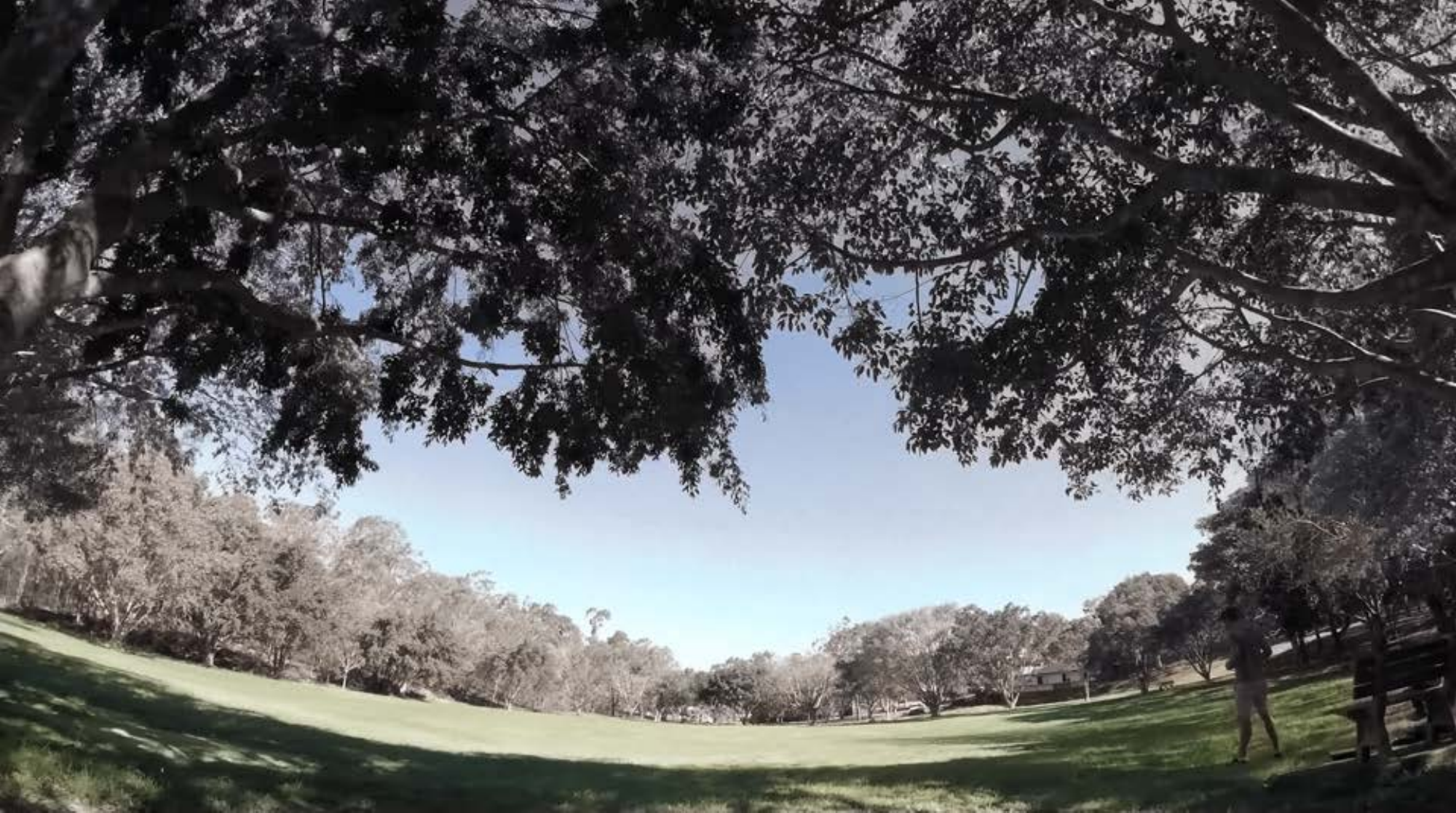} &
        \includegraphics[width=0.238\textwidth, height = 0.165\textwidth]{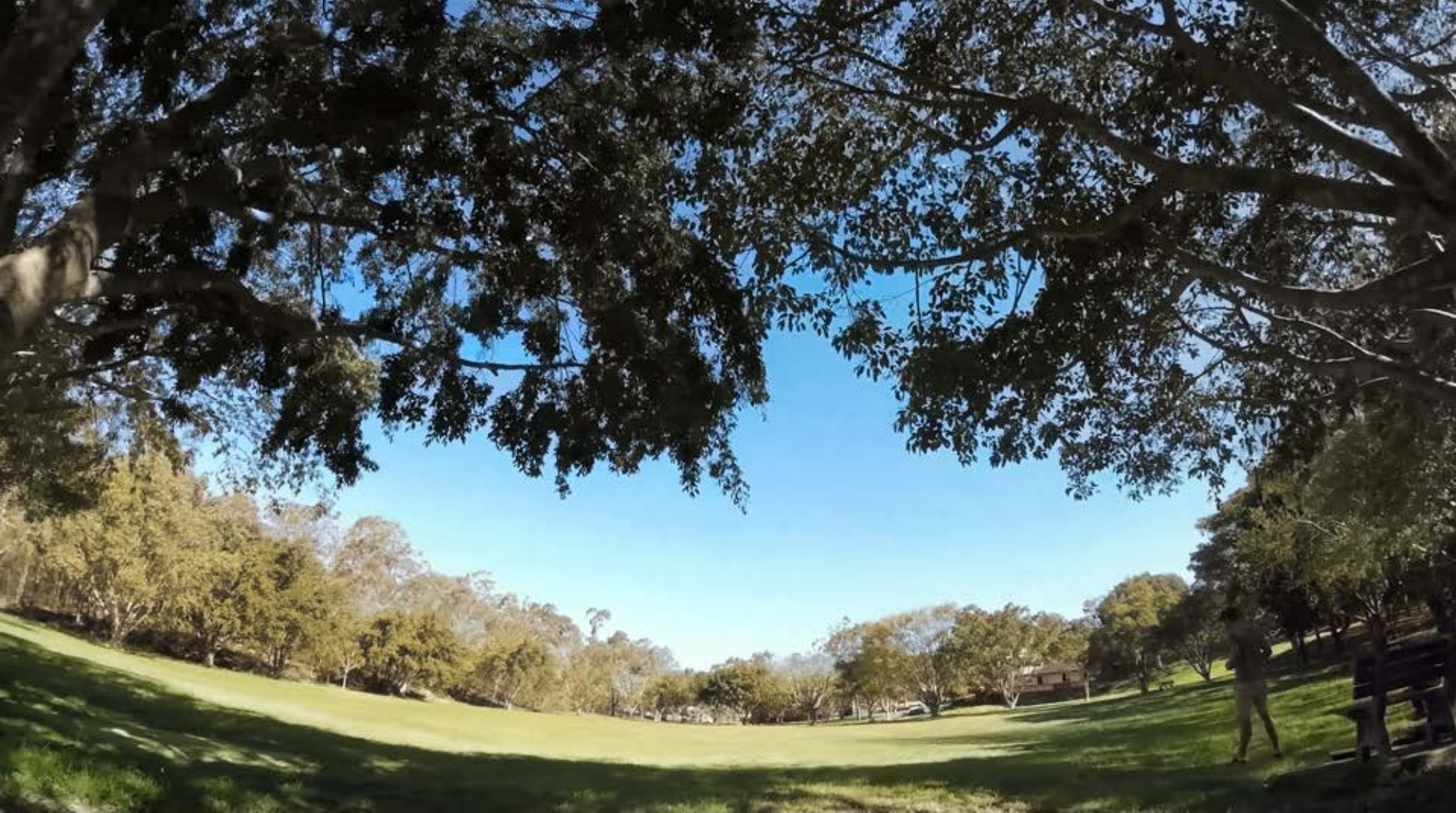}  &
        \includegraphics[width=0.238\textwidth, height = 0.165\textwidth]{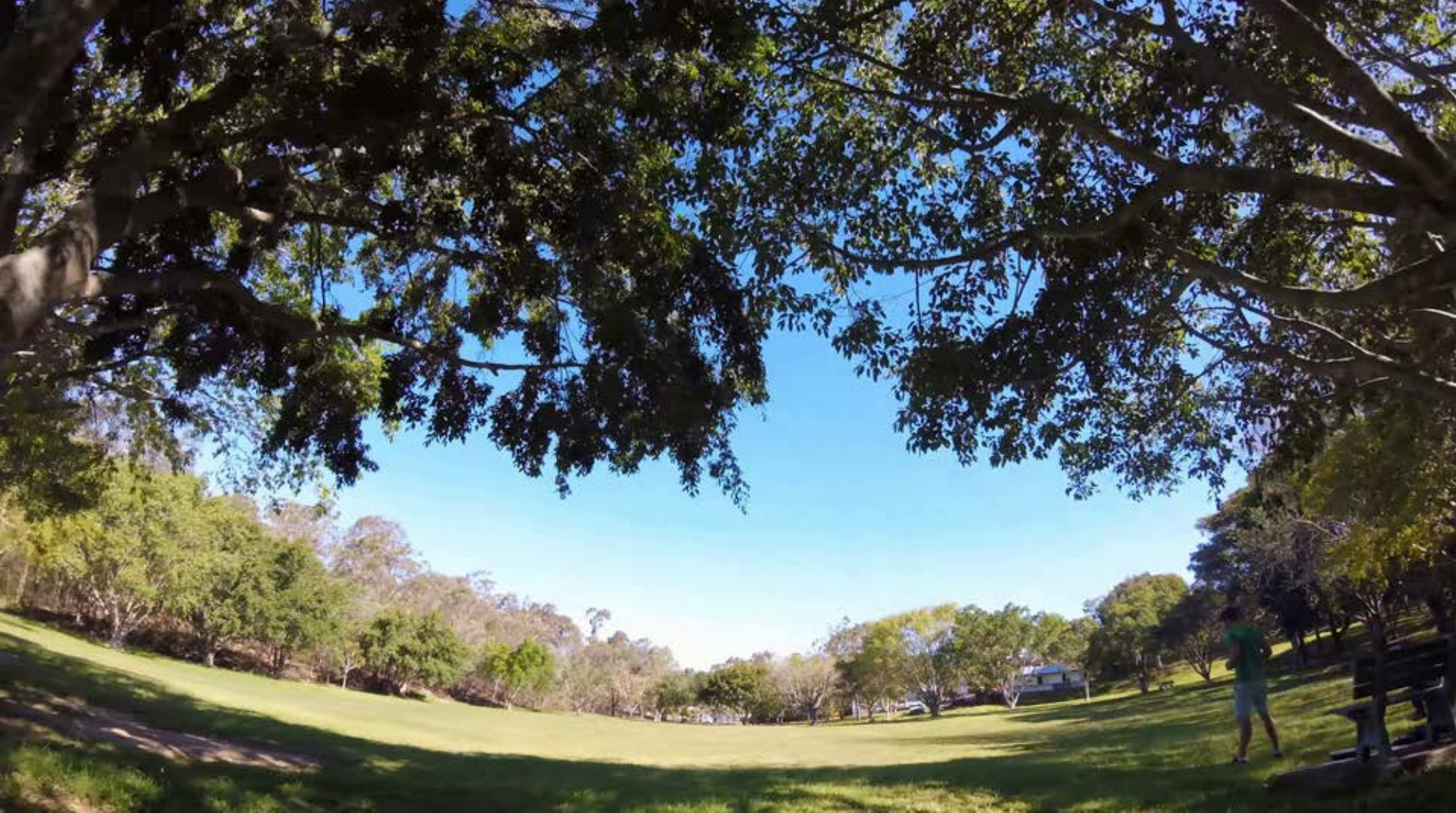}
        \vspace{-1mm}
        \\
        \vspace{-1mm}
        \makebox[0.112\textwidth]{\scriptsize (a) } &
        \makebox[0.112\textwidth]{\scriptsize (b)} &
        \makebox[0.112\textwidth]{\scriptsize (c) } &
        \makebox[0.112\textwidth]{\scriptsize (d) }
	\end{tabularx}
	\vspace{+1mm}
	\captionof{figure}{{Effectiveness of the proposed LA module for video colorization. (a) Input frame and exemplar image. (b) and (c) are the colorization results by ColorMNet$_{\text{w/o LA}}$ and ColorMNet (Ours). (d) Ground truth. Our approach is able to generate better-colorized result in (c).}}
 % \vspace{-1mm}
	\label{fig:la}

	\vspace{-3mm}

 \setlength\tabcolsep{1.0pt}
	\centering
	\small
	\begin{tabularx}{1.0\textwidth}{cccc}
        \includegraphics[width=0.238\textwidth, height = 0.165\textwidth] {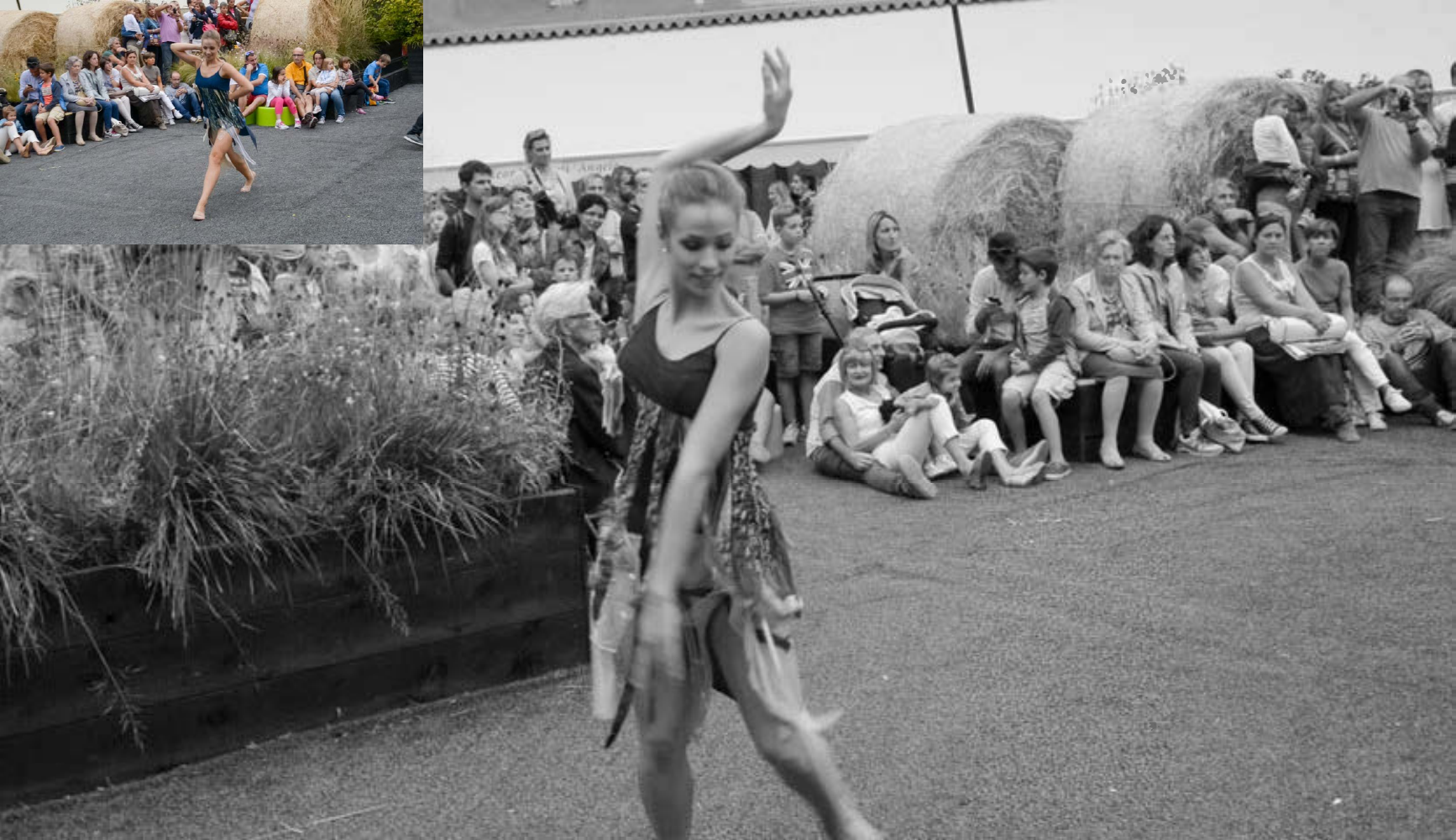} &
        \includegraphics[width=0.238\textwidth, height = 0.165\textwidth] {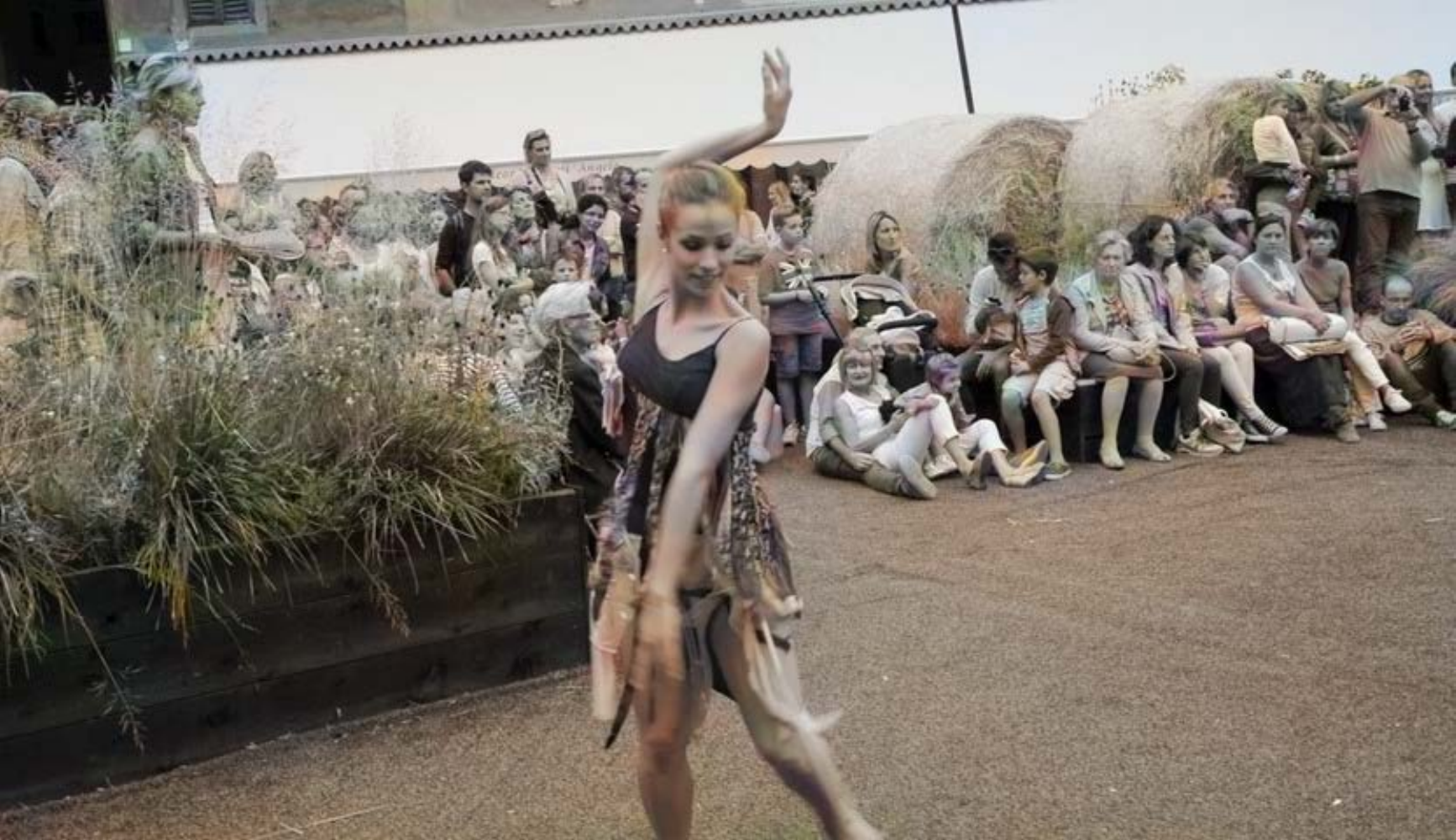} &
        \includegraphics[width=0.238\textwidth, height = 0.165\textwidth]{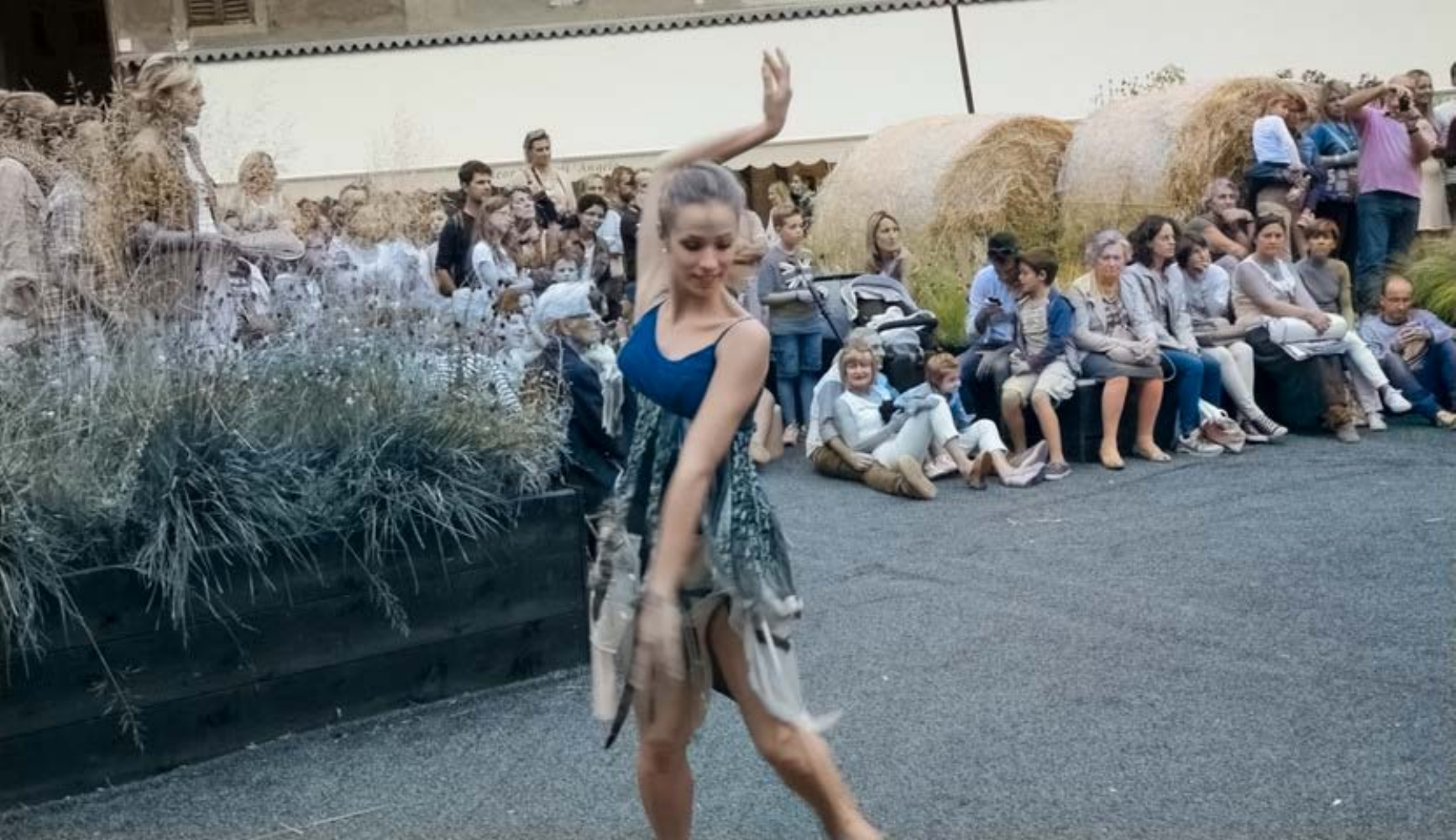}  &
        \includegraphics[width=0.238\textwidth, height = 0.165\textwidth]{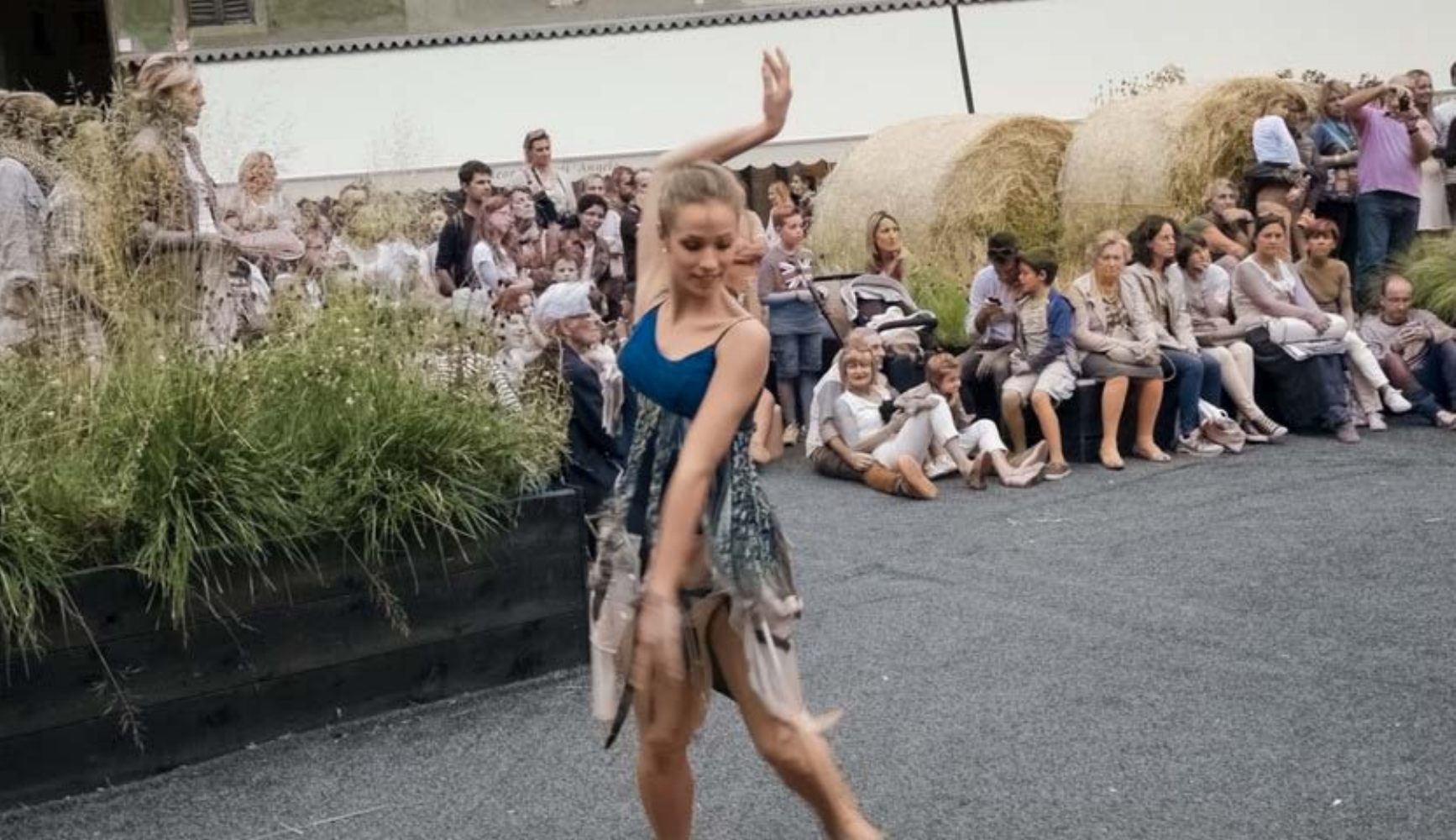}
        \vspace{-1mm}
        \\
        \vspace{-1mm}
        \makebox[0.112\textwidth]{\tiny (a) Inputs} &
        \makebox[0.112\textwidth]{\tiny (b) {\makecell{MAMBA\\~\cite{sun2021mamba}} }} &
        \makebox[0.112\textwidth]{\tiny (c) {\makecell{MeMOTR \\~\cite{gao2023memotr}} }} &
        \makebox[0.112\textwidth]{\tiny (d) Ours}
	\end{tabularx}
	\vspace{+1mm}
	\captionof{figure}{{Comparison results with closely-related methods on the DAVIS~\cite{Perazzi_CVPR_2016}.}}
 % \vspace{-1mm}
	\label{fig:closememory}
    \end{minipage}
    \vspace{-11mm}
\end{table}

\noindent \textbf{Effectiveness of LA.}
To demonstrate the effect of the proposed LA, we further compare with a baseline method that removes the LA module (ColorMNet$_{\text{w/o LA}}$ for short) in our implementation.
Table~\ref{tab:pvgfe} shows that our ColorMNet
using the LA generates better results with higher PSNR and SSIM values than the baseline method.
Figure~\ref{fig:la}(b) shows that the baseline method without the LA does not exploit the prior information among consecutive frames and thus cannot restore the colors on the sky and the leaves.
However, our approach generates vivid and realistic colors in Figure~\ref{fig:la}(c), which demonstrates that the proposed LA module is effective in capturing and leveraging better spatial-temporal features.

\noindent \textbf{Closely-related methods.}
To the best of our knowledge, we are the first to optimize a memory bank strategy suitable for colorization, yet it should be acknowledged that related strategies have been explored in some video processing works, \eg, MAMBA~\cite{sun2021mamba} constructs a memory bank to solve video object detection by employing random selection strategy, MeMOTR~\cite{gao2023memotr} introduces a long-term memory to solve video object tracking by assigning exponentially decaying weights to it.
Unlike MAMBA which applies a randomized selection approach, treating every feature on par, or MeMOTR which updates past memorized features via exponentially decaying weights, our proposed MFP module stores features based on their importance which is determined by the frequency of usage, thus empowering the ability of global relation mining.

We further adopt the random selection in MAMBA and the decaying weights in MeMOTR to replace our MFP for comparison.
To ensure a fair comparison, the same training settings are kept for model testing.
Figure~\ref{fig:closememory} shows that our method can generate better colors for the dancing girl and the green grass.

\noindent \textbf{Limitations.}
Our proposed method aims to further enhance video colorization performance while reducing GPU memory usage.
However, there are limitations in the model complexity, \ie, our model requires 123.61 Million parameters.

\vspace{-3mm}
\section{Conclusion}
\vspace{-3mm}
\label{sec:conclusion}

We present an effective memory-based deep spatial-temporal feature propagation network for video colorization.
We develop a large-pretrained visual model guided feature estimation module to better explore robust spatial features.
To establish reliable connections from far-apart frames, we propose a memory-based feature propagation module.
We develop a local attention module to better utilize spatial-temporal priors.
Both quantitative and qualitative experimental results show that our method performs favorably against state-of-the-art methods.

\bibliographystyle{splncs04}
\bibliography{main}
\end{document}

%% TITLE
% \title{\paperTitle}
\title{ColorMNet: A Memory-based Deep Spatial-Temporal Feature Propagation Network for Video Colorization} 

\titlerunning{ColorMNet: A MSTFP Network for Video Colorization}

\author{
Yixin Yang \and
Jiangxin Dong \and
Jinhui Tang \and
Jinshan Pan}

\authorrunning{Y.~Yang \etal}
% If there are more than two authors, '\etal' is used.

\institute{Nanjing University of Science and Technology}

% \maketitlesupplementary
\maketitle

\vspace{-5mm}
% \hspace{-5mm}
% \appendix
{\section*{\centering Overview}}
\vspace{-3mm}
In this document, we first present the network details in Section~\ref{sec:Network_Details}. 
Then, we analyze the effectiveness of the
proposed large-pretrained visual model guided feature estimation (PVGFE) module and the memory-based feature propagation (MFP) module in Section~\ref{sec:epvgfe} and Section~\ref{sec:emfp}. 
To examine the effectiveness of the proposed local attention (LA) module on video colorization, we further analyze it in Section~\ref{sec:ela}.
In addition, we conduct a user study to investigate the subjective preference by human observers of each colorization method in Section~\ref{sec:User_Study}.
Finally, we show more visual comparisons on both synthetic datasets and real-world videos in Section~\ref{sec:More_Experimental_Results}.

\vspace{6mm}
\section{Network Details}
\label{sec:Network_Details}
As stated in Section 3 of the main manuscript, our method contains a large-pretrained visual model guided feature estimation module, a memory-based feature propagation module, and a local attention module for video colorization. 
We also show the network details of the proposed memory-based deep spatial-temporal feature propagation network for video colorization in Figures 2 of the main manuscript. 
In this document, we list the detailed architecture of our proposed ColorMNet in Table~\ref{tab:Network_Details}. The spatial resolution of the input image is $448\times 448$ pixels.

\begin{table*}[!t]
\caption{{Detailed architecture of our proposed ColorMNet}. $\begin{bmatrix}\text{Conv. 7$\times$7, 64, stride 2}\end{bmatrix}$ denotes a convolution with the filter size of $7\times7$ pixels with the filter number of 64 with stride 2, Embed dim. denotes the dimension of embedding, $\begin{bmatrix}\text{Interpolation, $\times2$}\end{bmatrix}$ denotes an interpolation operation with a scale factor equal to 2, $\begin{bmatrix}\text{ResBlock, 256}\end{bmatrix}$ denotes a ResBlock consisting of convolutions with the filter size of $3\times3$ pixels with the filter number of 256. }
\centering
\addtolength{\tabcolsep}{-2pt}
\vspace{-2mm}
\resizebox{\textwidth}{!}{
\begin{tabular}{c|c|cc}
\noalign{\hrule height 0.3mm}
& & \multicolumn{2}{c}{ColorMNet} \\
\hline
 & Output size & ResNet50~\cite{He_2016_CVPR} &  DINOv2~\cite{oquab2023dinov2}\\ \hline
\multirow{3}{*}{Key feature extractor (PVGFE)}&  {28$\times$28$\times$1024}  &  
 \makecell[c]{Conv. 7$\times$7, 64, stride 2 \\ 
 MaxPool, 3$\times$3, stride 2\\
 $\begin{bmatrix}\text{Conv. 1$\times$1, 64}\\\text{Conv. 3$\times$3, 64}\\\text{Conv. 1$\times$1, 256}\end{bmatrix}$ $\times$ 3 \\
 $\begin{bmatrix}\text{Conv. 1$\times$1, 128}\\\text{Conv. 3$\times$3, 128}\\\text{Conv. 1$\times$1, 512}\end{bmatrix}$ $\times$ 4  \\
 $\begin{bmatrix}\text{Conv. 1$\times$1, 256}\\\text{Conv. 3$\times$3, 256}\\\text{Conv. 1$\times$1, 1024}\end{bmatrix}$ $\times$ 6 } & \makecell[c]{Patch Embedding\\
 $\begin{bmatrix}\text{Transformer block}\\\text{Patch size = 14}\\\text{Embed dim. = 384} \\\text{Heads = 6} \\\text{Blocks = 12} \\\text{FFN layer = MLP} \end{bmatrix}$ $\times$ 12 \\
 Concat features from last 4 layers\\
 Conv. 1$\times$1, 1536\\
 Interpolation, $\times 14 /16$\\
 Conv. 3$\times$3, 1024
 }
\\ \cline{2-4}
& 28$\times$28$\times$64 & Conv. 3$\times$3, 64 & Conv. 3$\times$3, 64 \\ \cline{2-4}
& 28$\times$28$\times$64 & \multicolumn{2}{c}{\makecell[c]{Cross-channel attention~\cite{zamir2022restormer}}}\\ \hline
\multirow{2}{*}{Value feature extractor (ResNet18~\cite{He_2016_CVPR})}& 28$\times$28$\times$256  &  \multicolumn{2}{c}{
 \makecell[c]{ResNet18~\cite{He_2016_CVPR}\\Conv. 7$\times$7, 64, stride 2 \\ 
 MaxPool, 3$\times$3, stride 2\\
 $\begin{bmatrix}\text{Conv. 1$\times$1, 64}\\\text{Conv. 3$\times$3, 64}\end{bmatrix}$ $\times$ 2 \\
 $\begin{bmatrix}\text{Conv. 1$\times$1, 128}\\\text{Conv. 3$\times$3, 128}\end{bmatrix}$ $\times$ 2  \\
 $\begin{bmatrix}\text{Conv. 1$\times$1, 256}\\\text{Conv. 3$\times$3, 256}\end{bmatrix}$ $\times$ 2}}  \\ \cline{2-4}
& 28$\times$28$\times$512 & \multicolumn{2}{c}{Conv. 3$\times$3, 512}  \\ \hline
\multirow{4}{*}{Decoder} &  112$\times$112$\times$256  &  
 \multicolumn{2}{c}{\makecell{Interpolation, $\times 2$\\ ResBlock, 256 \\ Interpolation, $\times 2$ \\ResBlock, 256 }}  \\
\cline{2-4}
& 112$\times$112$\times$2 & \multicolumn{2}{c}{Conv. 1$\times$1, 2}\\ \cline{2-4}
& 112$\times$112$\times$2 & \multicolumn{2}{c}{GRU~\cite{cho2014gru}}\\ \cline{2-4}
& 448$\times$448$\times$2& \multicolumn{2}{c}{Interpolation, $\times 4$}\\
\noalign{\hrule height 0.3mm}
\end{tabular}
}
\vspace{-1mm}
% \normalsize
\label{tab:Network_Details}
% \vspace{-3mm}
\end{table*}

\section{Effectiveness of the Large-Pretrained Visual Model Guided Feature Estimation Module}
\label{sec:epvgfe}
As stated in Section 5 of the manuscript, we have analyzed the effectiveness of the large-pretrained visual model guided feature estimation (PVGFE) module.
In this supplemental material, we further show more visual comparisons to demonstrate the effectiveness of the PVGFE module. In \textit{`ComparisonsWithSOTA.mp4'}, we show that our proposed ColorMNet with using the PVGFE is able to generate better-colorized videos.

\section{Effectiveness of the Memory-based Feature Propagation Module}
\label{sec:emfp}
As stated in Section 5 of the manuscript, we have analyzed the effectiveness of the memory-based feature propagation (MFP) module.
In this supplemental material, we further show more visual comparisons to demonstrate the effectiveness of the MFP module. In \textit{`ComparisonsWithSOTA.mp4'}, we show that our proposed ColorMNet with using the MFP is able to generate better-colorized videos compared with the method without using the MFP.

% NTIRE testset teaser
\begin{figure*}[!t]\tiny
	% \vspace{-3.7in} 
     % \hspace{-5mm}
	\begin{minipage}{\textwidth}
    % \hspace{-2mm}
	\centering
        \setlength{\tabcolsep}{0.8mm}
    % \hspace{-4mm}
	 \begin{tabular}{ccccc}
    \multicolumn{2}{c}{\multirow{3}{*}[41pt]{\includegraphics[width=0.5\linewidth,scale=1.00]{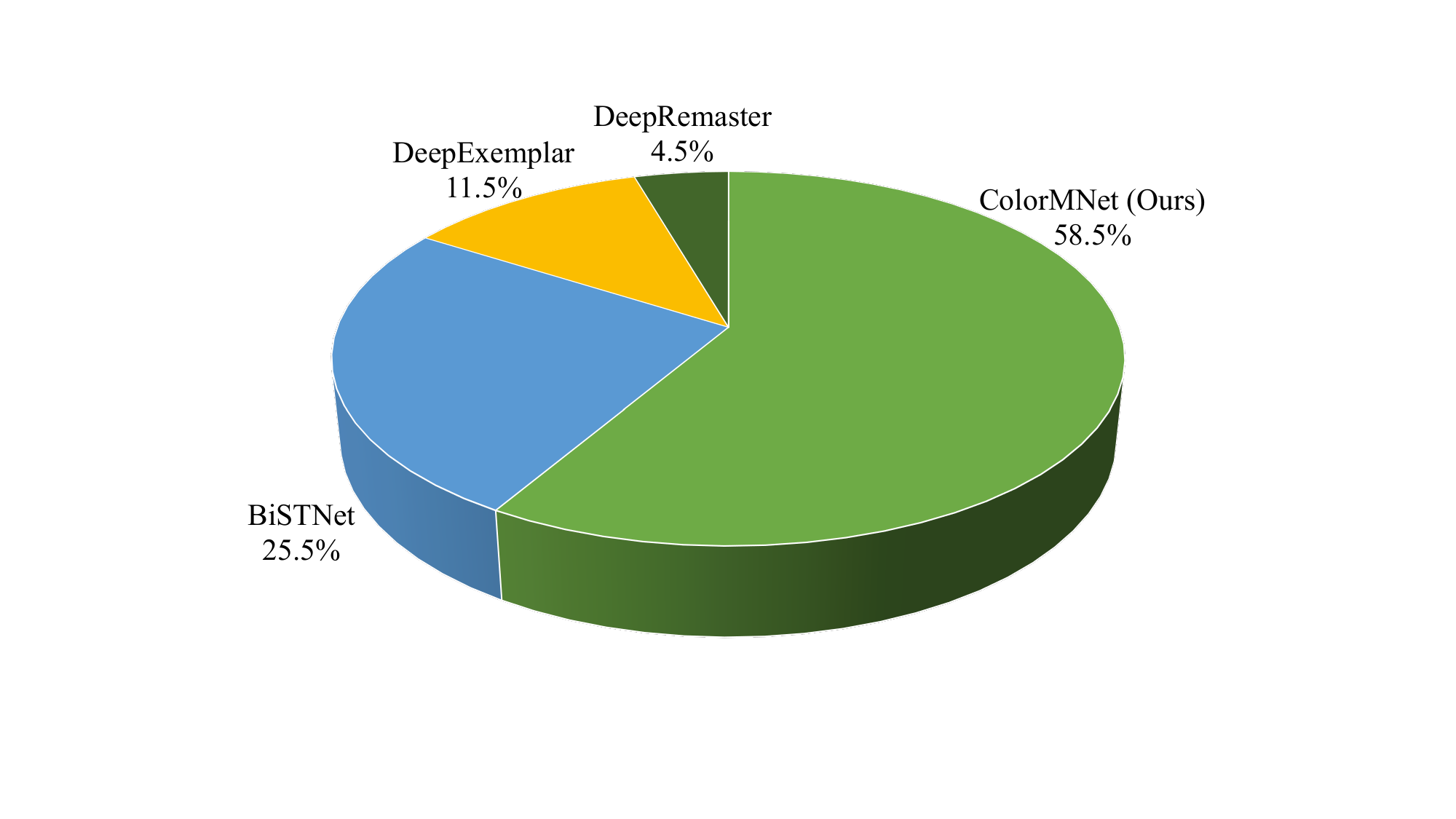}}}  & 
    \includegraphics[width=0.143\linewidth, height = 0.125\linewidth]{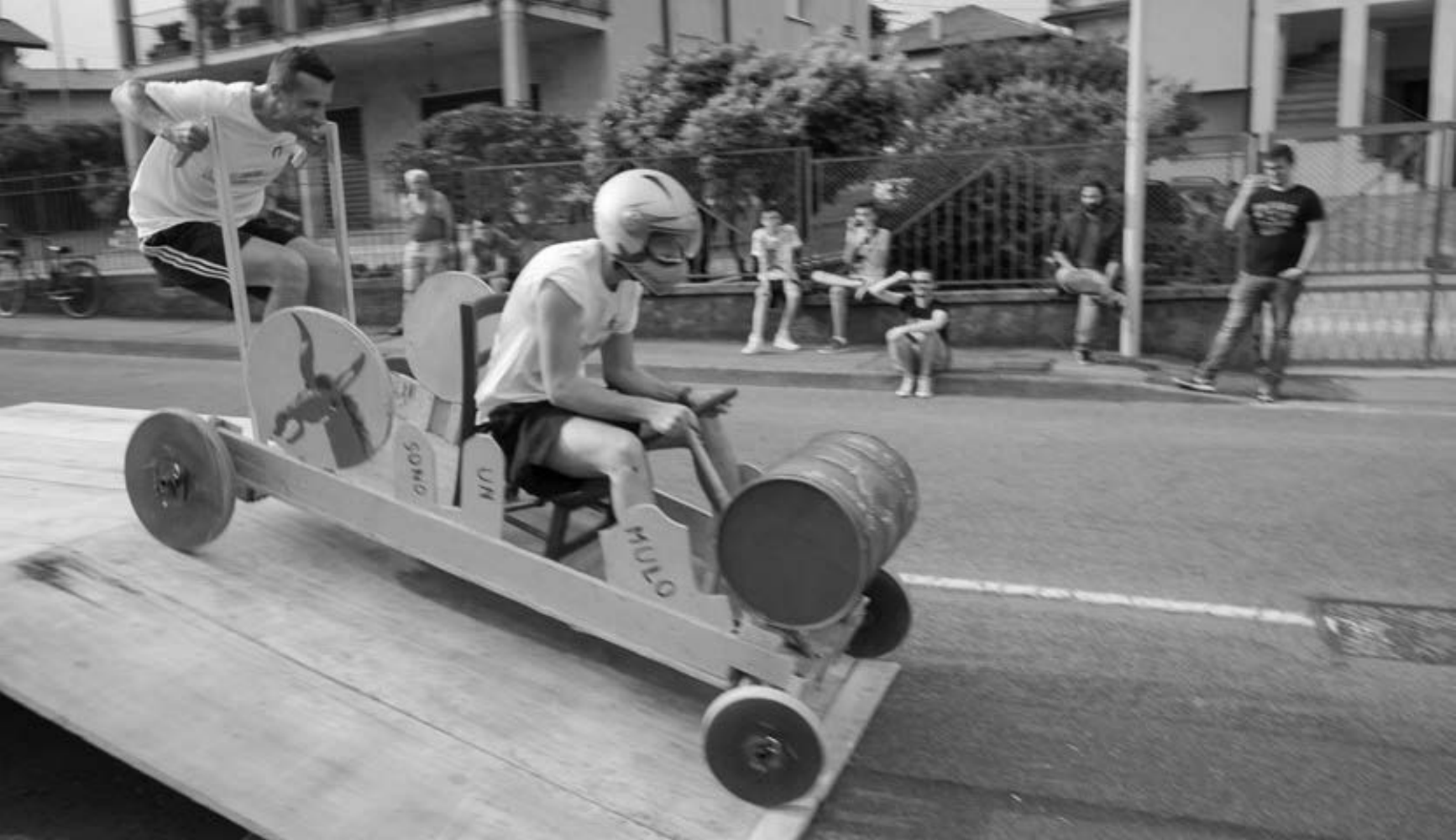} & 
     % \hspace{+2mm}
    {\includegraphics[width=0.143\linewidth, height = 0.125\linewidth]{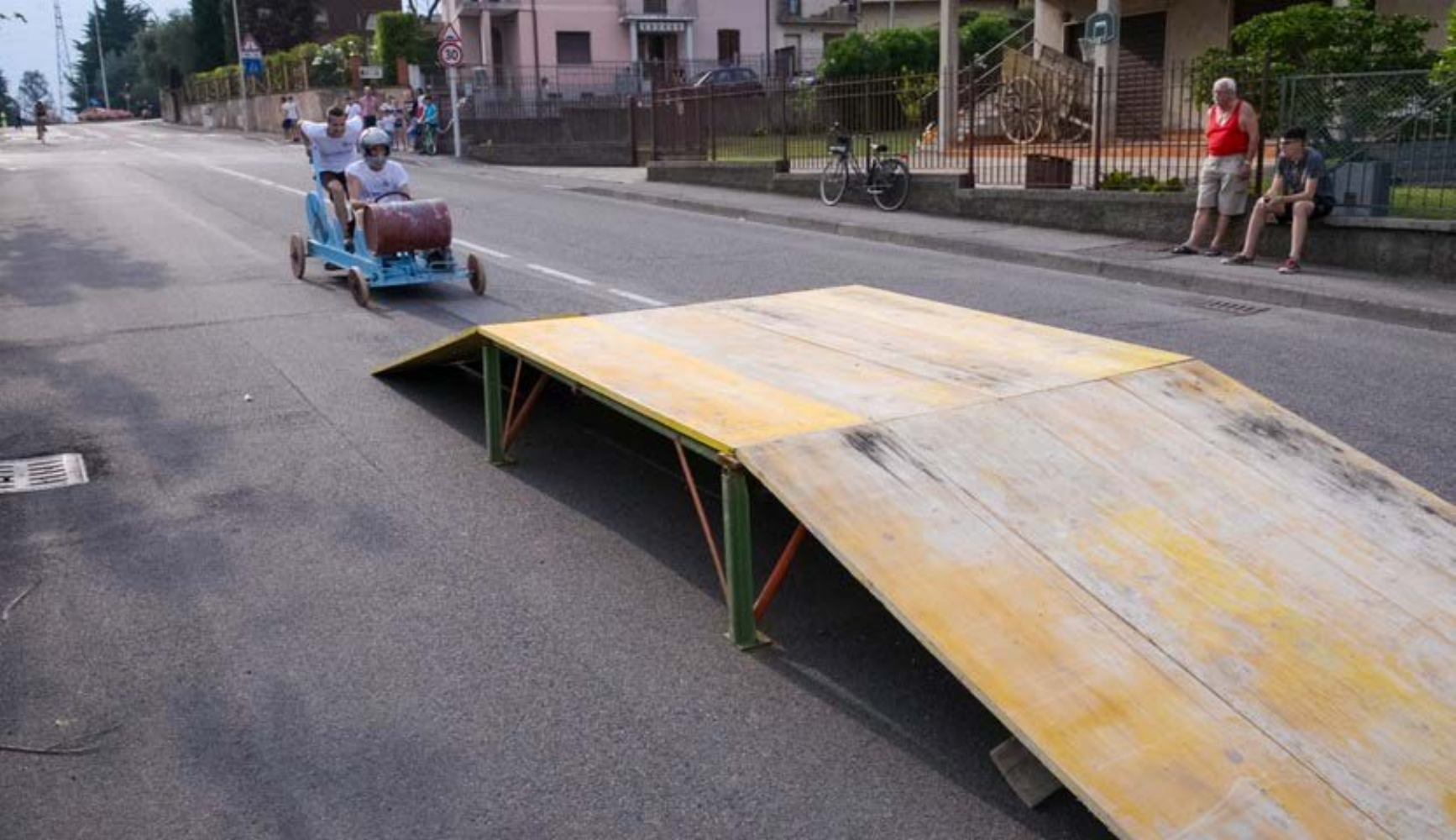}} & 
    {\includegraphics[width=0.143\linewidth, height = 0.125\linewidth]{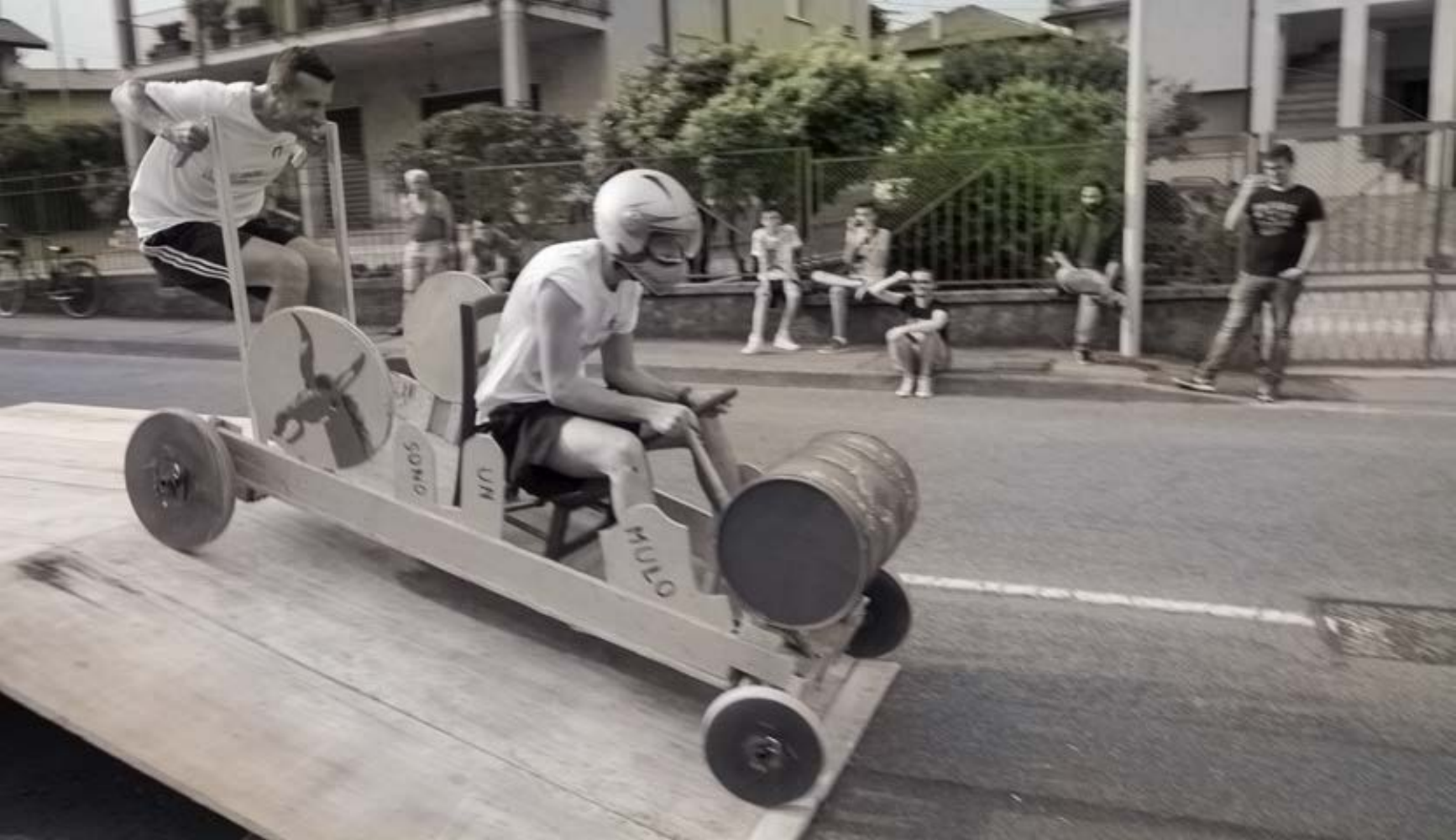}}
    \\
    \multicolumn{2}{c}{} &  \makebox[0.153\textwidth]{(b) Input video}  &  \makebox[0.153\textwidth]{(c) Exemplar} & 
    \makebox[0.153\textwidth]{(d)~\cite{IizukaSIGGRAPHASIA2019}}
    \\
    \multicolumn{2}{c}{} & \includegraphics[width=0.143\linewidth, height = 0.125\linewidth]{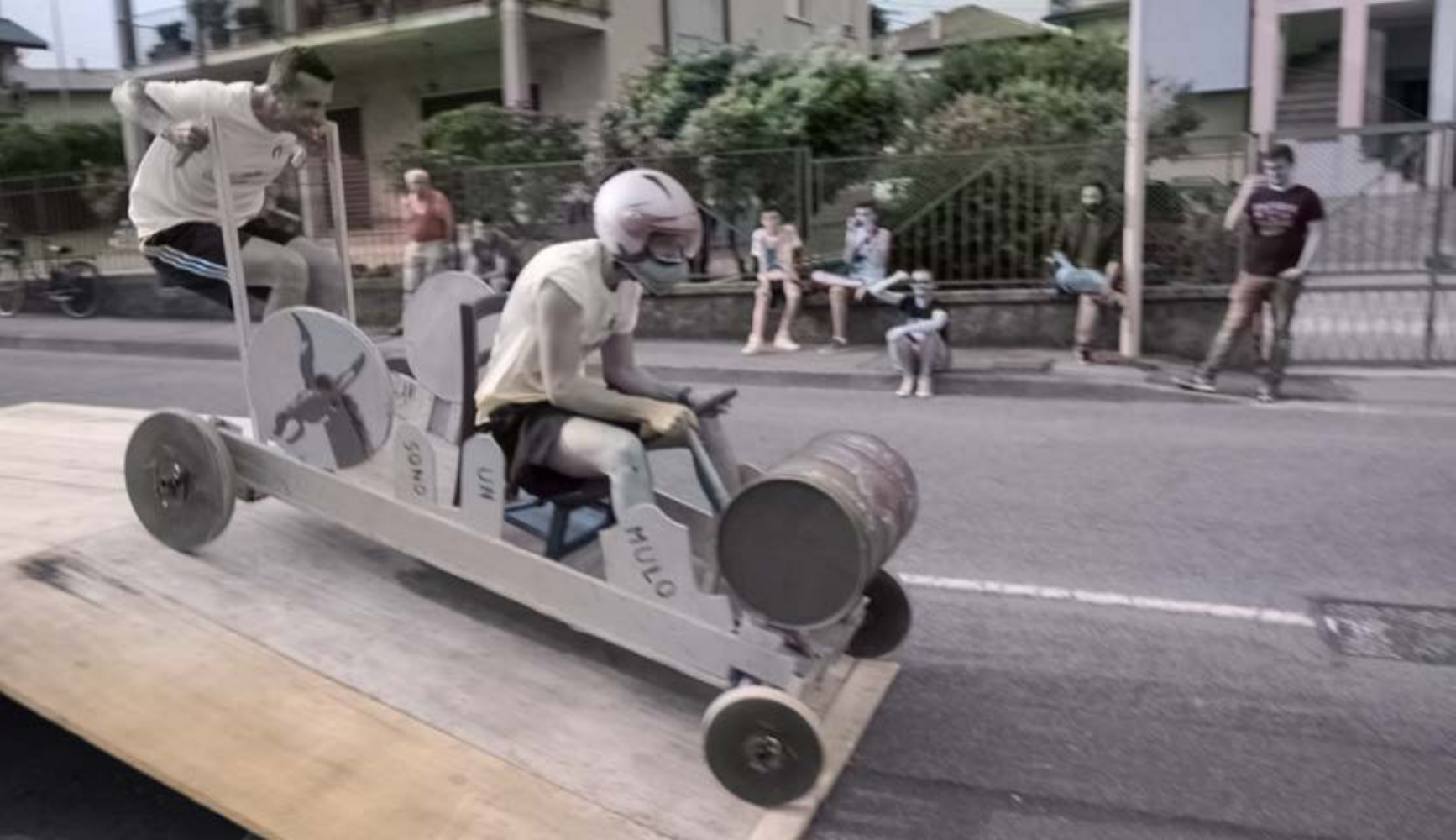} & 
    {\includegraphics[width=0.143\linewidth, height = 0.125\linewidth]{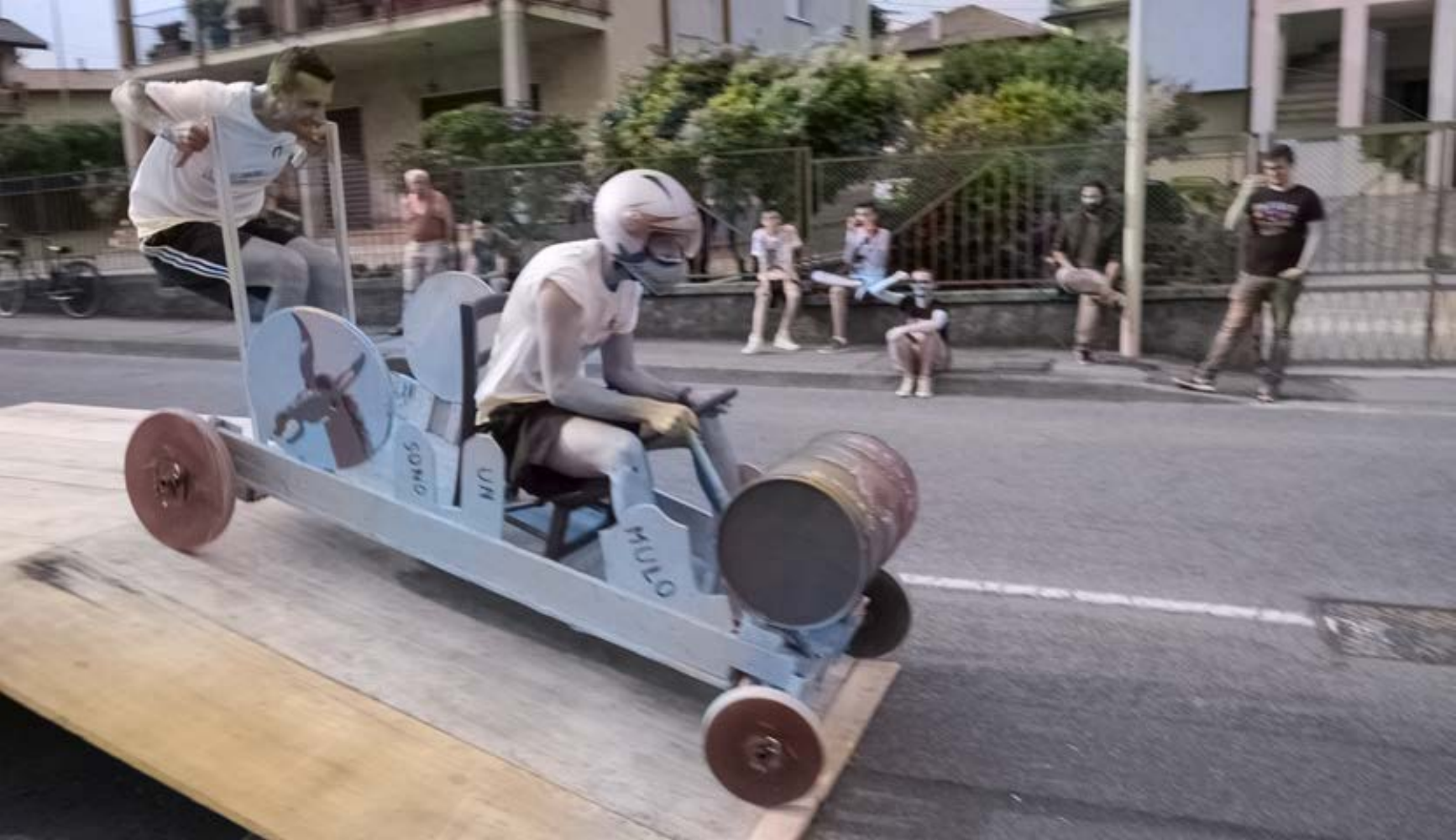}} & 
    {\includegraphics[width=0.143\linewidth, height = 0.125\linewidth]{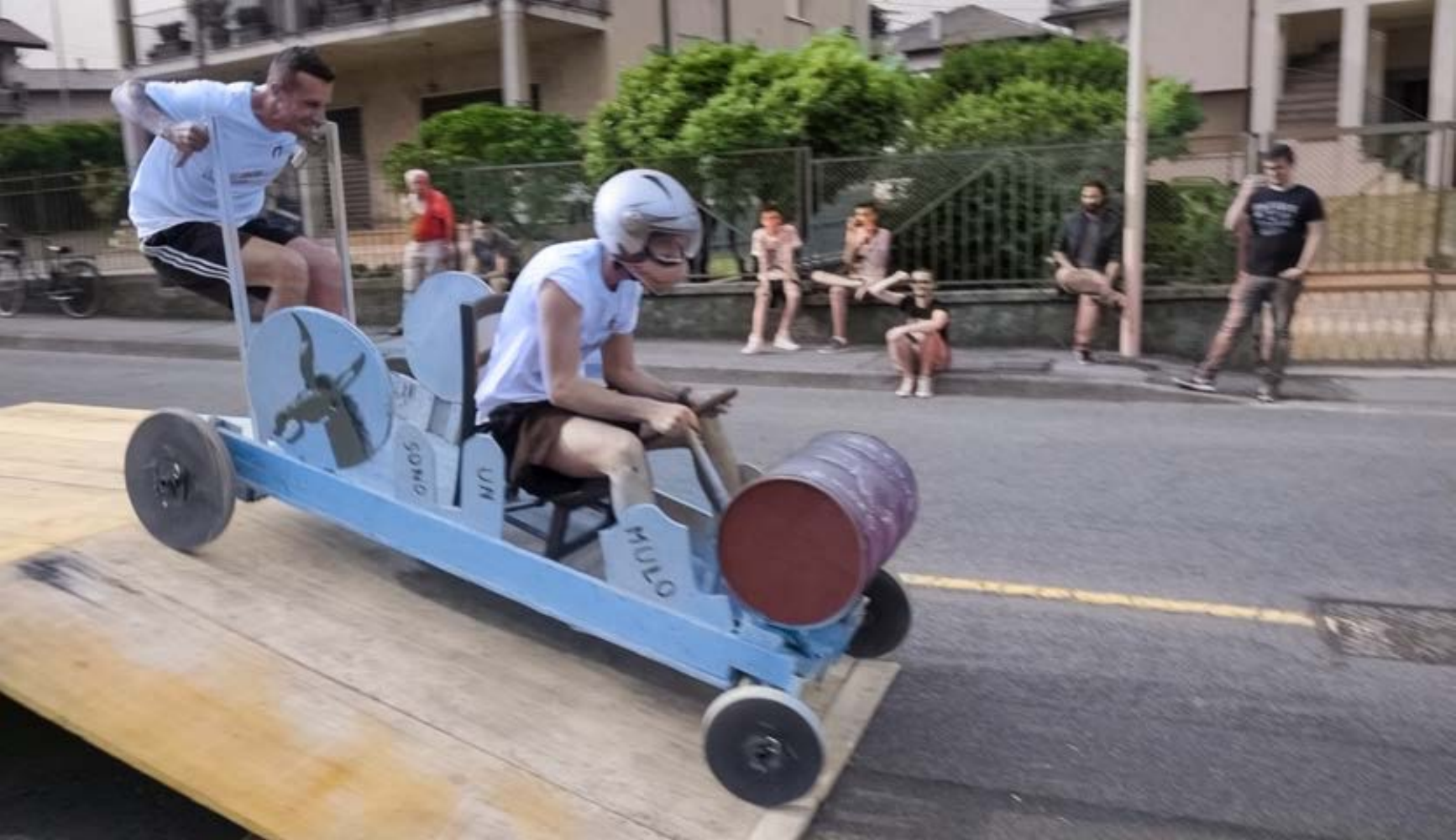}}
    \\
    \multicolumn{2}{c}{\makebox[0.5\textwidth]{(a) Averaged selection percentage of user study}} &  \makebox[0.153\textwidth]{(e)~\cite{zhang2019deep}}  & \makebox[0.153\textwidth]{(f)~\cite{bistnet}} & 
    \makebox[0.153\textwidth]{(g) Ours}
	\end{tabular}
	\vspace{-2mm}
	\caption{{User study result and an example of a group of results displayed to human observers in the user study. (a) shows that our proposed ColorMNet achieves obviously higher score than other state-of-the-art methods, which demonstrates its subjective advantages. (b)-(g) are the input video, the exemplar image, the colorized videos by DeepRemaster~\cite{IizukaSIGGRAPHASIA2019}, DeepExemplar~\cite{zhang2019deep}, BiSTNet$^\dagger$~\cite{bistnet} and ColorMNet (Ours), respectively. We make the methods anonymous and randomly sort the videos in (d)-(g) to ensure fairness. $^\dagger$ denotes that two exemplars are used.}}
	\label{fig:User_Study}
	\end{minipage}
	\vspace{-5mm}
\end{figure*}

\section{Effectiveness of the Local Attention Module}
\label{sec:ela}
As stated in Section 5 of the manuscript, we have analyzed the effectiveness of the local attention (LA) module.
We empirically set $\lambda = 7$ for $\lambda \times \lambda$ patch $\mathcal{N}(p)$.
In this supplemental material, we further show more visual comparisons to demonstrate the effectiveness of the LA module. In \textit{`ComparisonsWithSOTA.mp4'}, we show that our proposed ColorMNet with using the LA is able to generate better-colorized videos.

\section{User Study}
\label{sec:User_Study}
To evaluate whether our results are favored by human observers, we further conduct user study experiments. 
Specifically, we compare our method with exemplar-based methods, i.e., BiSTNet~\cite{bistnet}, DeepExemplar~\cite{zhang2019deep} and DeepRemaster~\cite{IizukaSIGGRAPHASIA2019}.
We randomly select 10 input videos from the DAVIS~\cite{Perazzi_CVPR_2016} validation set, the Videvo~\cite{Lai2018videvo} validation set and the NVCC2023~\cite{Kang_2023_CVPR} validation set together with the colorization results and the exemplar images displayed to 20 online observers without constraints. 
We make the methods anonymous and randomly sort the videos in each group to ensure fairness.
Observers are asked to choose the most visually pleasing results from a group of videos. 
Figure~\ref{fig:User_Study} shows that our method is preferred by a wider range of users than other state-of-the-art methods.

\section{More Experimental Results}
\label{sec:More_Experimental_Results}
\begin{figure*}[!htb]
	\setlength\tabcolsep{1.0pt}
	\centering
	\small
	\begin{tabularx}{1\textwidth}{cc}
        \includegraphics[width=0.495\textwidth, height = 0.28\textwidth] {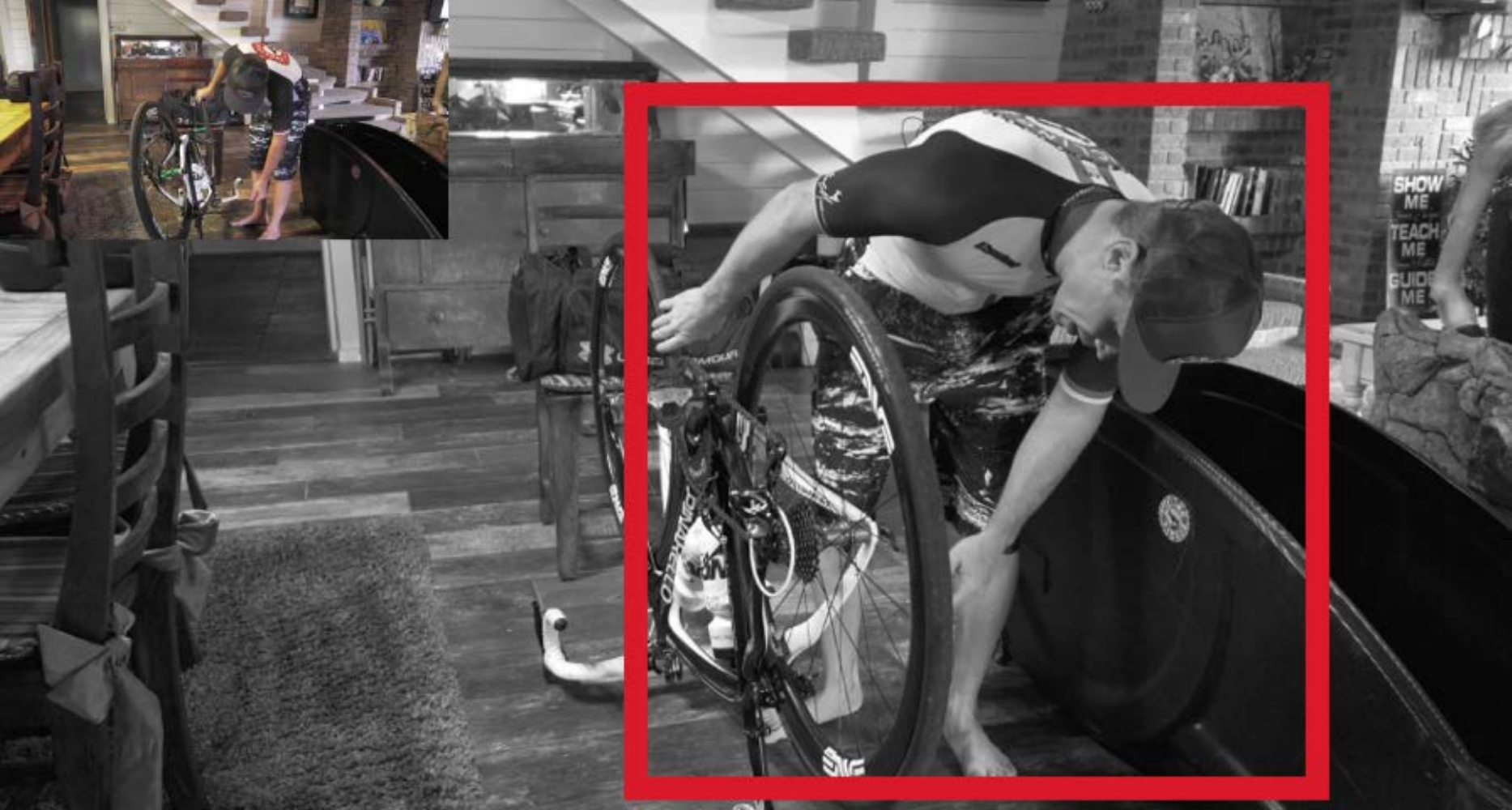} &
        \includegraphics[width=0.495\textwidth, height = 0.28\textwidth] {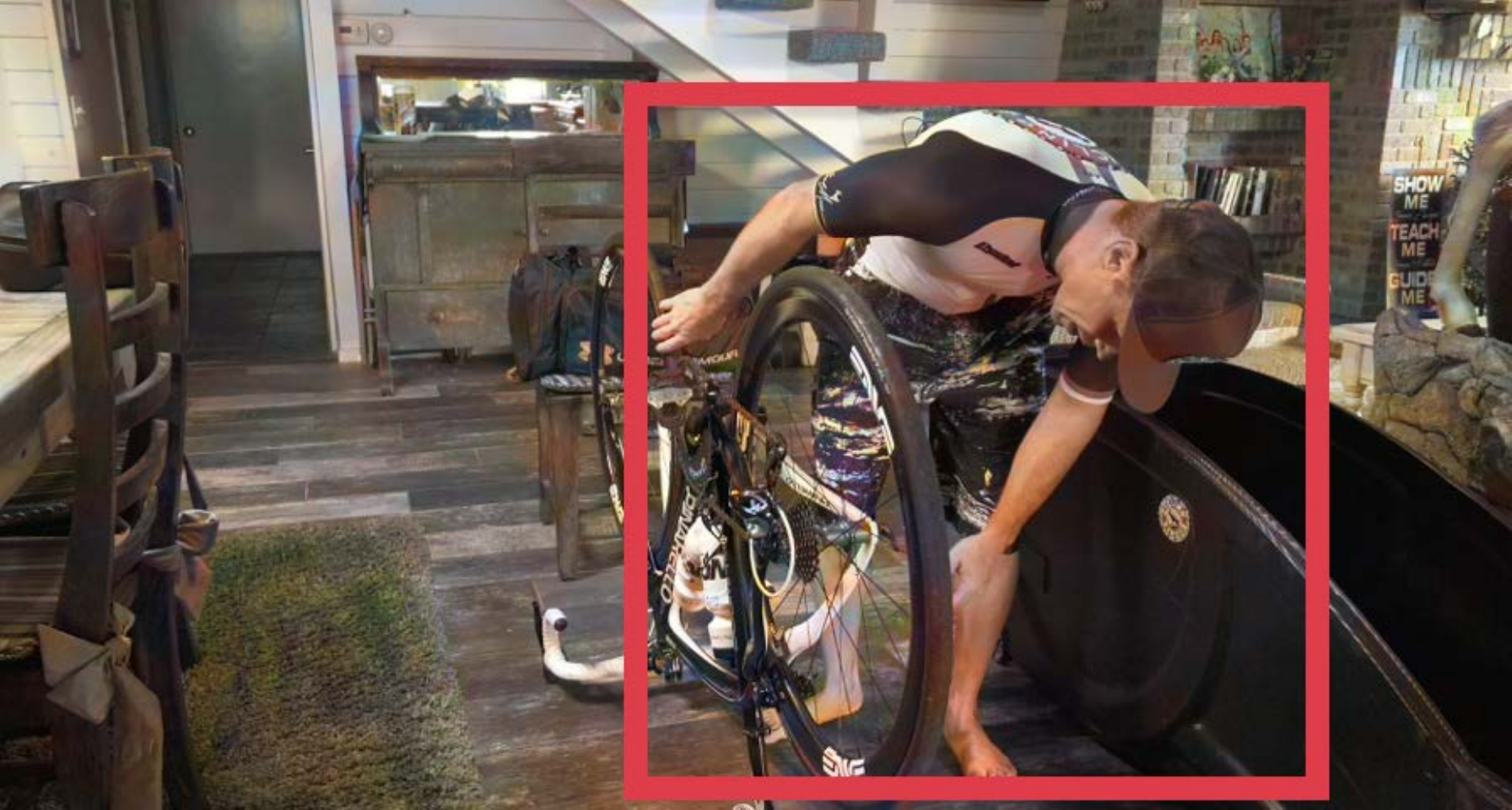} 
        \\
        \makebox[0.45\textwidth]{ (a) Input frame and exemplar image} &
        \makebox[0.45\textwidth]{ (b) DDColor~\cite{kang2022ddcolor}} 
        \\ 
        \includegraphics[width=0.495\textwidth, height = 0.28\textwidth] {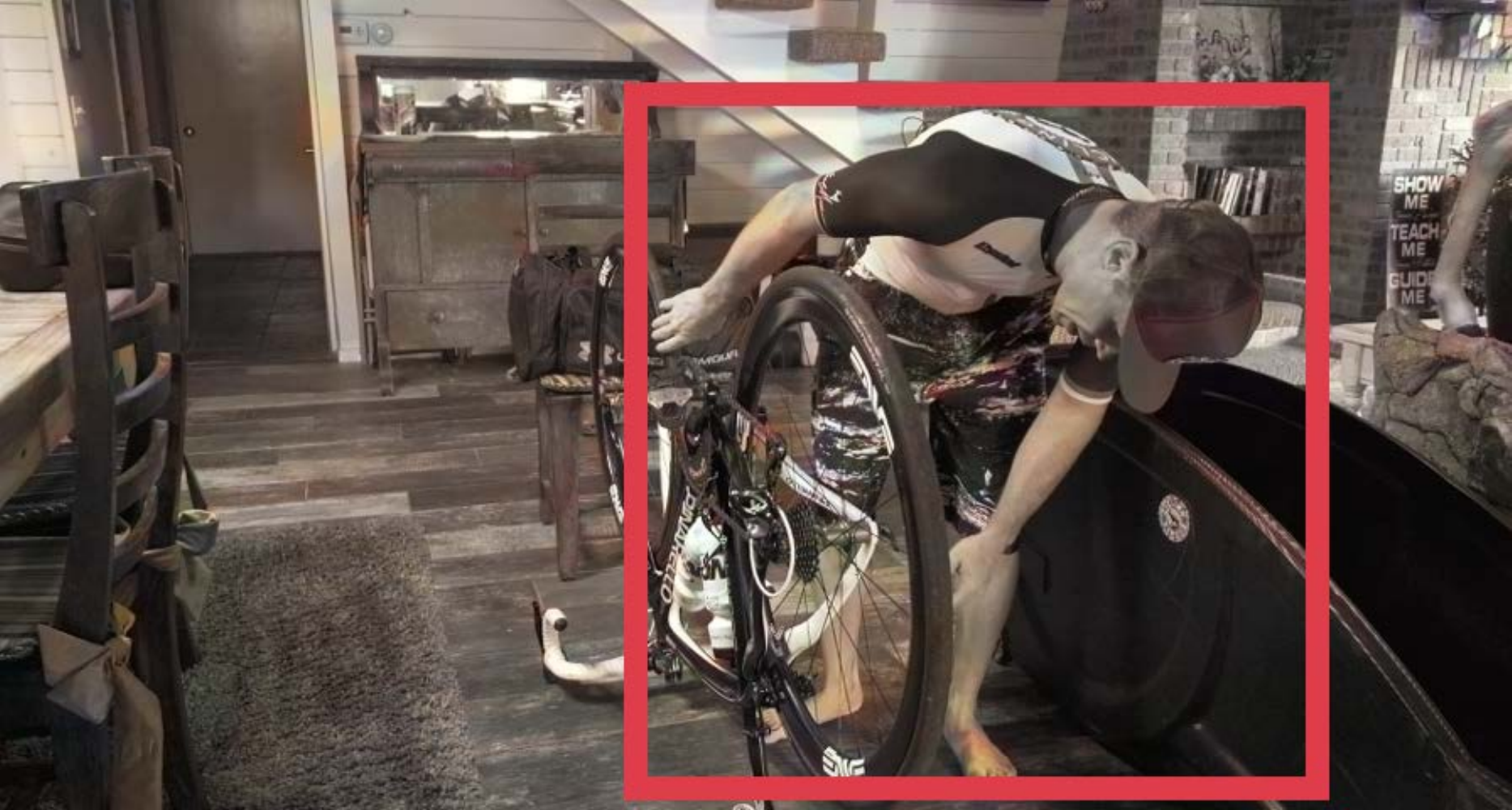} &
        \includegraphics[width=0.495\textwidth, height = 0.28\textwidth] {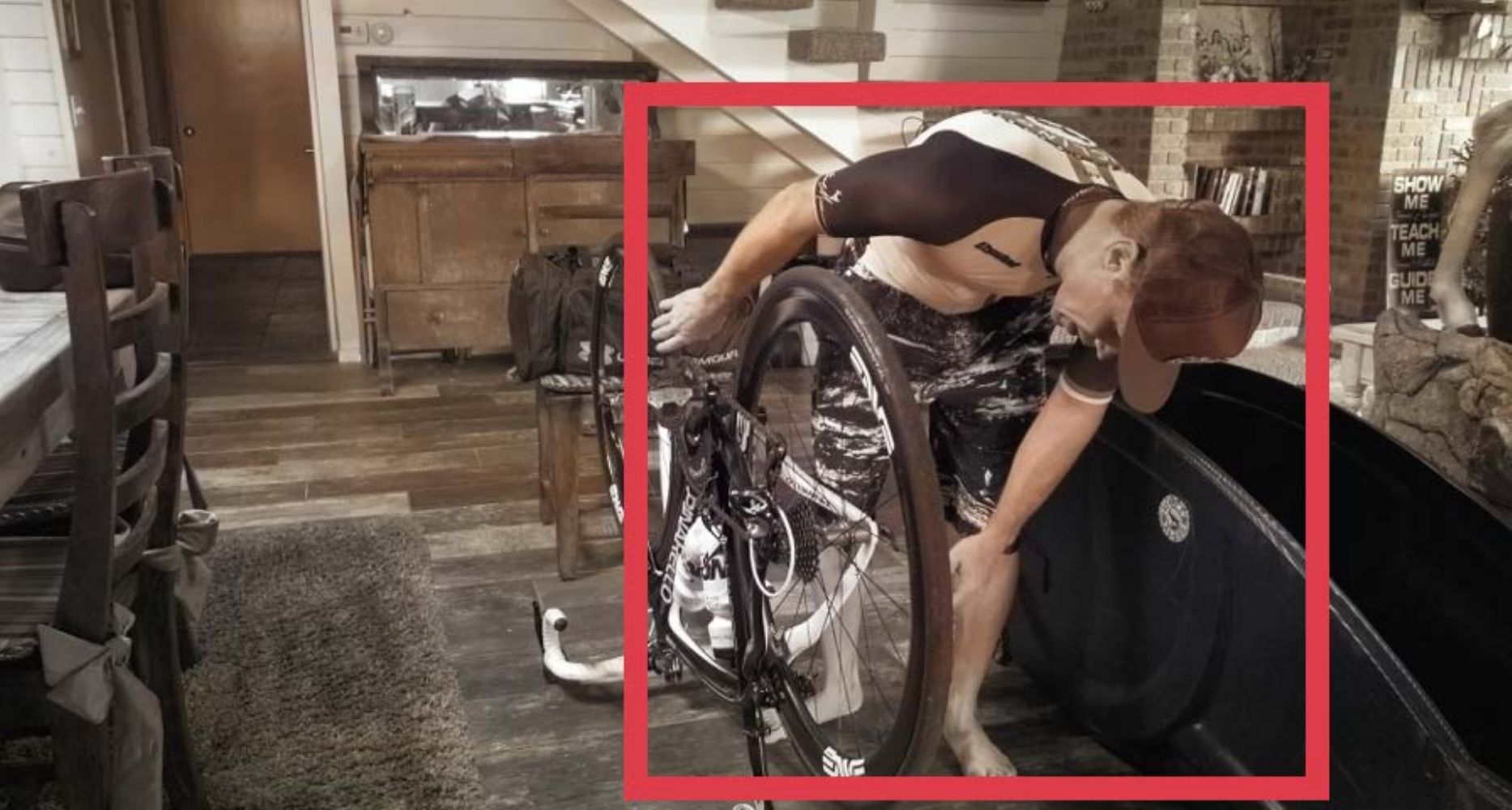} 
        \\ 
        \makebox[0.45\textwidth]{ (c) Color2Embed~\cite{zhao2021color2embed}} &
        \makebox[0.45\textwidth]{ (d) TCVC~\cite{liu2021temporally}} \\
        \includegraphics[width=0.495\textwidth, height = 0.28\textwidth] {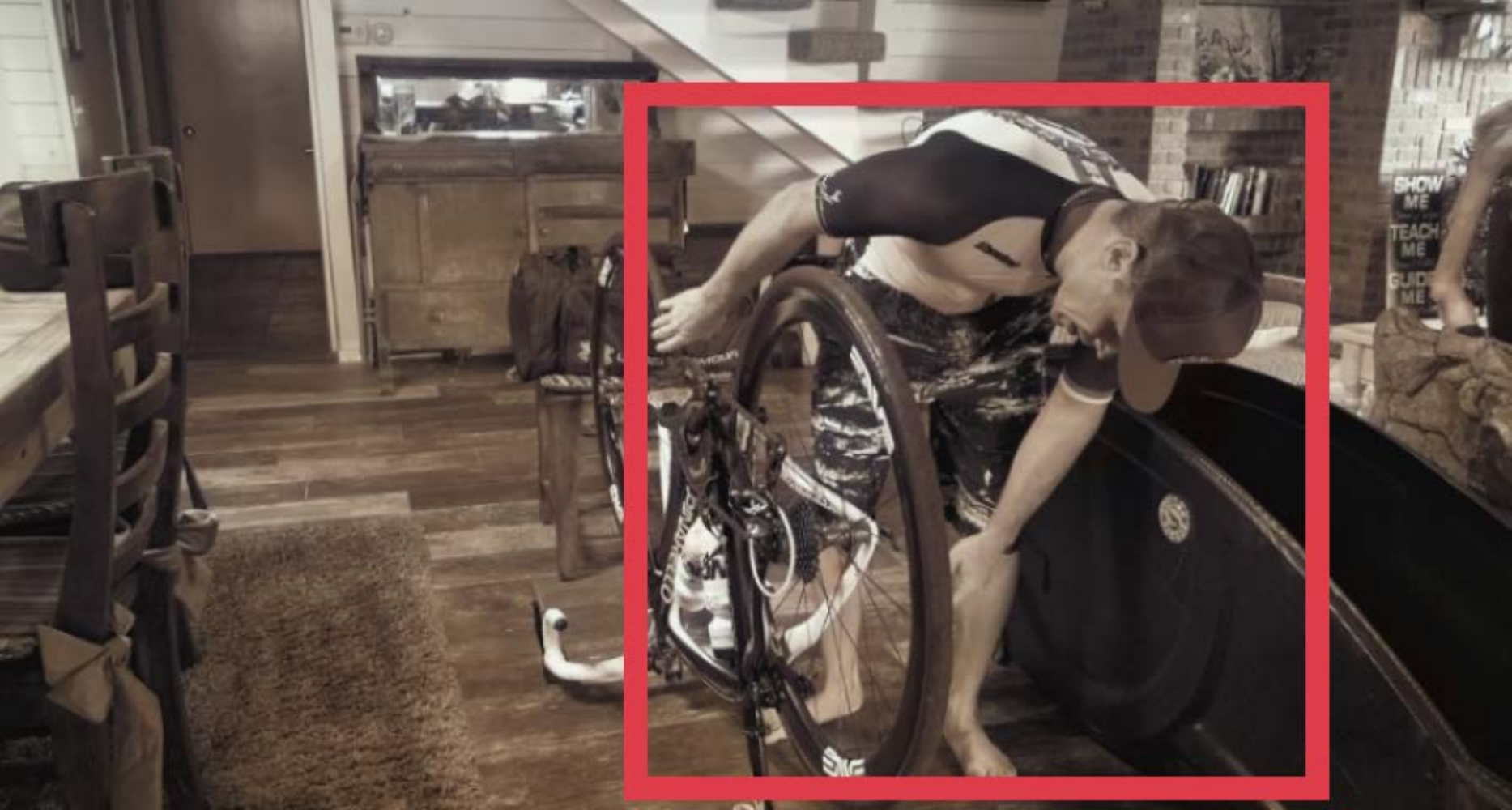} &
        \includegraphics[width=0.495\textwidth, height = 0.28\textwidth] {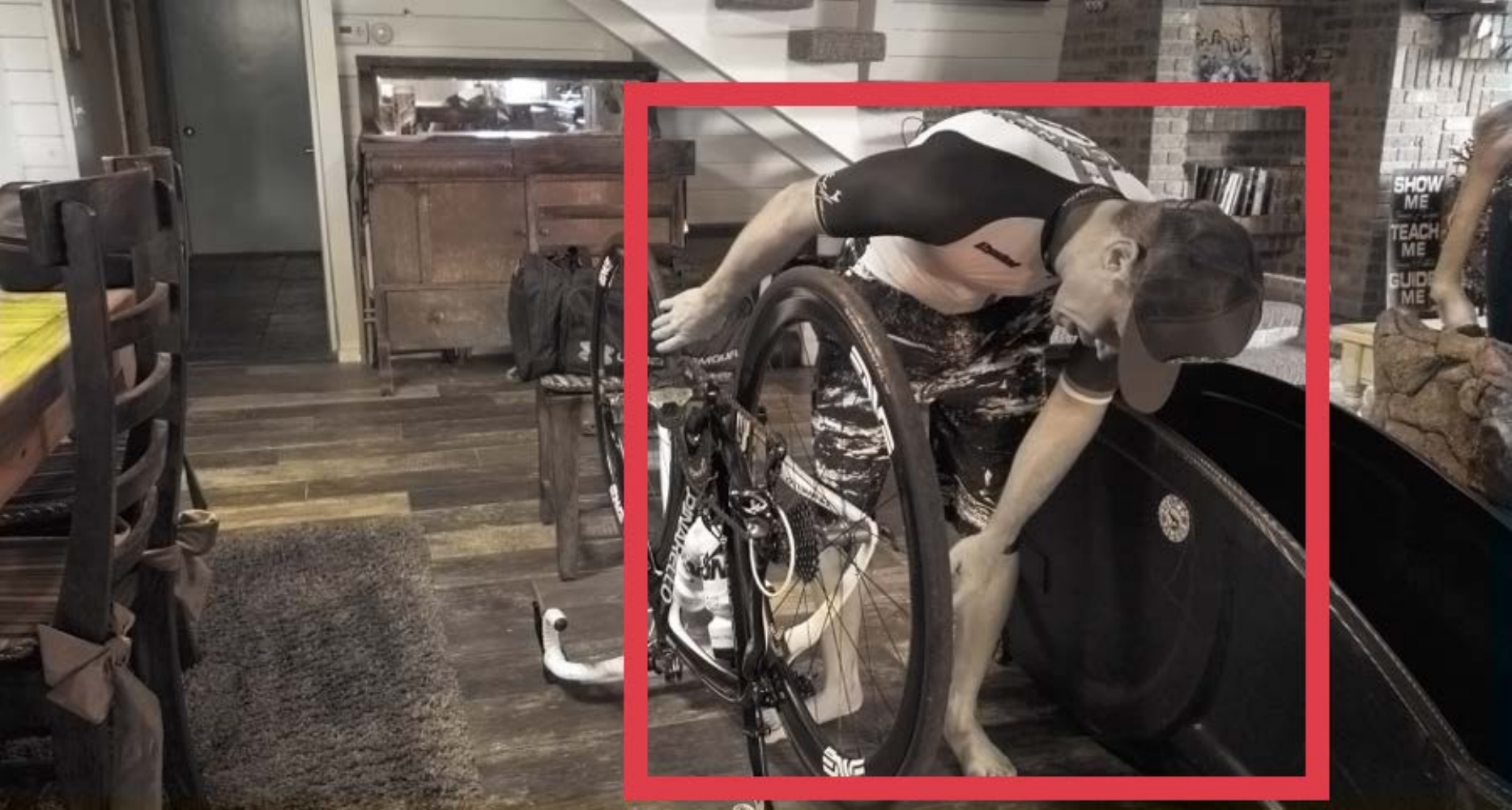} 
        \\
        \makebox[0.45\textwidth]{ (e) VCGAN~\cite{vcgan}} &
        \makebox[0.45\textwidth]{ (f) DeepRemaster~\cite{IizukaSIGGRAPHASIA2019}} 
        \\ 
        \includegraphics[width=0.495\textwidth, height = 0.28\textwidth] {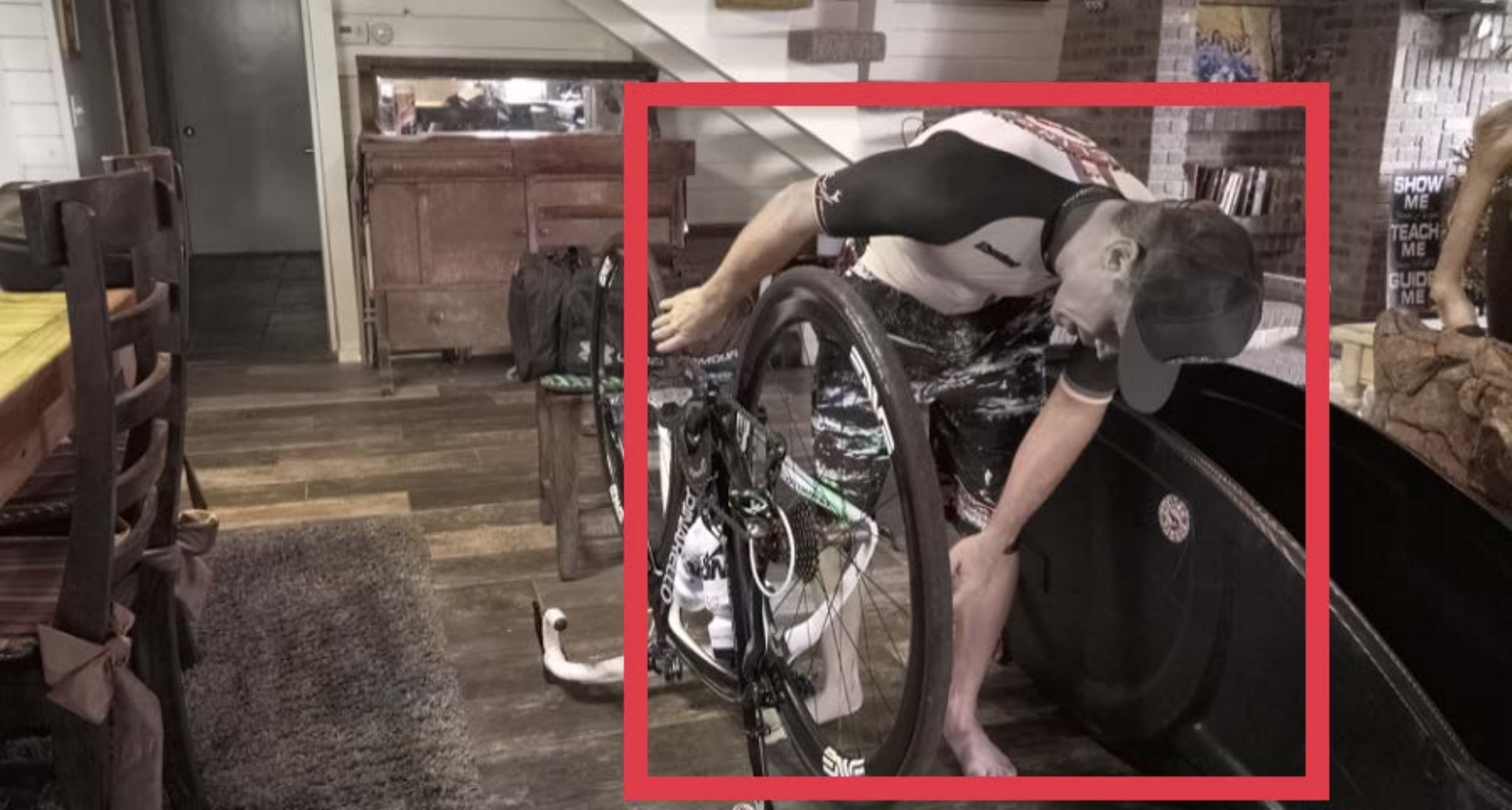} &
        \includegraphics[width=0.495\textwidth, height = 0.28\textwidth] {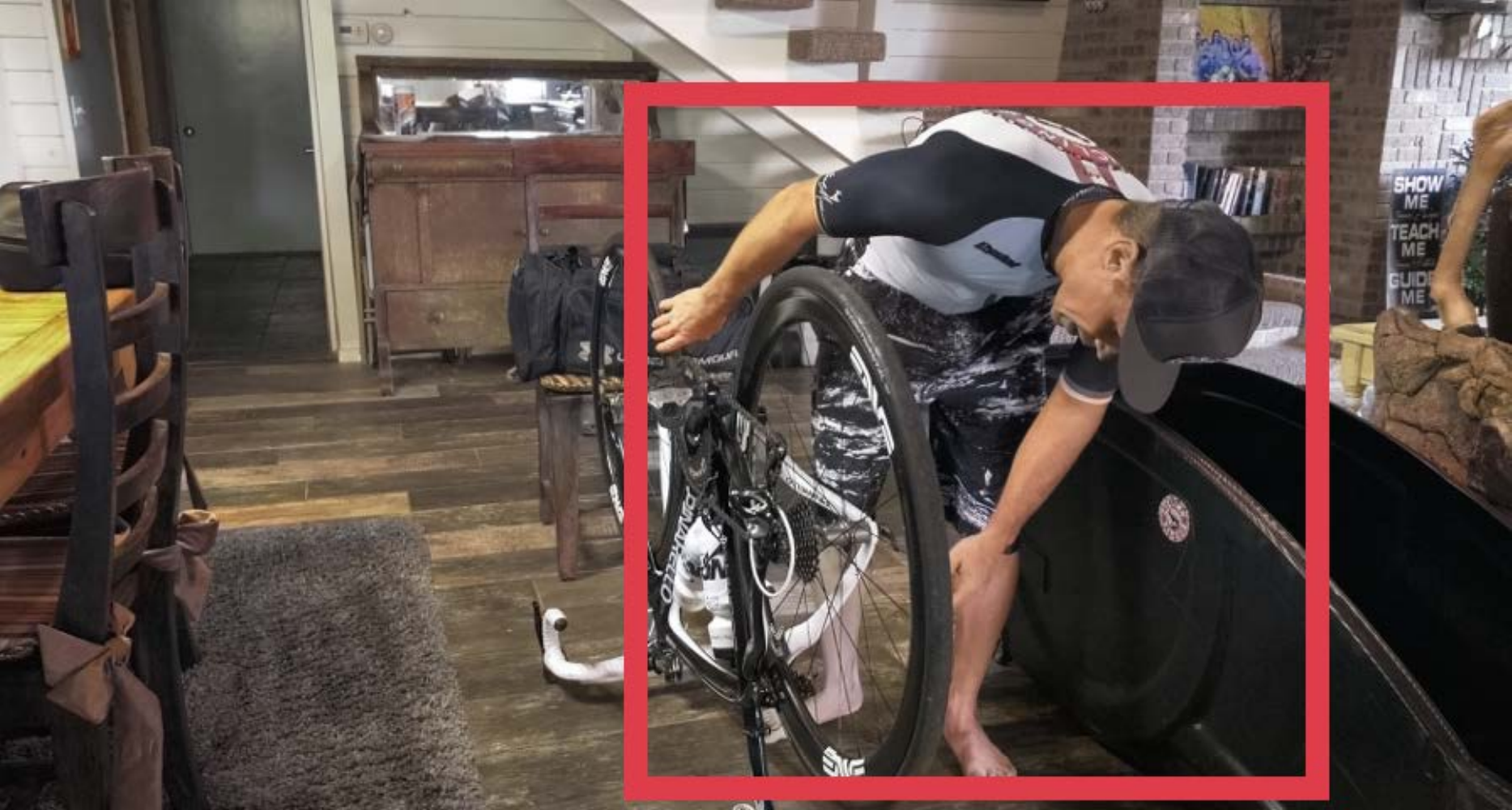} 
        \\
        \makebox[0.45\textwidth]{ (g) DeepExemplar~\cite{zhang2019deep}} &
        \makebox[0.45\textwidth]{ (h) ColorMNet (Ours)}  
         %& \vspace{-0.7em}
	\end{tabularx}
	\vspace{-0.8em}
	\caption{{Colorization results on clip \textit{bike-packing} from the DAVIS validation dataset~\cite{Perazzi_CVPR_2016}. The results shown in (b) and (c) still contain significant color-bleeding artifacts.~\cite{zhang2019deep, liu2021temporally, vcgan, IizukaSIGGRAPHASIA2019} do not recover the man well. In contrast, our proposed method generates a better-colorized frame, where the man is restored well and the colors look better.}}
	\label{fig:1}
	%\vspace{-3mm}
\end{figure*}

\begin{figure*}[!htb]
	\setlength\tabcolsep{1.0pt}
	\centering
	\small
	\begin{tabularx}{1\textwidth}{cc}
        \includegraphics[width=0.495\textwidth, height = 0.28\textwidth] {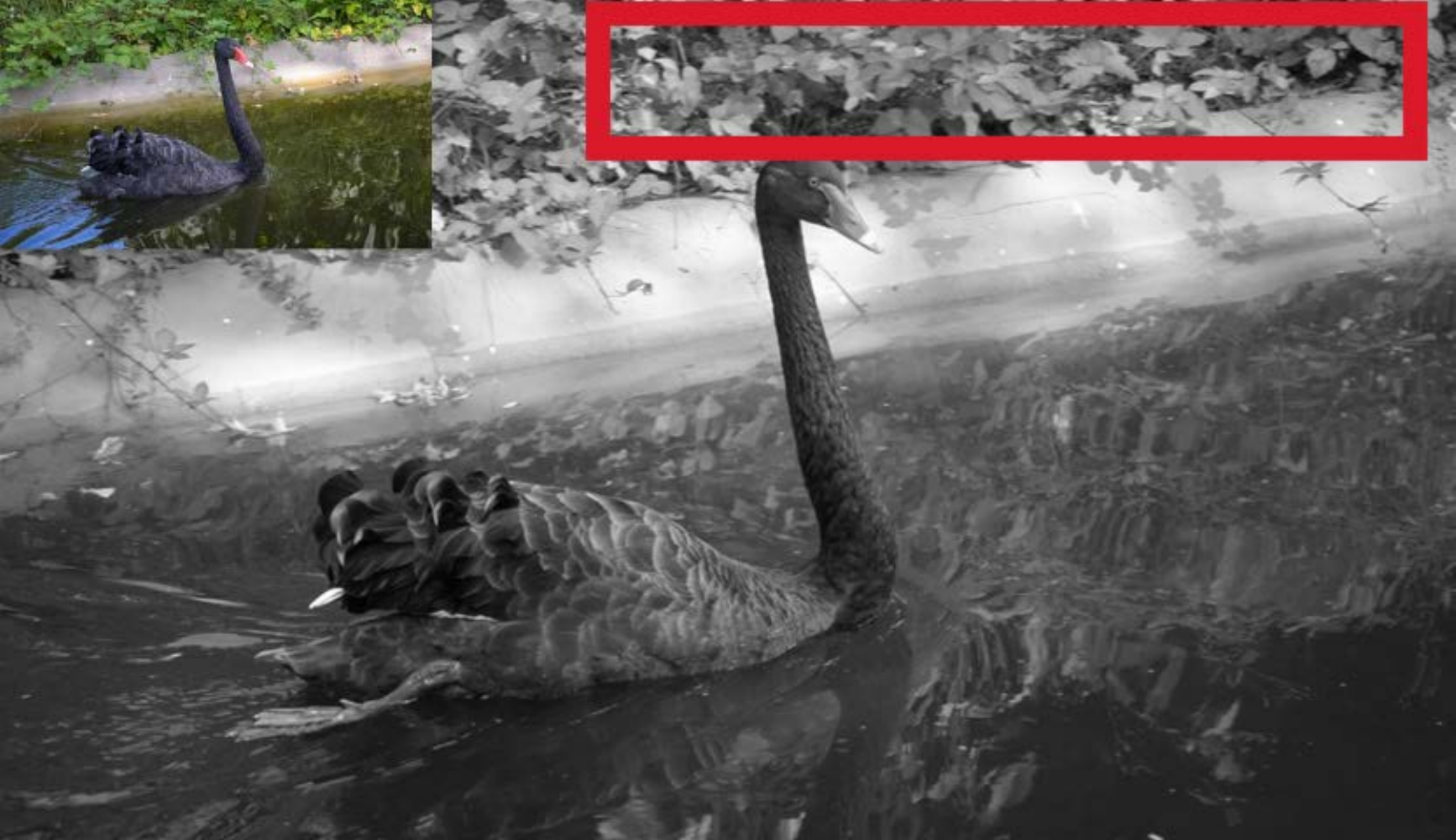} &
        \includegraphics[width=0.495\textwidth, height = 0.28\textwidth] {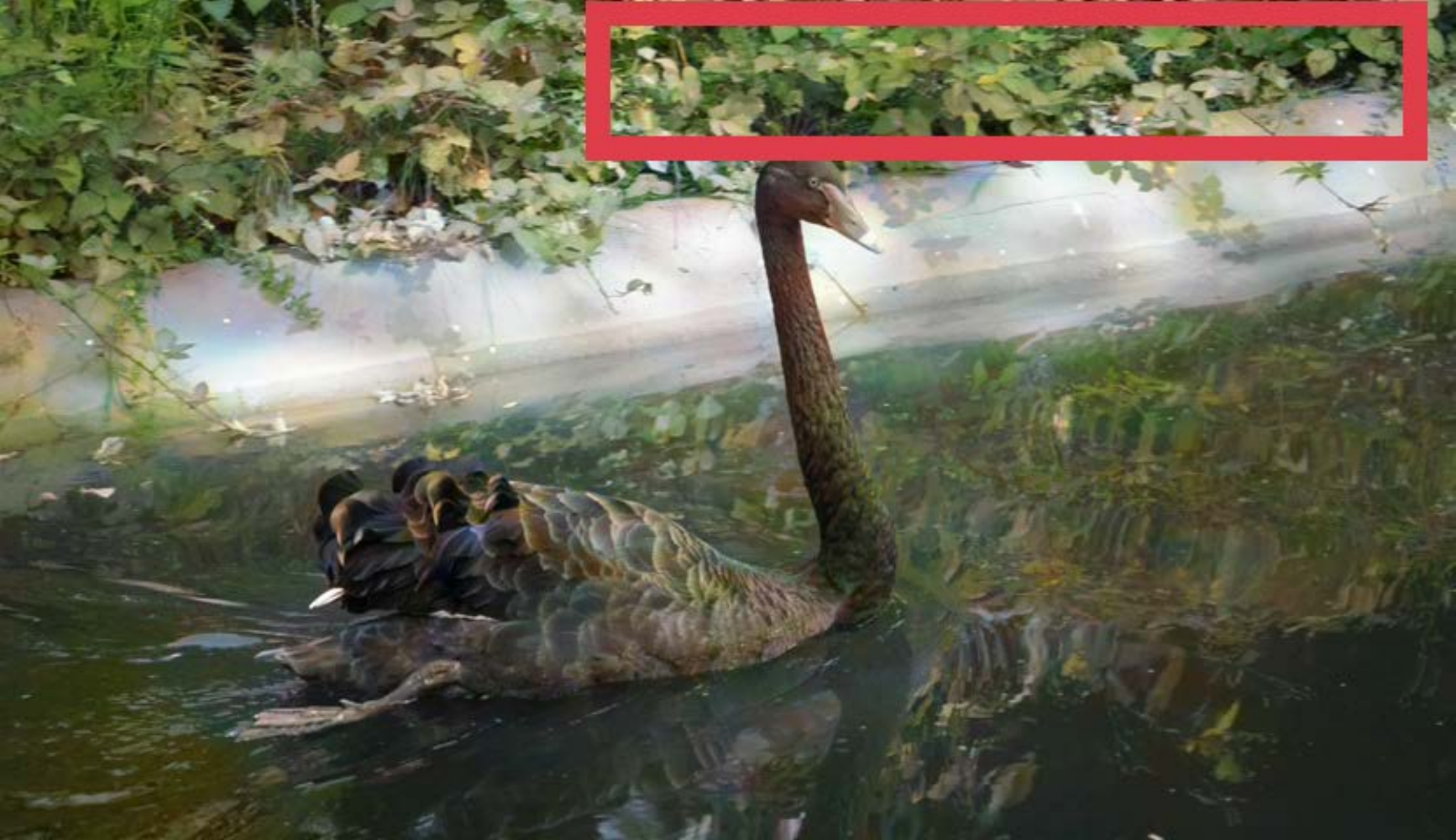} 
        \\
        \makebox[0.45\textwidth]{ (a) Input frame and exemplar image} &
        \makebox[0.45\textwidth]{ (b) DDColor~\cite{kang2022ddcolor}} 
        \\ 
        \includegraphics[width=0.495\textwidth, height = 0.28\textwidth] {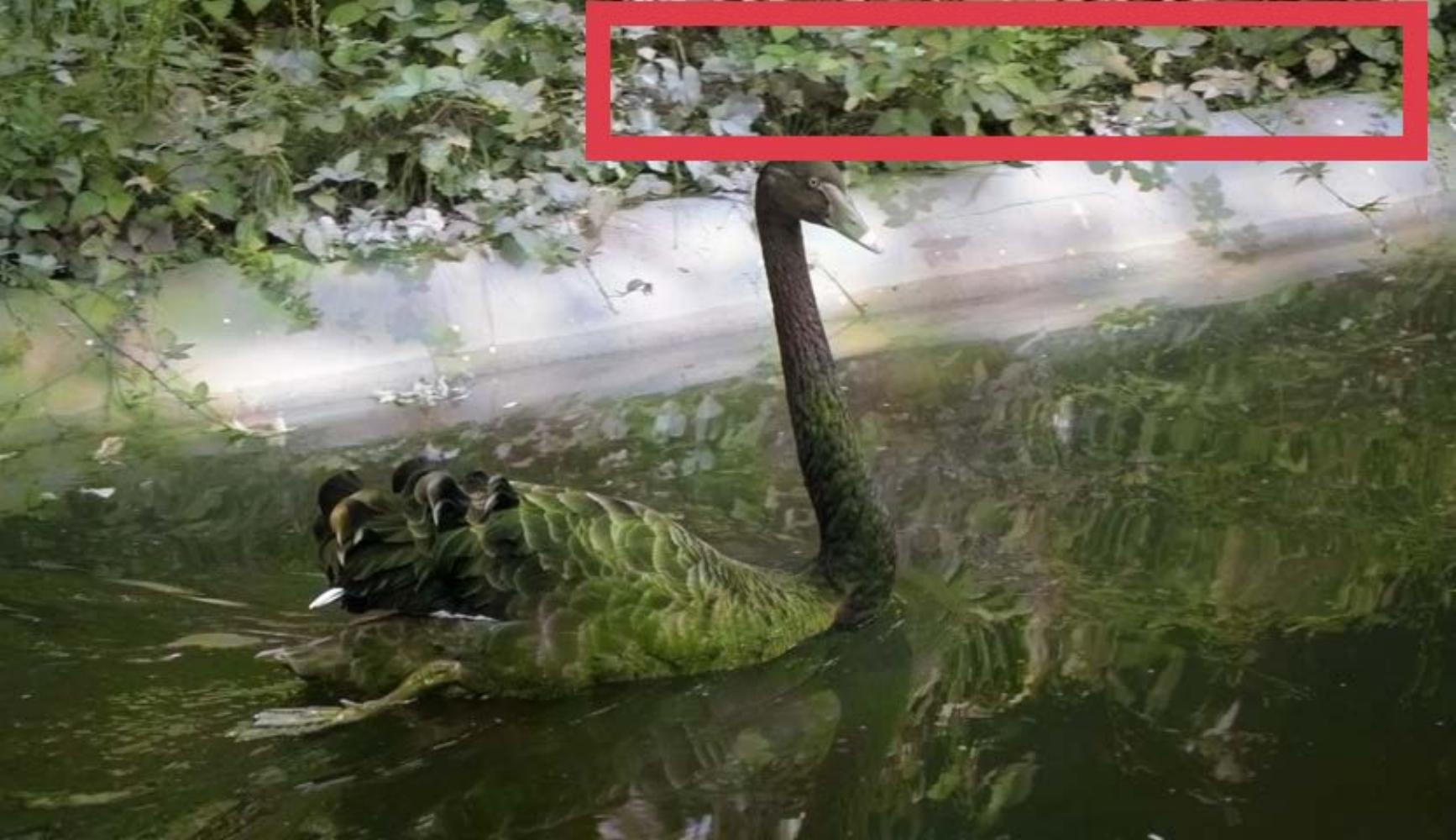} &
        \includegraphics[width=0.495\textwidth, height = 0.28\textwidth] {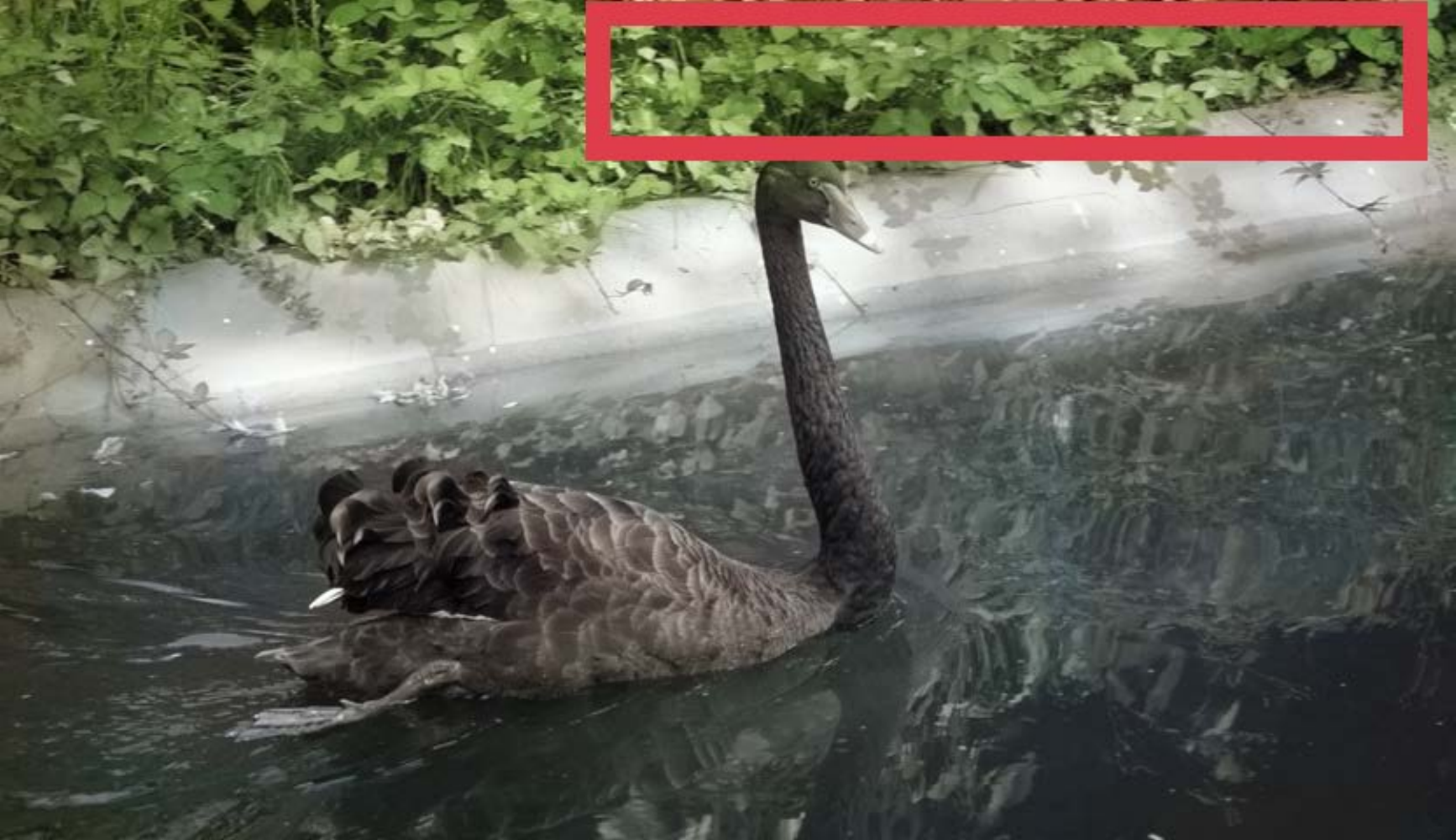} 
        \\ 
        \makebox[0.45\textwidth]{ (c) Color2Embed~\cite{zhao2021color2embed}} &
        \makebox[0.45\textwidth]{ (d) TCVC~\cite{liu2021temporally}} \\
        \includegraphics[width=0.495\textwidth, height = 0.28\textwidth] {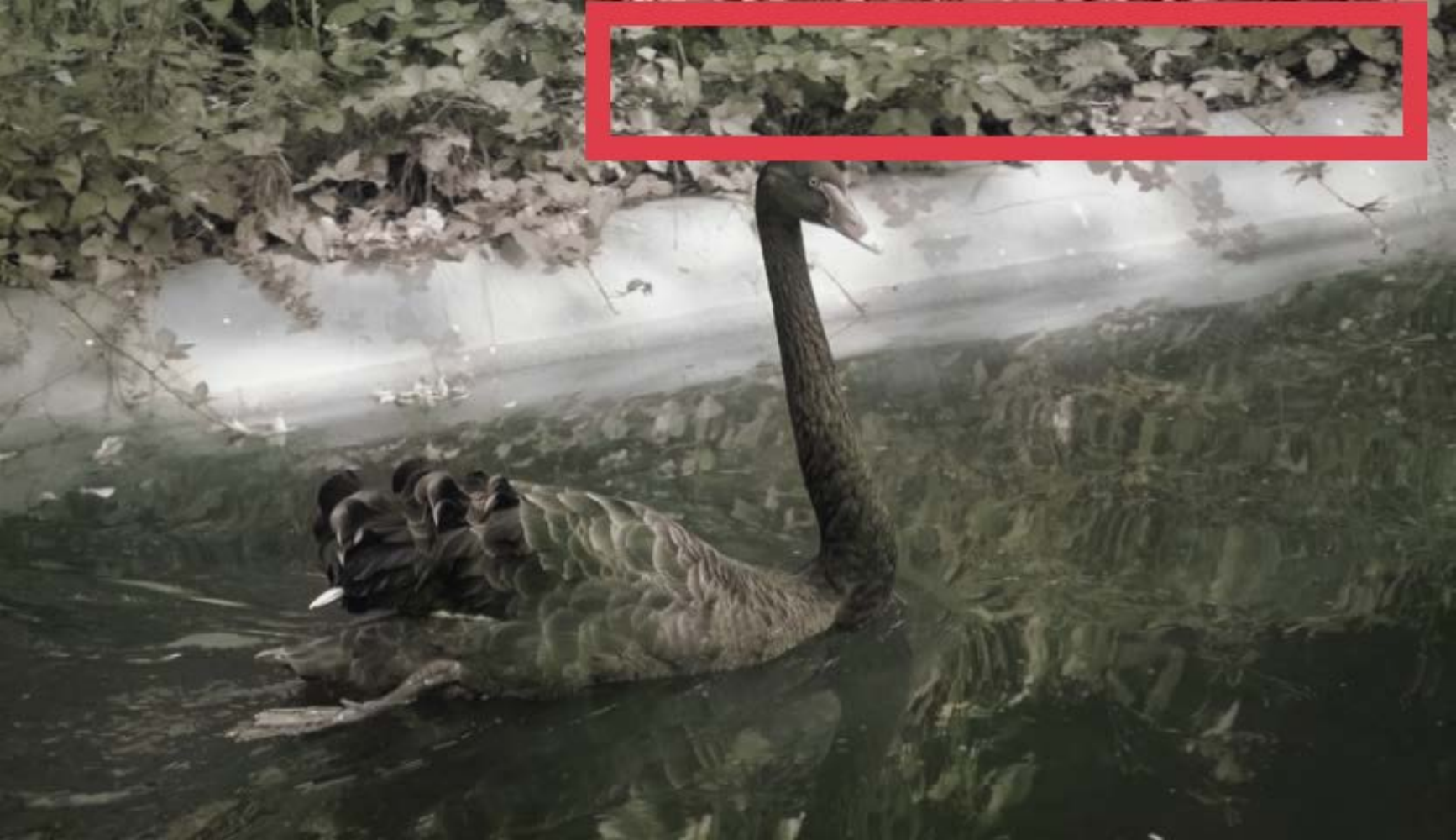} &
        \includegraphics[width=0.495\textwidth, height = 0.28\textwidth] {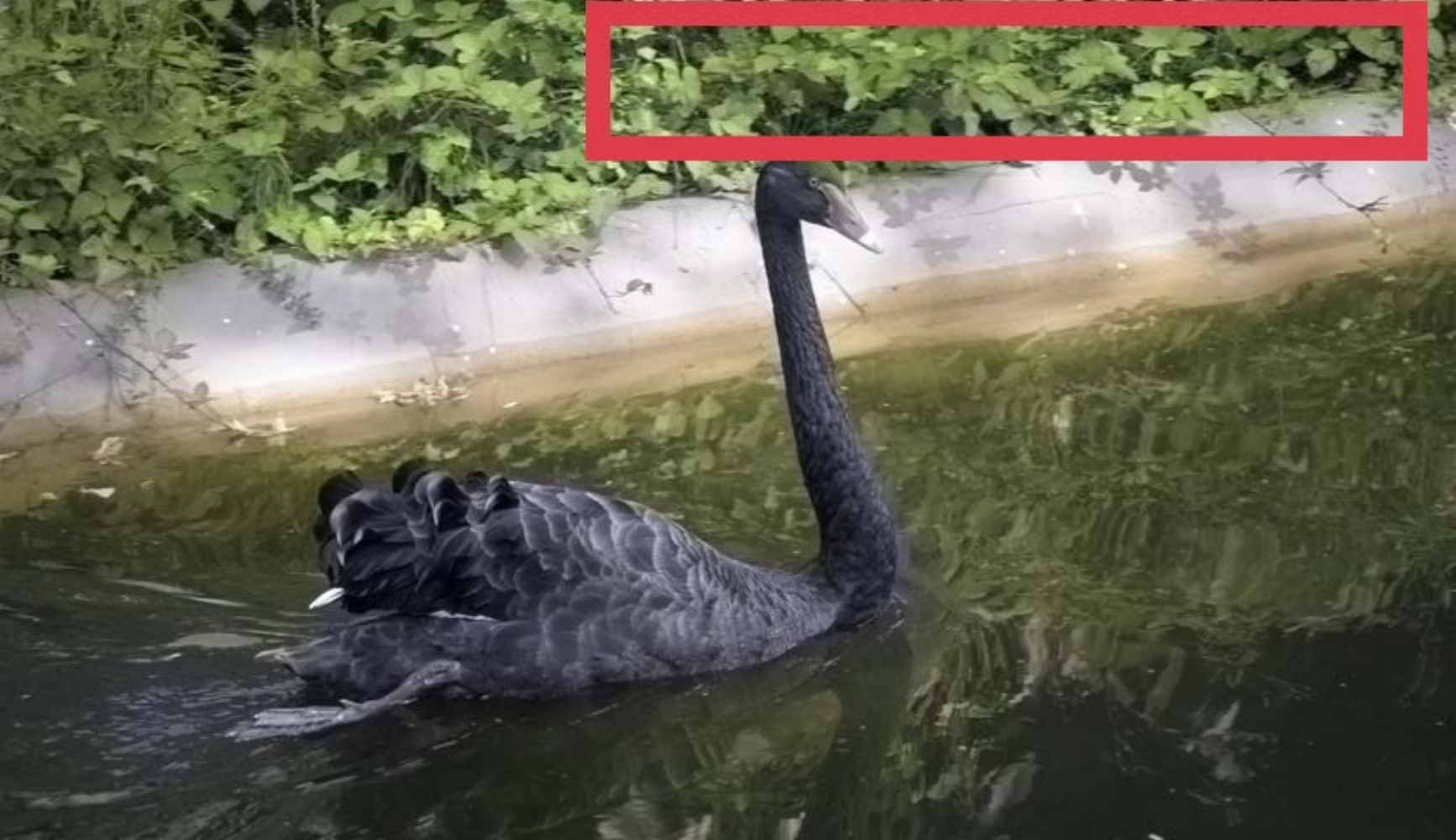} 
        \\
        \makebox[0.45\textwidth]{ (e) VCGAN~\cite{vcgan}} &
        \makebox[0.45\textwidth]{ (f) DeepRemaster~\cite{IizukaSIGGRAPHASIA2019}} 
        \\ 
        \includegraphics[width=0.495\textwidth, height = 0.28\textwidth] {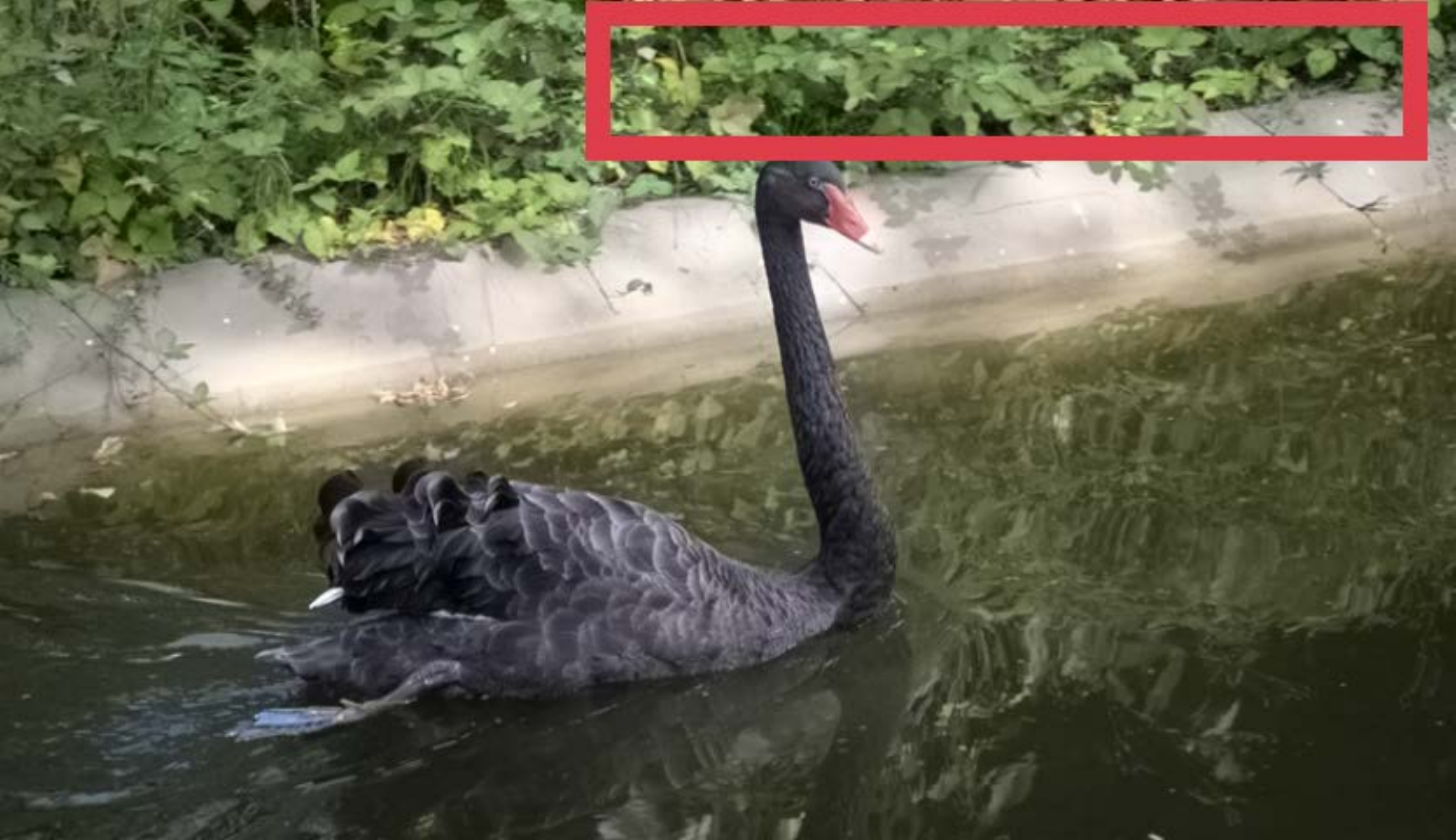} &
        \includegraphics[width=0.495\textwidth, height = 0.28\textwidth] {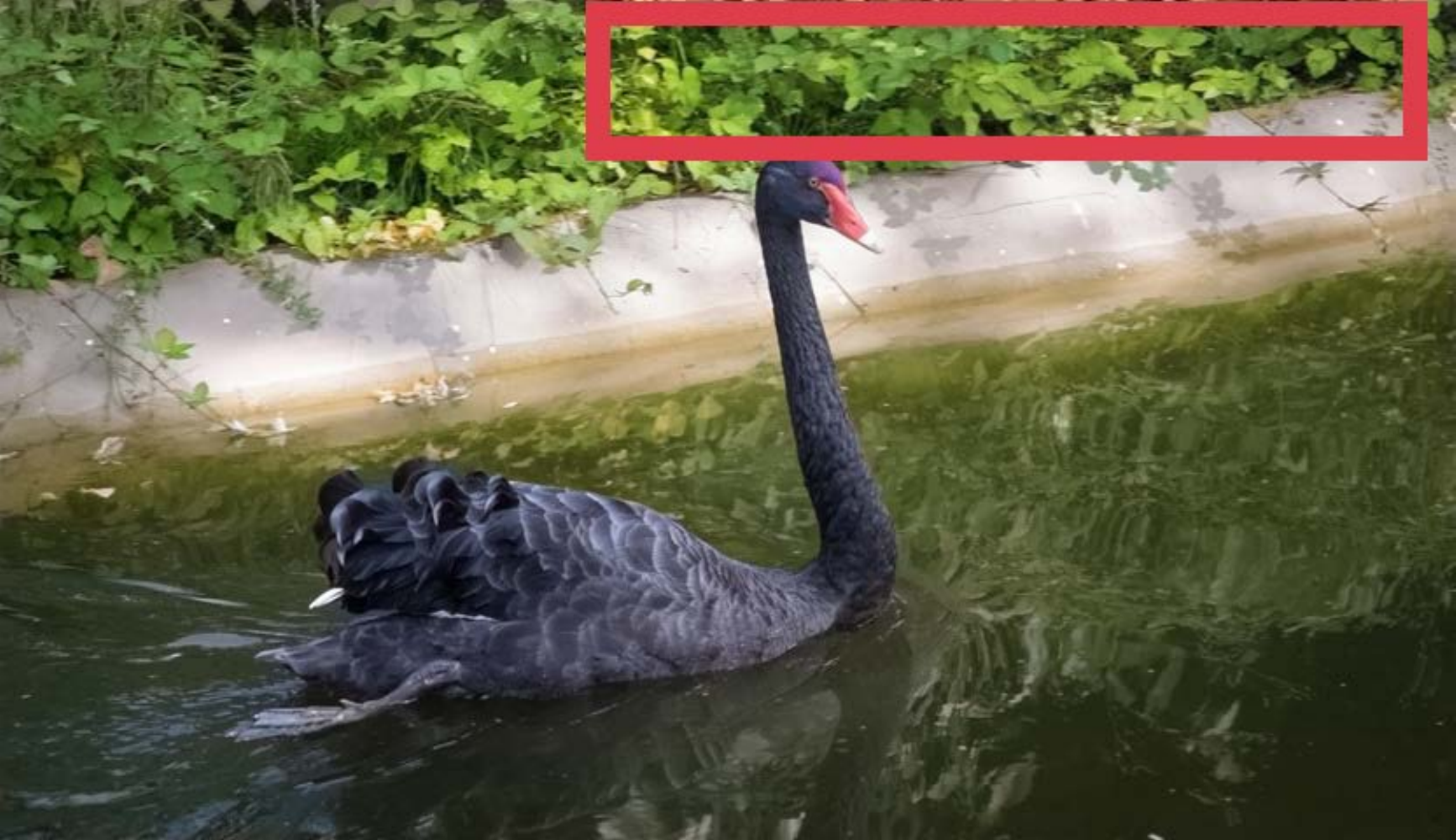} 
        \\
        \makebox[0.45\textwidth]{ (g) DeepExemplar~\cite{zhang2019deep}} &
        \makebox[0.45\textwidth]{ (h) ColorMNet (Ours)}  
         %& \vspace{-0.7em}
	\end{tabularx}
	\vspace{-0.8em}
	\caption{{Colorization results on clip \textit{blackswan} from the DAVIS validation dataset~\cite{Perazzi_CVPR_2016}. The results shown in (b) and (c) still contain significant color-bleeding artifacts. In contrast, our proposed method generates a better-colorized frame.}}
	\label{fig:2}
	%\vspace{-3mm}
\end{figure*}

\begin{figure*}[!htb]
	\setlength\tabcolsep{1.0pt}
	\centering
	\small
	\begin{tabularx}{1\textwidth}{cc}
        \includegraphics[width=0.495\textwidth, height = 0.28\textwidth] {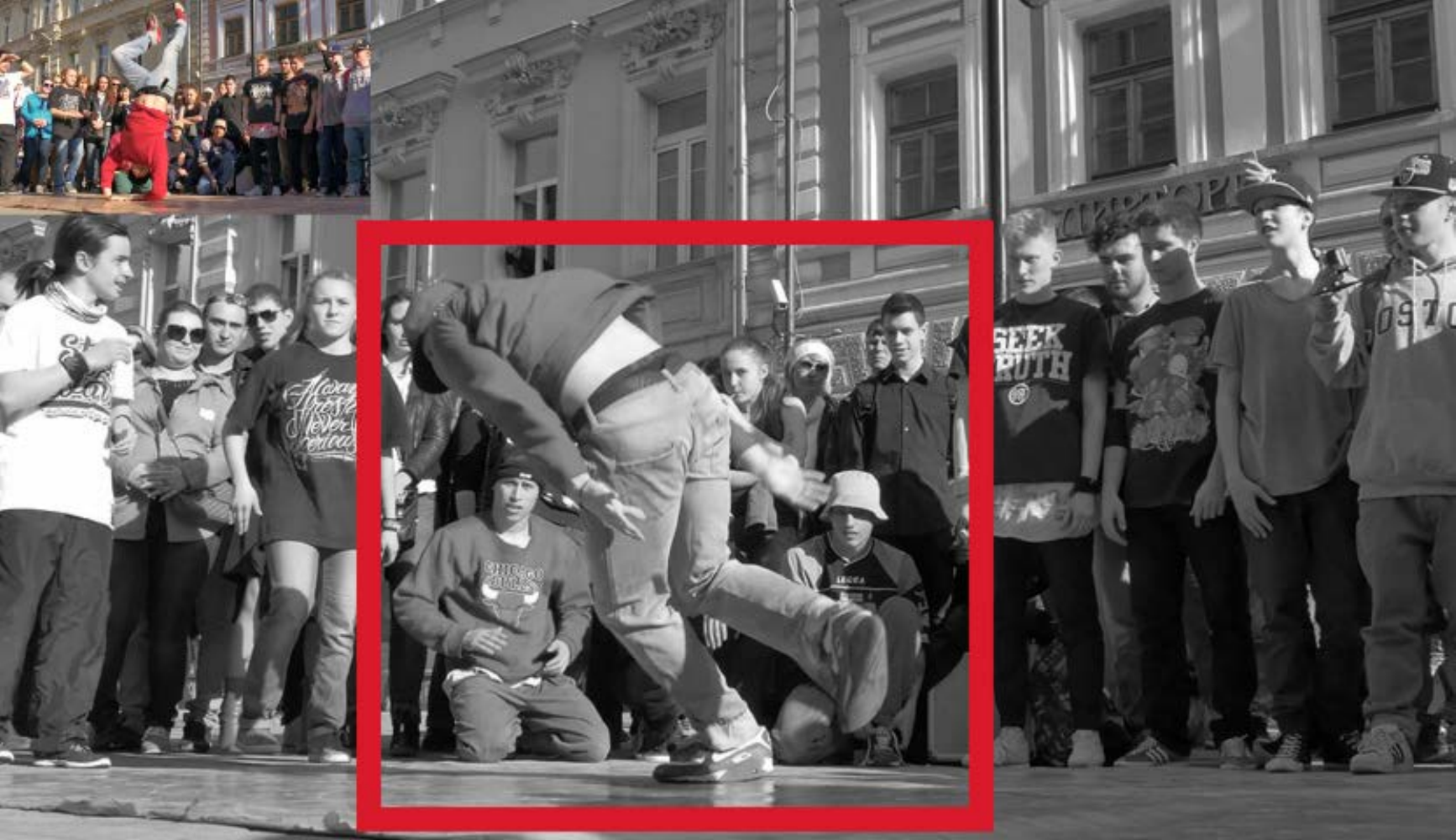} &
        \includegraphics[width=0.495\textwidth, height = 0.28\textwidth] {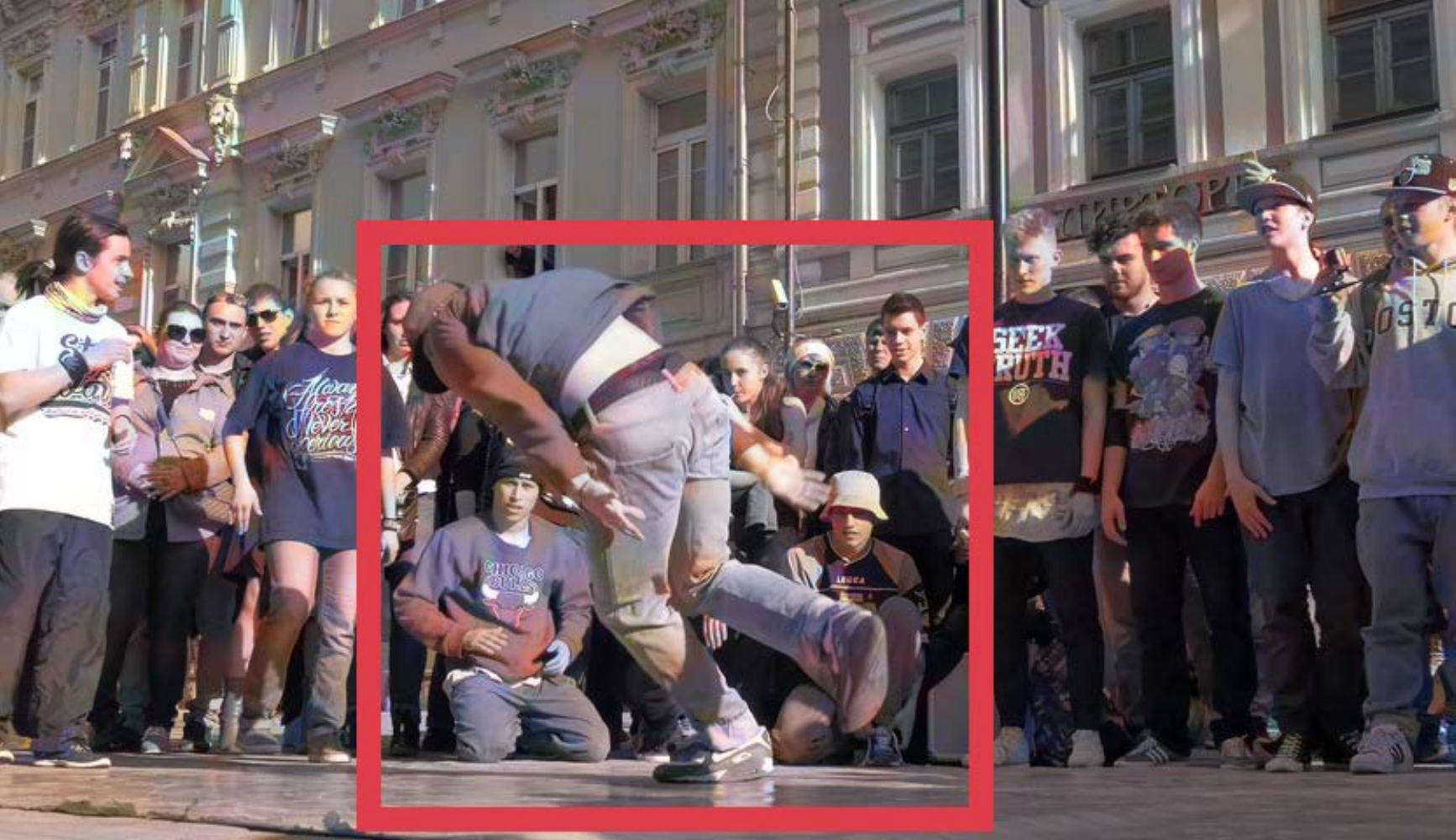} 
        \\
        \makebox[0.45\textwidth]{ (a) Input frame and exemplar image} &
        \makebox[0.45\textwidth]{ (b) DDColor~\cite{kang2022ddcolor}} 
        \\ 
        \includegraphics[width=0.495\textwidth, height = 0.28\textwidth] {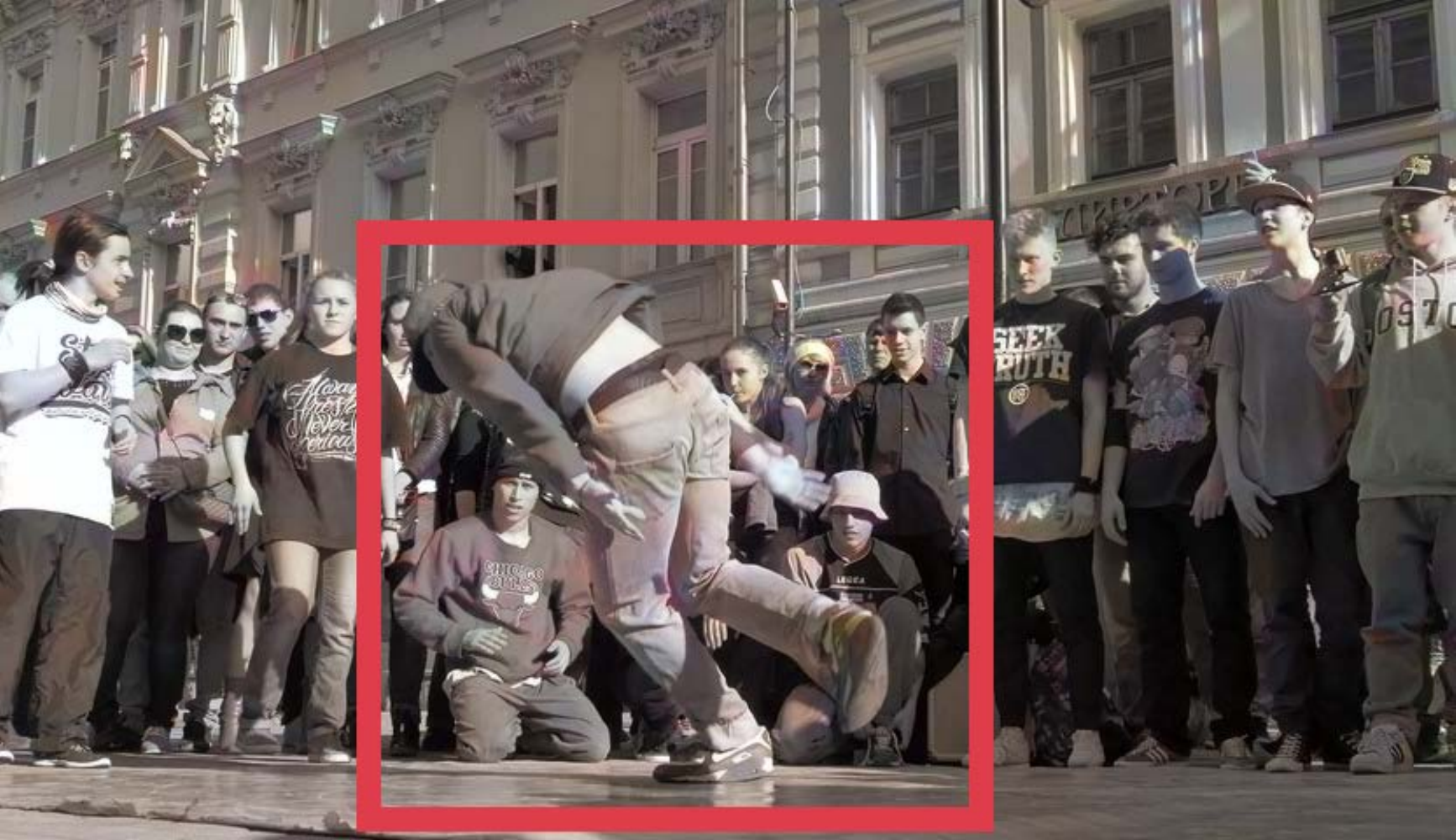} &
        \includegraphics[width=0.495\textwidth, height = 0.28\textwidth] {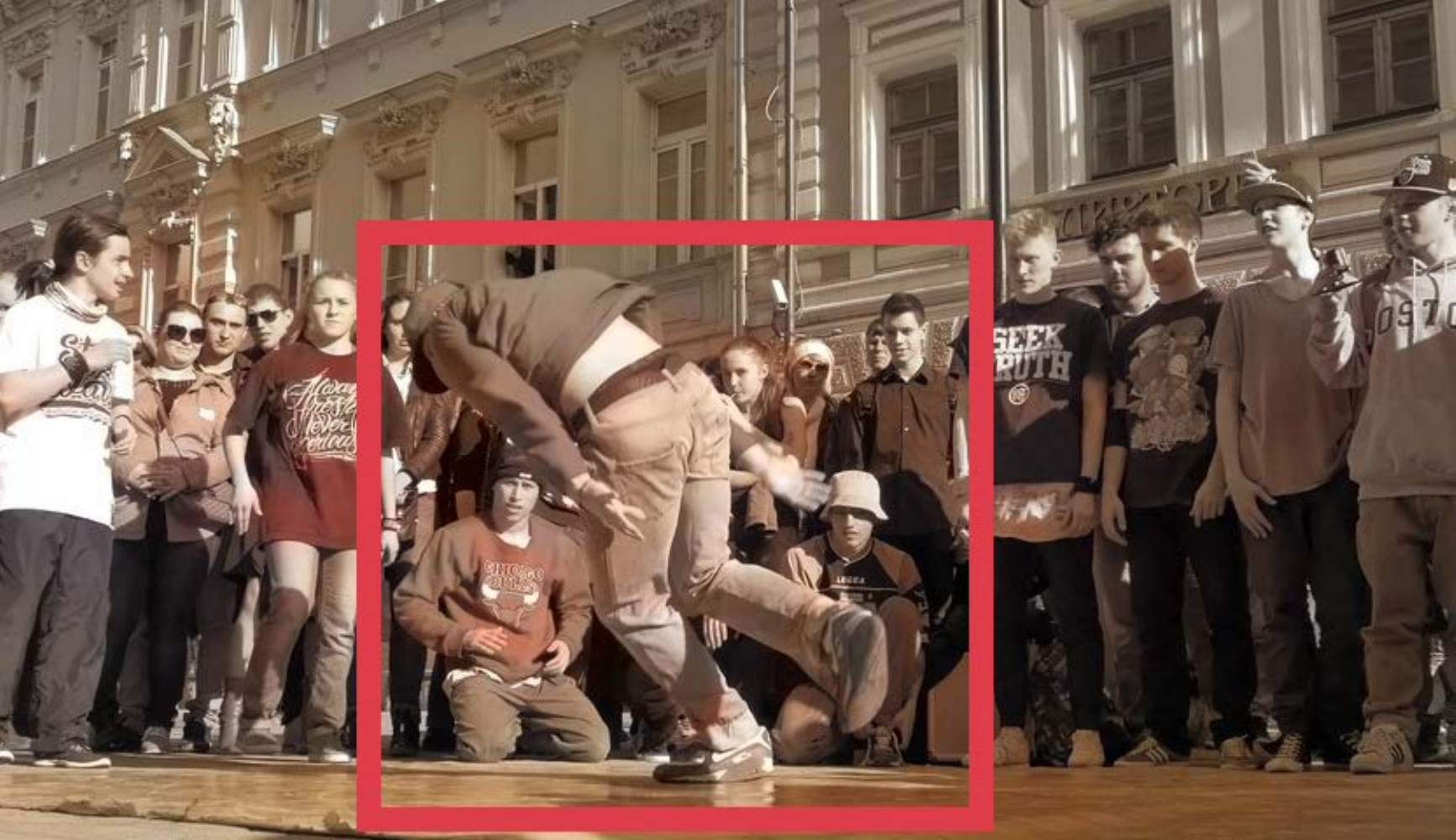} 
        \\ 
        \makebox[0.45\textwidth]{ (c) Color2Embed~\cite{zhao2021color2embed}} &
        \makebox[0.45\textwidth]{ (d) TCVC~\cite{liu2021temporally}} \\
        \includegraphics[width=0.495\textwidth, height = 0.28\textwidth] {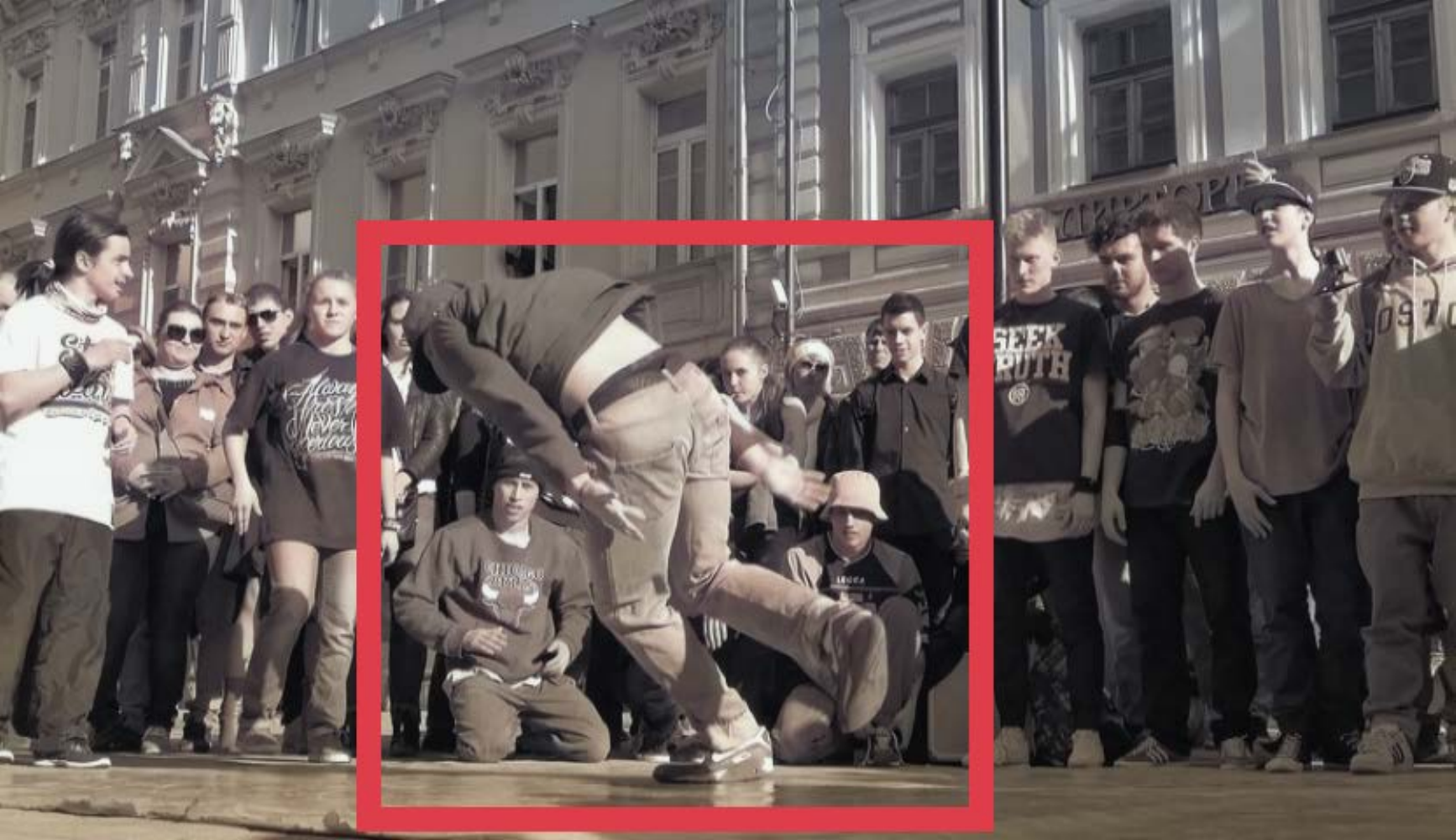} &
        \includegraphics[width=0.495\textwidth, height = 0.28\textwidth] {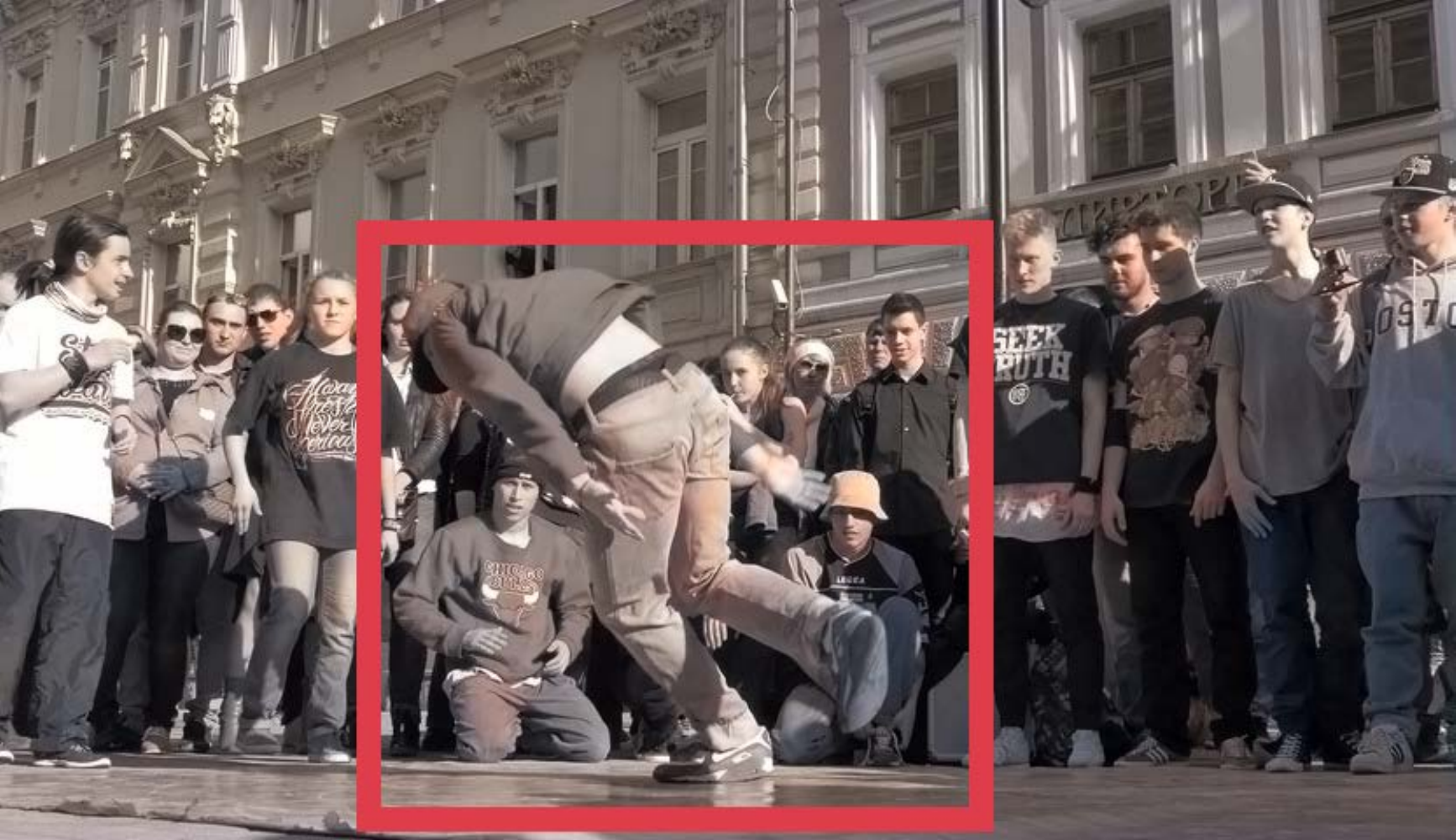} 
        \\
        \makebox[0.45\textwidth]{ (e) VCGAN~\cite{vcgan}} &
        \makebox[0.45\textwidth]{ (f) DeepRemaster~\cite{IizukaSIGGRAPHASIA2019}} 
        \\ 
        \includegraphics[width=0.495\textwidth, height = 0.28\textwidth] {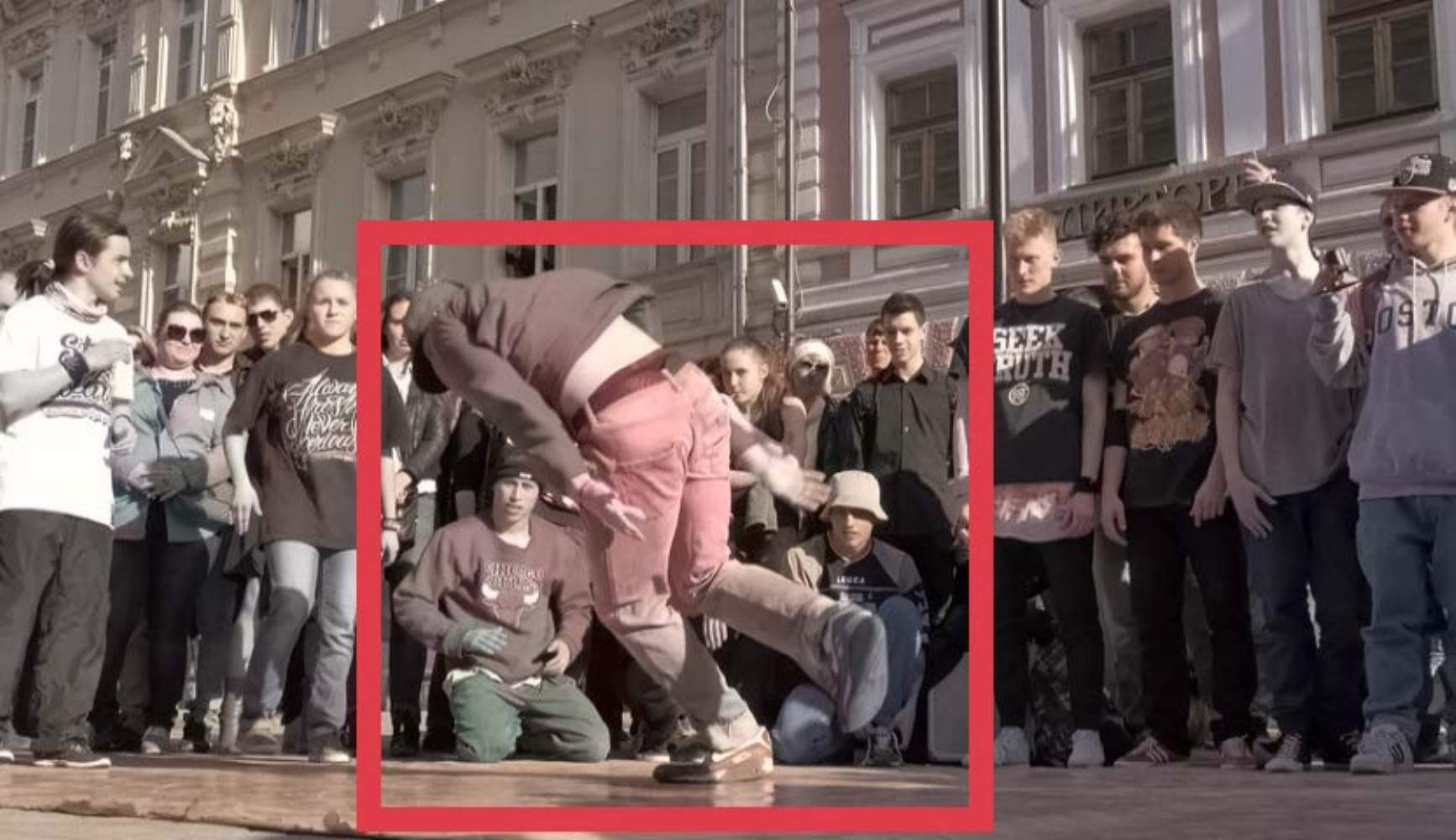} &
        \includegraphics[width=0.495\textwidth, height = 0.28\textwidth] {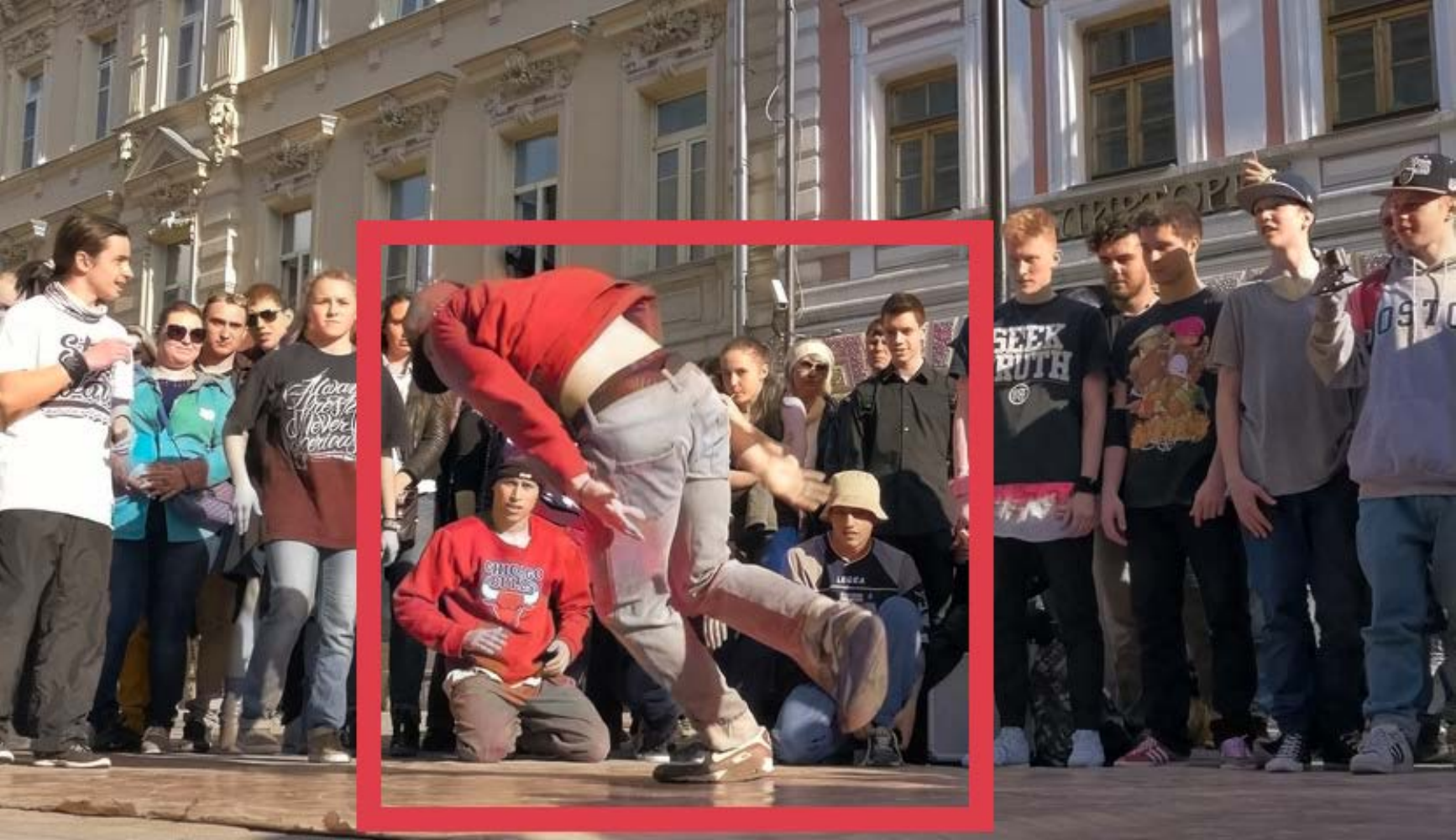} 
        \\
        \makebox[0.45\textwidth]{ (g) DeepExemplar~\cite{zhang2019deep}} &
        \makebox[0.45\textwidth]{ (h) ColorMNet (Ours)}  
         %& \vspace{-0.7em}
	\end{tabularx}
	\vspace{-0.8em}
	\caption{{Colorization results on clip \textit{breakdance} from the DAVIS validation dataset~\cite{Perazzi_CVPR_2016}. The results shown in (b) and (c) still contain significant color-bleeding artifacts. In contrast, our proposed method generates a better-colorized frame, where the colors of the dancer are restored well.}}
	\label{fig:3}
	%\vspace{-3mm}
\end{figure*}

\begin{figure*}[!htb]
	\setlength\tabcolsep{1.0pt}
	\centering
	\small
	\begin{tabularx}{1\textwidth}{cc}
        \includegraphics[width=0.495\textwidth, height = 0.28\textwidth] {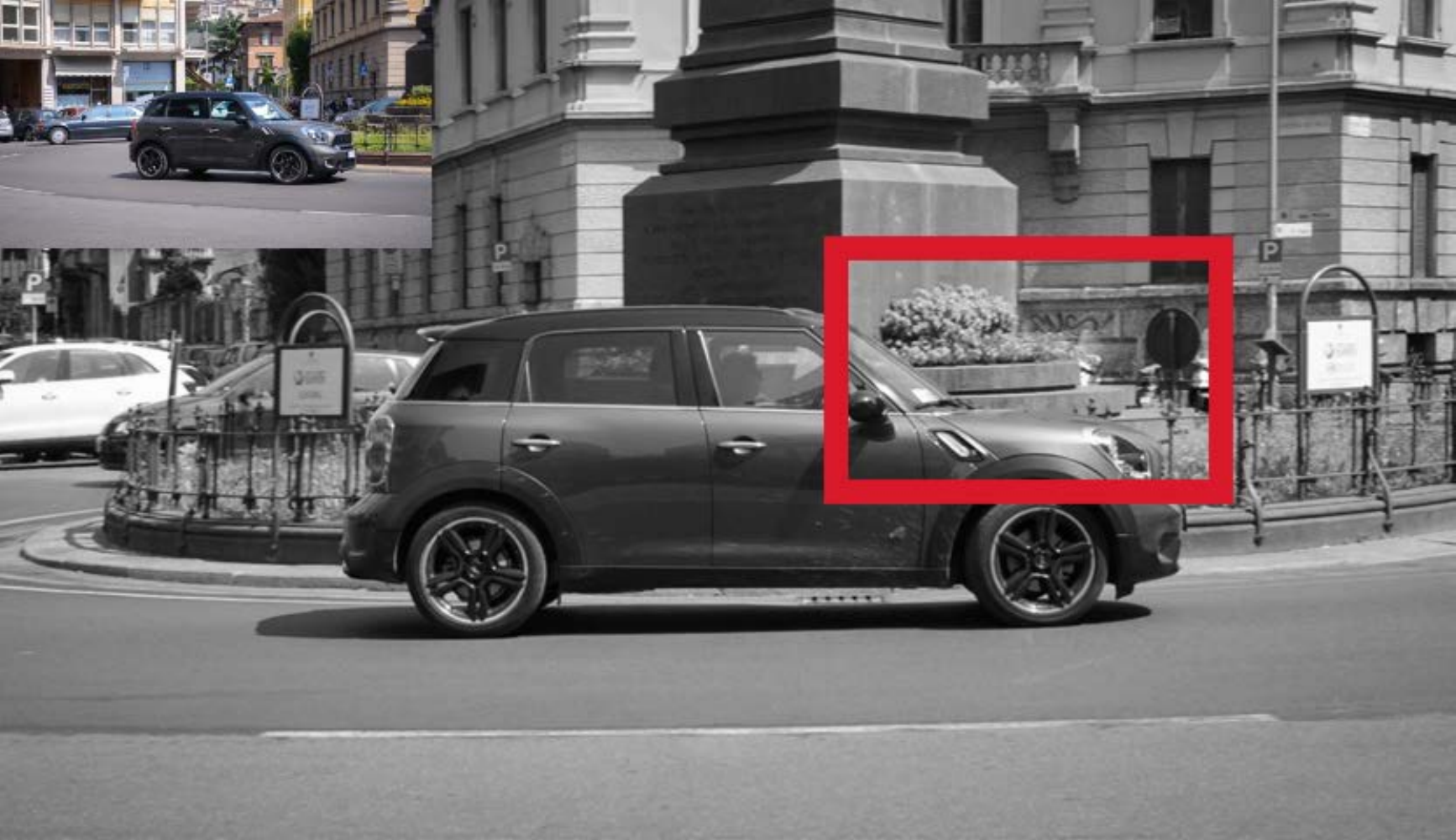} &
        \includegraphics[width=0.495\textwidth, height = 0.28\textwidth] {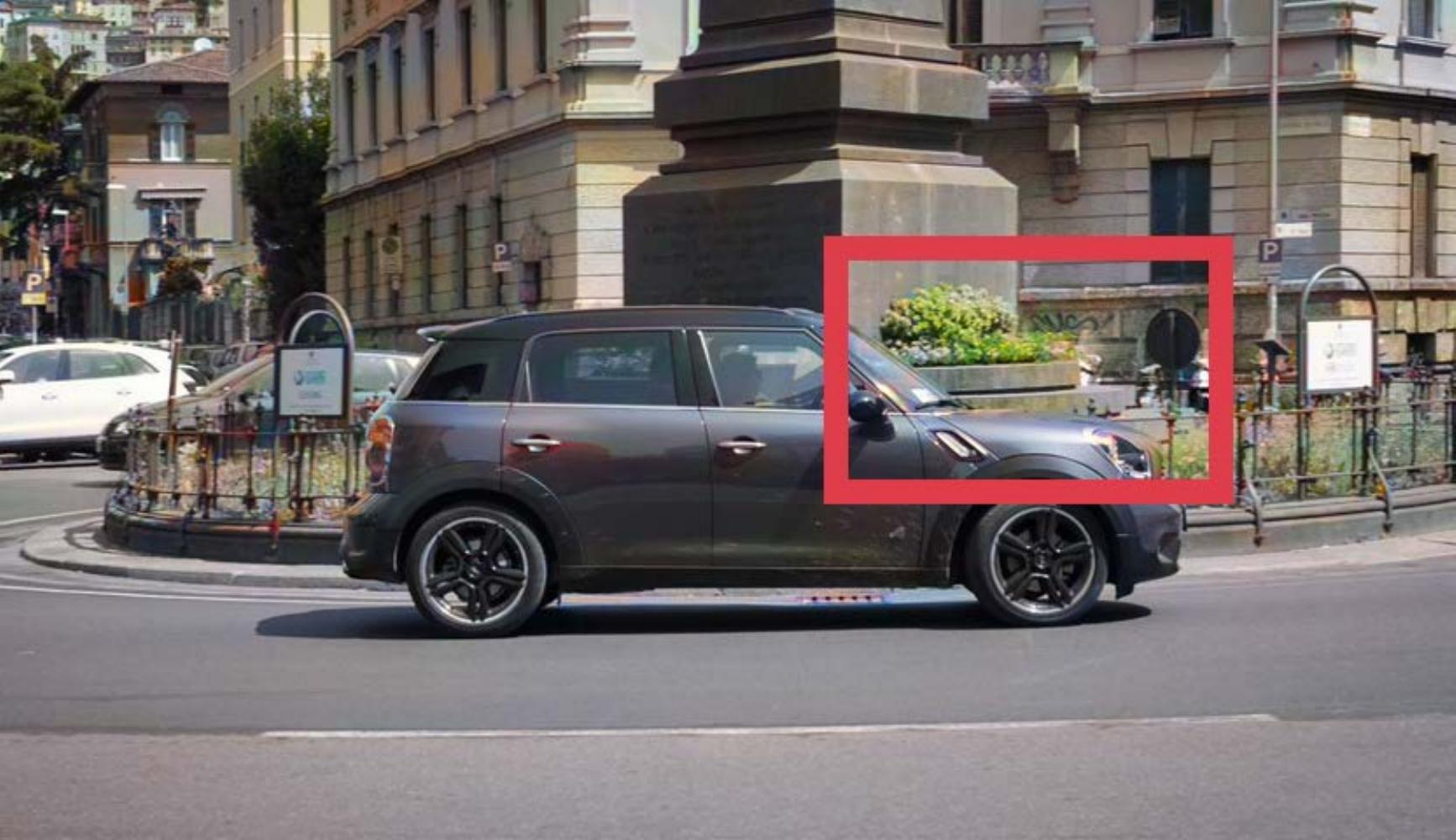} 
        \\
        \makebox[0.45\textwidth]{ (a) Input frame and exemplar image} &
        \makebox[0.45\textwidth]{ (b) DDColor~\cite{kang2022ddcolor}} 
        \\ 
        \includegraphics[width=0.495\textwidth, height = 0.28\textwidth] {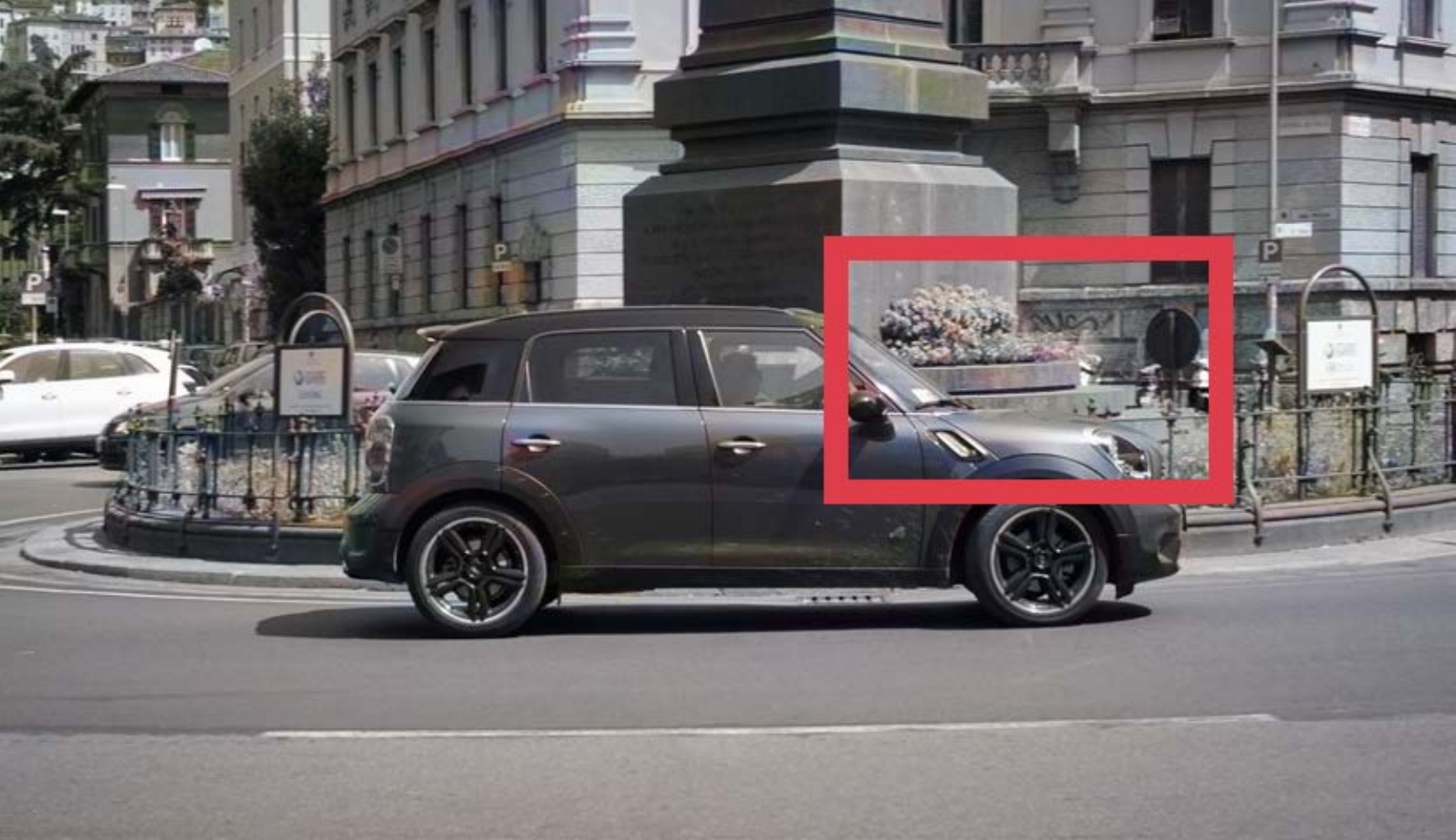} &
        \includegraphics[width=0.495\textwidth, height = 0.28\textwidth] {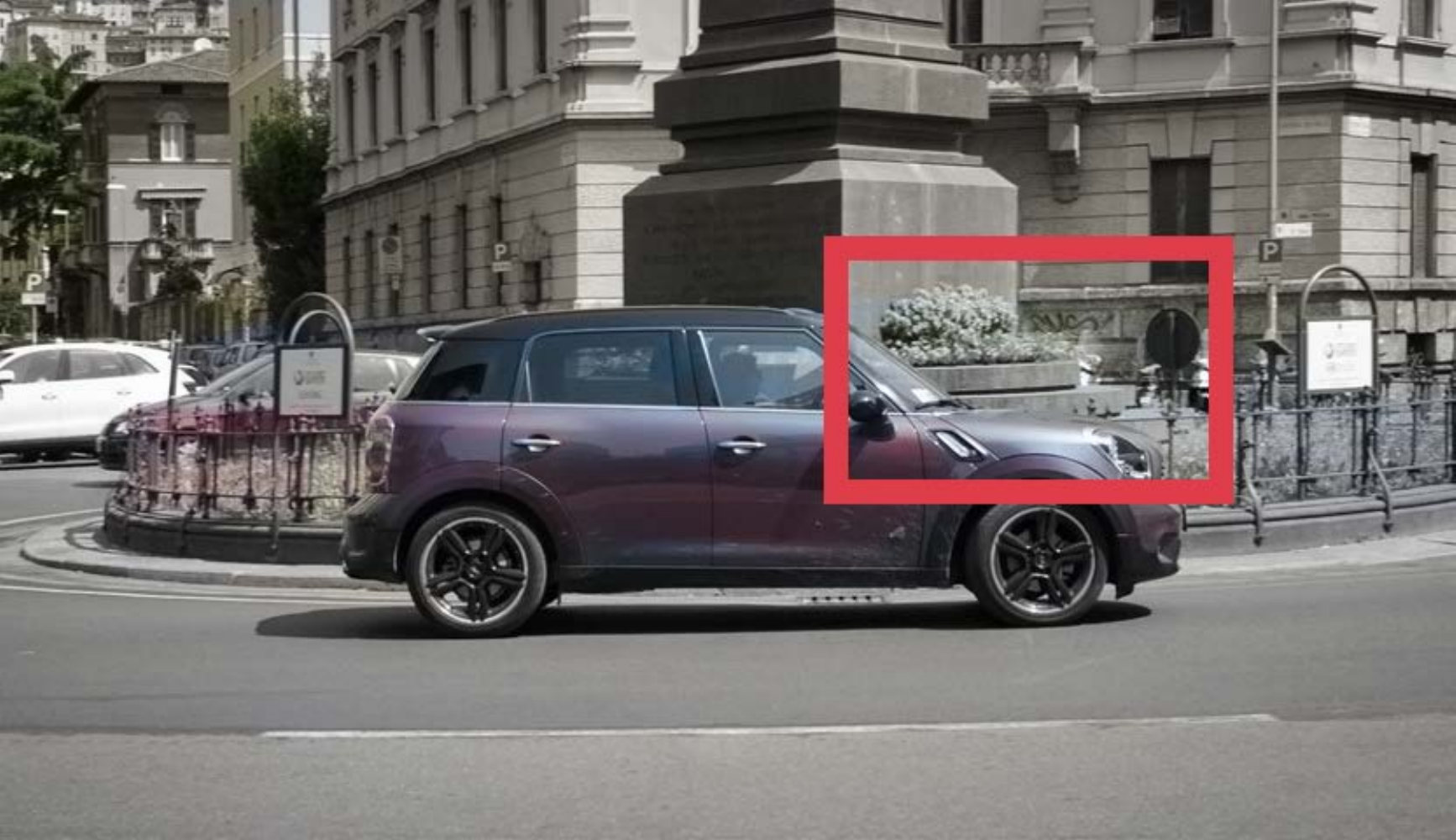} 
        \\ 
        \makebox[0.45\textwidth]{ (c) Color2Embed~\cite{zhao2021color2embed}} &
        \makebox[0.45\textwidth]{ (d) TCVC~\cite{liu2021temporally}} \\
        \includegraphics[width=0.495\textwidth, height = 0.28\textwidth] {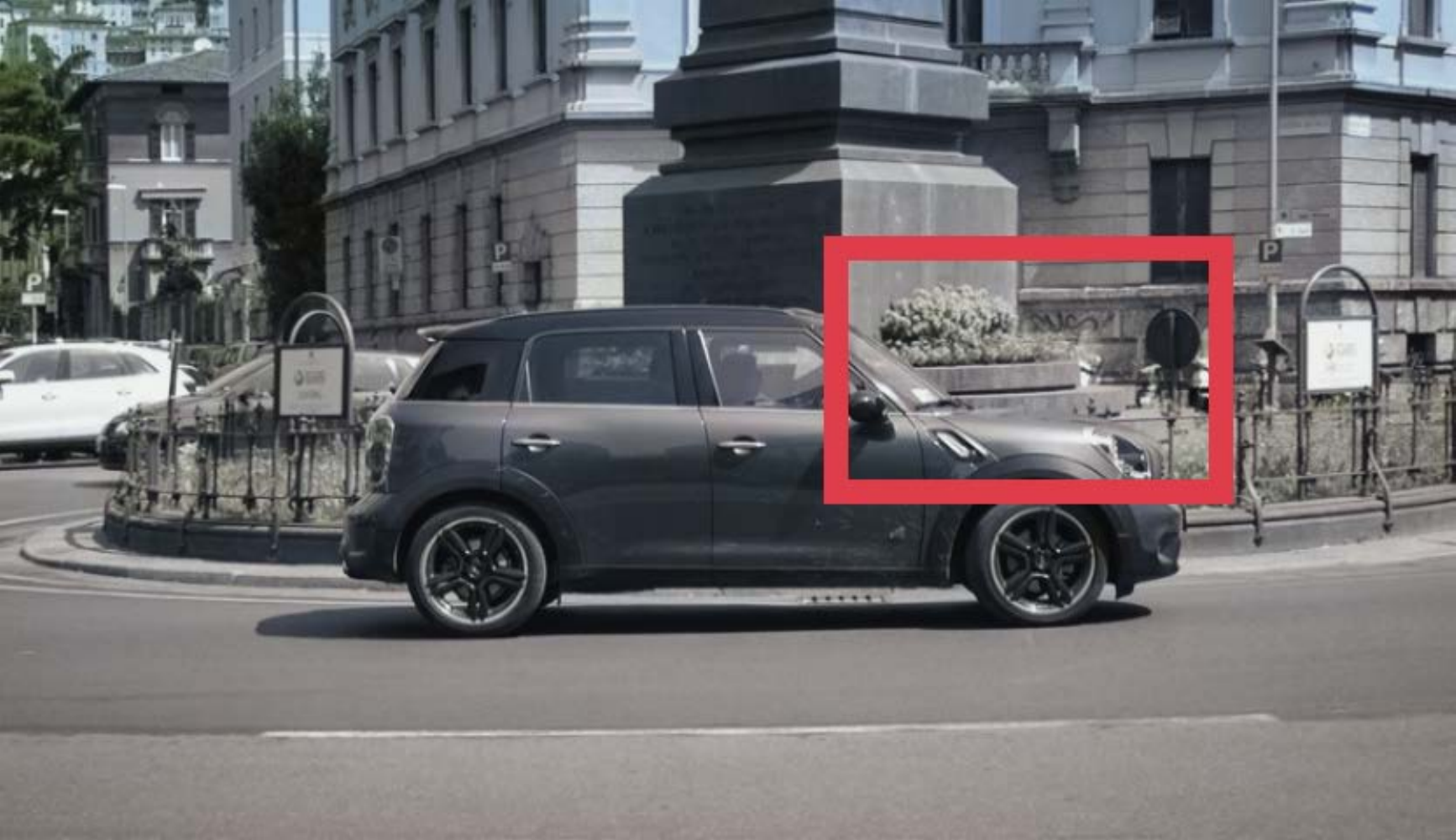} &
        \includegraphics[width=0.495\textwidth, height = 0.28\textwidth] {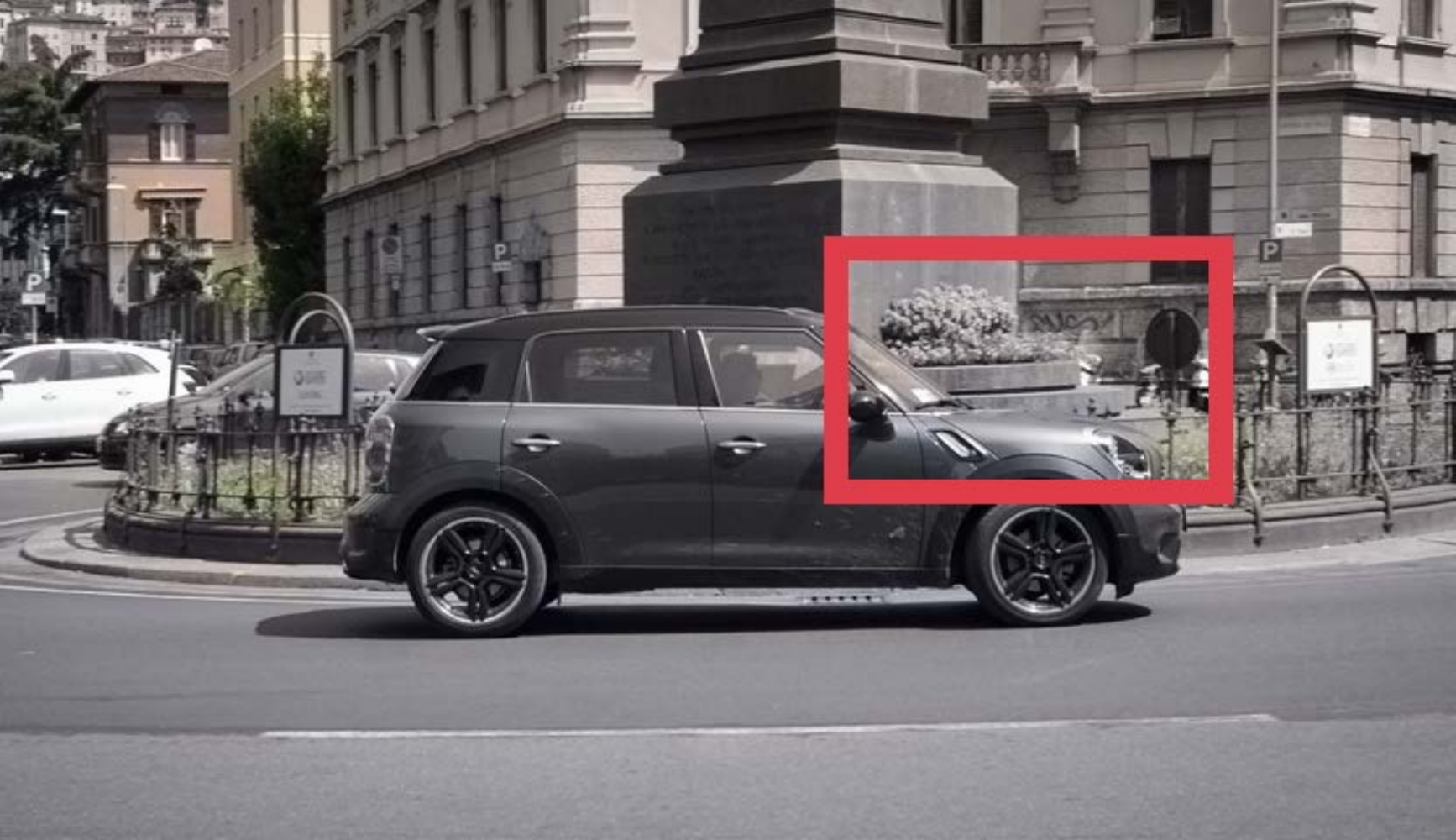} 
        \\
        \makebox[0.45\textwidth]{ (e) VCGAN~\cite{vcgan}} &
        \makebox[0.45\textwidth]{ (f) DeepRemaster~\cite{IizukaSIGGRAPHASIA2019}} 
        \\ 
        \includegraphics[width=0.495\textwidth, height = 0.28\textwidth] {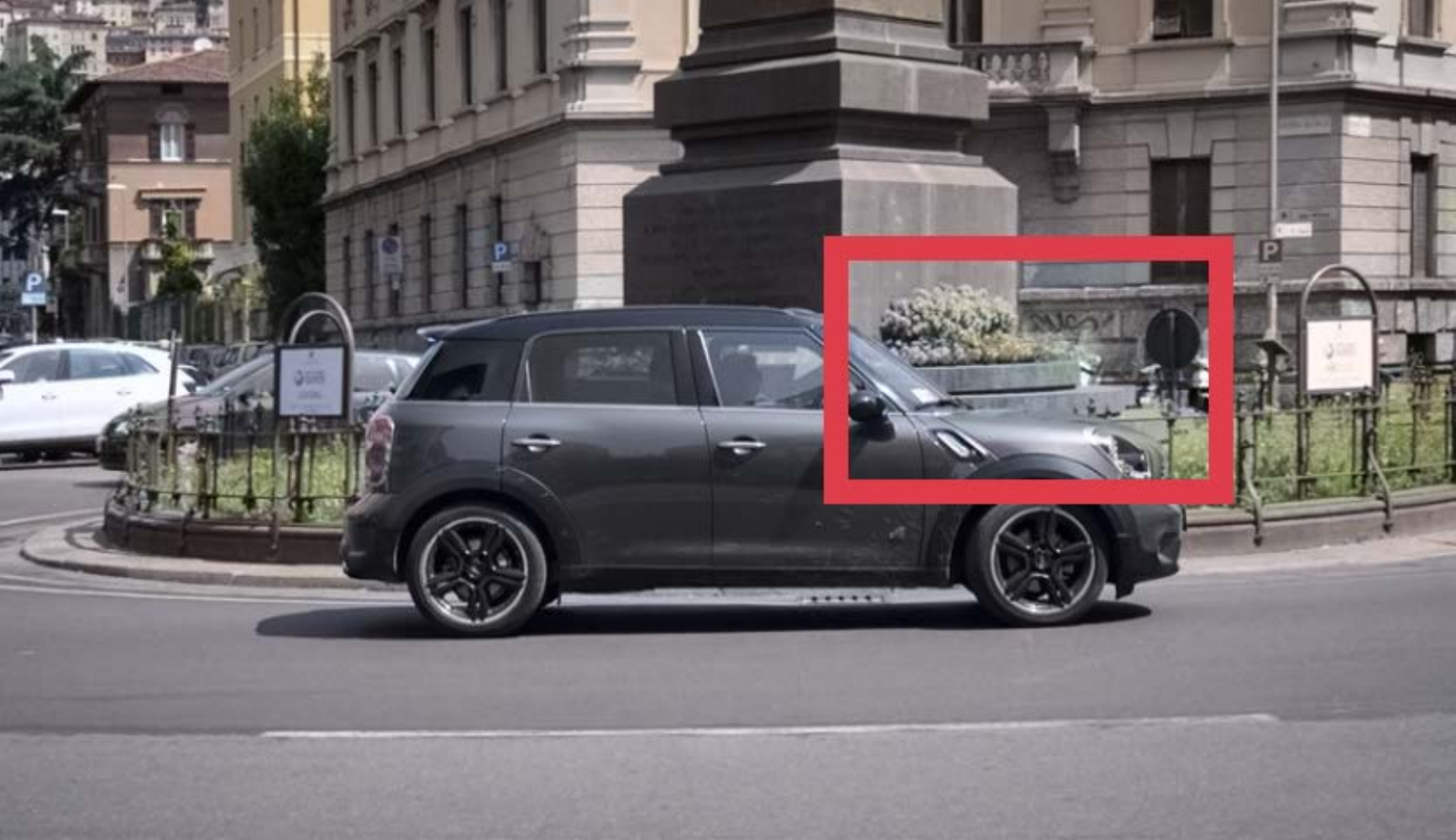} &
        \includegraphics[width=0.495\textwidth, height = 0.28\textwidth] {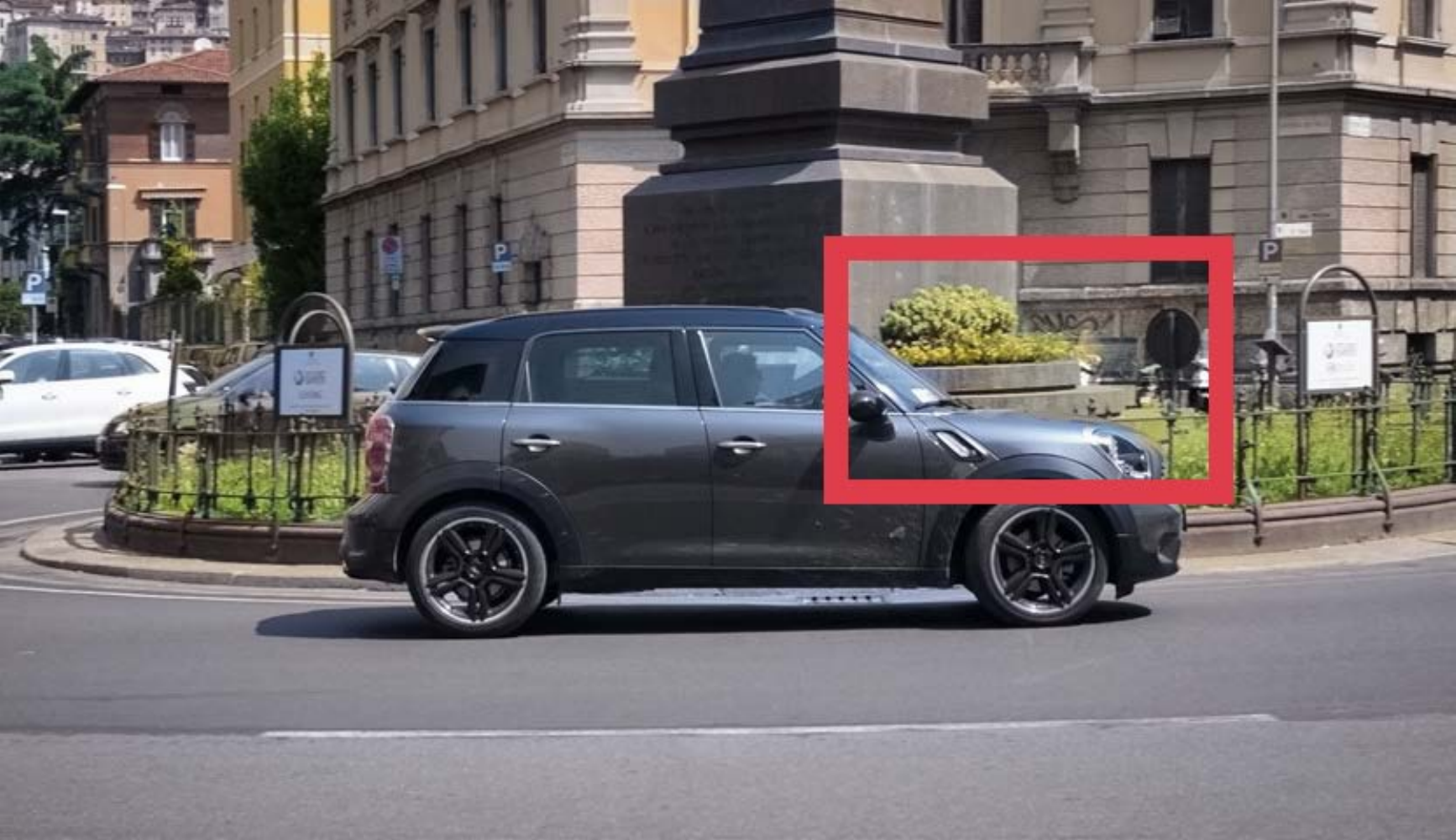} 
        \\
        \makebox[0.45\textwidth]{ (g) DeepExemplar~\cite{zhang2019deep}} &
        \makebox[0.45\textwidth]{ (h) ColorMNet (Ours)}  
         %& \vspace{-0.7em}
	\end{tabularx}
	\vspace{-0.8em}
	\caption{{Colorization results on clip \textit{car-roundabout} from the DAVIS validation dataset~\cite{Perazzi_CVPR_2016}. The results shown in (b) and (c) still contain significant color-bleeding artifacts. In contrast, our proposed method restores the colors of the flowerbed and generates a better-colorized frame.}}
	\label{fig:4}
	%\vspace{-3mm}
\end{figure*}

\begin{figure*}[!htb]
	\setlength\tabcolsep{1.0pt}
	\centering
	\small
	\begin{tabularx}{1\textwidth}{cc}
        \includegraphics[width=0.495\textwidth, height = 0.28\textwidth] {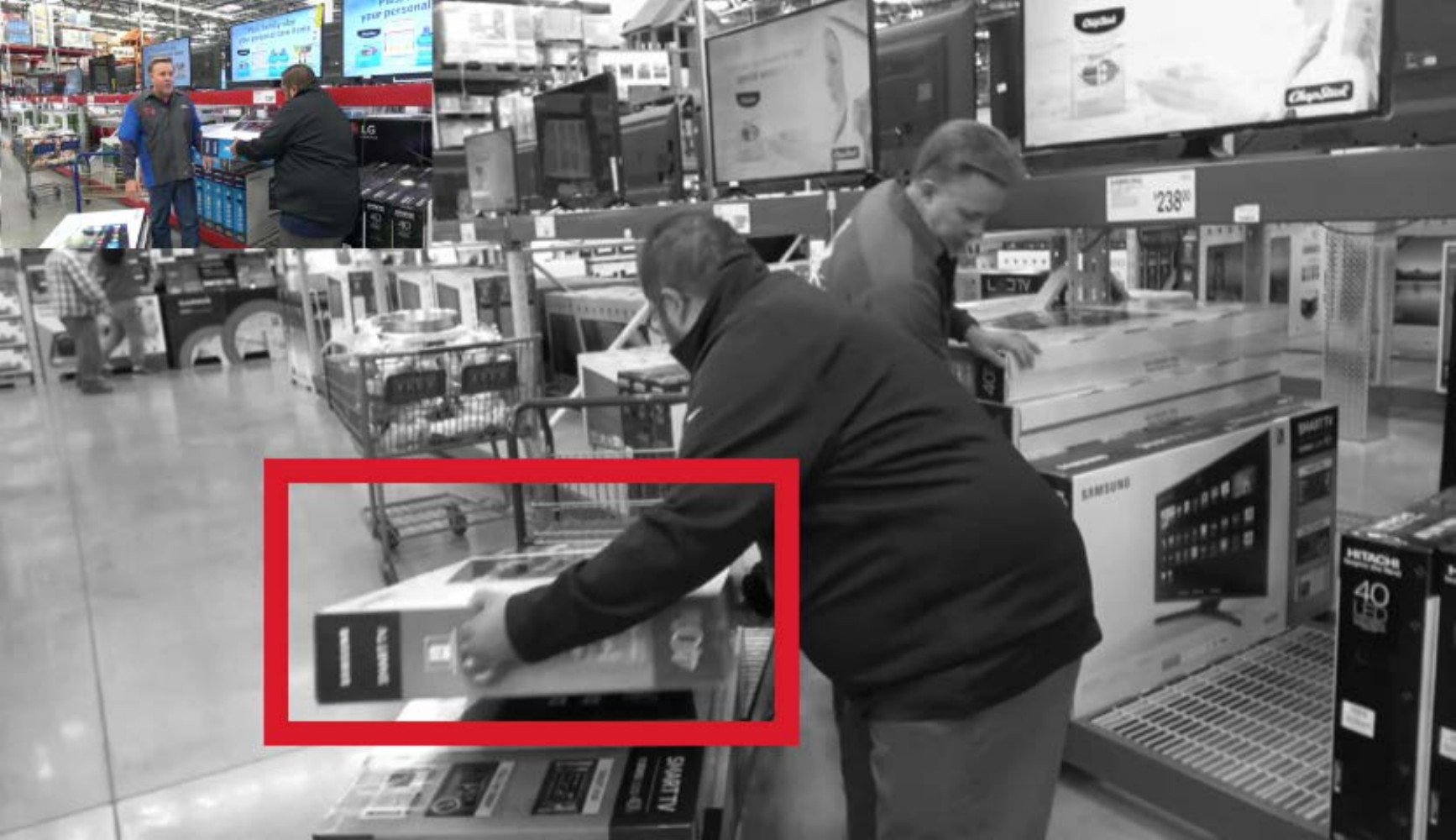} &
        \includegraphics[width=0.495\textwidth, height = 0.28\textwidth] {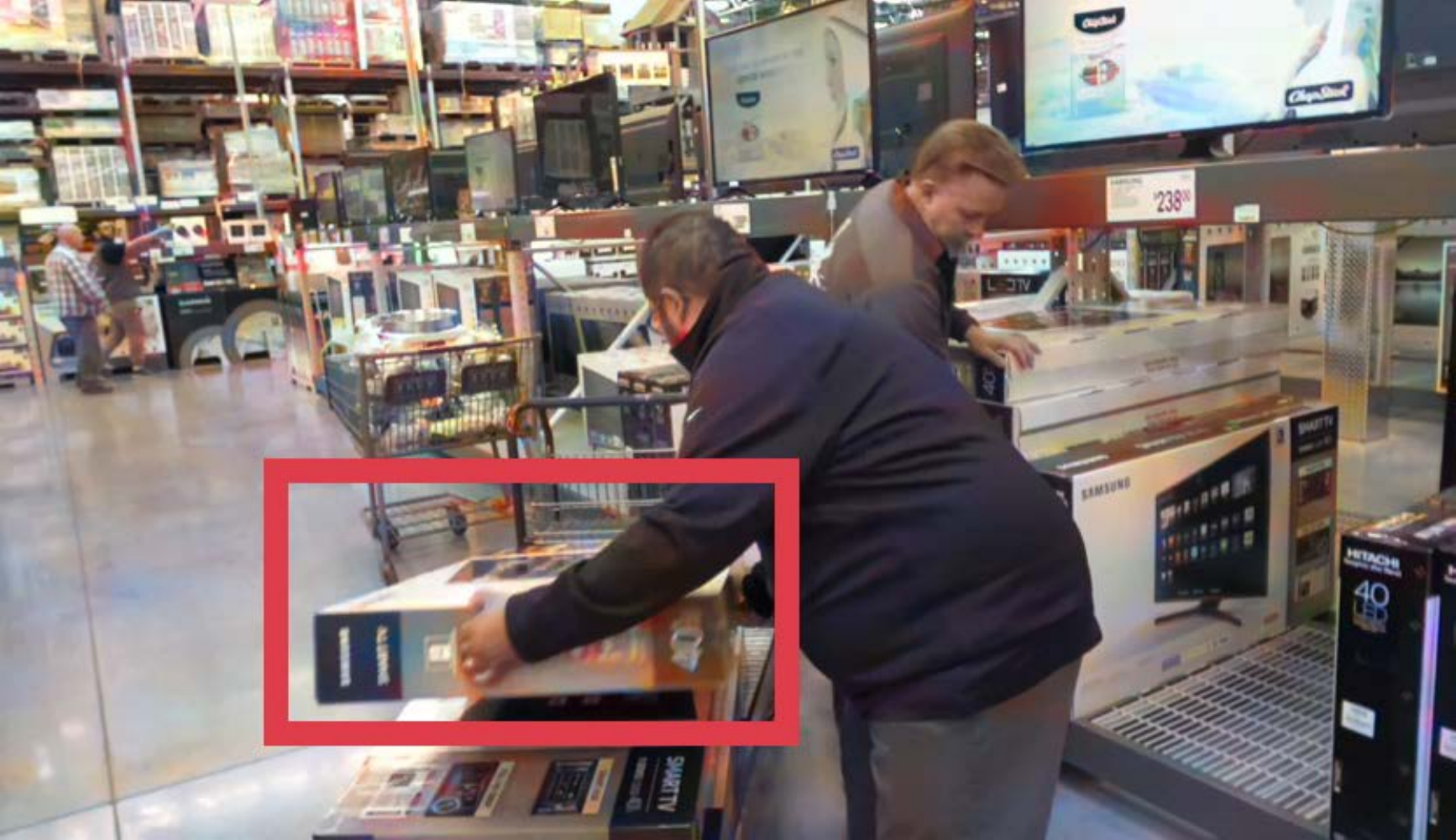} 
        \\
        \makebox[0.45\textwidth]{ (a) Input frame and exemplar image} &
        \makebox[0.45\textwidth]{ (b) DDColor~\cite{kang2022ddcolor}} 
        \\ 
        \includegraphics[width=0.495\textwidth, height = 0.28\textwidth] {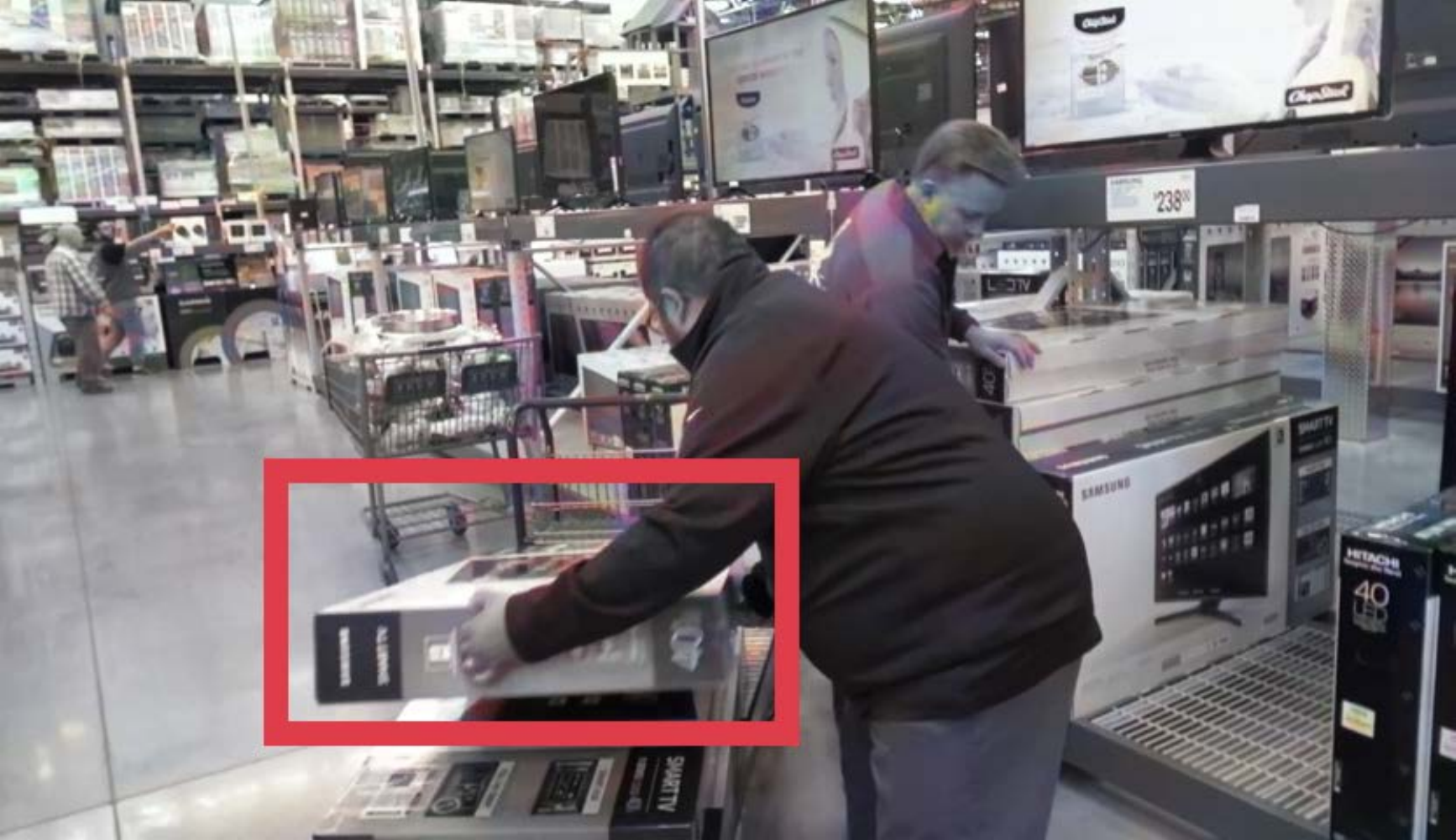} &
        \includegraphics[width=0.495\textwidth, height = 0.28\textwidth] {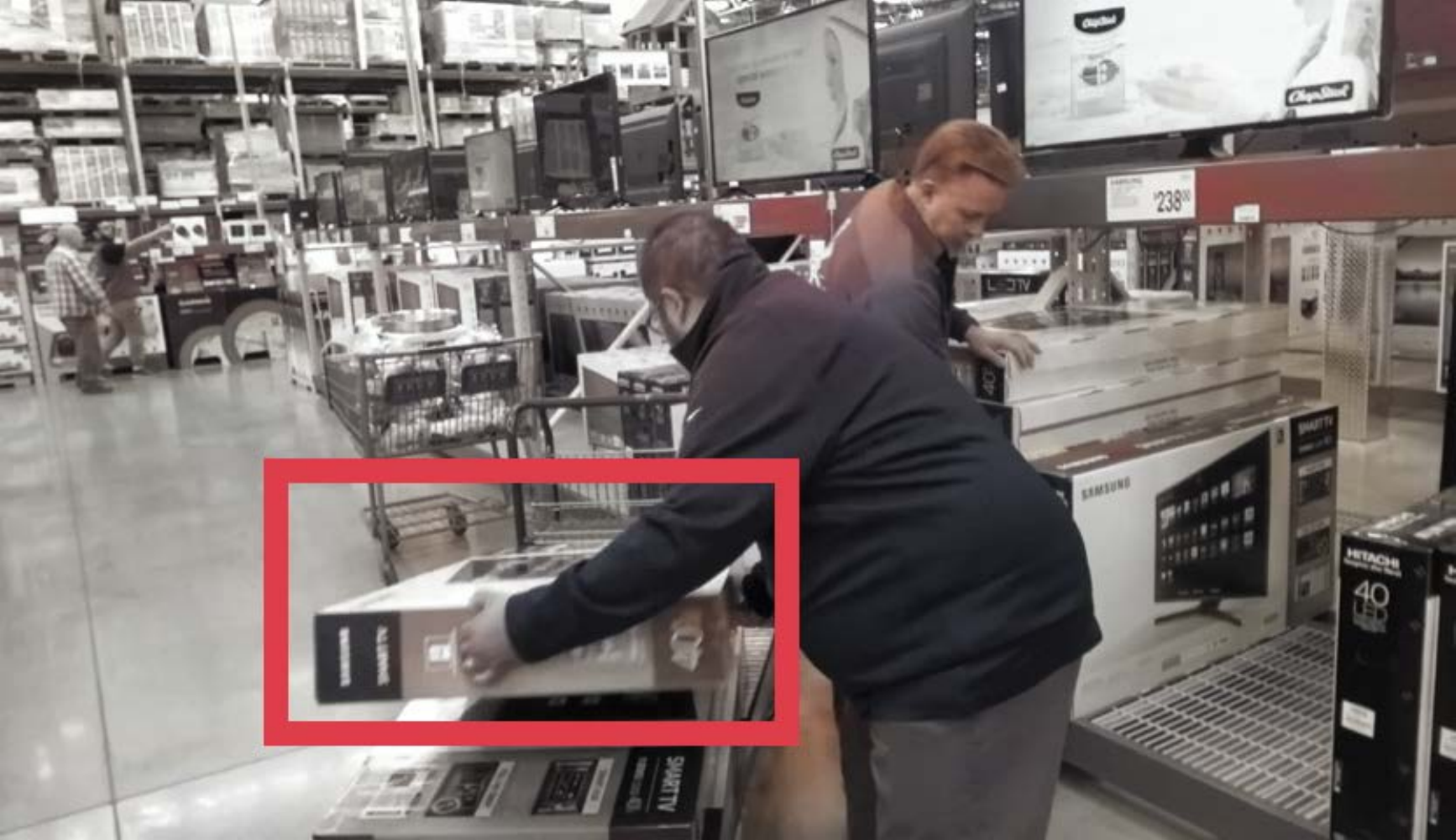} 
        \\ 
        \makebox[0.45\textwidth]{ (c) Color2Embed~\cite{zhao2021color2embed}} &
        \makebox[0.45\textwidth]{ (d) TCVC~\cite{liu2021temporally}} \\
        \includegraphics[width=0.495\textwidth, height = 0.28\textwidth] {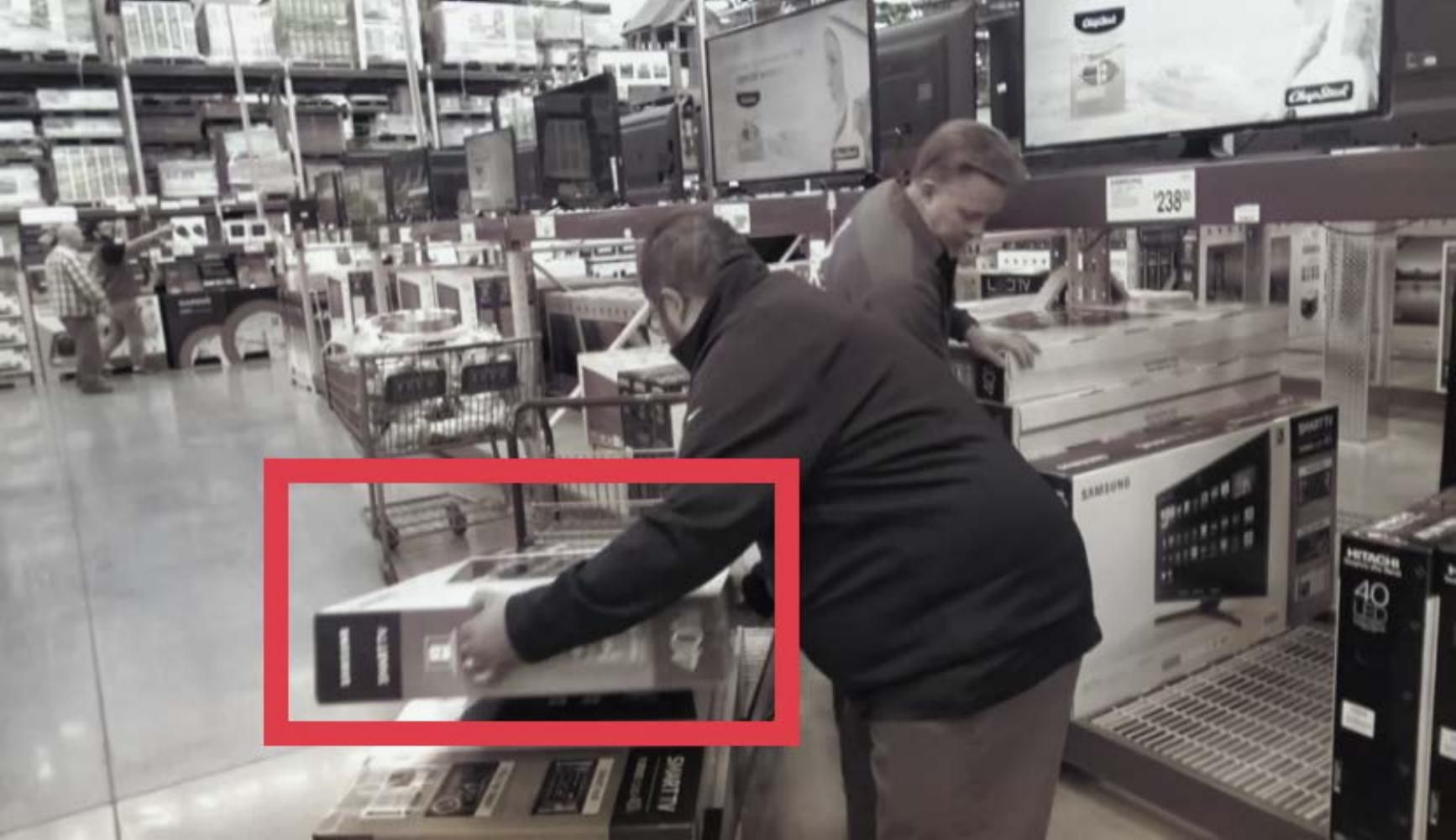} &
        \includegraphics[width=0.495\textwidth, height = 0.28\textwidth] {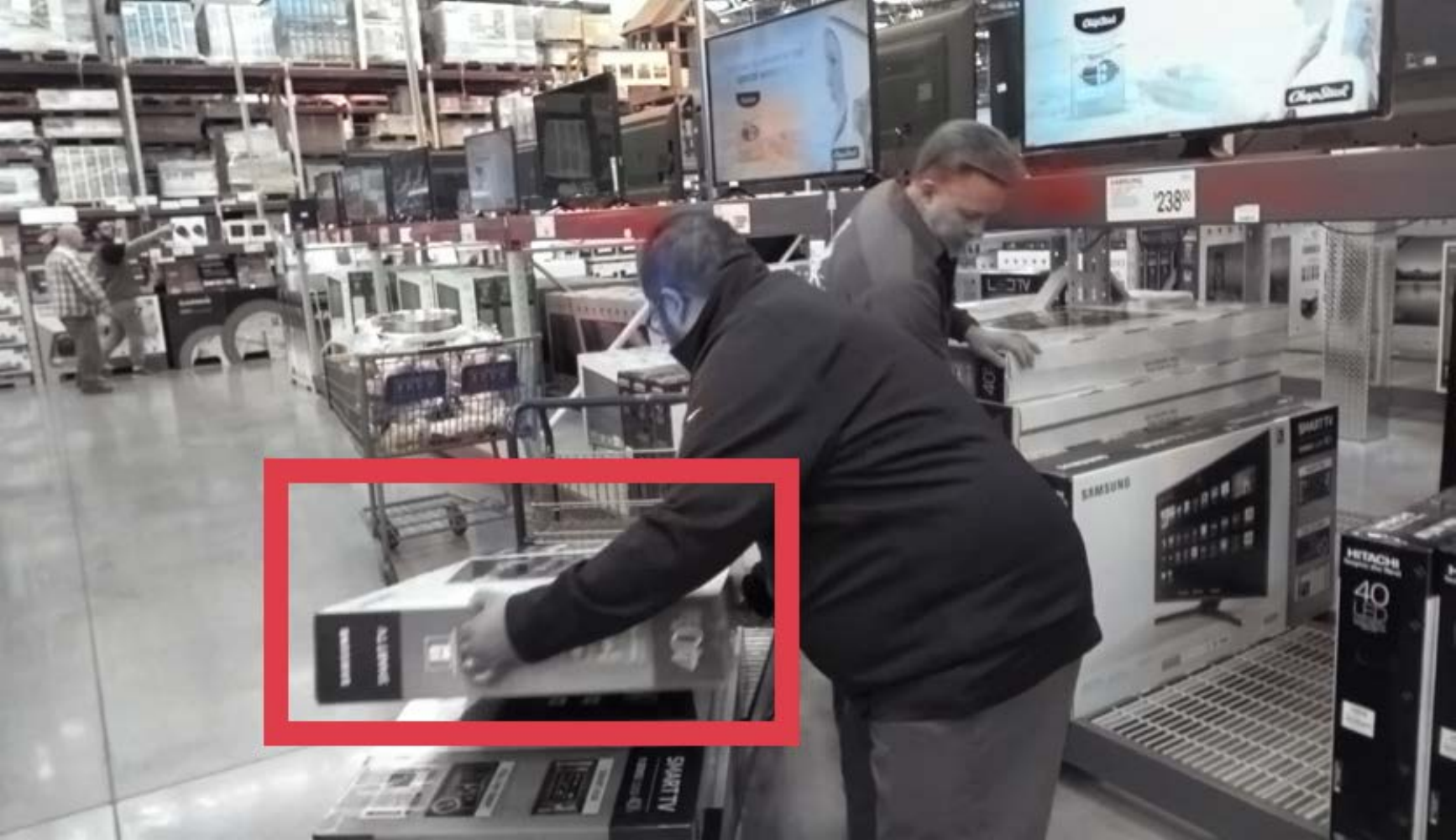} 
        \\
        \makebox[0.45\textwidth]{ (e) VCGAN~\cite{vcgan}} &
        \makebox[0.45\textwidth]{ (f) DeepRemaster~\cite{IizukaSIGGRAPHASIA2019}} 
        \\ 
        \includegraphics[width=0.495\textwidth, height = 0.28\textwidth] {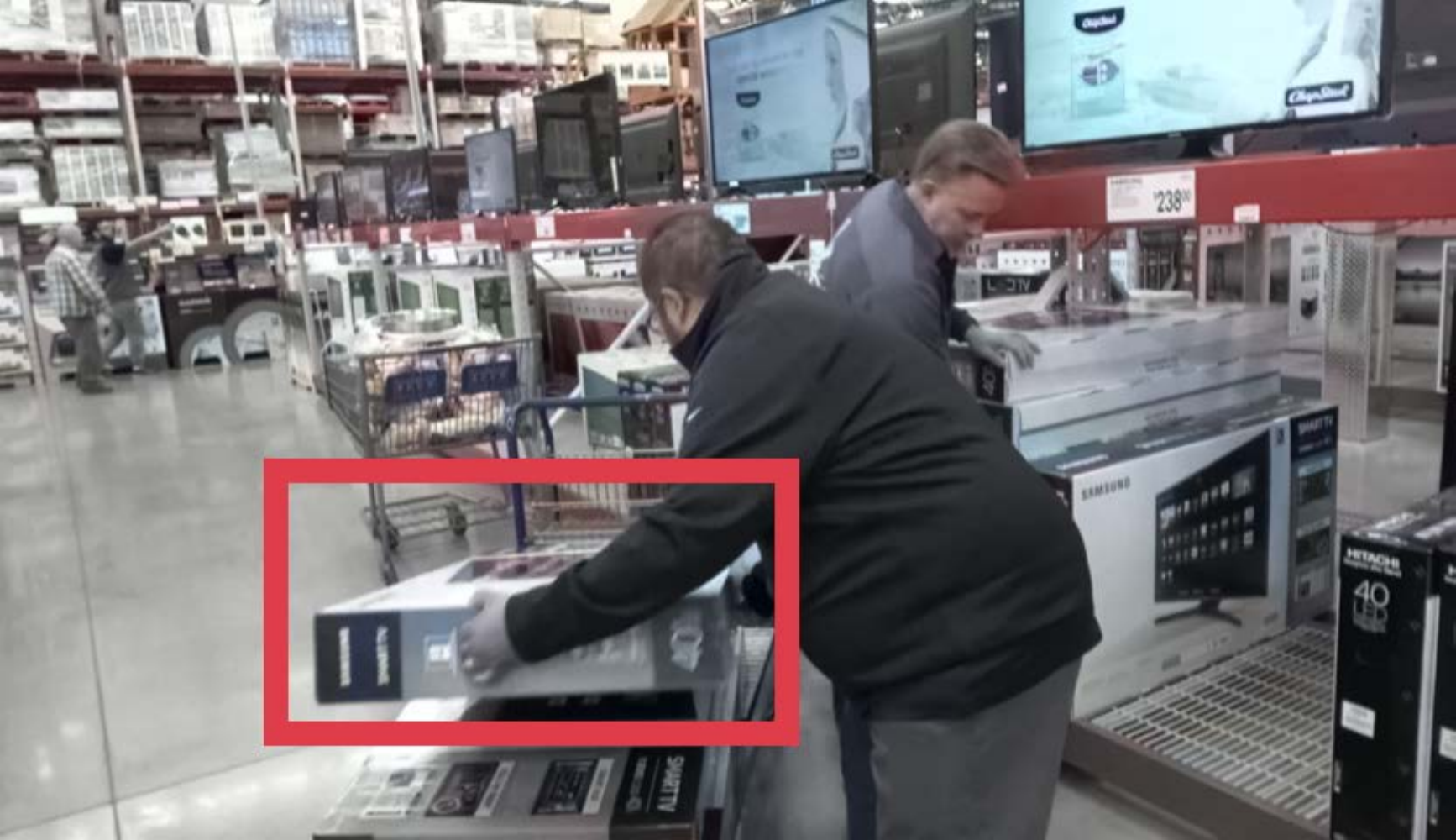} &
        \includegraphics[width=0.495\textwidth, height = 0.28\textwidth] {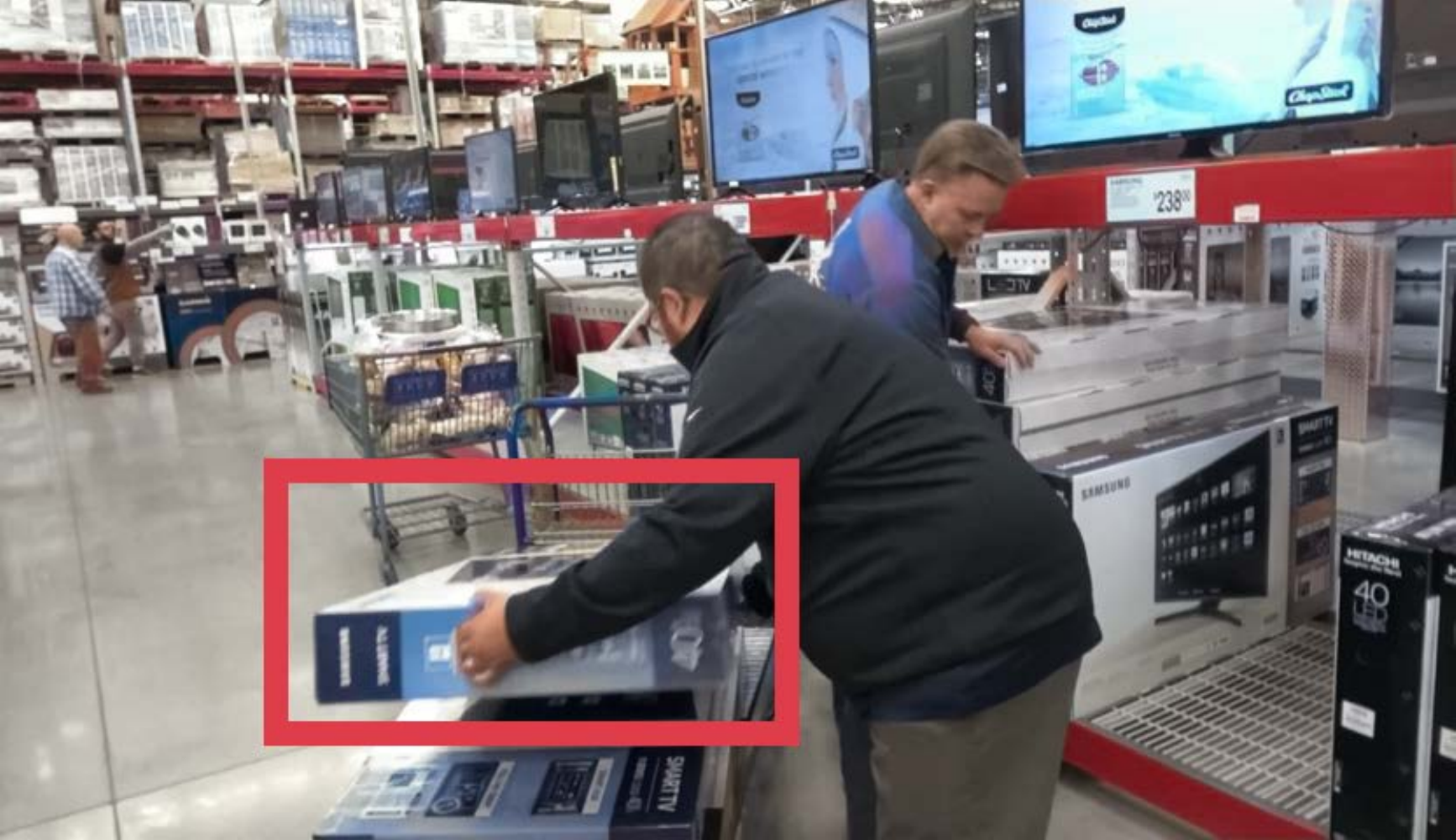} 
        \\
        \makebox[0.45\textwidth]{ (g) DeepExemplar~\cite{zhang2019deep}} &
        \makebox[0.45\textwidth]{ (h) ColorMNet (Ours)}  
         %& \vspace{-0.7em}
	\end{tabularx}
	\vspace{-0.8em}
	\caption{{Colorization results on clip \textit{loading} from the DAVIS validation dataset~\cite{Perazzi_CVPR_2016}. The results shown in (b) and (c) still contain significant color-bleeding artifacts. In contrast, our proposed method generates a vivid and realistic frame, where the colors of the box and the man's hands are better restored.}}
	\label{fig:5}
	%\vspace{-3mm}
\end{figure*}

\begin{figure*}[!htb]
	\setlength\tabcolsep{1.0pt}
	\centering
	\small
	\begin{tabularx}{1\textwidth}{cc}
        \includegraphics[width=0.495\textwidth, height = 0.28\textwidth] {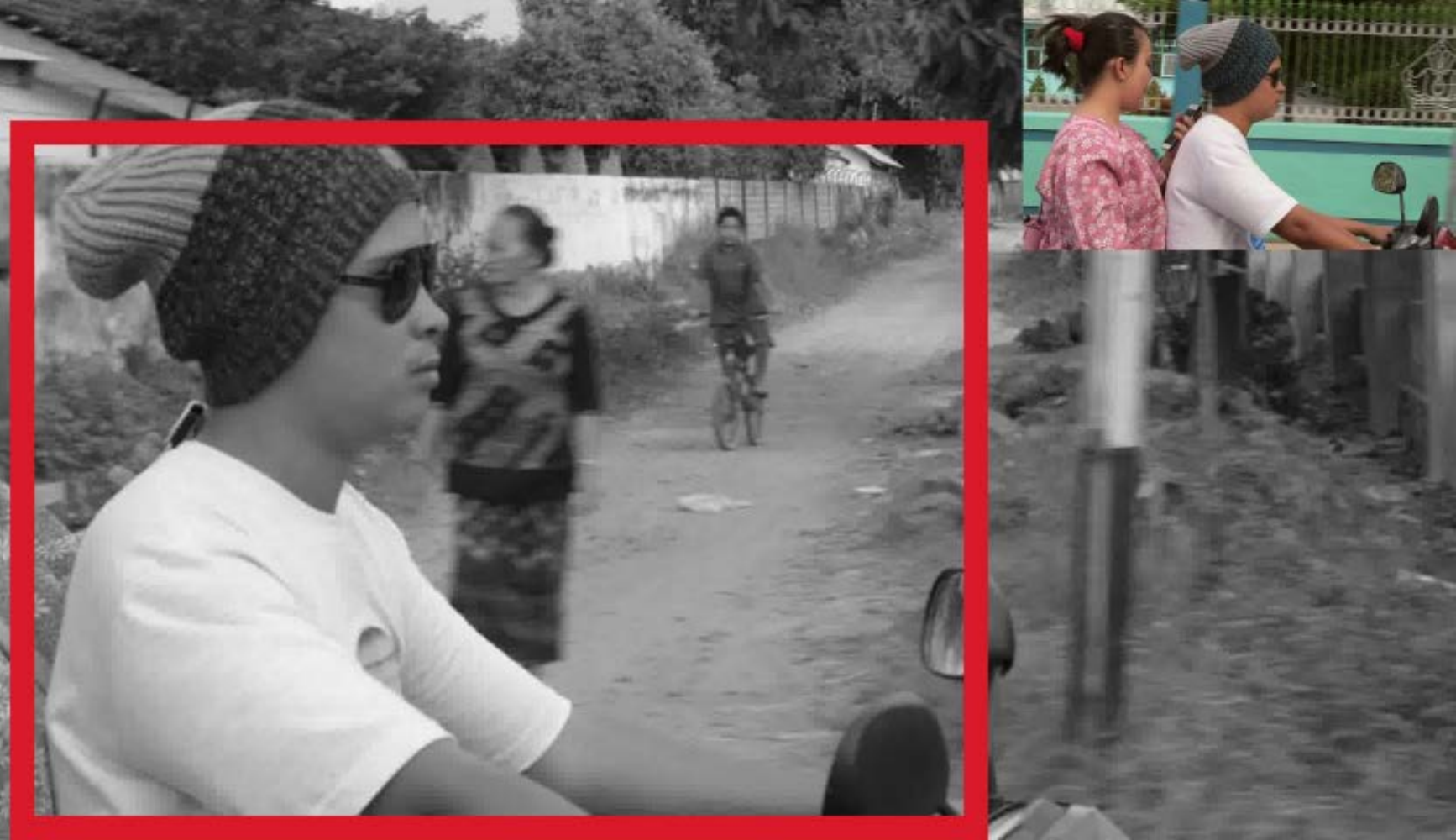} &
        \includegraphics[width=0.495\textwidth, height = 0.28\textwidth] {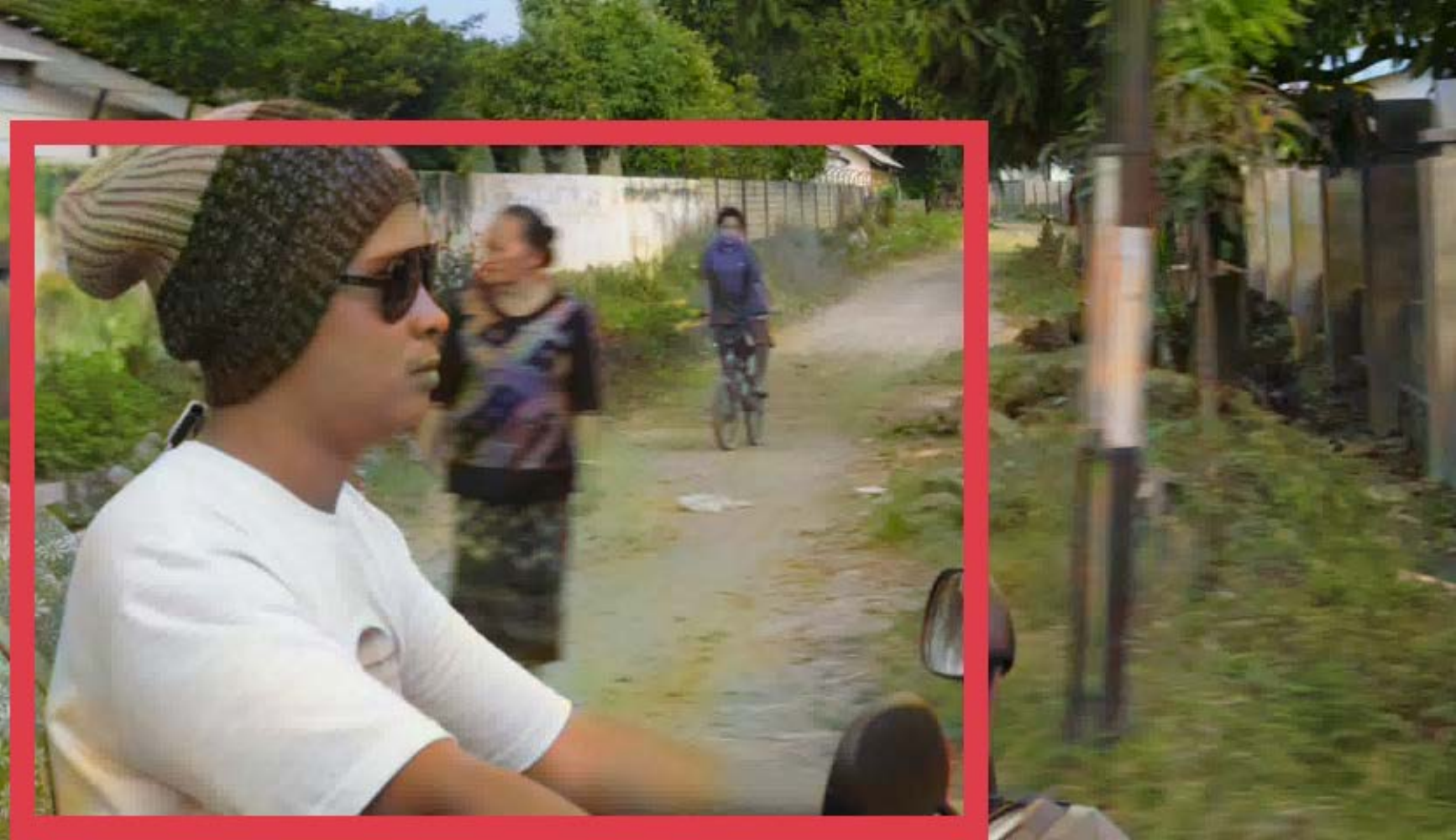} 
        \\
        \makebox[0.45\textwidth]{ (a) Input frame and exemplar image} &
        \makebox[0.45\textwidth]{ (b) DDColor~\cite{kang2022ddcolor}} 
        \\ 
        \includegraphics[width=0.495\textwidth, height = 0.28\textwidth] {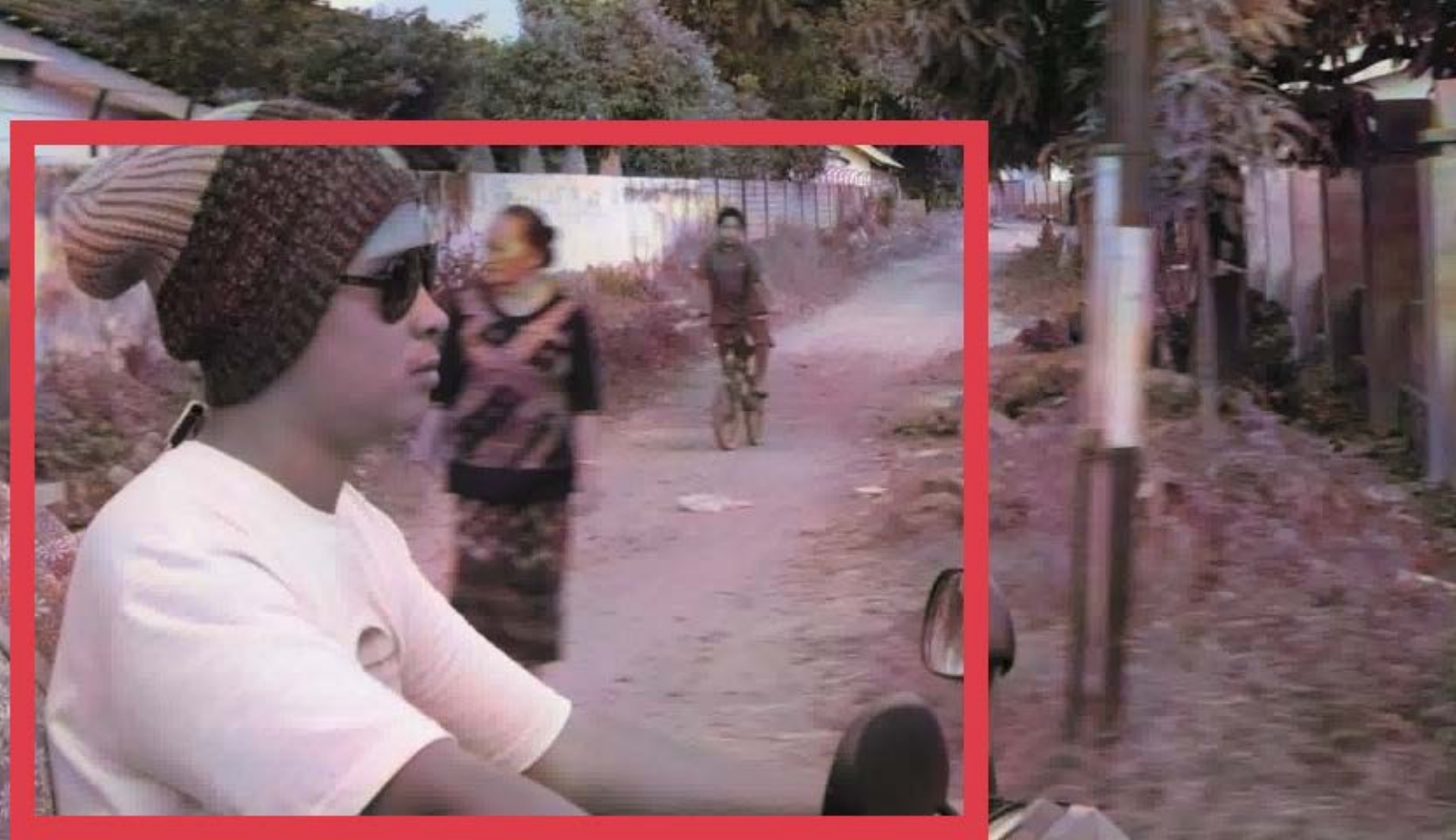} &
        \includegraphics[width=0.495\textwidth, height = 0.28\textwidth] {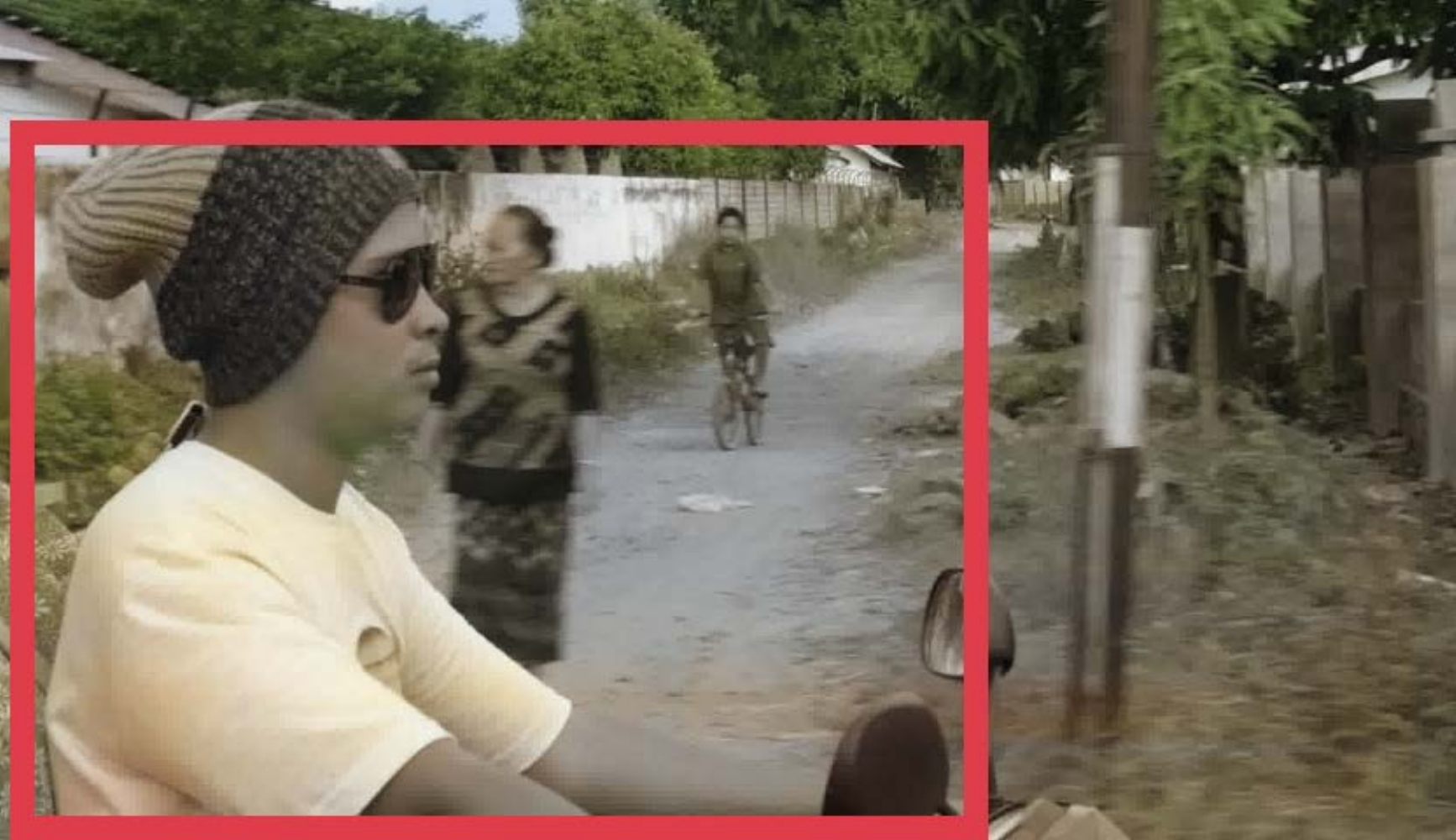} 
        \\ 
        \makebox[0.45\textwidth]{ (c) Color2Embed~\cite{zhao2021color2embed}} &
        \makebox[0.45\textwidth]{ (d) TCVC~\cite{liu2021temporally}} \\
        \includegraphics[width=0.495\textwidth, height = 0.28\textwidth] {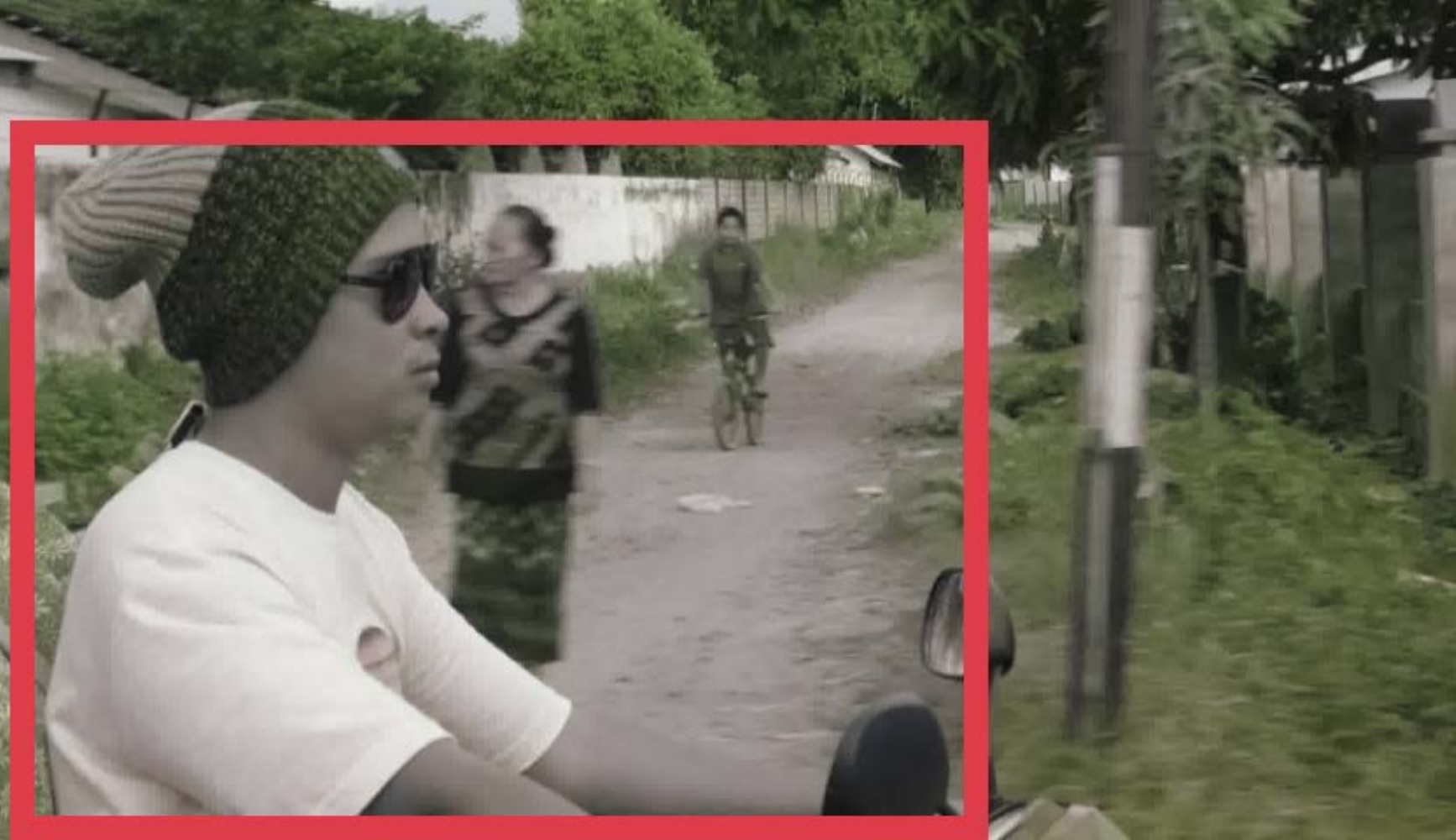} &
        \includegraphics[width=0.495\textwidth, height = 0.28\textwidth] {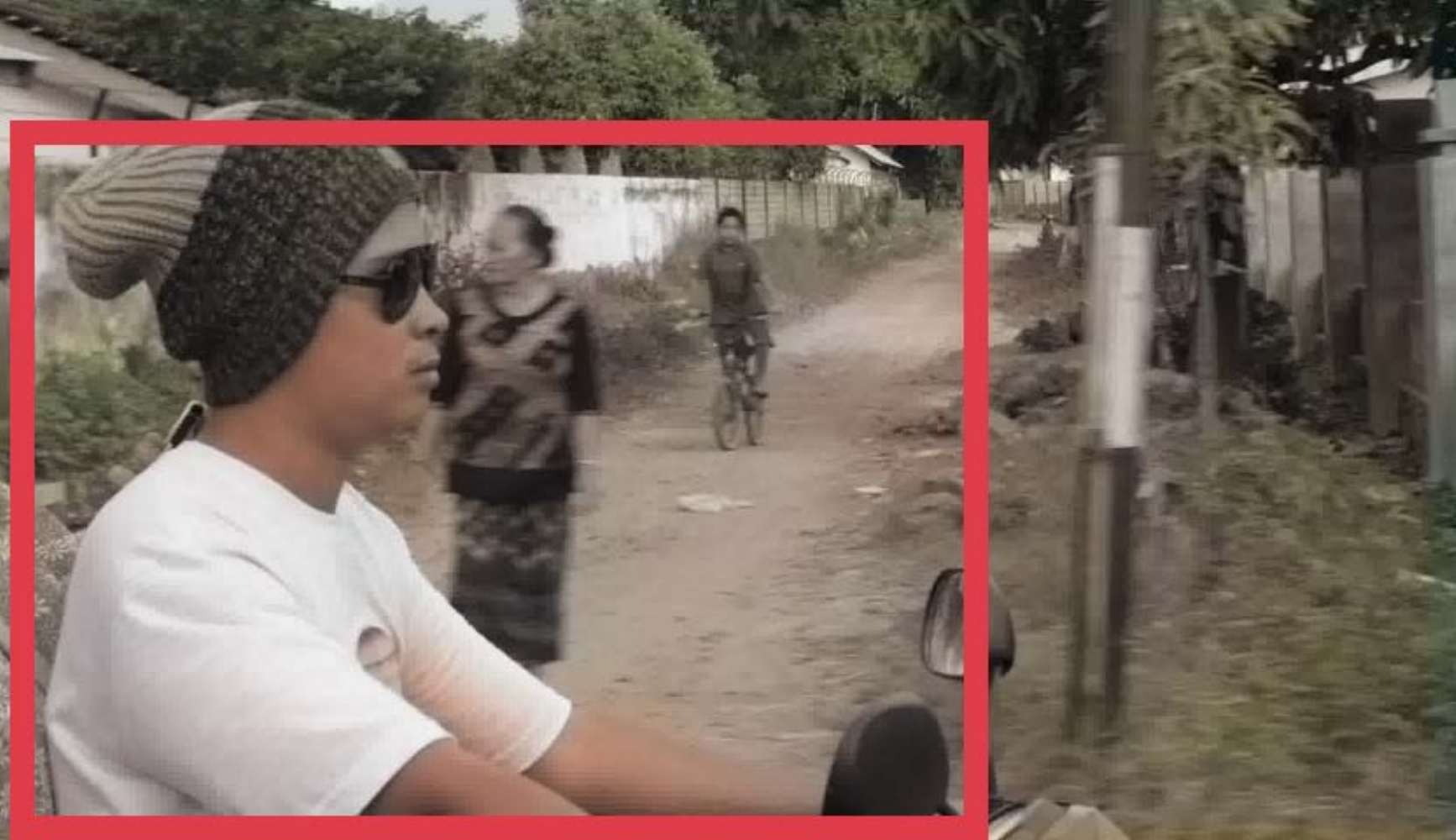} 
        \\
        \makebox[0.45\textwidth]{ (e) VCGAN~\cite{vcgan}} &
        \makebox[0.45\textwidth]{ (f) DeepRemaster~\cite{IizukaSIGGRAPHASIA2019}} 
        \\ 
        \includegraphics[width=0.495\textwidth, height = 0.28\textwidth] {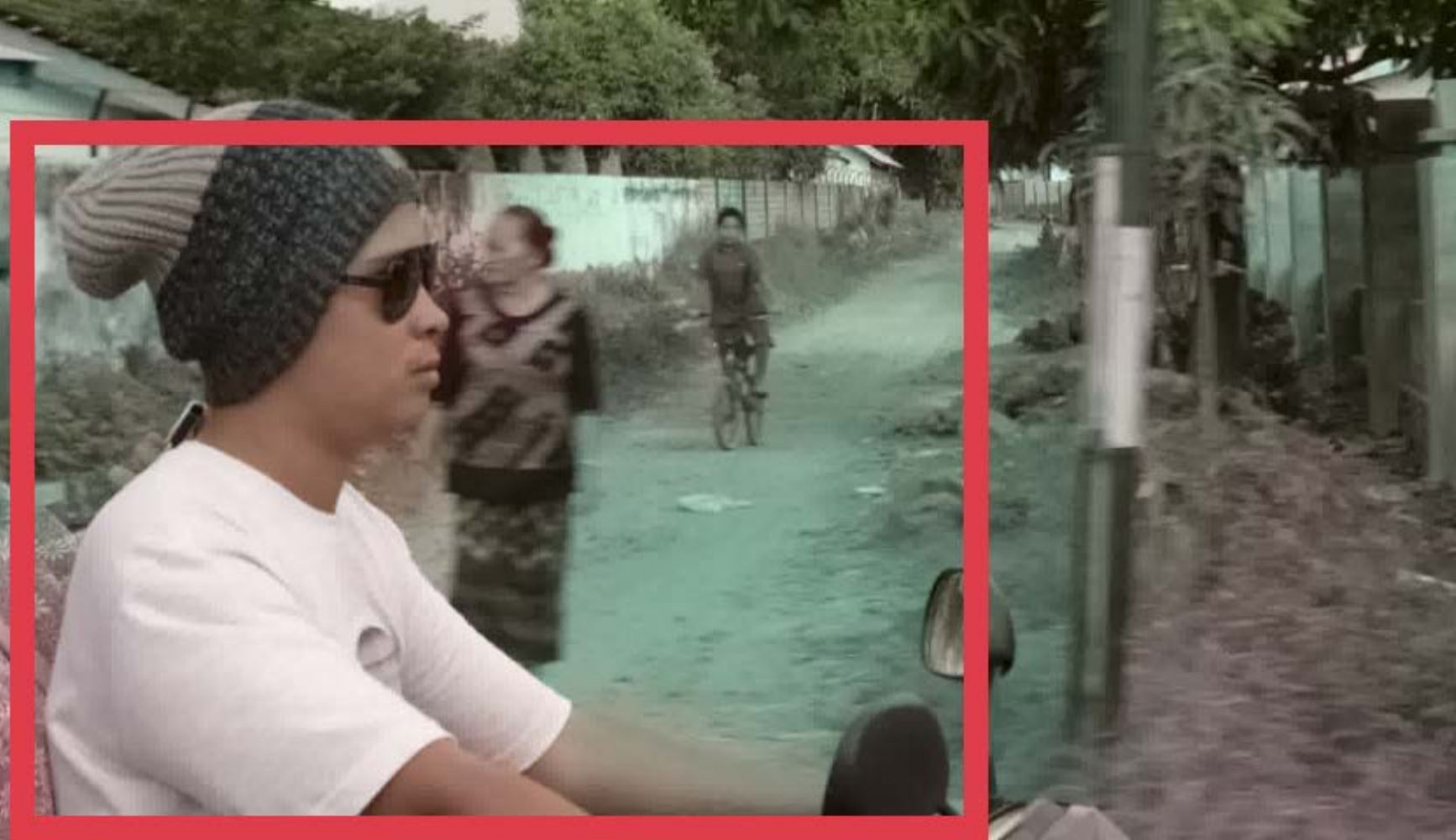} &
        \includegraphics[width=0.495\textwidth, height = 0.28\textwidth] {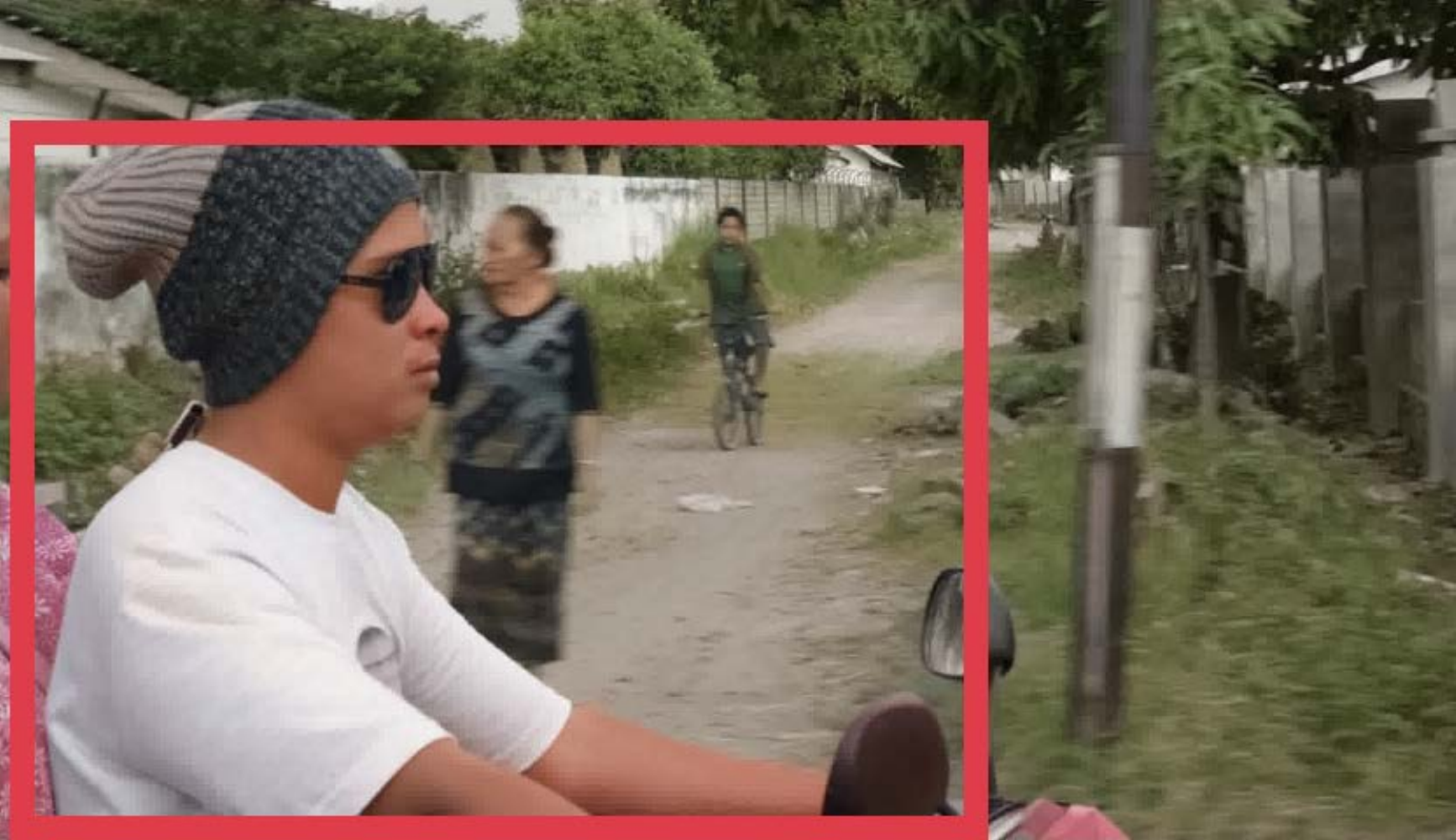} 
        \\
        \makebox[0.45\textwidth]{ (g) DeepExemplar~\cite{zhang2019deep}} &
        \makebox[0.45\textwidth]{ (h) ColorMNet (Ours)}  
         %& \vspace{-0.7em}
	\end{tabularx}
	\vspace{-0.8em}
	\caption{{Colorization results on clip \textit{CoupleRidingMotorbike} from the Videvo validation dataset~\cite{Lai2018videvo}. The results shown in (b) and (c) still contain significant color-bleeding artifacts. In contrast, our proposed method generates a realistic frame that is faithful to the exemplar image.}}
	\label{fig:6}
	%\vspace{-3mm}
\end{figure*}

\begin{figure*}[!htb]
	\setlength\tabcolsep{1.0pt}
	\centering
	\small
	\begin{tabularx}{1\textwidth}{cc}
        \includegraphics[width=0.495\textwidth, height = 0.28\textwidth] {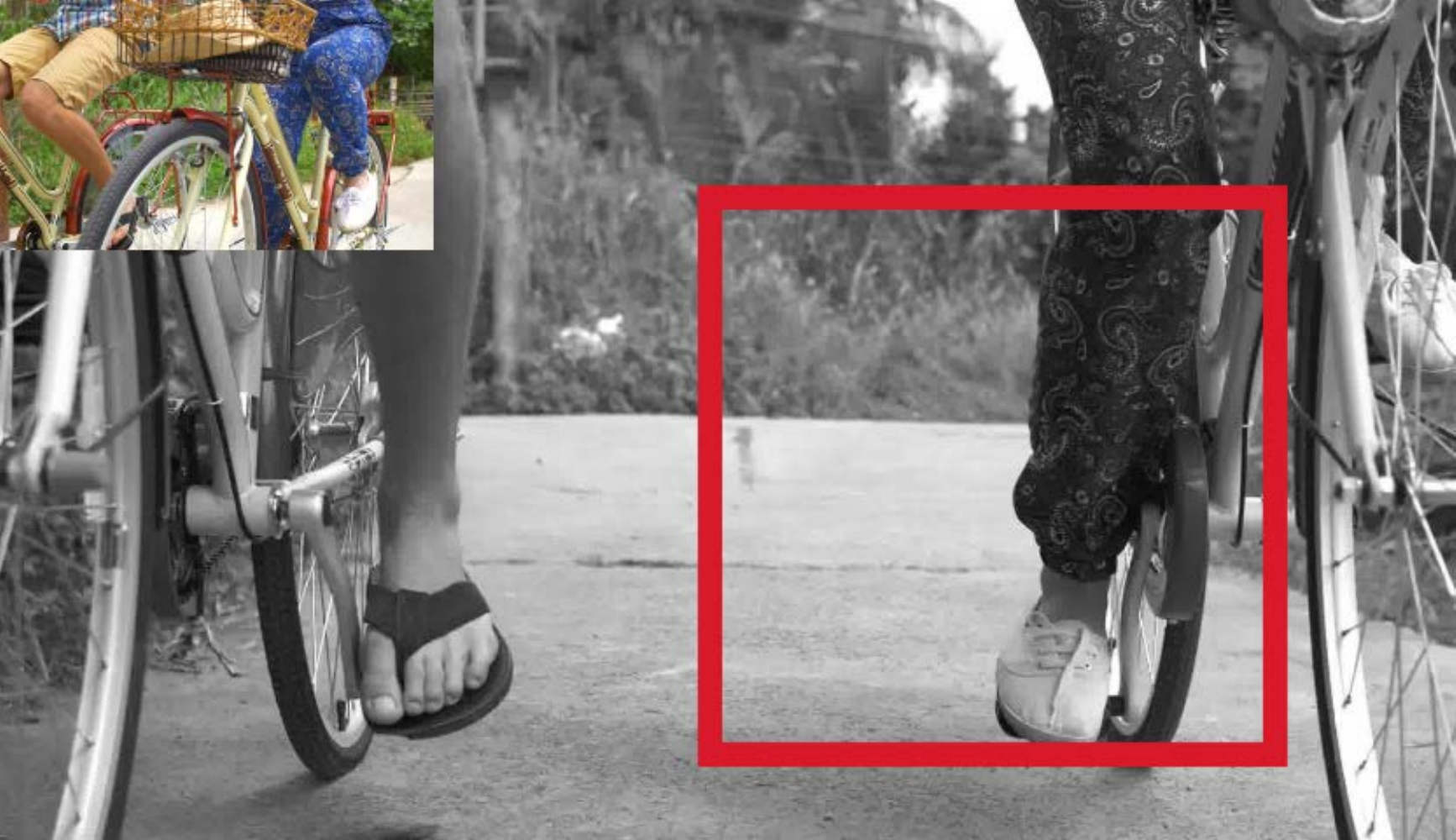} &
        \includegraphics[width=0.495\textwidth, height = 0.28\textwidth] {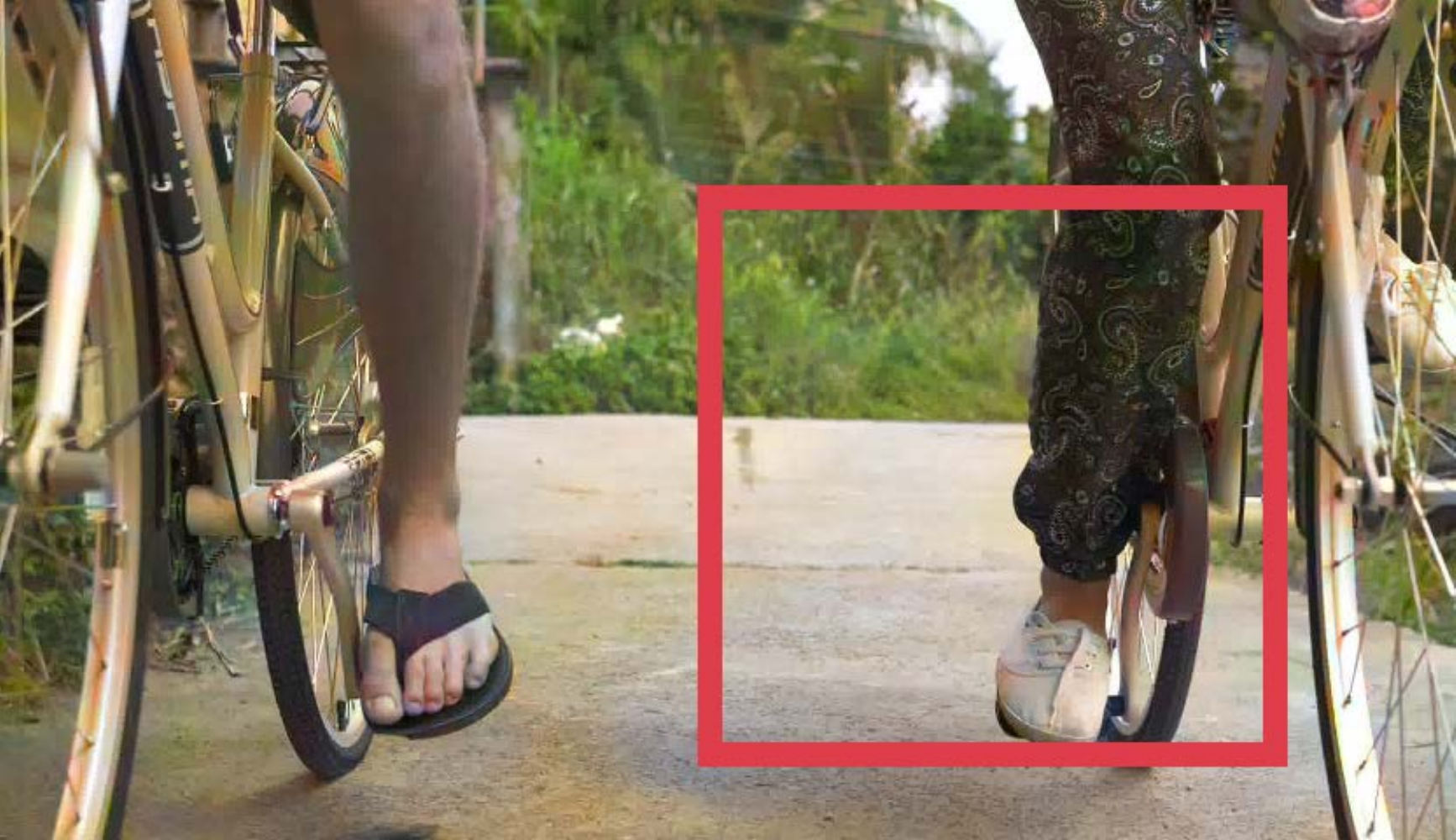} 
        \\
        \makebox[0.45\textwidth]{ (a) Input frame and exemplar image} &
        \makebox[0.45\textwidth]{ (b) DDColor~\cite{kang2022ddcolor}} 
        \\ 
        \includegraphics[width=0.495\textwidth, height = 0.28\textwidth] {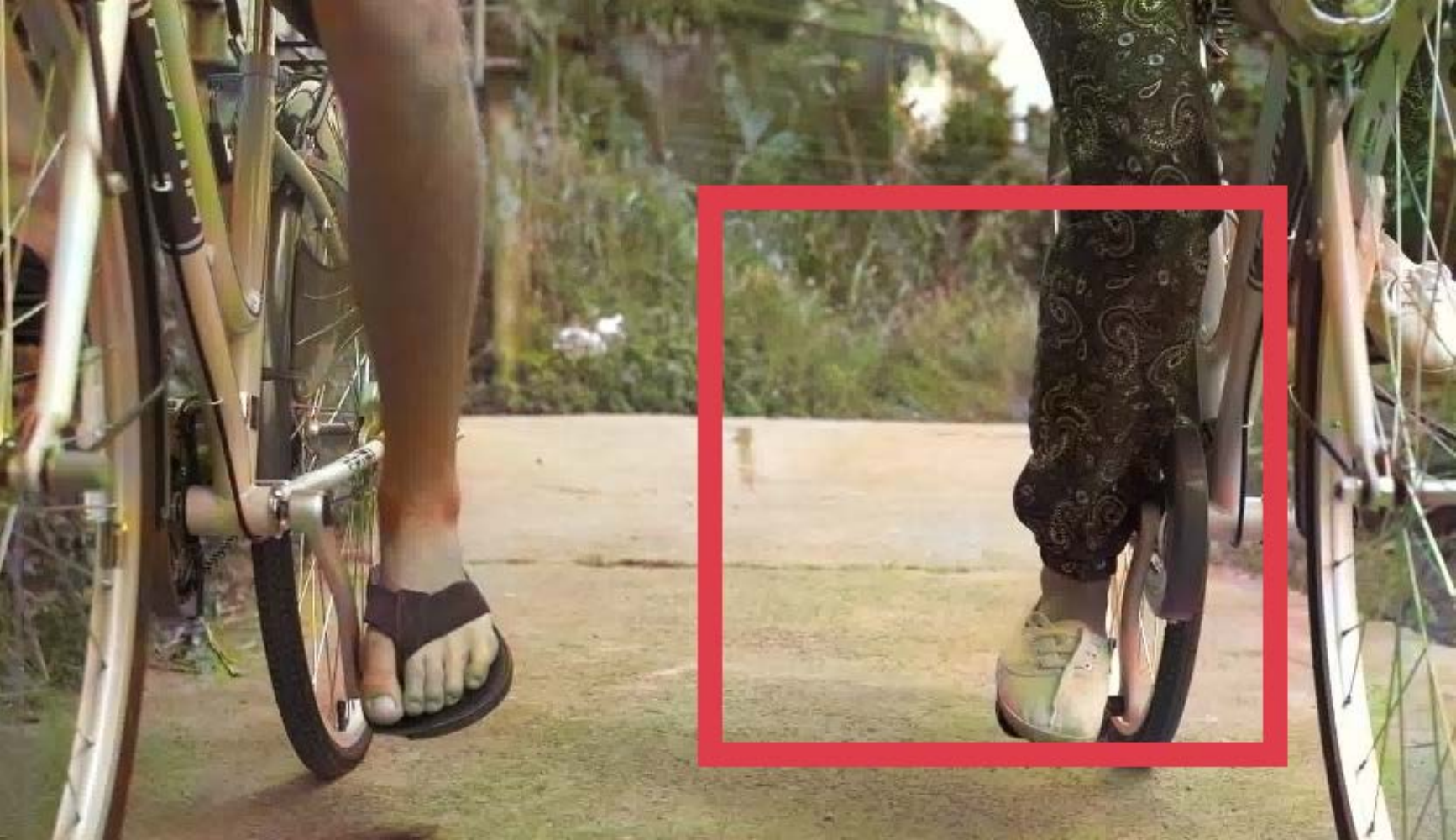} &
        \includegraphics[width=0.495\textwidth, height = 0.28\textwidth] {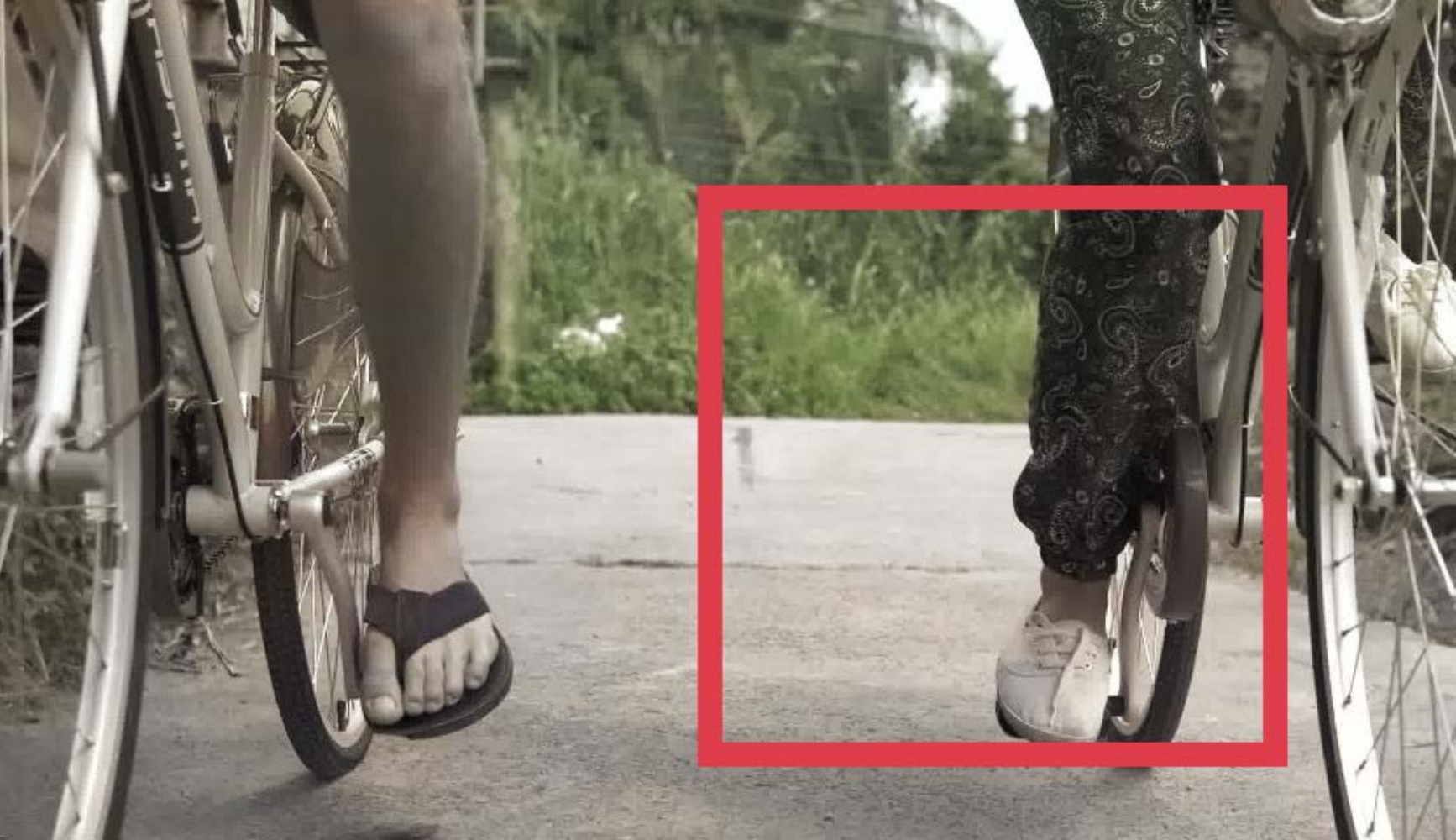} 
        \\ 
        \makebox[0.45\textwidth]{ (c) Color2Embed~\cite{zhao2021color2embed}} &
        \makebox[0.45\textwidth]{ (d) TCVC~\cite{liu2021temporally}} \\
        \includegraphics[width=0.495\textwidth, height = 0.28\textwidth] {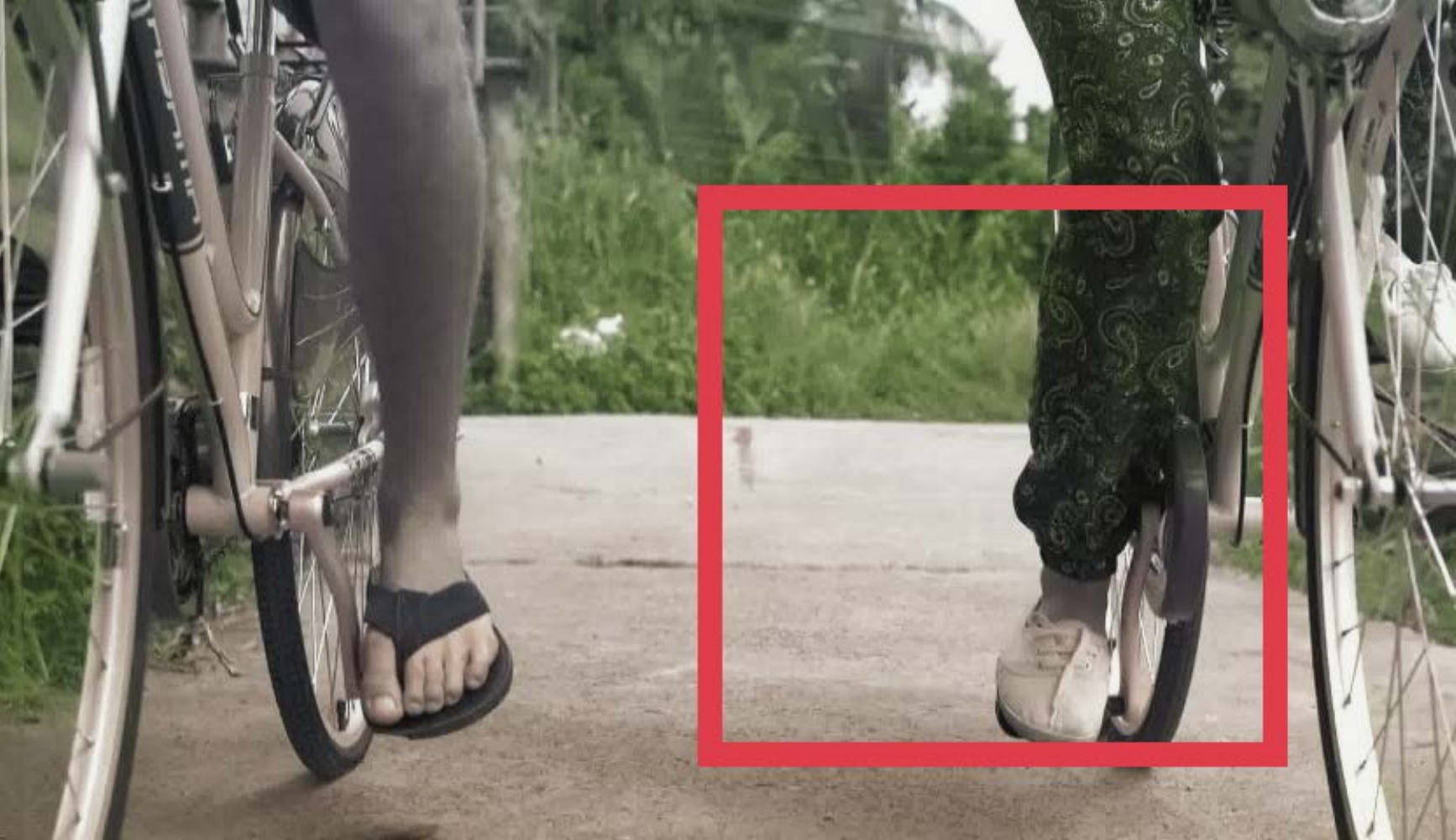} &
        \includegraphics[width=0.495\textwidth, height = 0.28\textwidth] {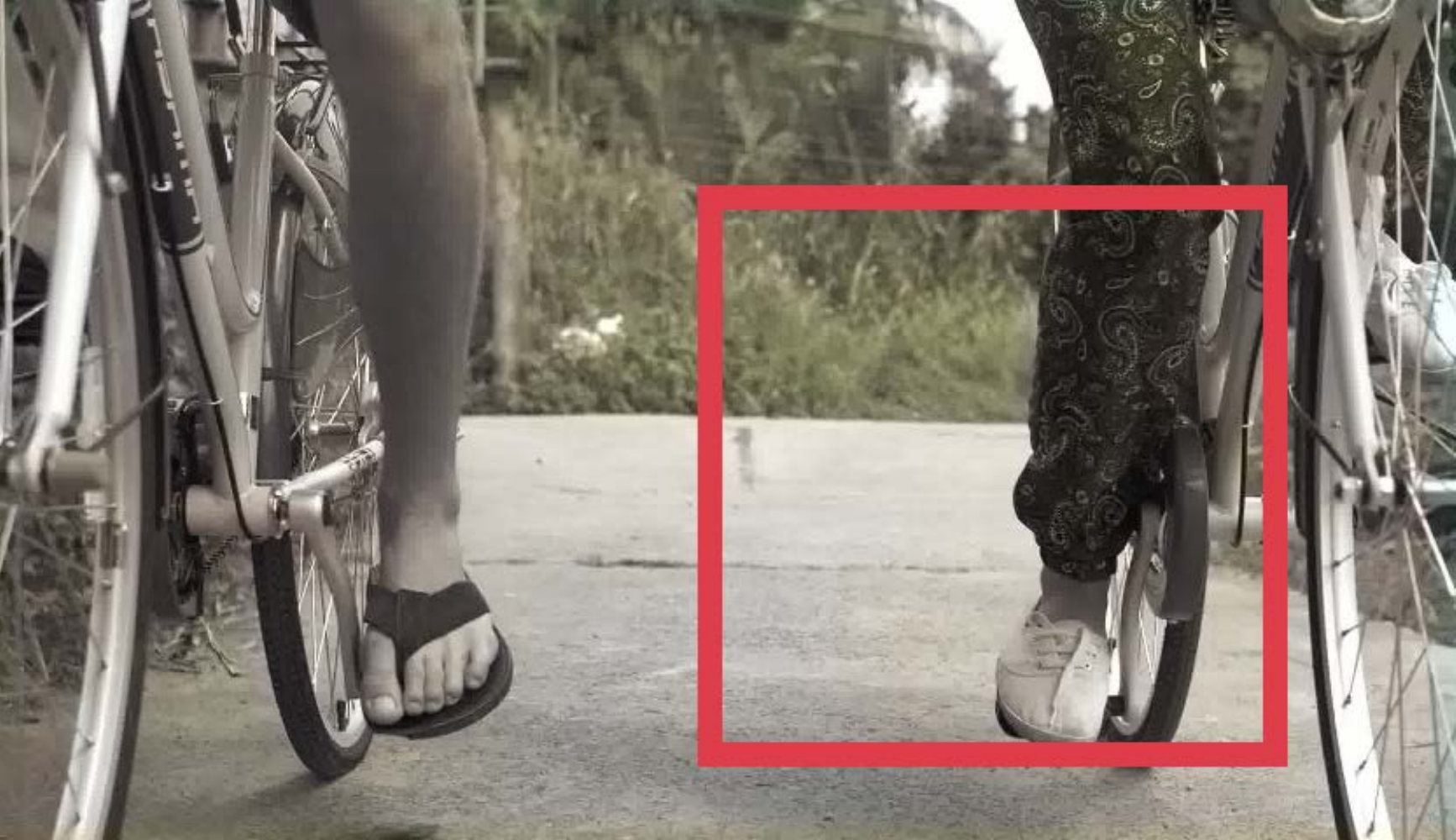} 
        \\
        \makebox[0.45\textwidth]{ (e) VCGAN~\cite{vcgan}} &
        \makebox[0.45\textwidth]{ (f) DeepRemaster~\cite{IizukaSIGGRAPHASIA2019}} 
        \\ 
        \includegraphics[width=0.495\textwidth, height = 0.28\textwidth] {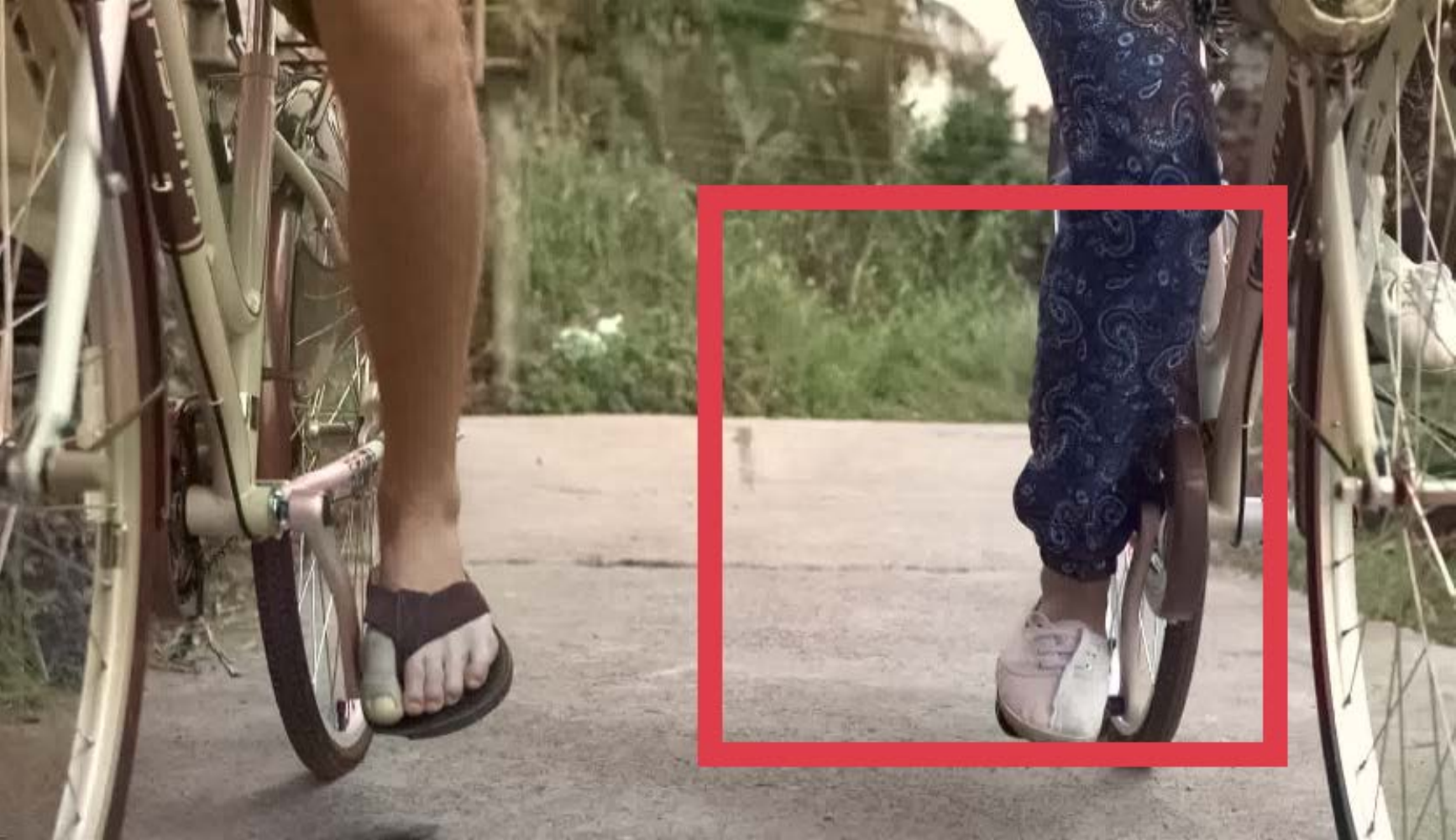} &
        \includegraphics[width=0.495\textwidth, height = 0.28\textwidth] {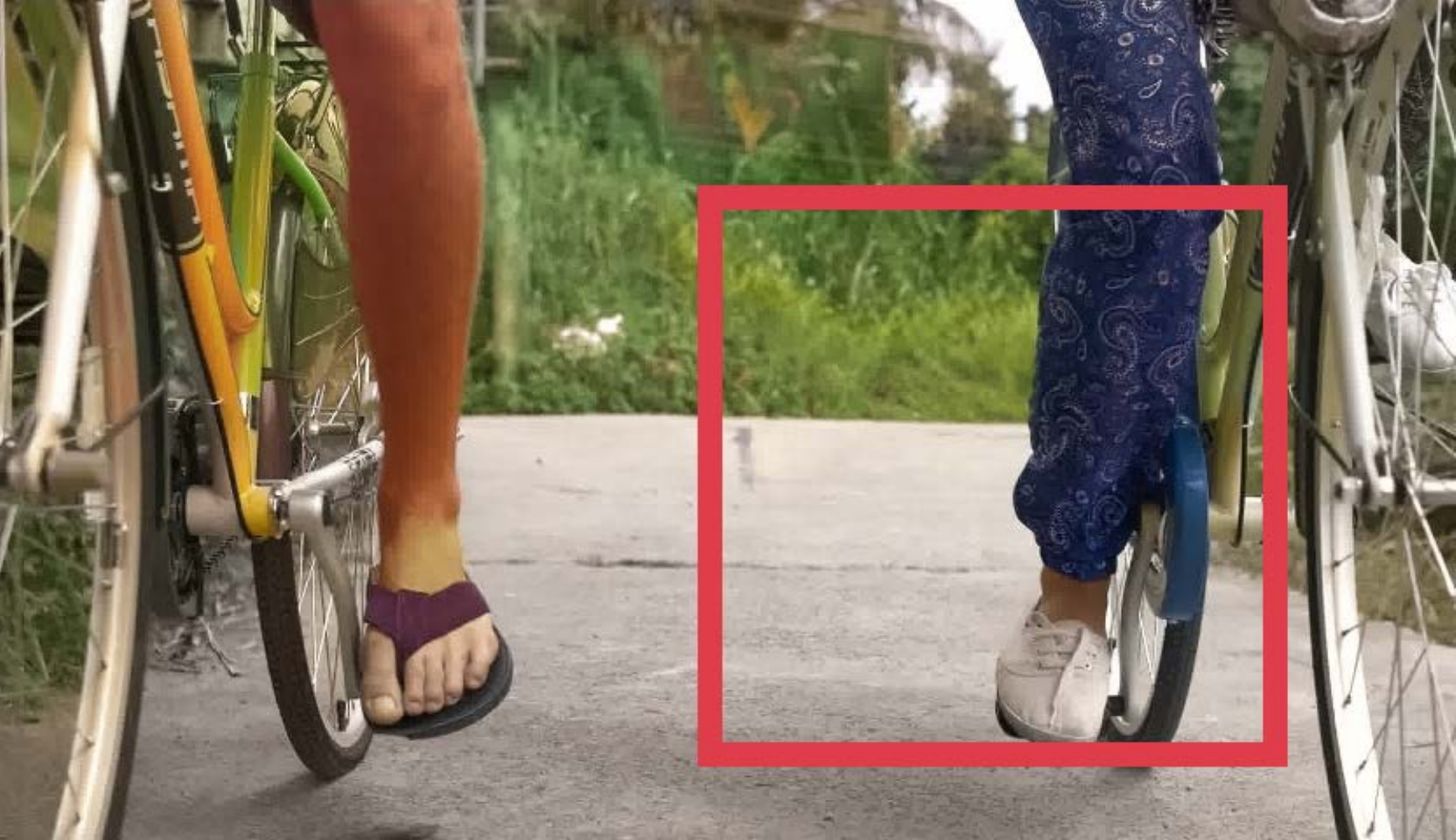} 
        \\
        \makebox[0.45\textwidth]{ (g) DeepExemplar~\cite{zhang2019deep}} &
        \makebox[0.45\textwidth]{ (h) ColorMNet (Ours)}  
         %& \vspace{-0.7em}
	\end{tabularx}
	\vspace{-0.8em}
	\caption{{Colorization results on clip \textit{Cycling} from the Videvo validation dataset~\cite{Lai2018videvo}. The results shown in (b) and (c) still contain significant color-bleeding artifacts. In contrast, our proposed method generates a vivid frame than other stage-of-the-art methods.}}
	\label{fig:7}
	%\vspace{-3mm}
\end{figure*}

\begin{figure*}[!htb]
	\setlength\tabcolsep{1.0pt}
	\centering
	\small
	\begin{tabularx}{1\textwidth}{cc}
        \includegraphics[width=0.495\textwidth, height = 0.28\textwidth] {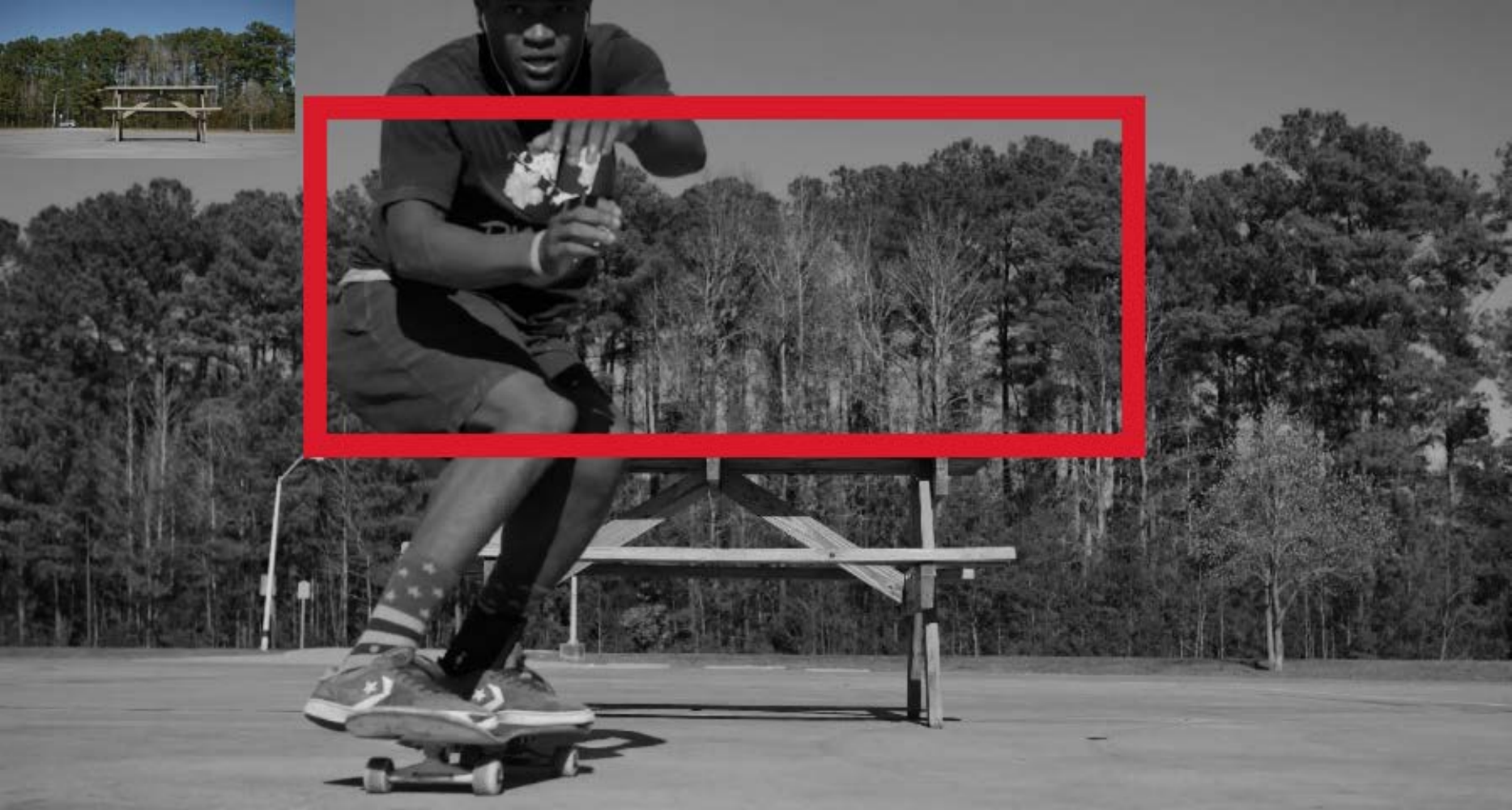} &
        \includegraphics[width=0.495\textwidth, height = 0.28\textwidth] {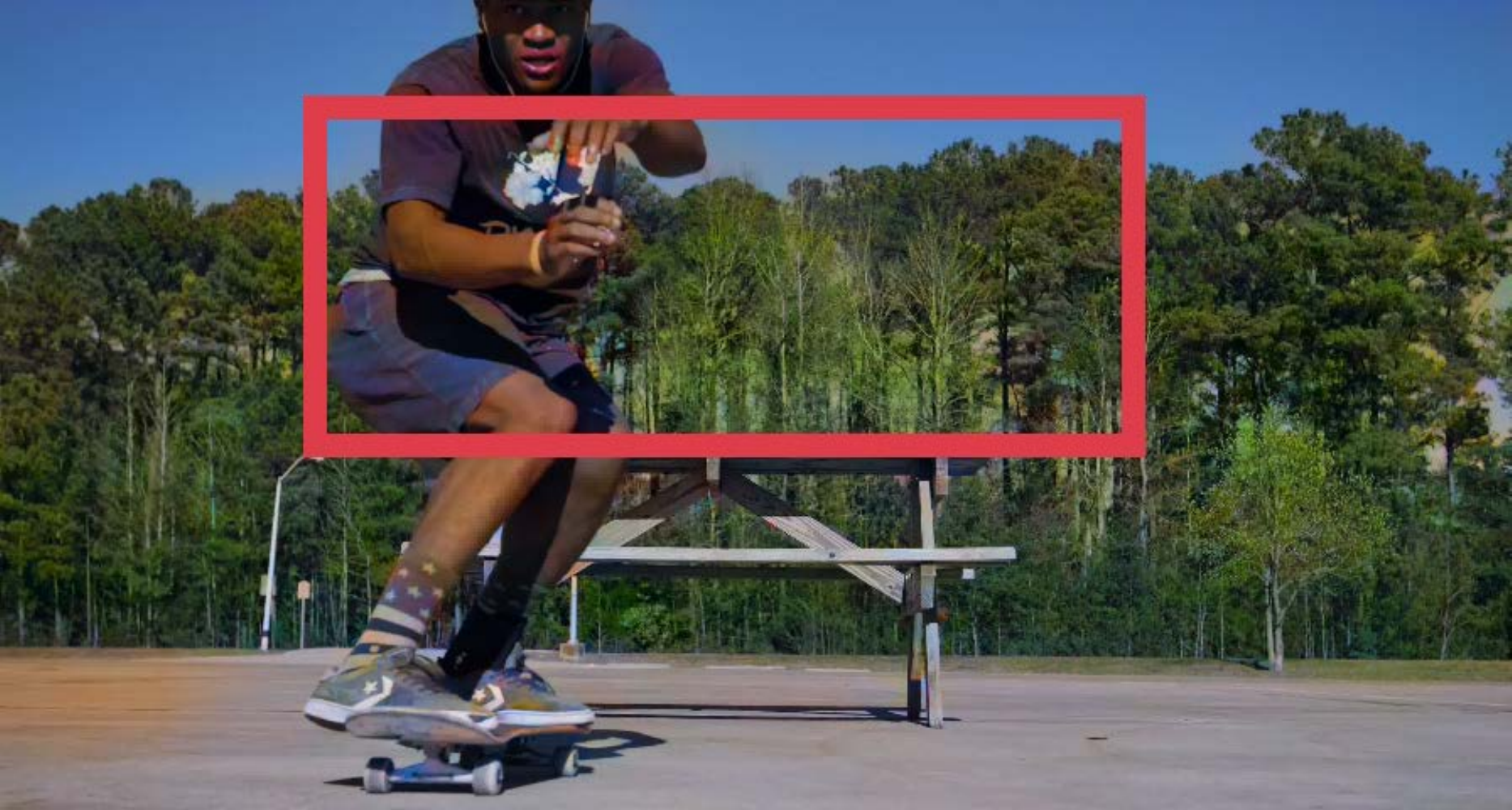} 
        \\
        \makebox[0.45\textwidth]{ (a) Input frame and exemplar image} &
        \makebox[0.45\textwidth]{ (b) DDColor~\cite{kang2022ddcolor}} 
        \\ 
        \includegraphics[width=0.495\textwidth, height = 0.28\textwidth] {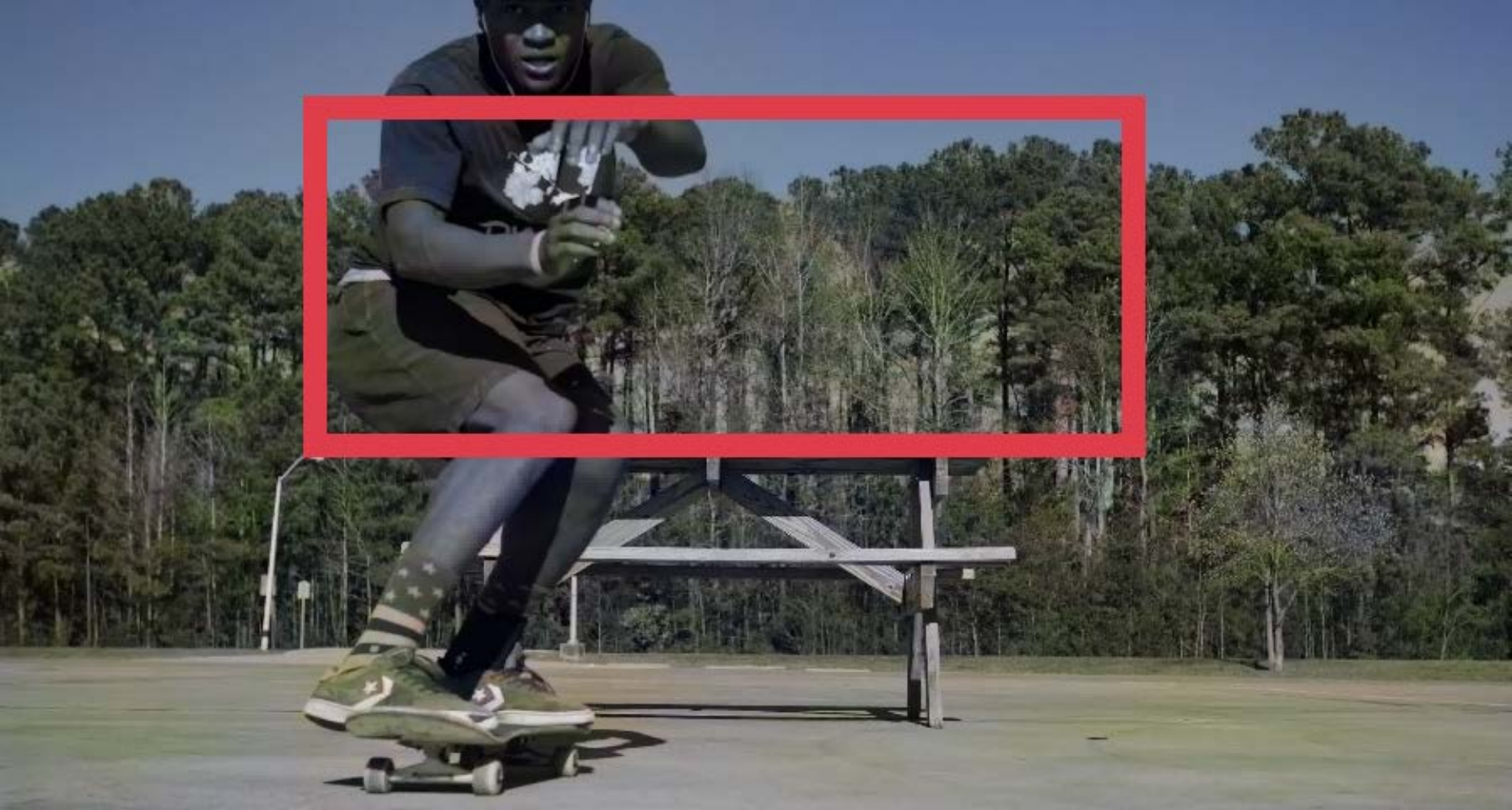} &
        \includegraphics[width=0.495\textwidth, height = 0.28\textwidth] {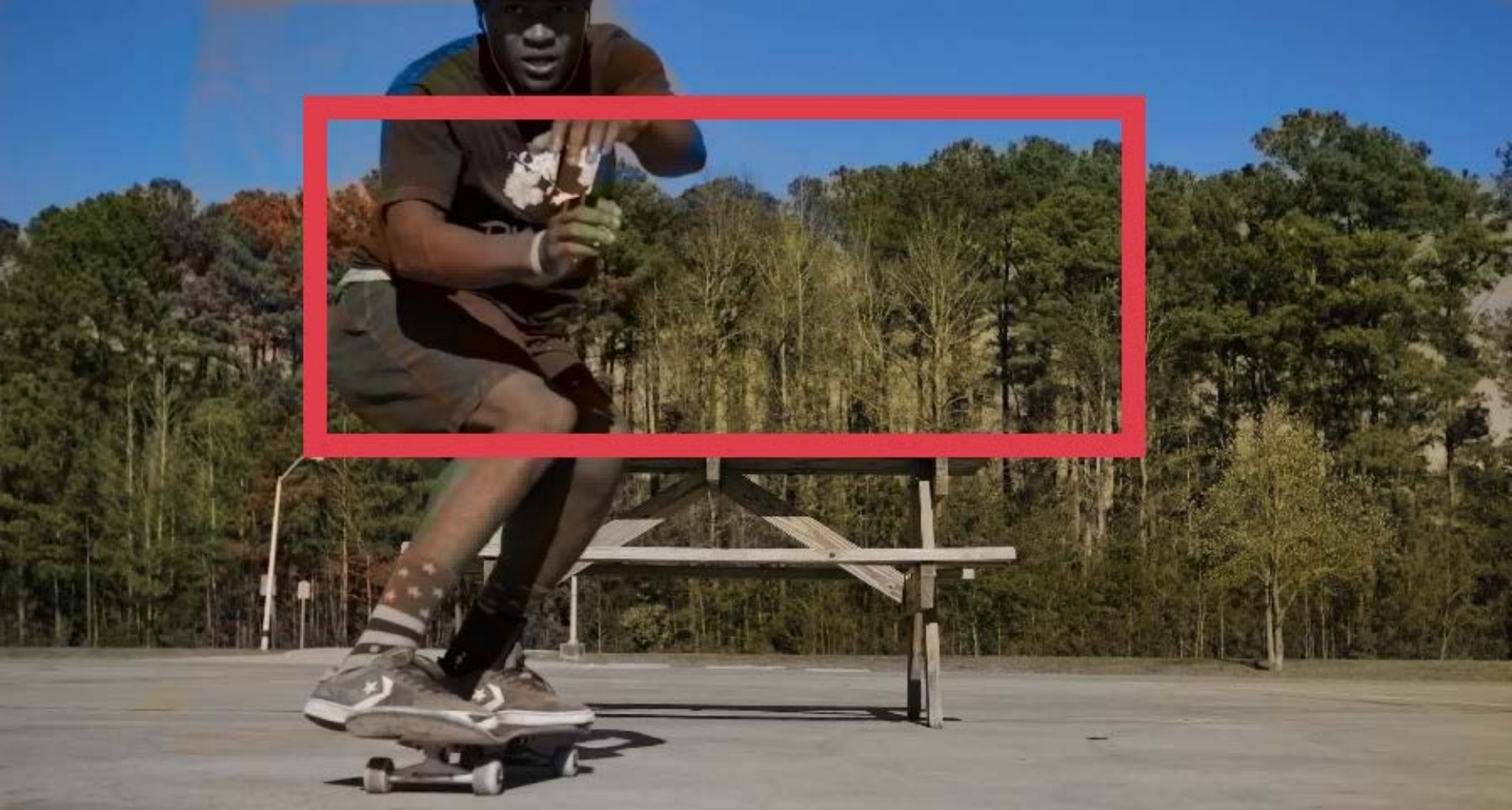} 
        \\ 
        \makebox[0.45\textwidth]{ (c) Color2Embed~\cite{zhao2021color2embed}} &
        \makebox[0.45\textwidth]{ (d) TCVC~\cite{liu2021temporally}} \\
        \includegraphics[width=0.495\textwidth, height = 0.28\textwidth] {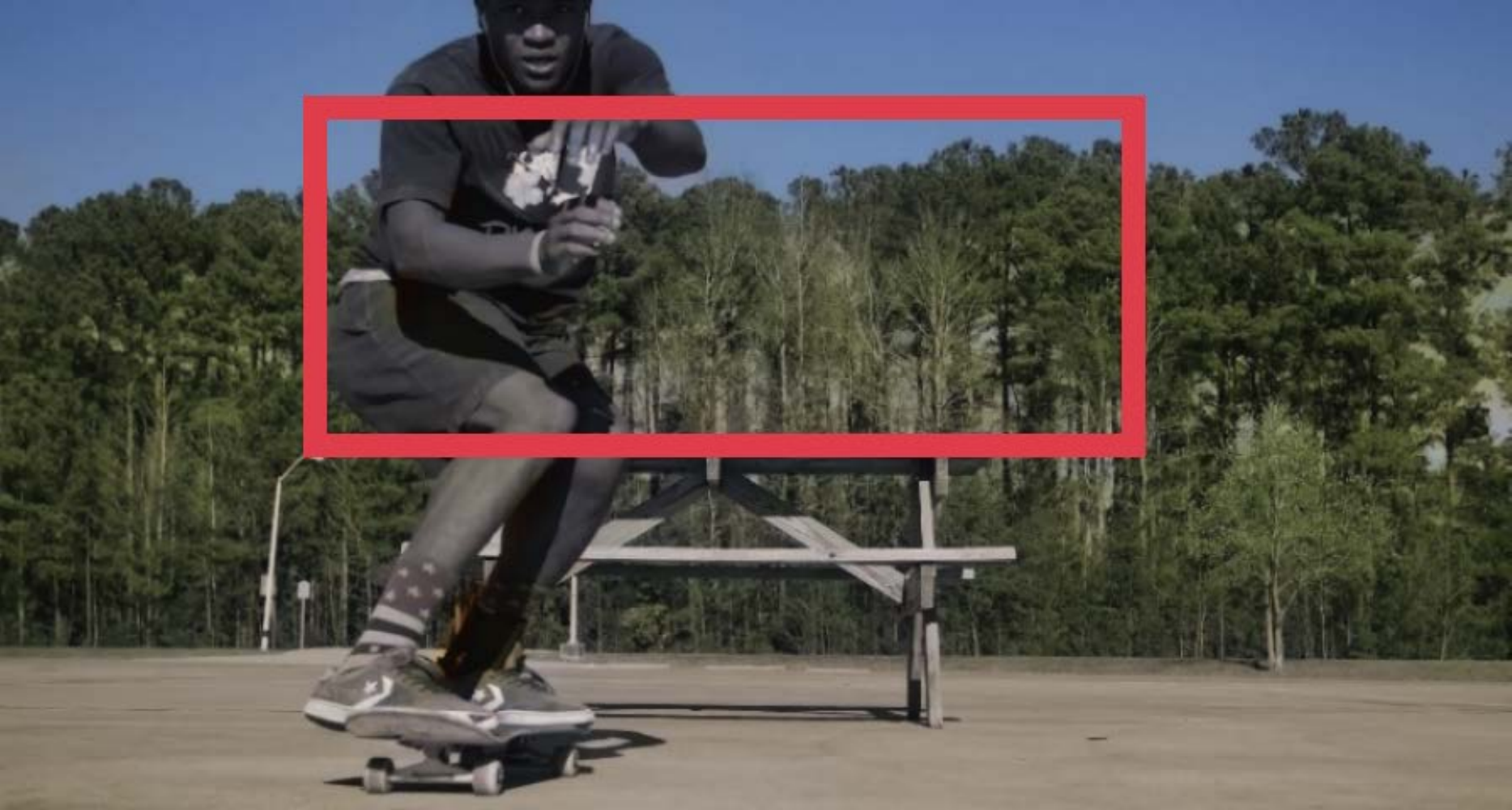} &
        \includegraphics[width=0.495\textwidth, height = 0.28\textwidth] {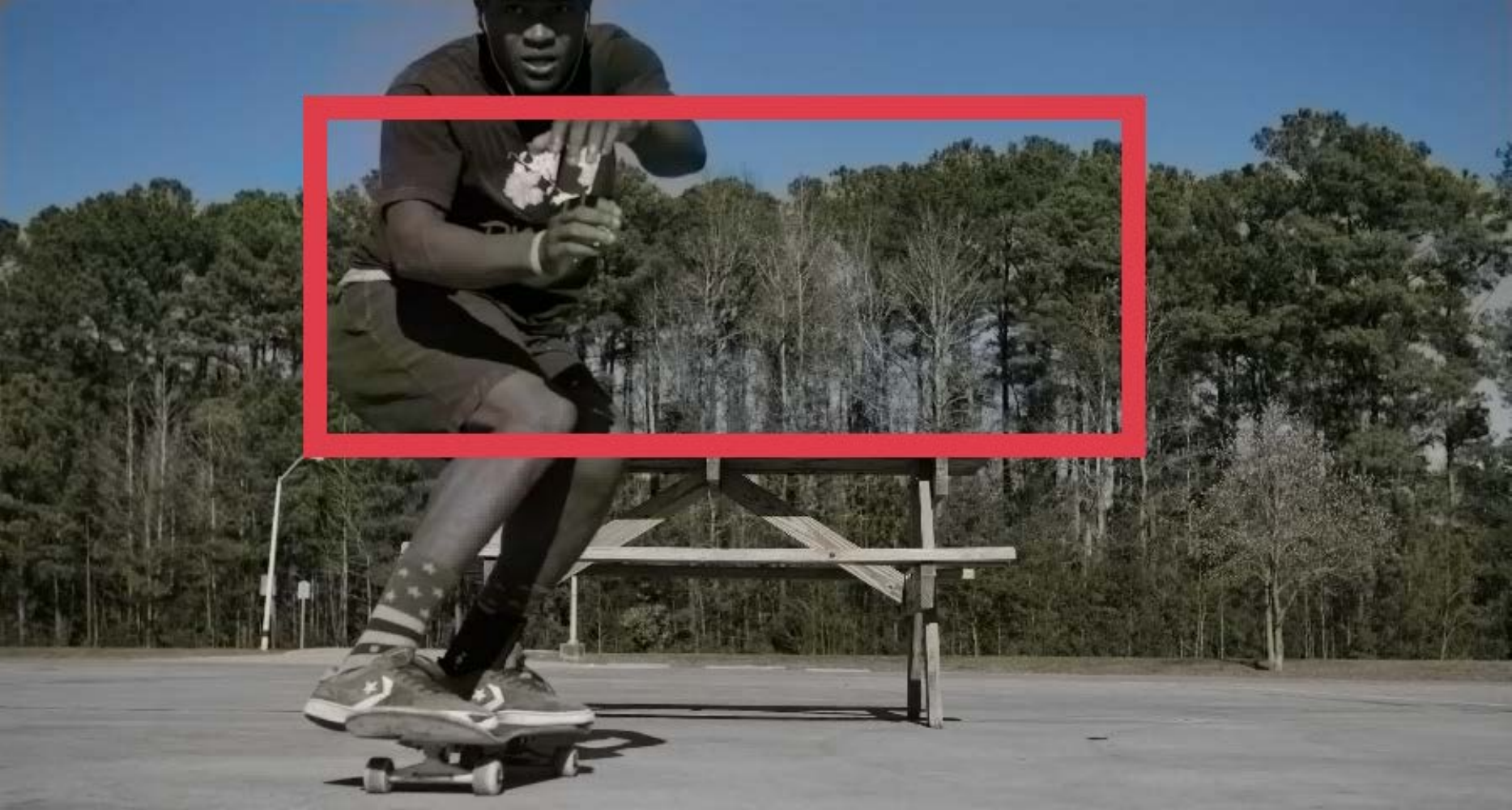} 
        \\
        \makebox[0.45\textwidth]{ (e) VCGAN~\cite{vcgan}} &
        \makebox[0.45\textwidth]{ (f) DeepRemaster~\cite{IizukaSIGGRAPHASIA2019}} 
        \\ 
        \includegraphics[width=0.495\textwidth, height = 0.28\textwidth] {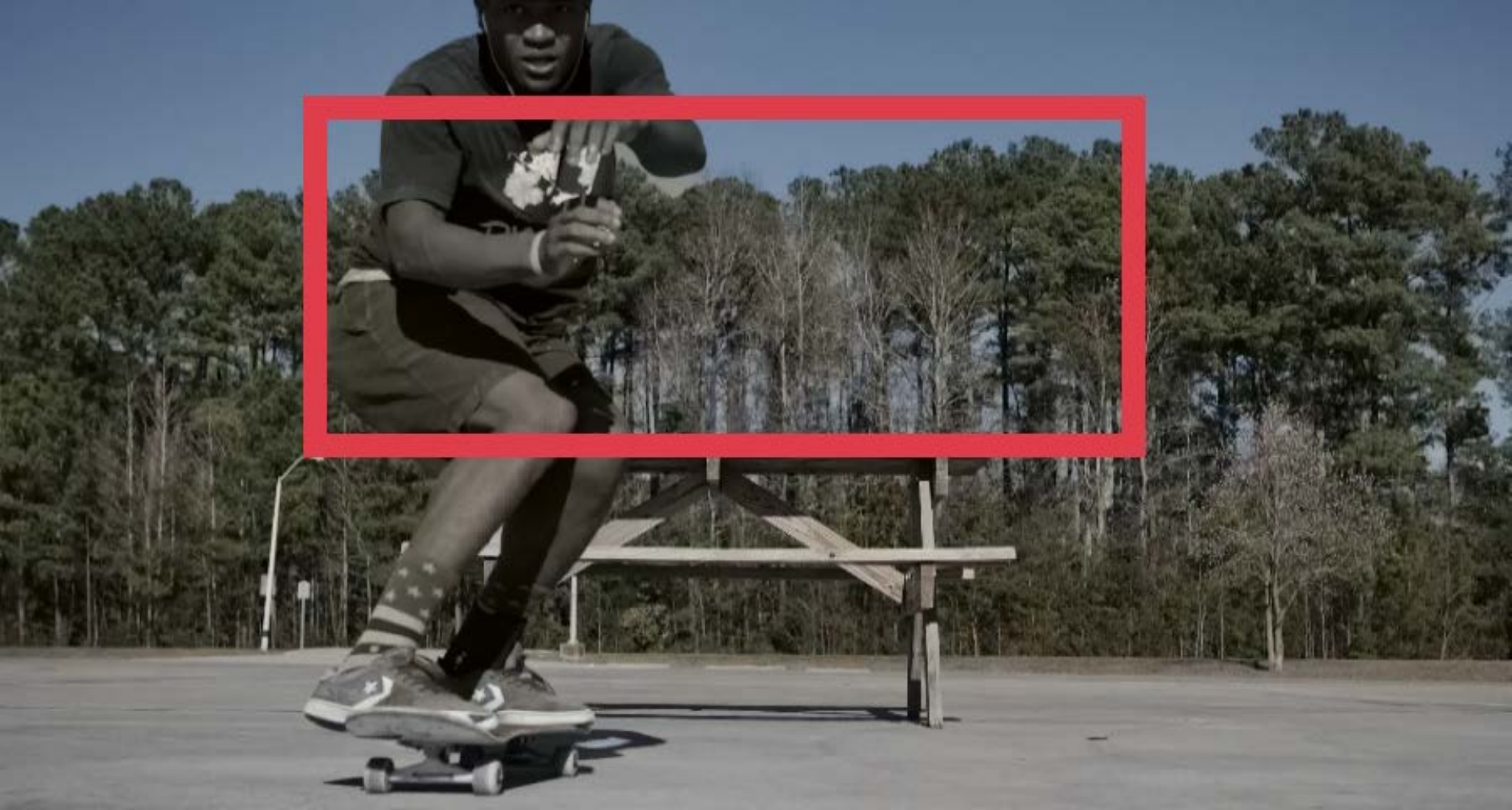} &
        \includegraphics[width=0.495\textwidth, height = 0.28\textwidth] {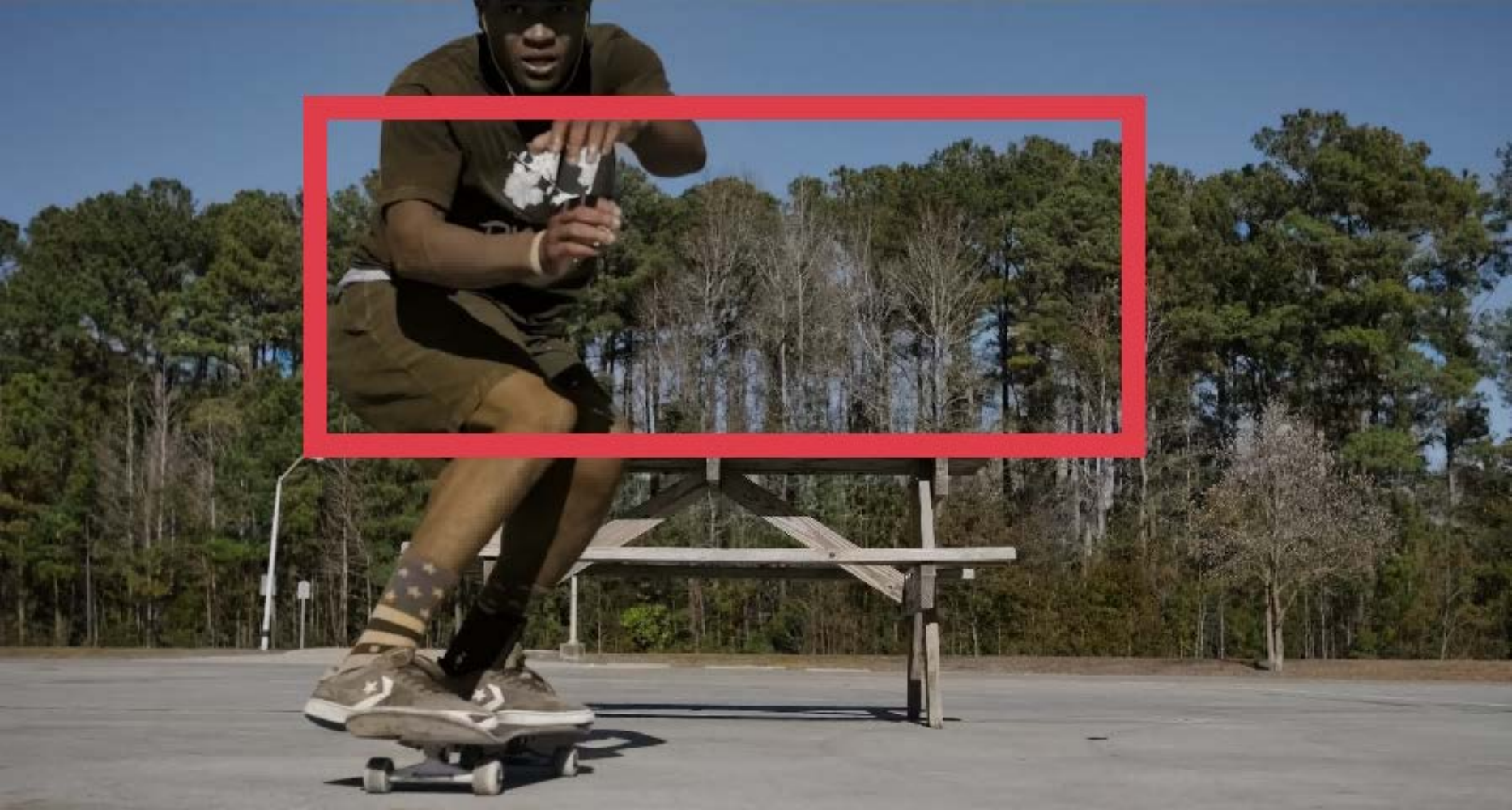} 
        \\
        \makebox[0.45\textwidth]{ (g) DeepExemplar~\cite{zhang2019deep}} &
        \makebox[0.45\textwidth]{ (h) ColorMNet (Ours)}  
         %& \vspace{-0.7em}
	\end{tabularx}
	\vspace{-0.8em}
	\caption{{Colorization results on clip \textit{SkateboarderTableJump} from the Videvo validation dataset~\cite{Lai2018videvo}. The results shown in (b) and (c) still contain significant color-bleeding artifacts. In contrast, our proposed method generates a realistic frame, where the colors of the skateboard man and the trees are better restored.}}
	\label{fig:8}
	%\vspace{-3mm}
\end{figure*}

\begin{figure*}[!htb]
	\setlength\tabcolsep{1.0pt}
	\centering
	\small
	\begin{tabularx}{1\textwidth}{cc}
        \includegraphics[width=0.495\textwidth, height = 0.28\textwidth] {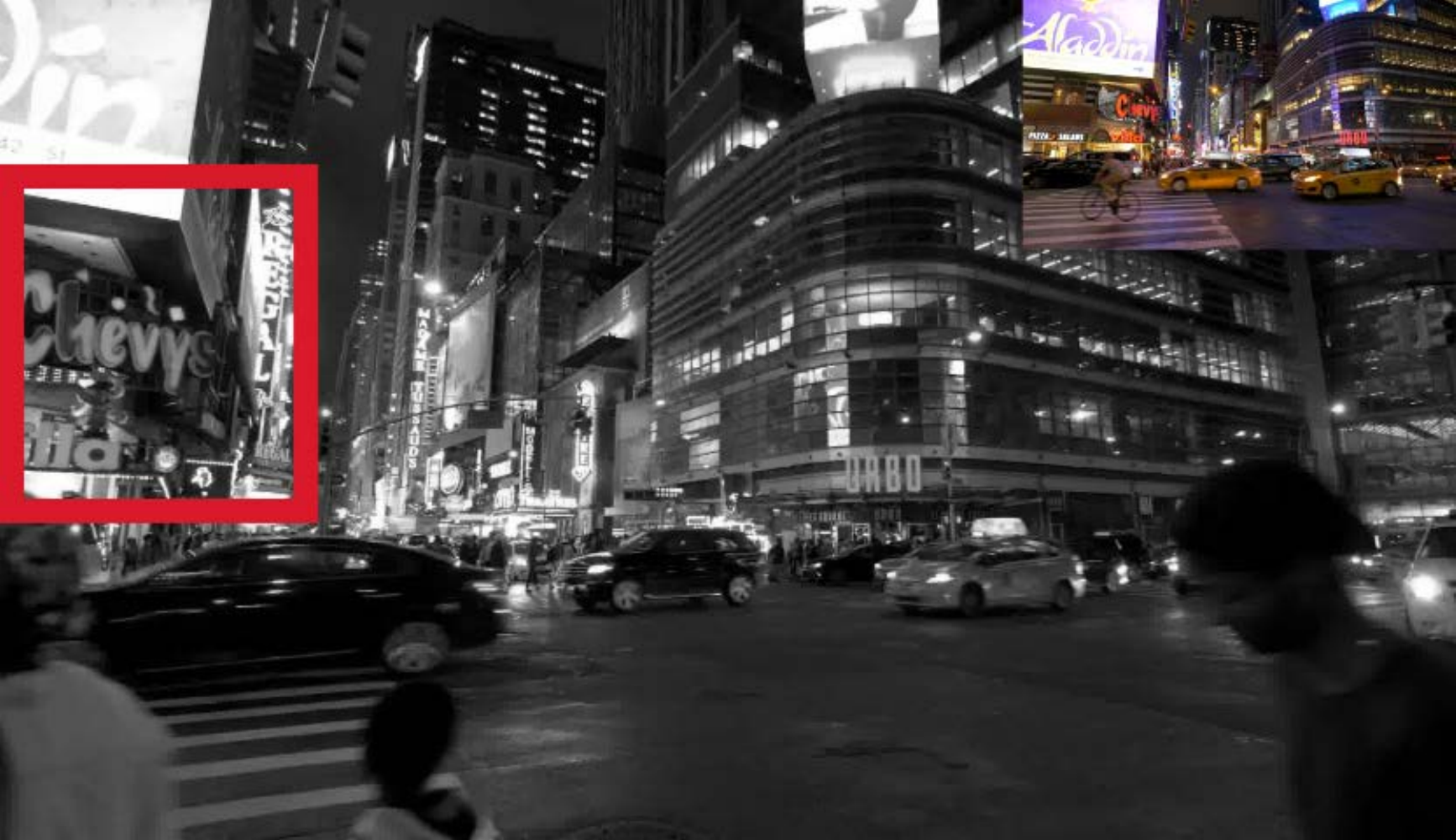} &
        \includegraphics[width=0.495\textwidth, height = 0.28\textwidth] {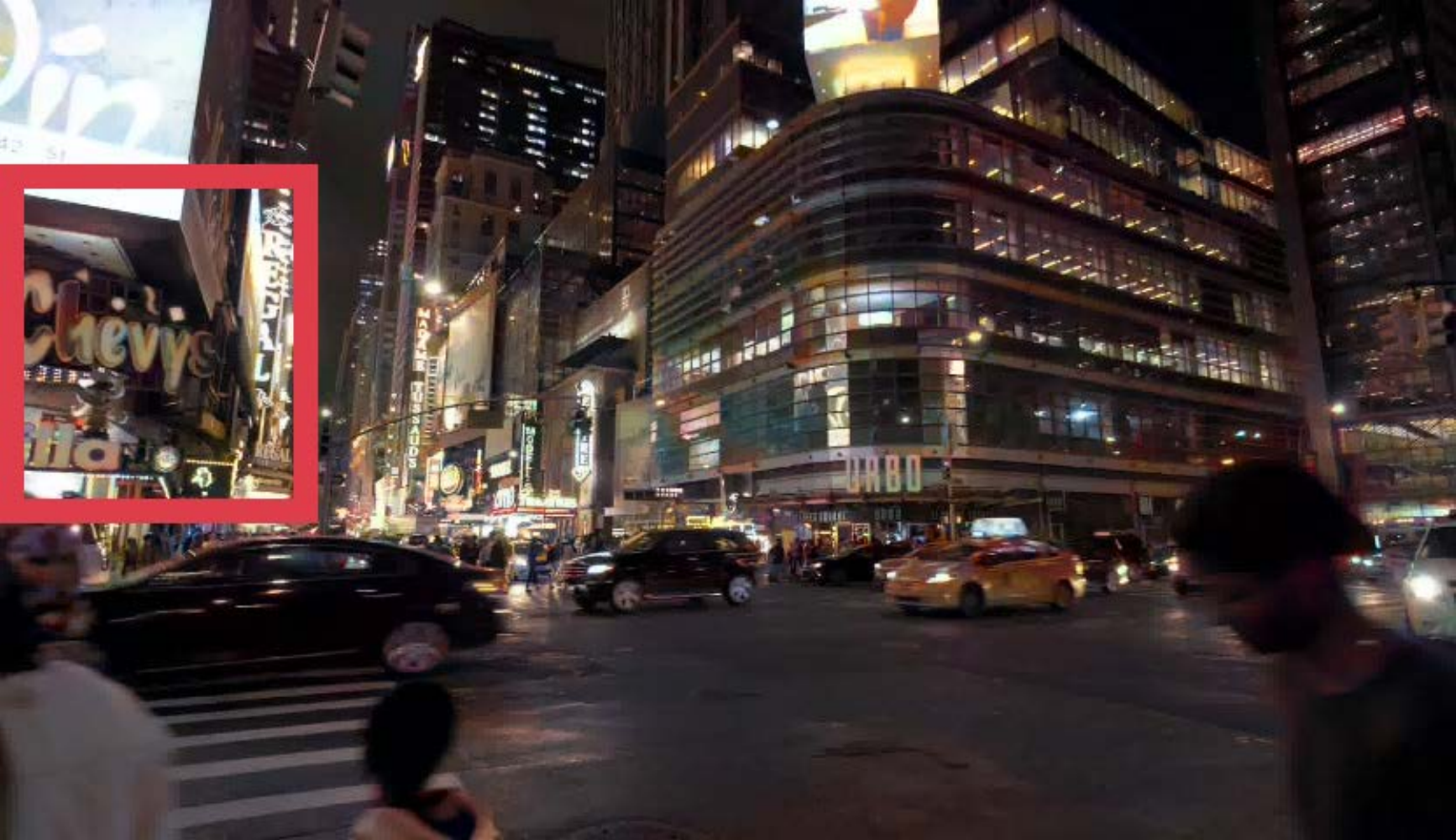} 
        \\
        \makebox[0.45\textwidth]{ (a) Input frame and exemplar image} &
        \makebox[0.45\textwidth]{ (b) DDColor~\cite{kang2022ddcolor}} 
        \\ 
        \includegraphics[width=0.495\textwidth, height = 0.28\textwidth] {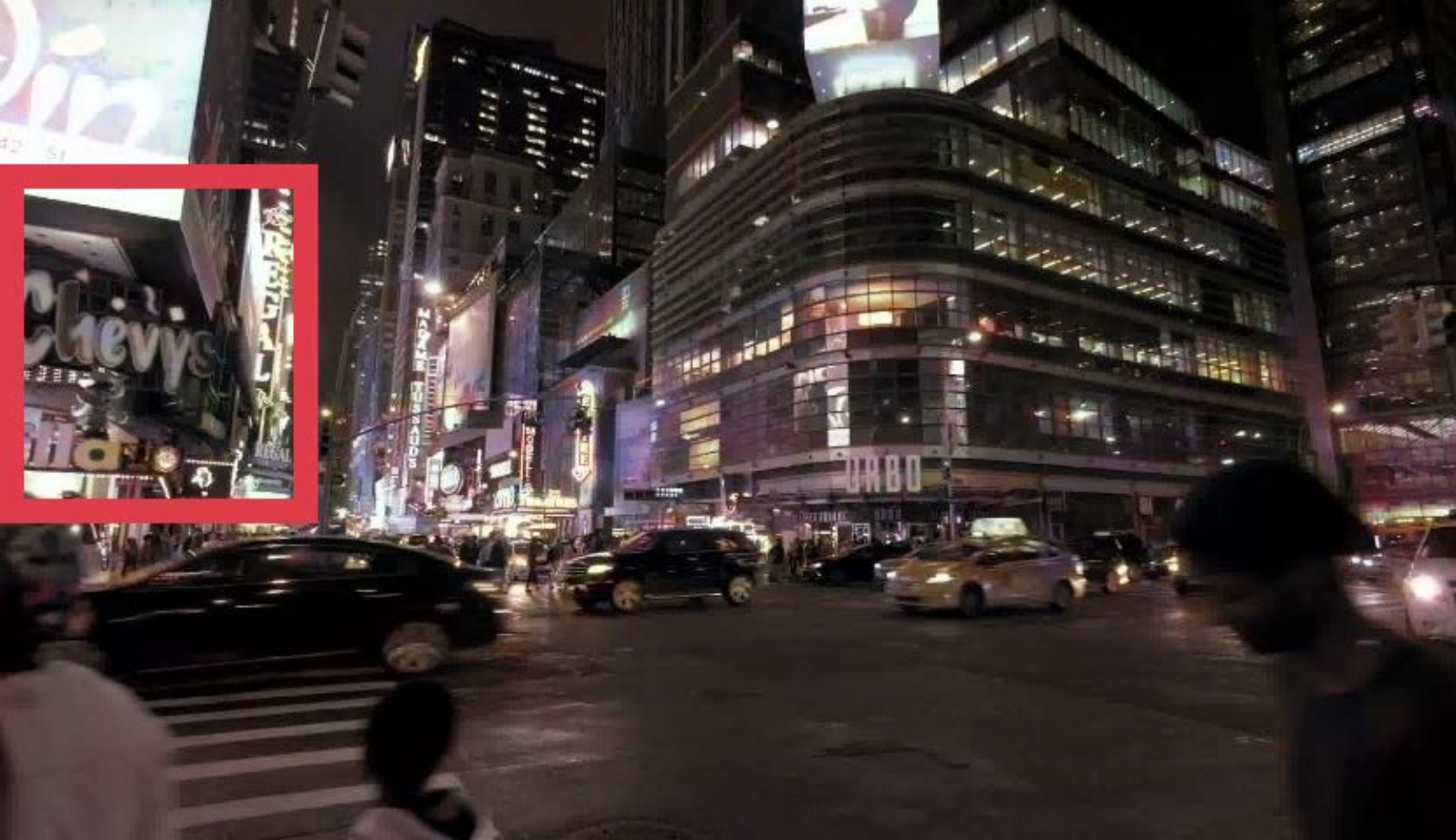} &
        \includegraphics[width=0.495\textwidth, height = 0.28\textwidth] {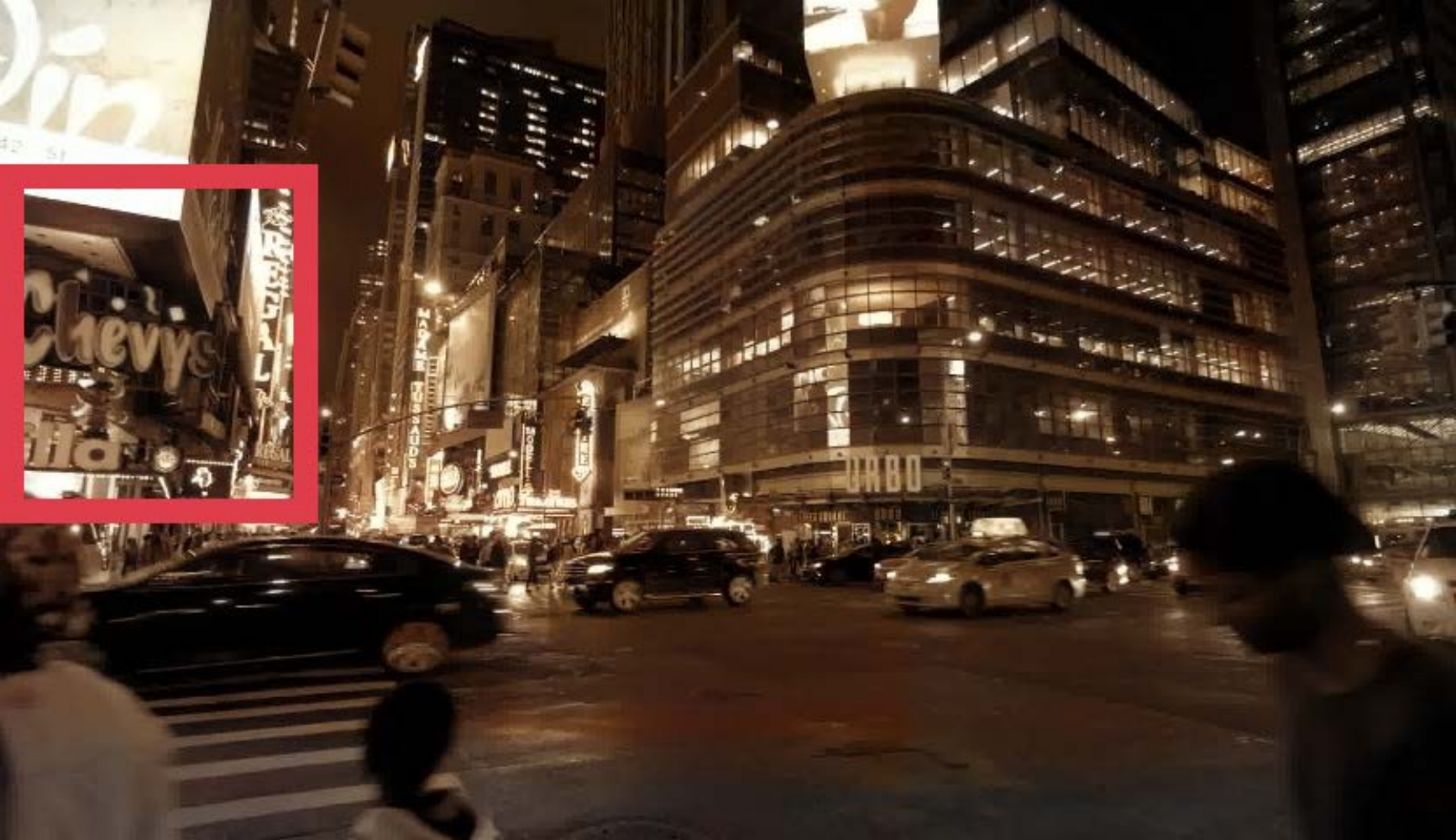} 
        \\ 
        \makebox[0.45\textwidth]{ (c) Color2Embed~\cite{zhao2021color2embed}} &
        \makebox[0.45\textwidth]{ (d) TCVC~\cite{liu2021temporally}} \\
        \includegraphics[width=0.495\textwidth, height = 0.28\textwidth] {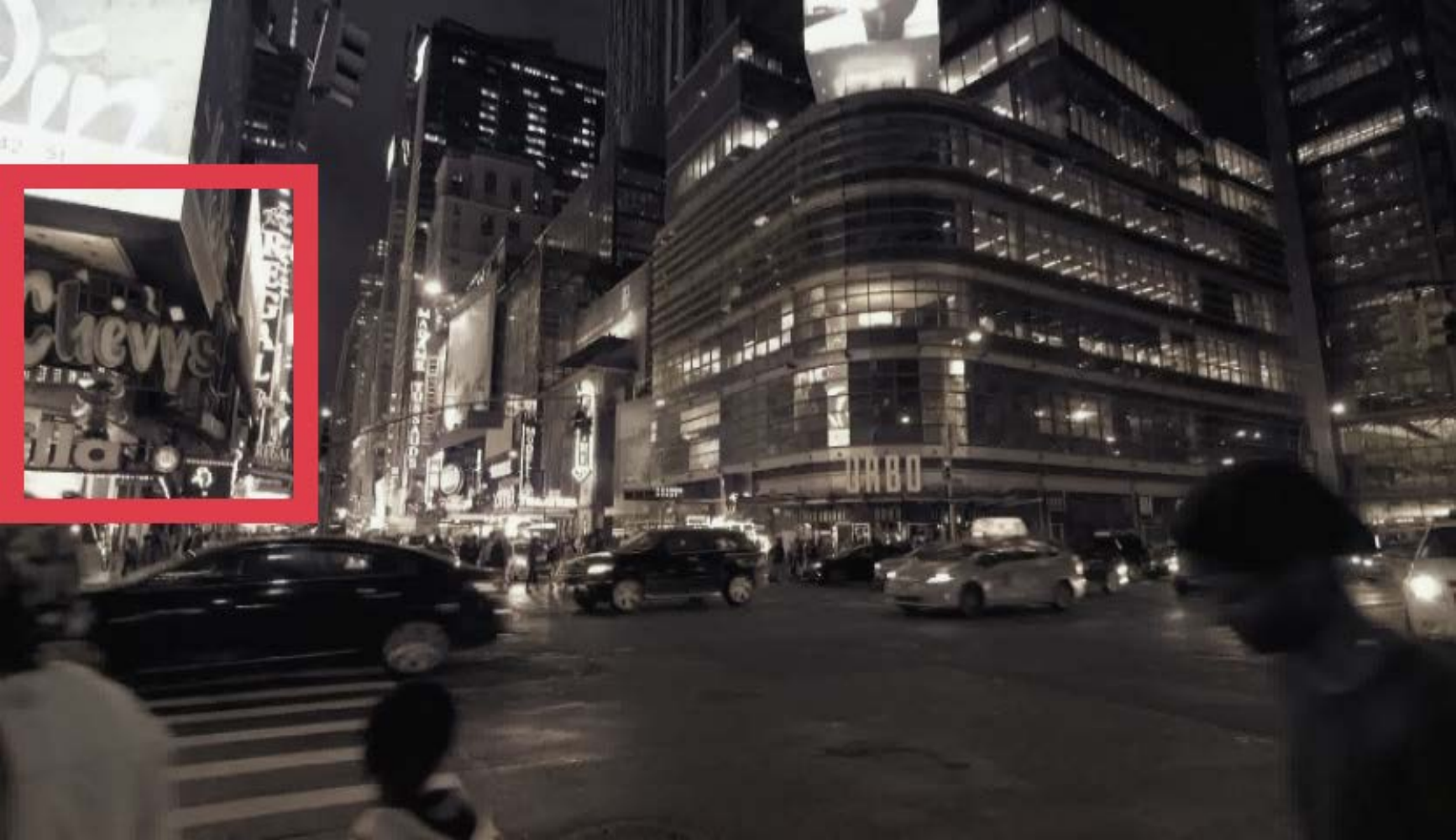} &
        \includegraphics[width=0.495\textwidth, height = 0.28\textwidth] {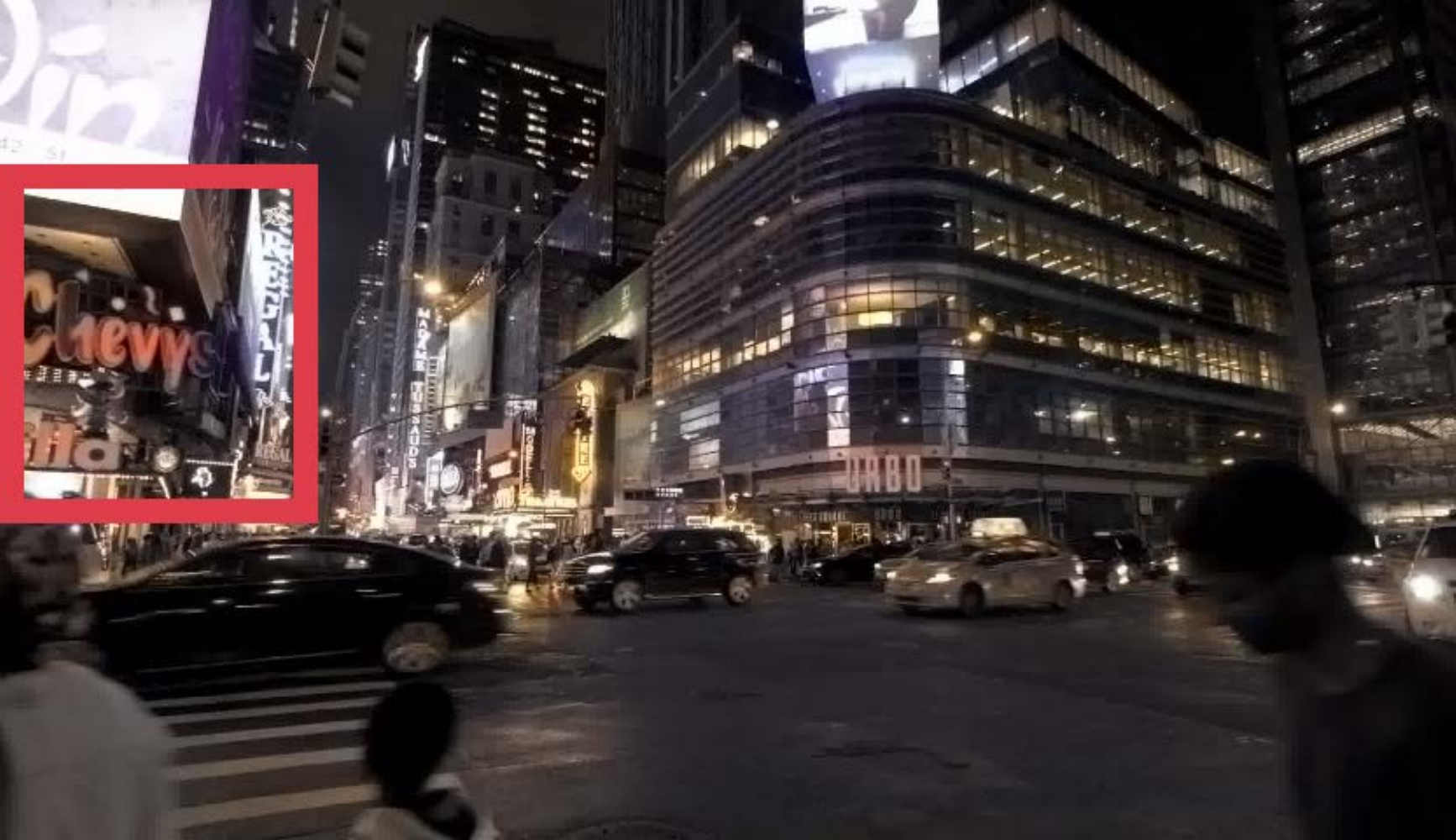} 
        \\
        \makebox[0.45\textwidth]{ (e) VCGAN~\cite{vcgan}} &
        \makebox[0.45\textwidth]{ (f) DeepRemaster~\cite{IizukaSIGGRAPHASIA2019}} 
        \\ 
        \includegraphics[width=0.495\textwidth, height = 0.28\textwidth] {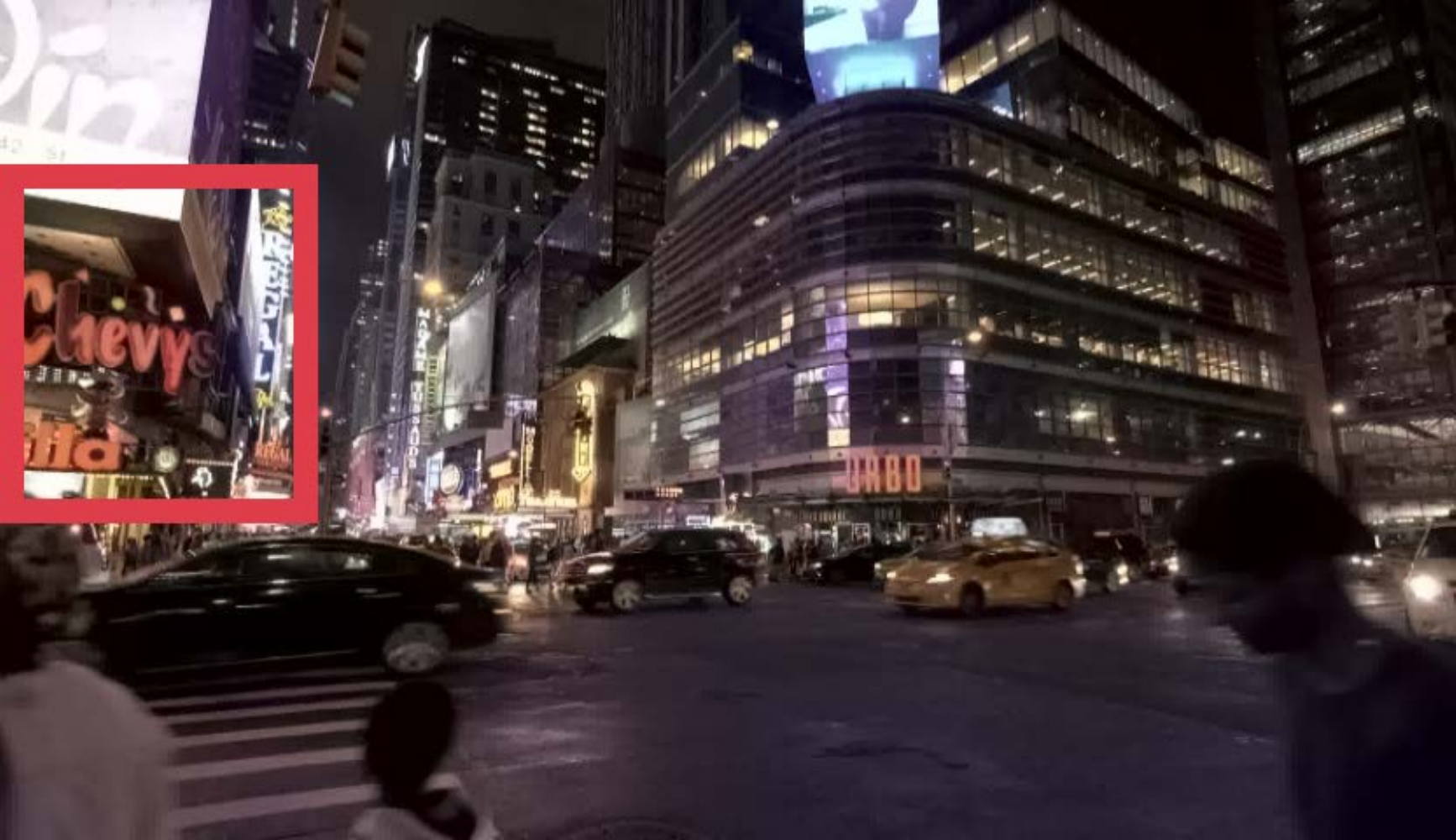} &
        \includegraphics[width=0.495\textwidth, height = 0.28\textwidth] {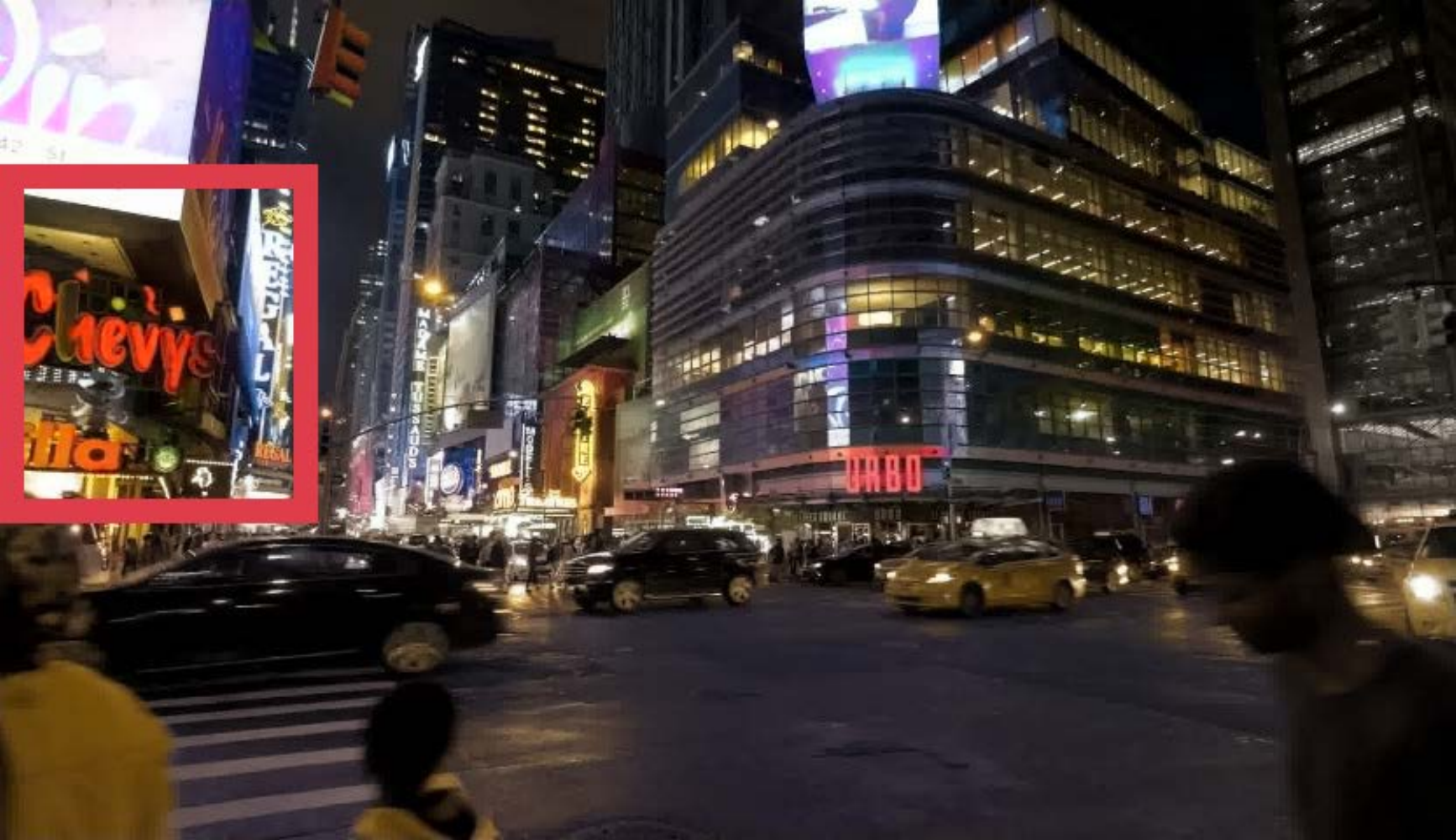} 
        \\
        \makebox[0.45\textwidth]{ (g) DeepExemplar~\cite{zhang2019deep}} &
        \makebox[0.45\textwidth]{ (h) ColorMNet (Ours)}  
         %& \vspace{-0.7em}
	\end{tabularx}
	\vspace{-0.8em}
	\caption{{Colorization results on clip \textit{TimeSquareTraffic} from the Videvo validation dataset~\cite{Lai2018videvo}. The results shown in (b) and (c) still contain significant color-bleeding artifacts. In contrast, our proposed method generates a vivid and realistic frame against other stage-of-the-art methods.}}
	\label{fig:9}
	%\vspace{-3mm}
\end{figure*}

\begin{figure*}[!htb]
	\setlength\tabcolsep{1.0pt}
	\centering
	\small
	\begin{tabularx}{1\textwidth}{cc}
        \includegraphics[width=0.495\textwidth, height = 0.28\textwidth] {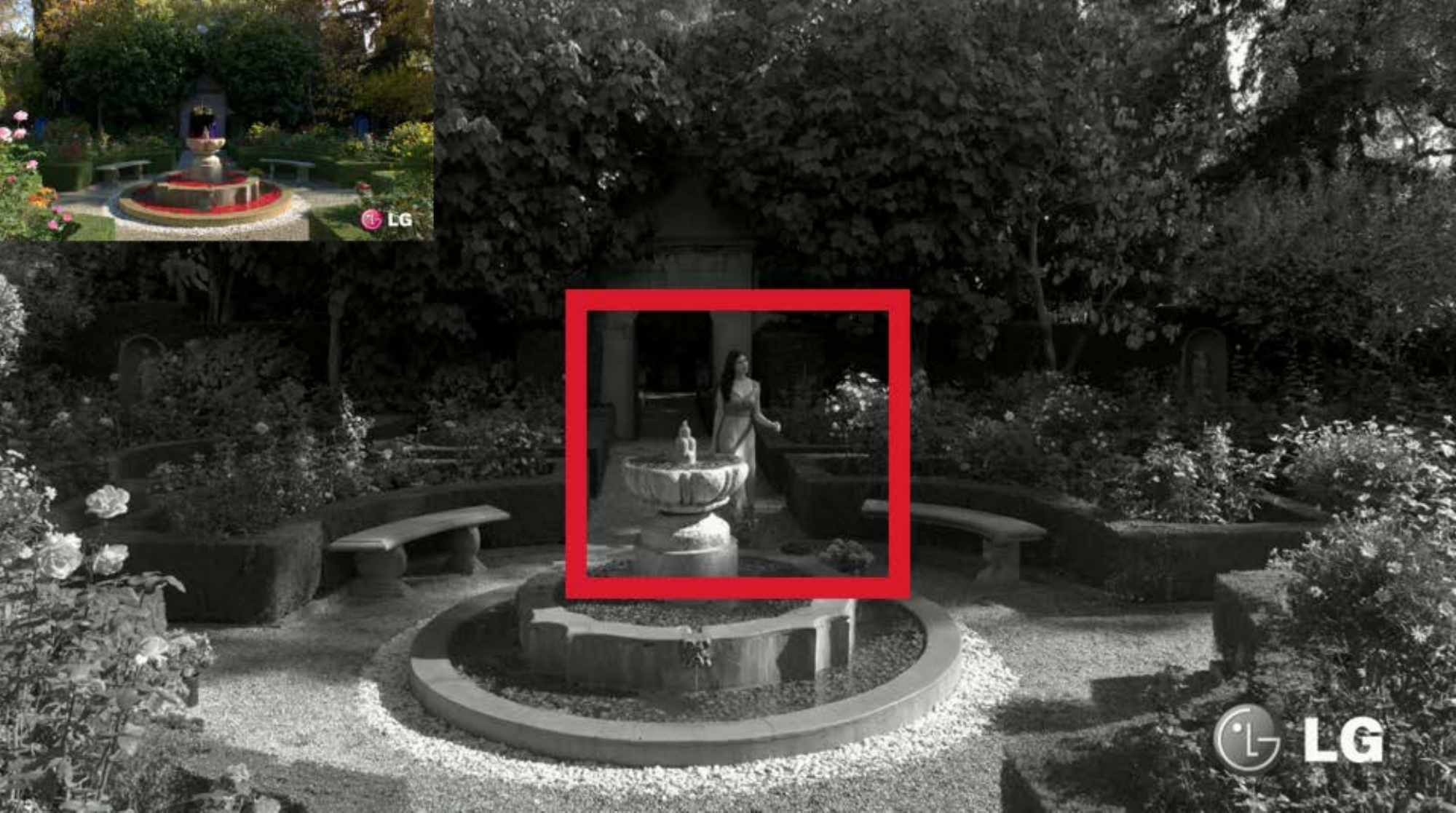} &
        \includegraphics[width=0.495\textwidth, height = 0.28\textwidth] {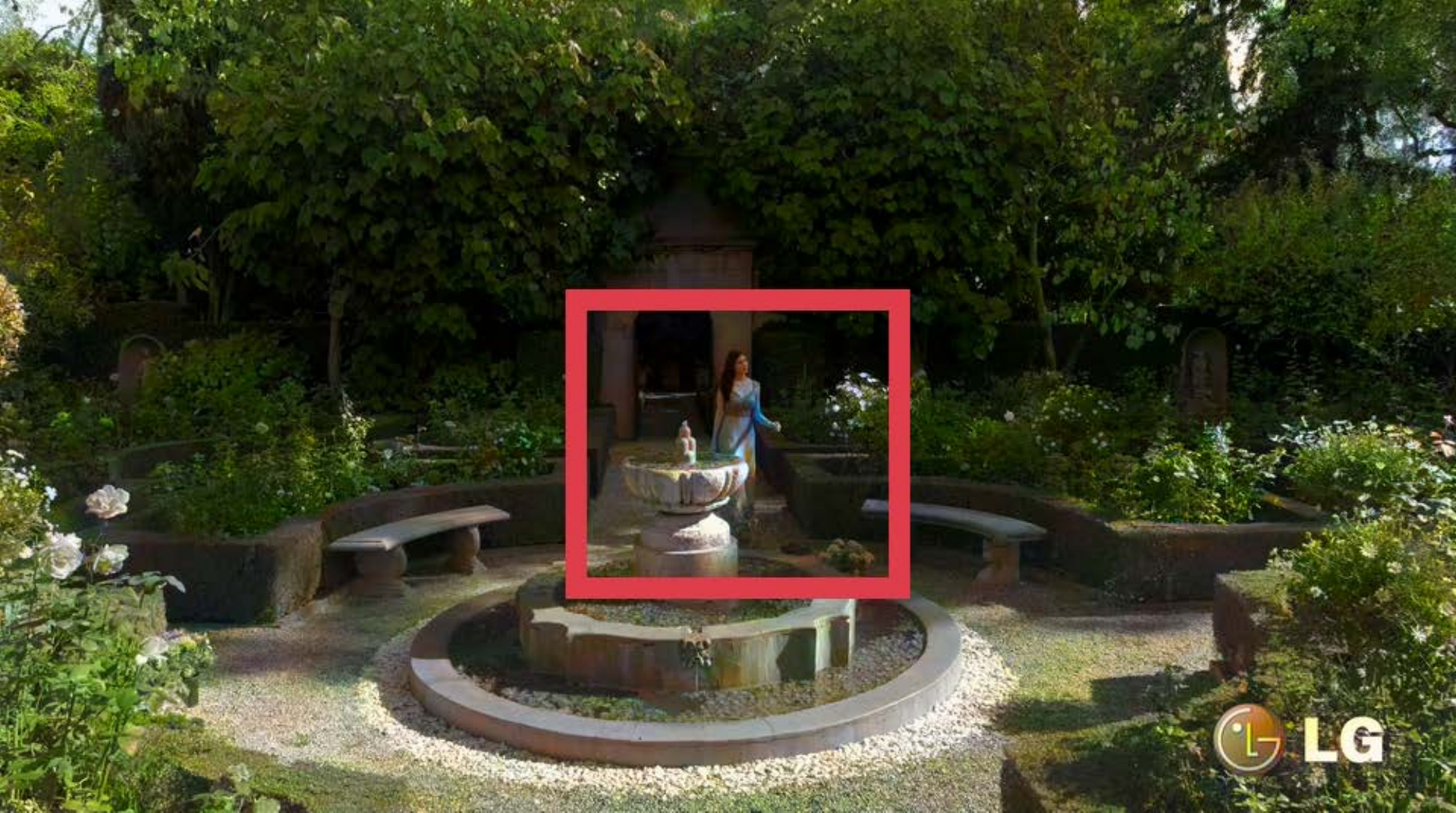} 
        \\
        \makebox[0.45\textwidth]{ (a) Input frame and exemplar image} &
        \makebox[0.45\textwidth]{ (b) DDColor~\cite{kang2022ddcolor}} 
        \\ 
        \includegraphics[width=0.495\textwidth, height = 0.28\textwidth] {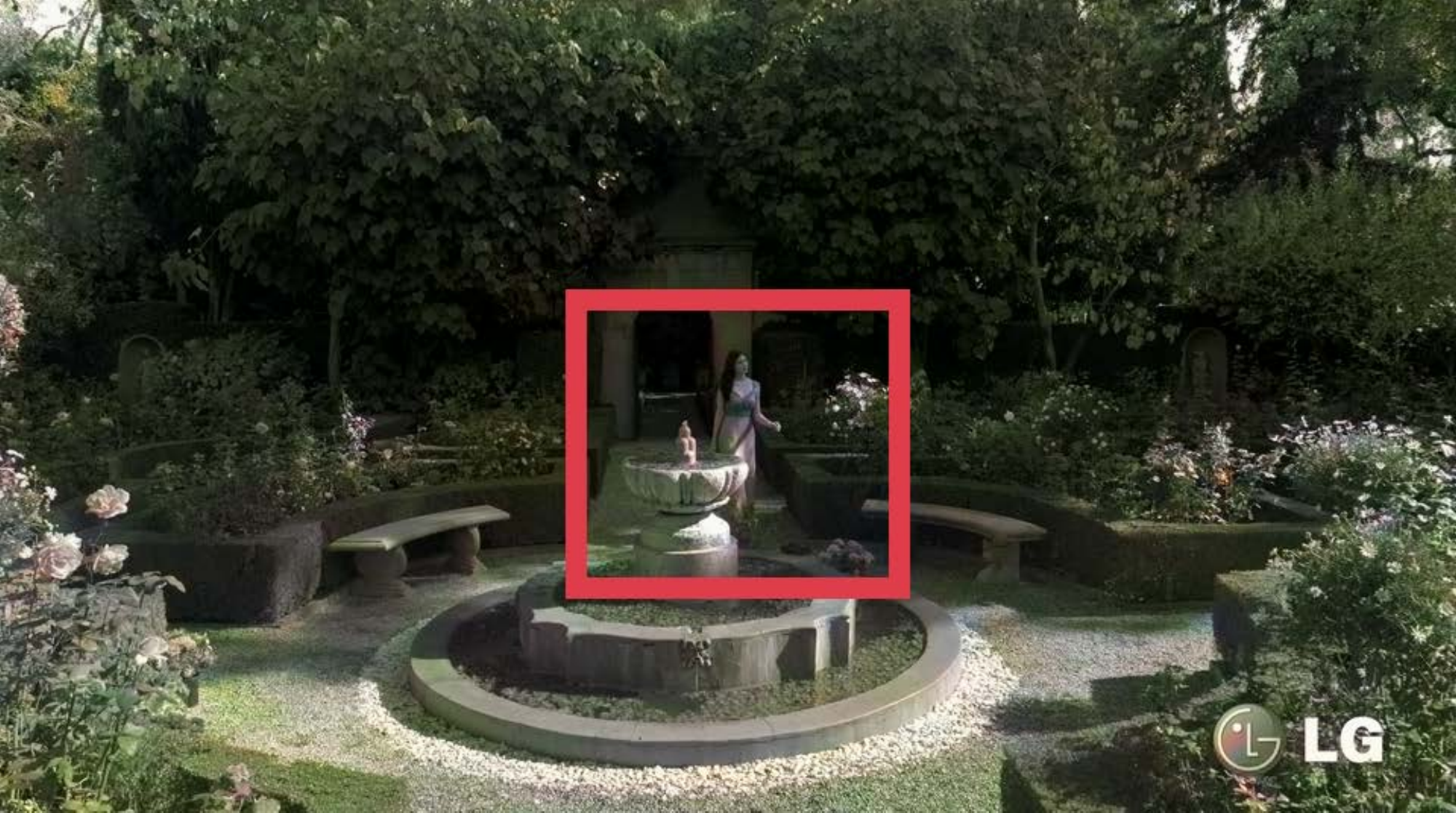} &
        \includegraphics[width=0.495\textwidth, height = 0.28\textwidth] {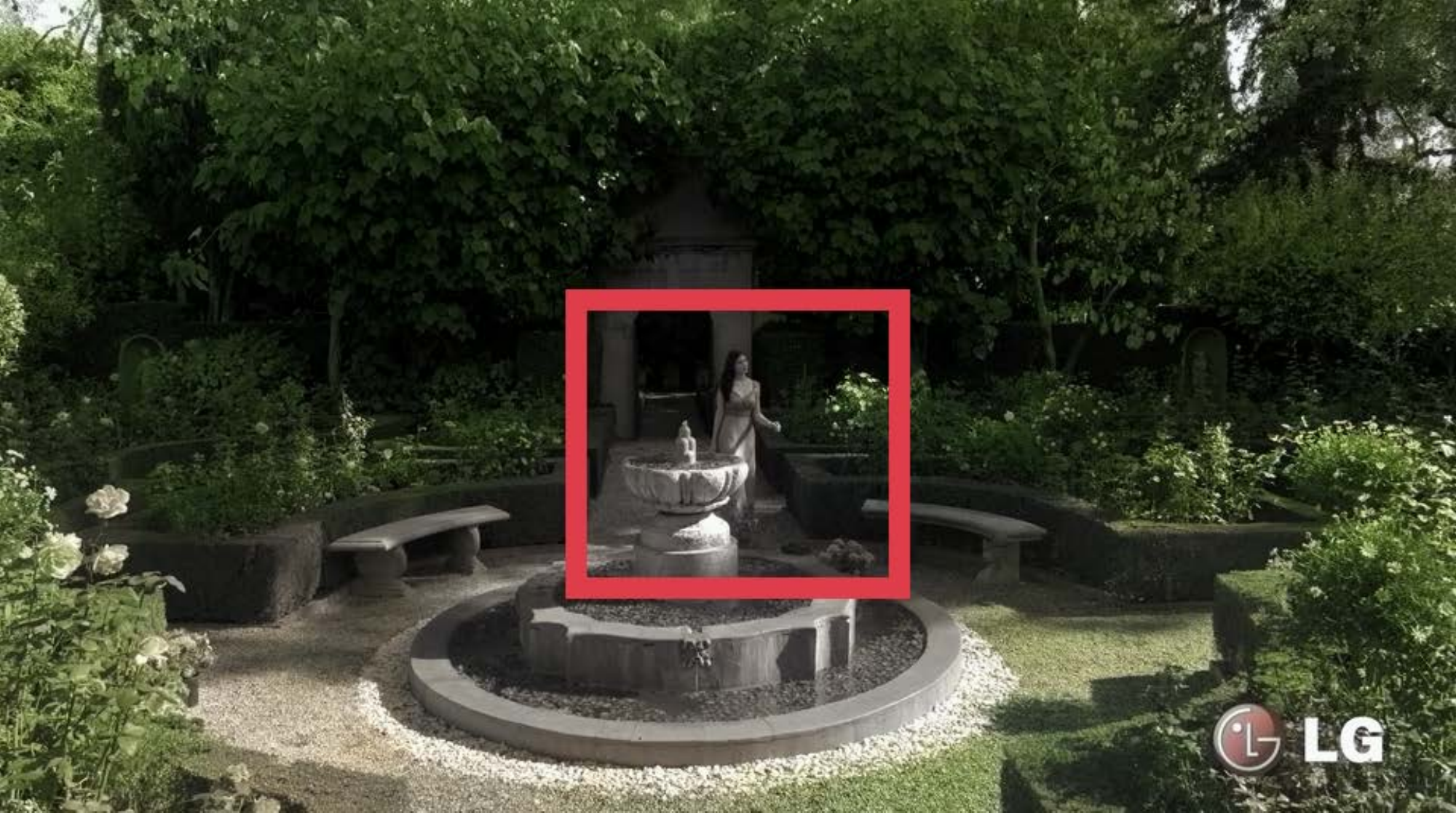} 
        \\ 
        \makebox[0.45\textwidth]{ (c) Color2Embed~\cite{zhao2021color2embed}} &
        \makebox[0.45\textwidth]{ (d) TCVC~\cite{liu2021temporally}} \\
        \includegraphics[width=0.495\textwidth, height = 0.28\textwidth] {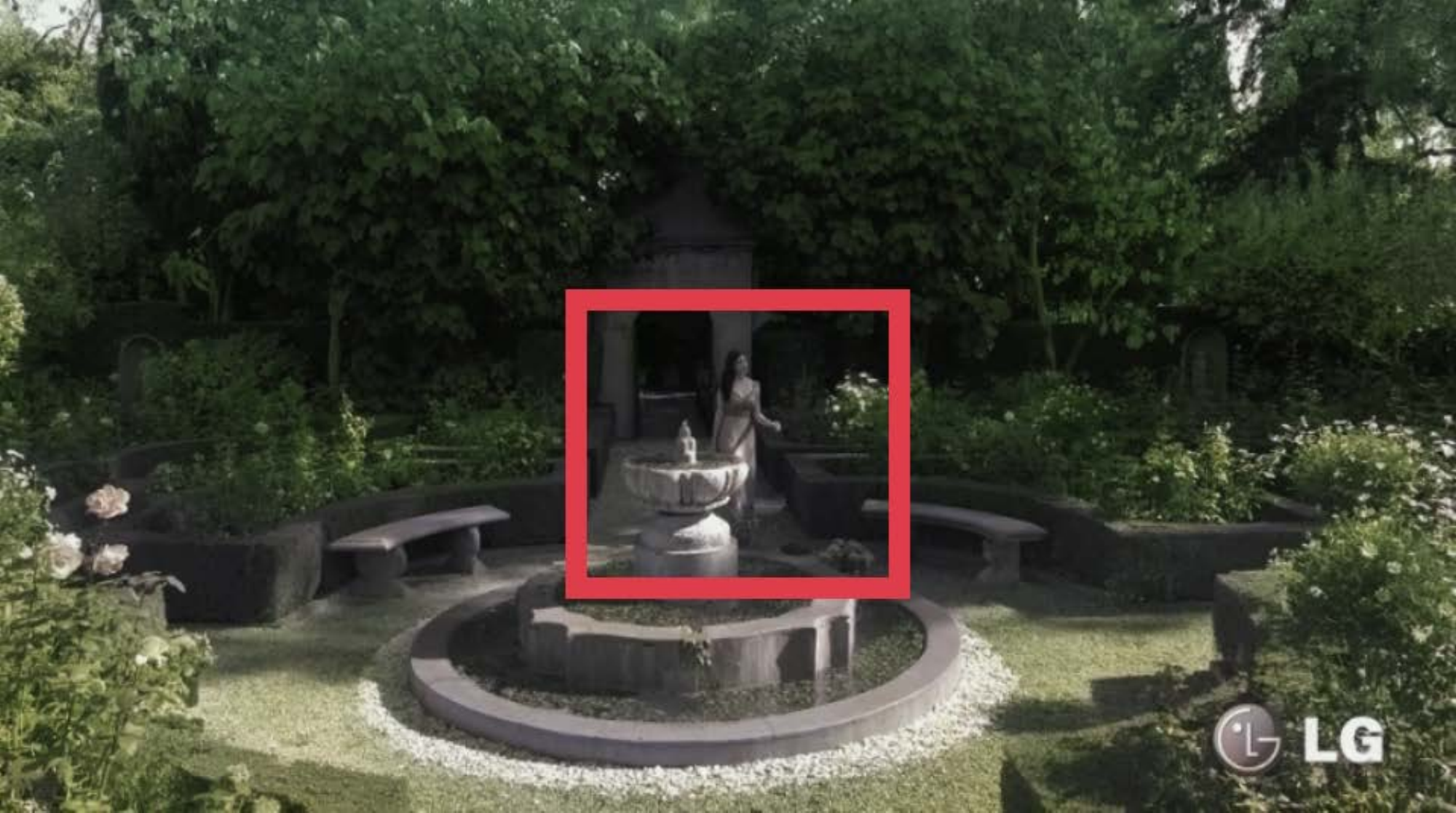} &
        \includegraphics[width=0.495\textwidth, height = 0.28\textwidth] {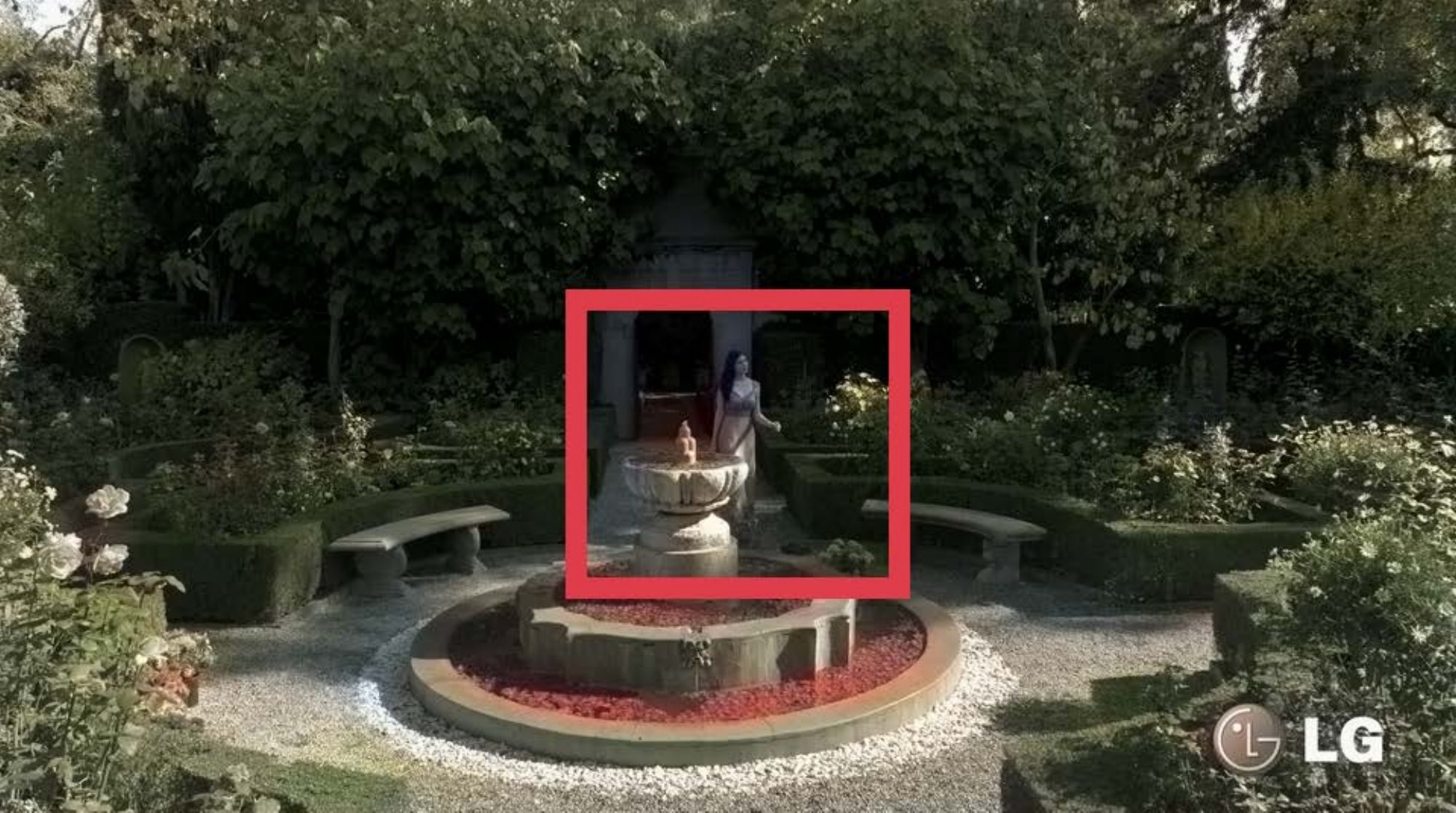} 
        \\
        \makebox[0.45\textwidth]{ (e) VCGAN~\cite{vcgan}} &
        \makebox[0.45\textwidth]{ (f) DeepRemaster~\cite{IizukaSIGGRAPHASIA2019}} 
        \\ 
        \includegraphics[width=0.495\textwidth, height = 0.28\textwidth] {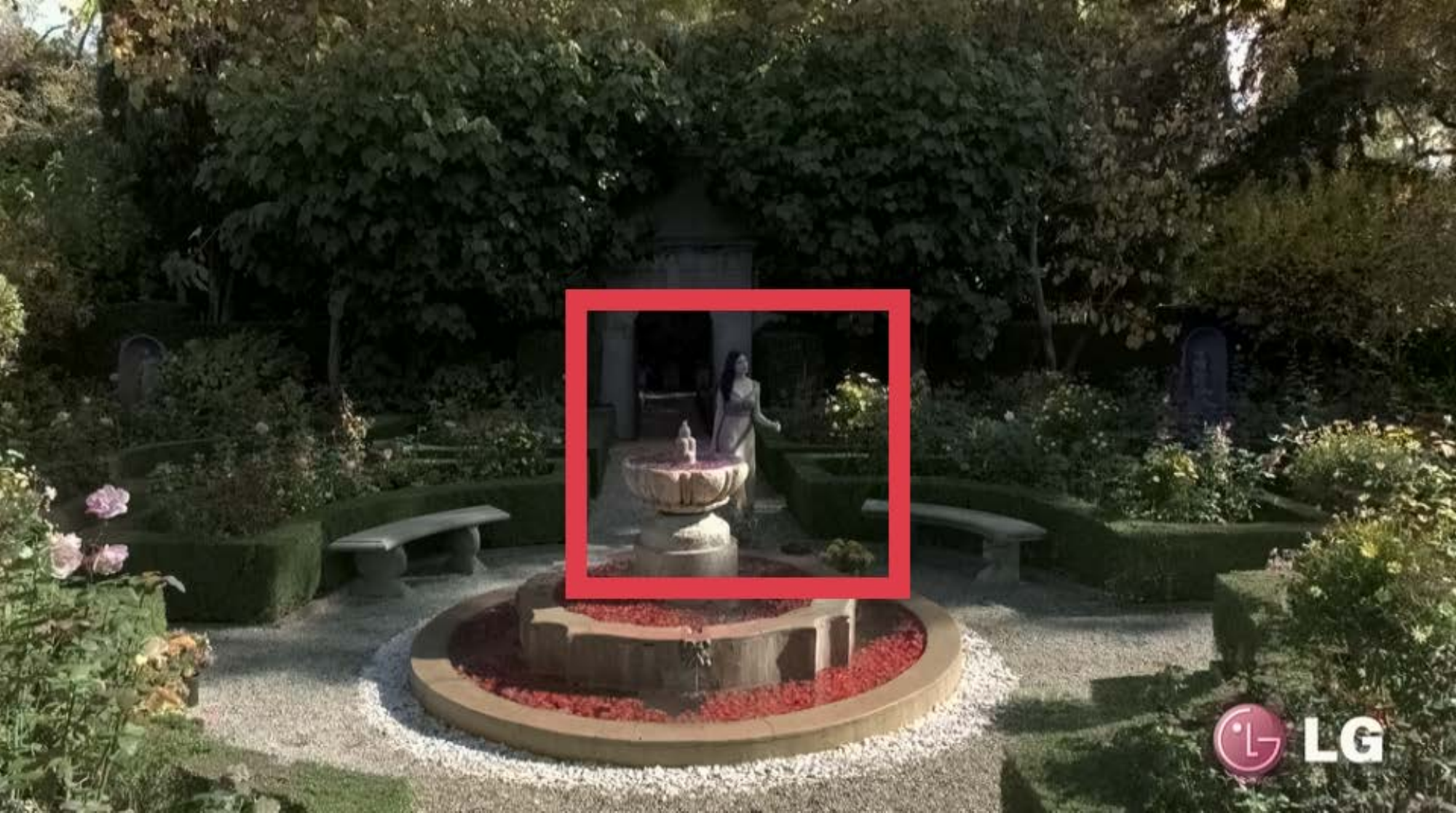} &
        \includegraphics[width=0.495\textwidth, height = 0.28\textwidth] {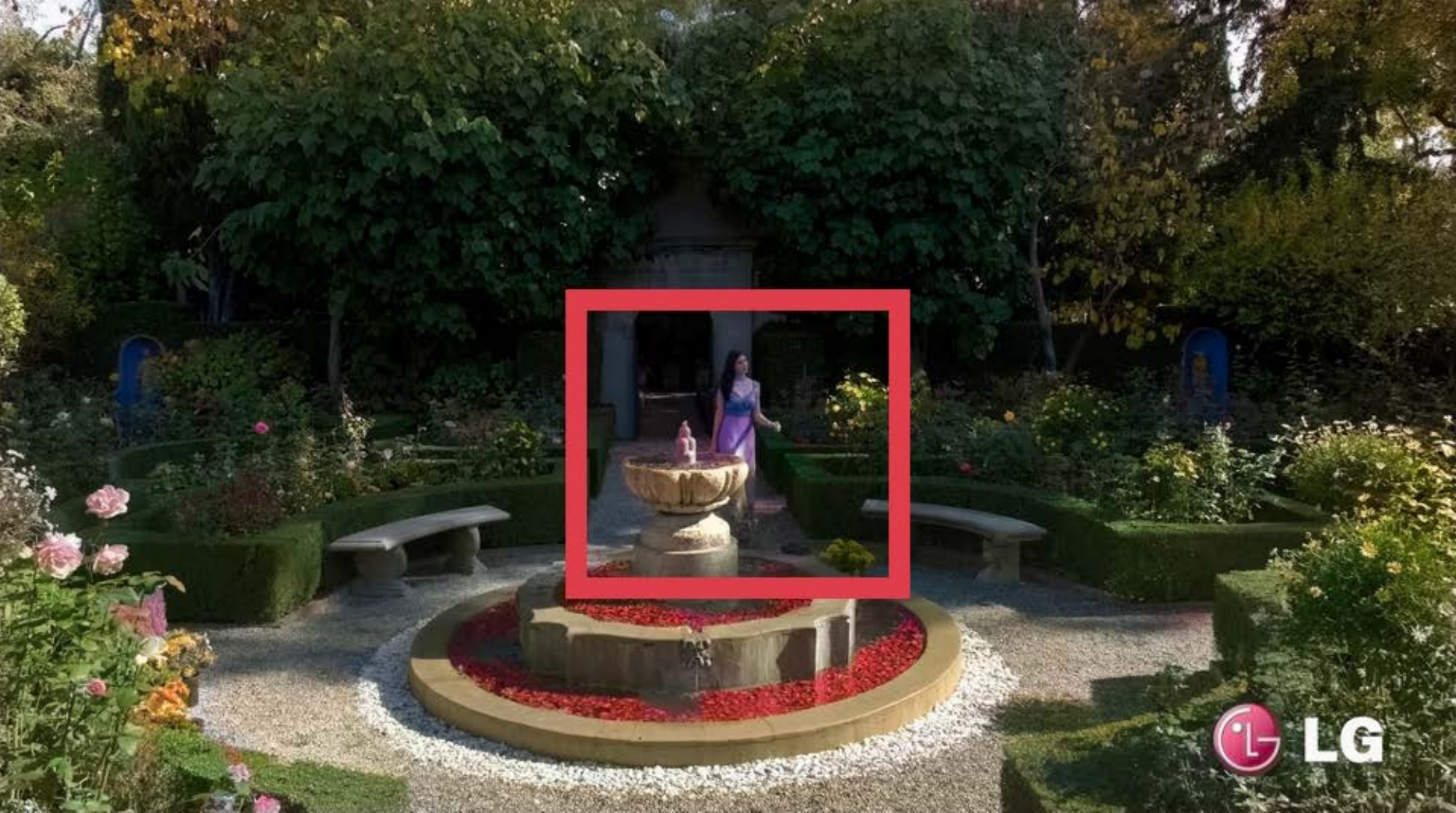} 
        \\
        \makebox[0.45\textwidth]{ (g) DeepExemplar~\cite{zhang2019deep}} &
        \makebox[0.45\textwidth]{ (h) ColorMNet (Ours)}  
         %& \vspace{-0.7em}
	\end{tabularx}
	\vspace{-0.8em}
	\caption{{Colorization results on clip \textit{001} from the NVCC2023 validation dataset~\cite{Kang_2023_CVPR}. The results shown in (b) and (c) still contain significant color-bleeding artifacts. In contrast, our proposed method generates a better-colorized frame, where the colors of the woman and the leaves are better restored.}}
	\label{fig:11}
	%\vspace{-3mm}
\end{figure*}

\begin{figure*}[!htb]
	\setlength\tabcolsep{1.0pt}
	\centering
	\small
	\begin{tabularx}{1\textwidth}{cc}
        \includegraphics[width=0.495\textwidth, height = 0.28\textwidth] {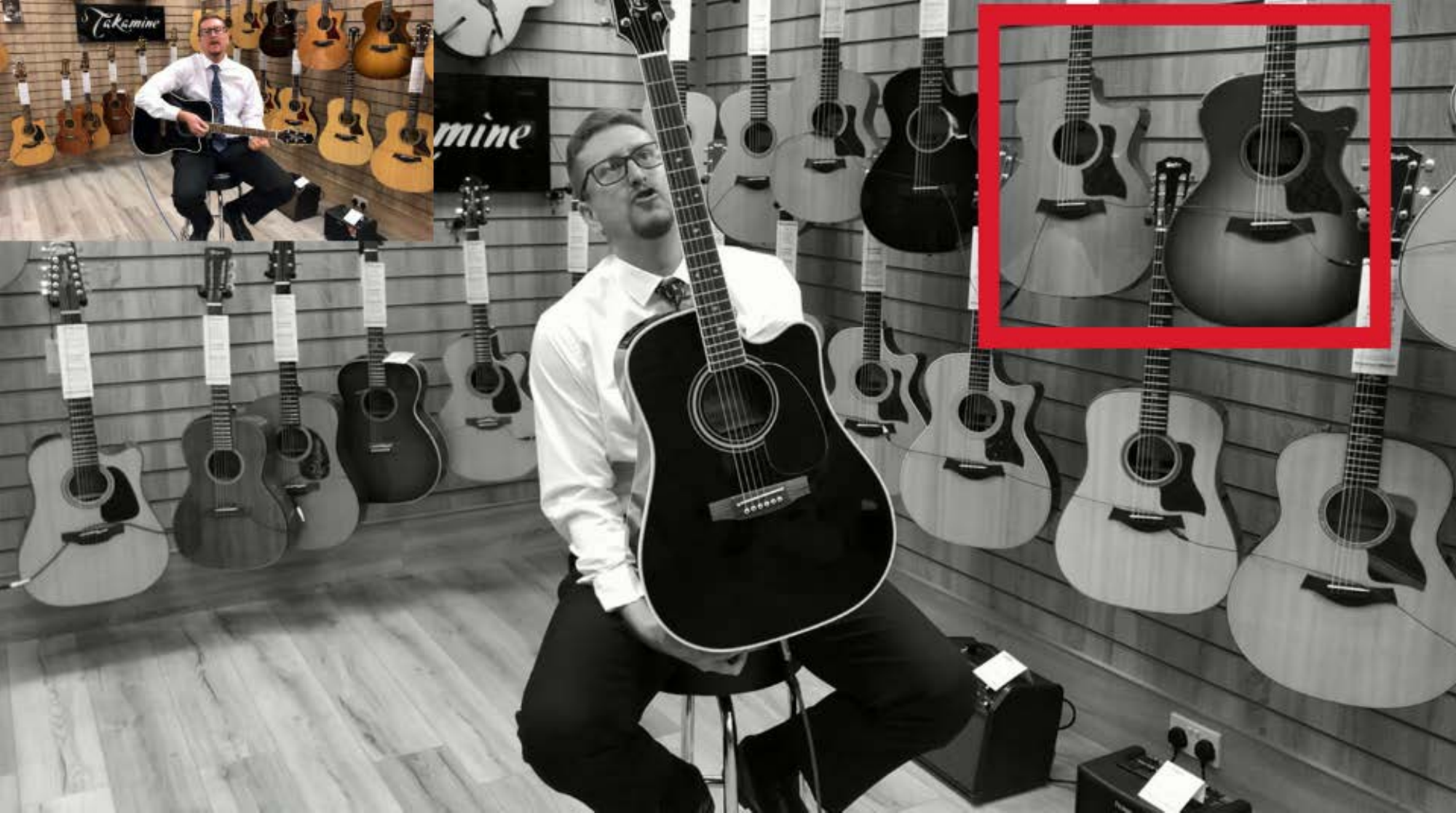} &
        \includegraphics[width=0.495\textwidth, height = 0.28\textwidth] {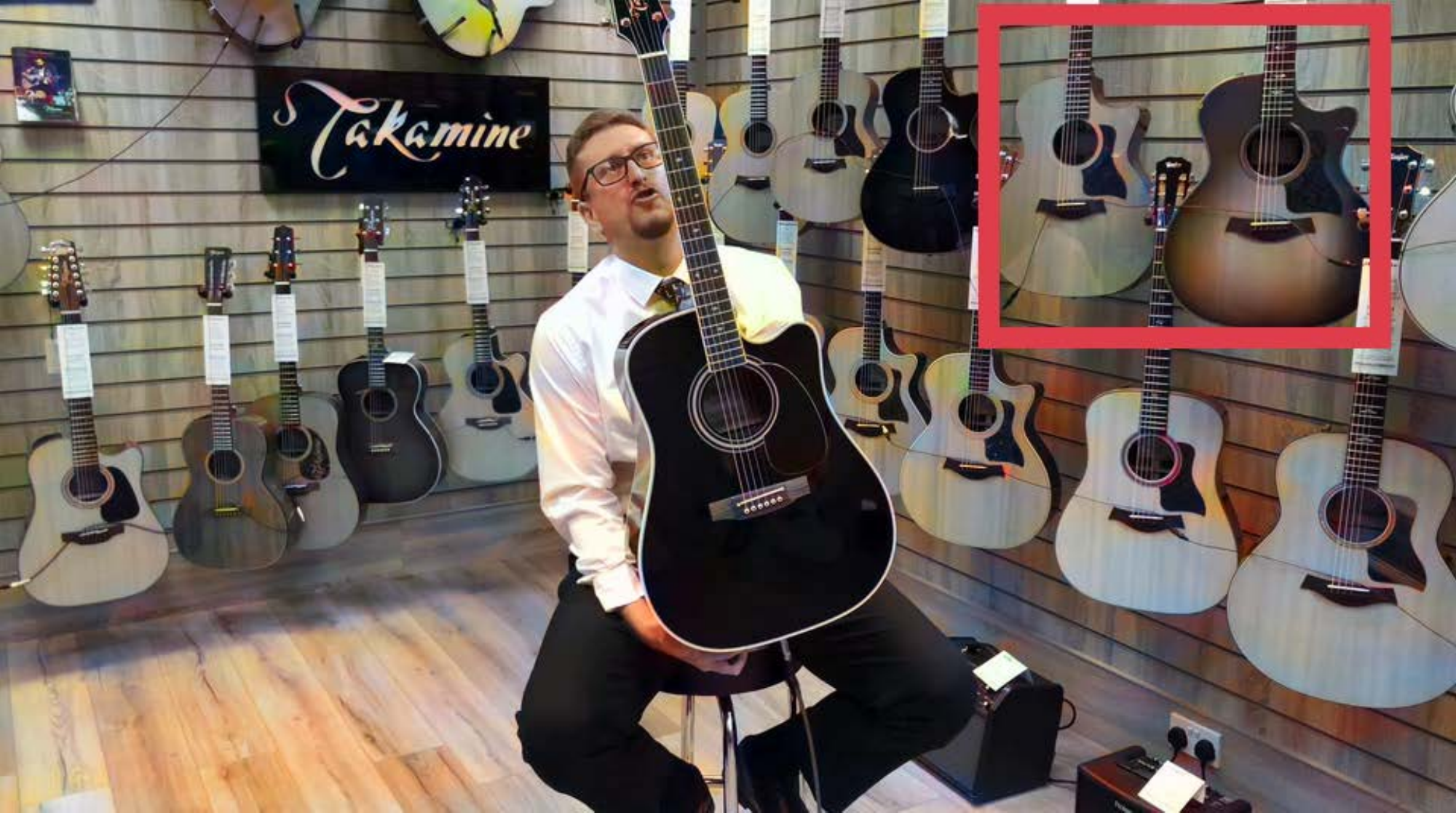} 
        \\
        \makebox[0.45\textwidth]{ (a) Input frame and exemplar image} &
        \makebox[0.45\textwidth]{ (b) DDColor~\cite{kang2022ddcolor}} 
        \\ 
        \includegraphics[width=0.495\textwidth, height = 0.28\textwidth] {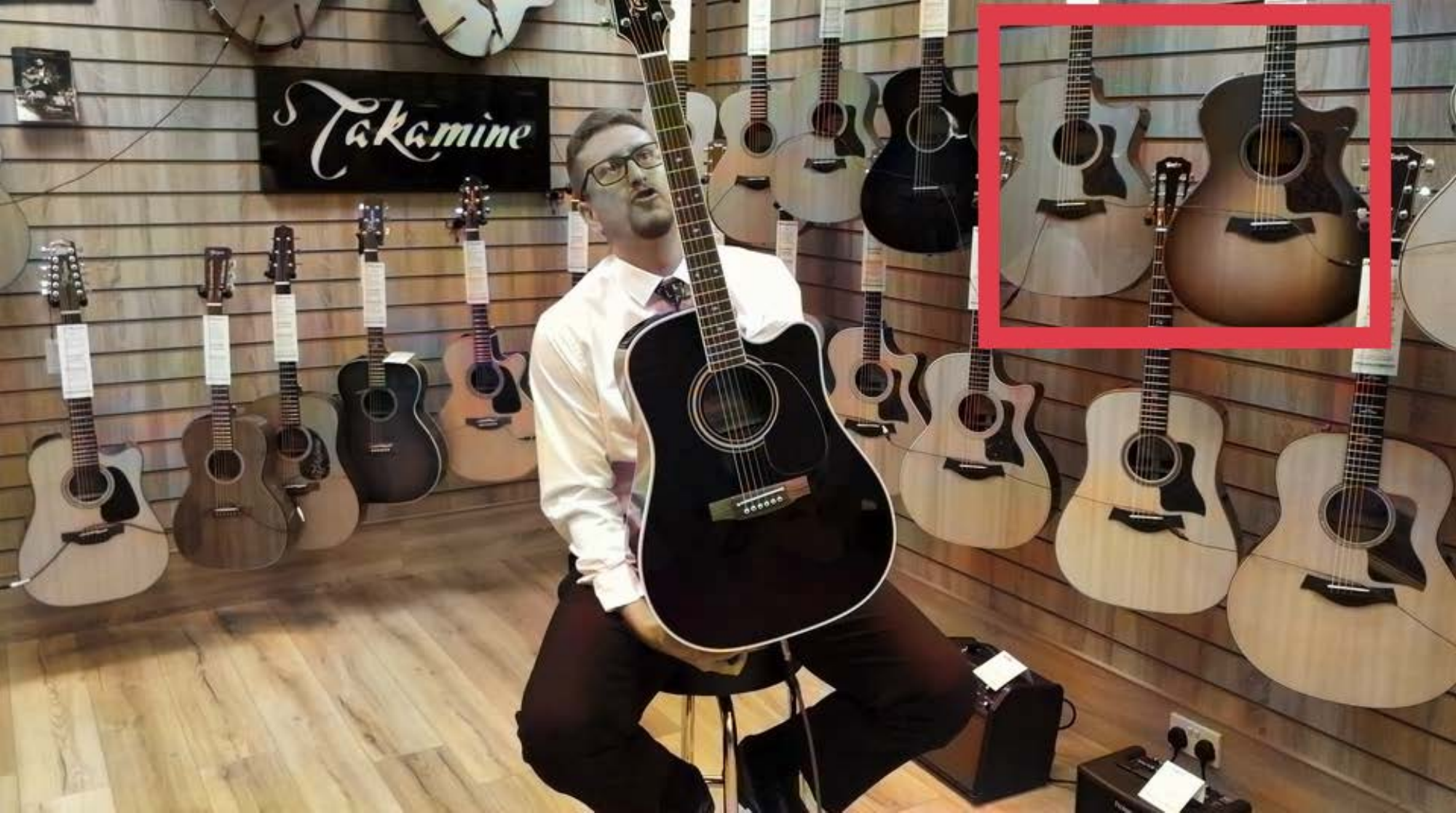} &
        \includegraphics[width=0.495\textwidth, height = 0.28\textwidth] {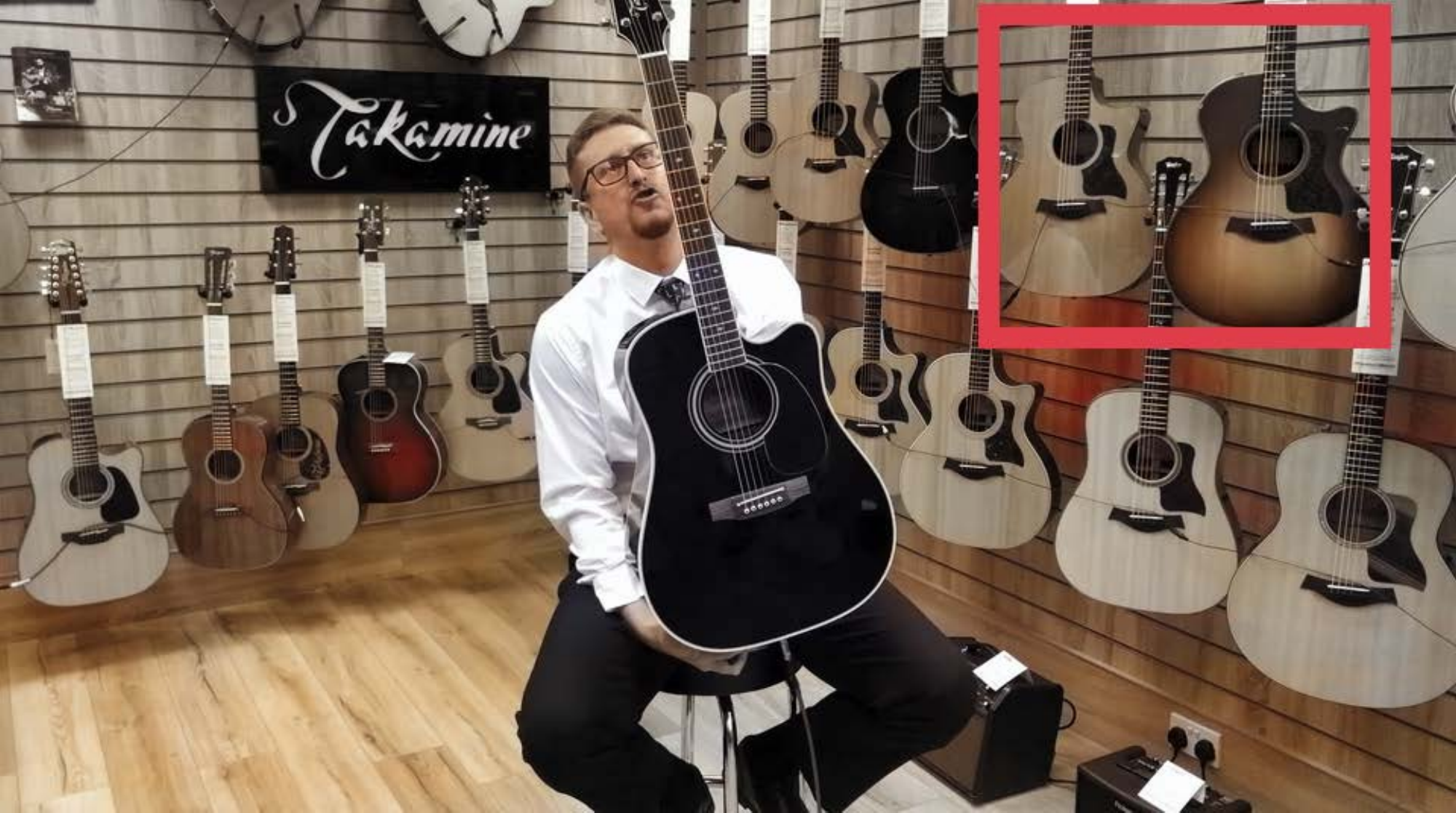} 
        \\ 
        \makebox[0.45\textwidth]{ (c) Color2Embed~\cite{zhao2021color2embed}} &
        \makebox[0.45\textwidth]{ (d) TCVC~\cite{liu2021temporally}} \\
        \includegraphics[width=0.495\textwidth, height = 0.28\textwidth] {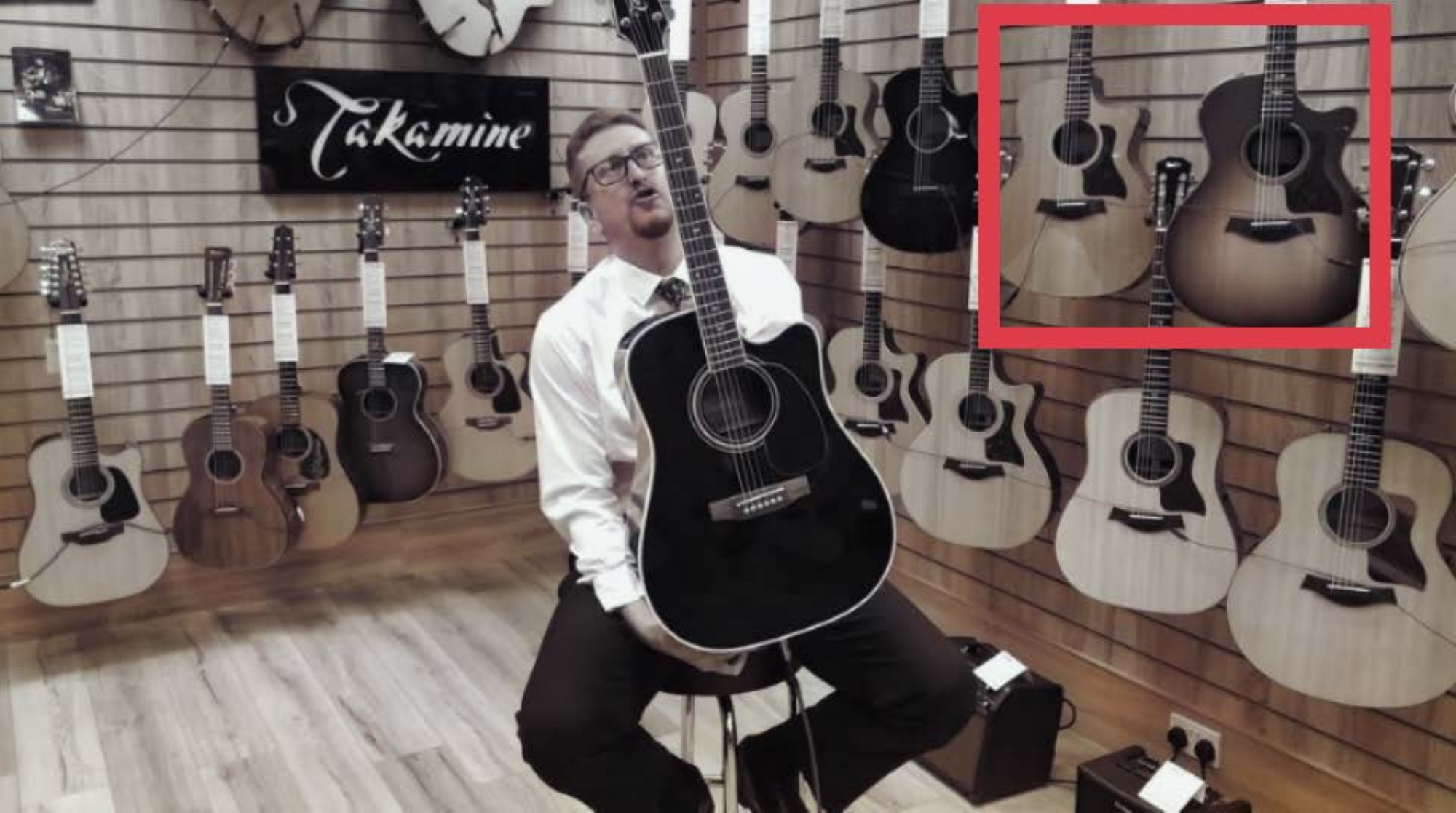} &
        \includegraphics[width=0.495\textwidth, height = 0.28\textwidth] {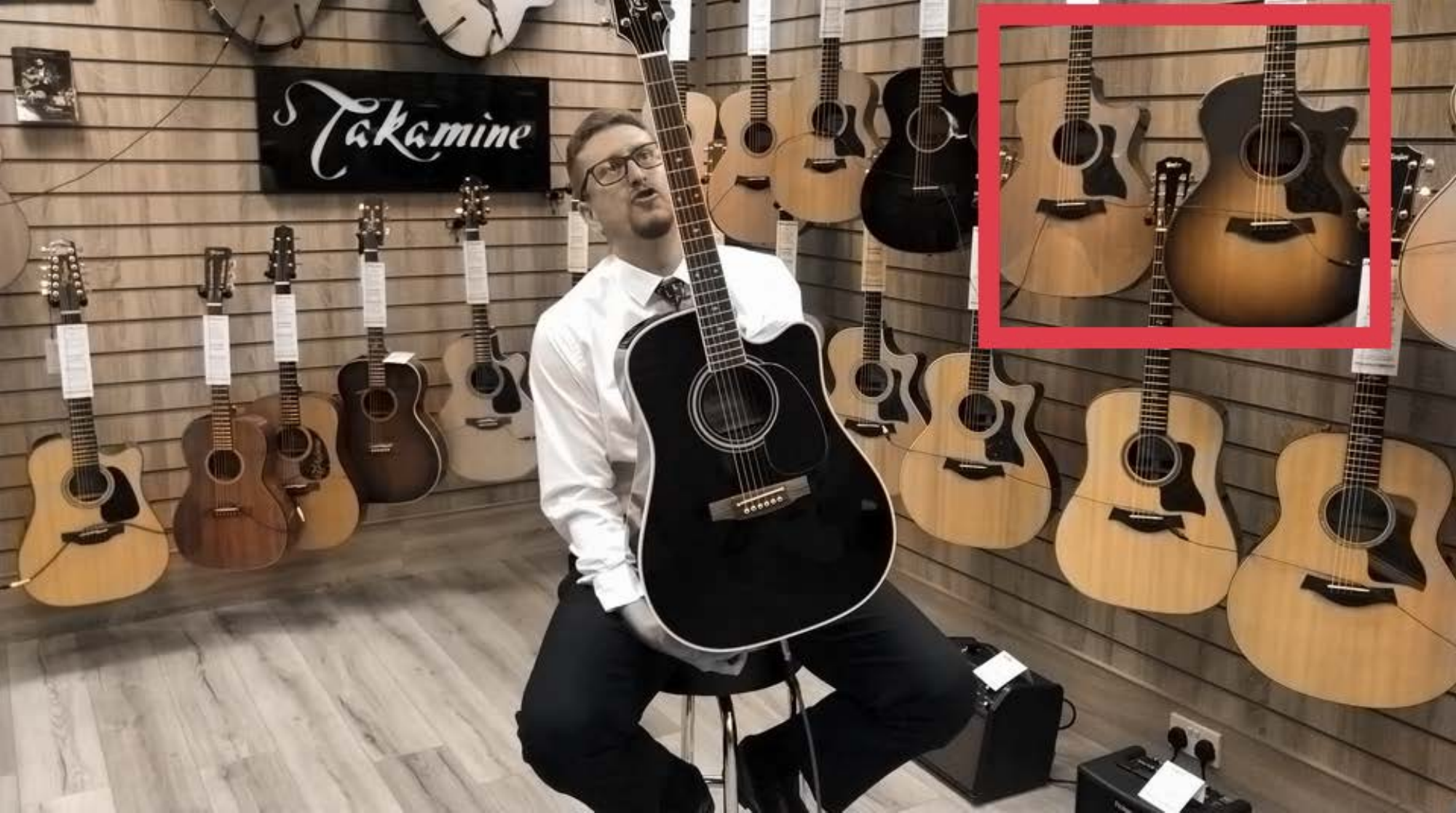} 
        \\
        \makebox[0.45\textwidth]{ (e) VCGAN~\cite{vcgan}} &
        \makebox[0.45\textwidth]{ (f) DeepRemaster~\cite{IizukaSIGGRAPHASIA2019}} 
        \\ 
        \includegraphics[width=0.495\textwidth, height = 0.28\textwidth] {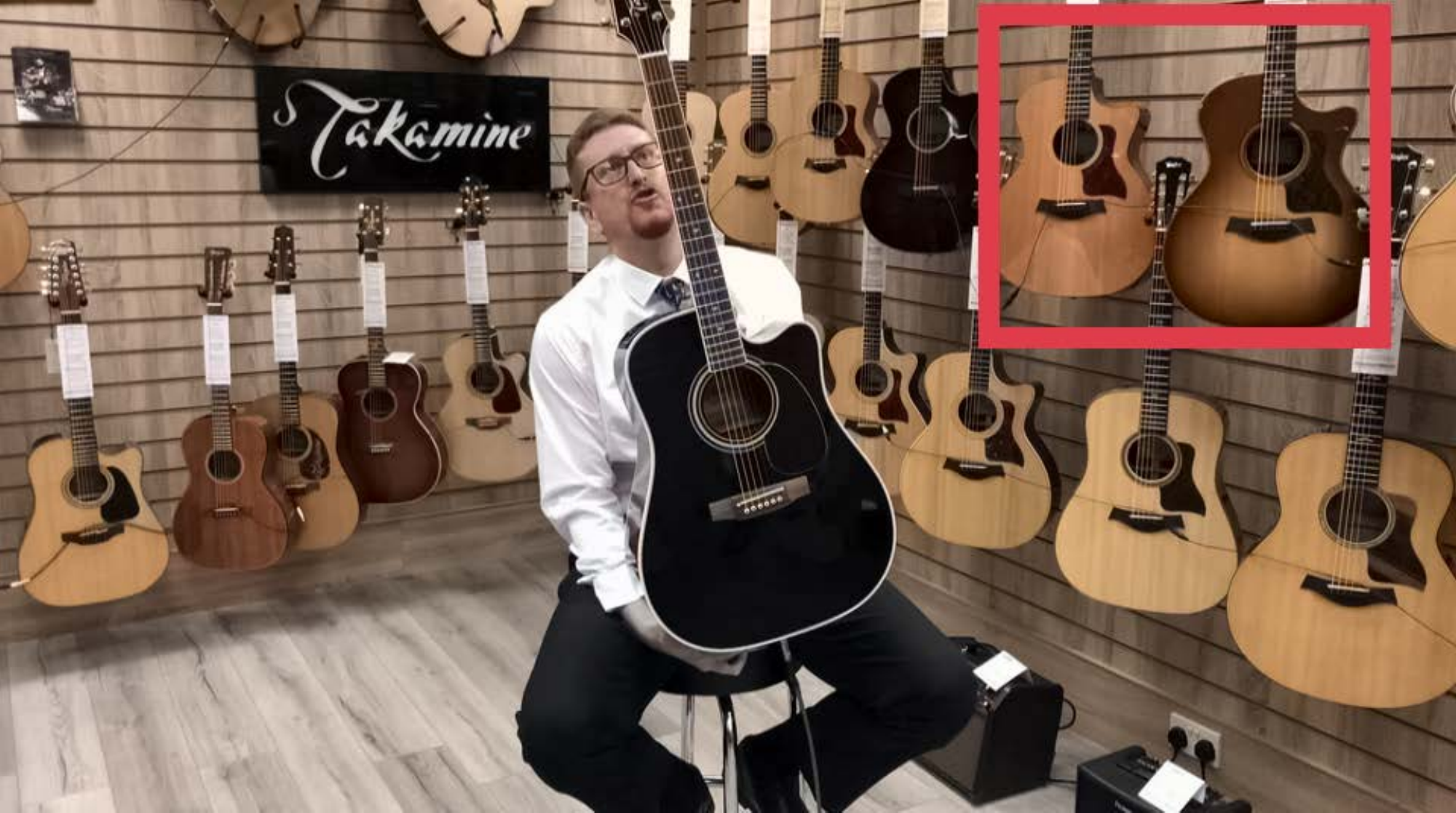} &
        \includegraphics[width=0.495\textwidth, height = 0.28\textwidth] {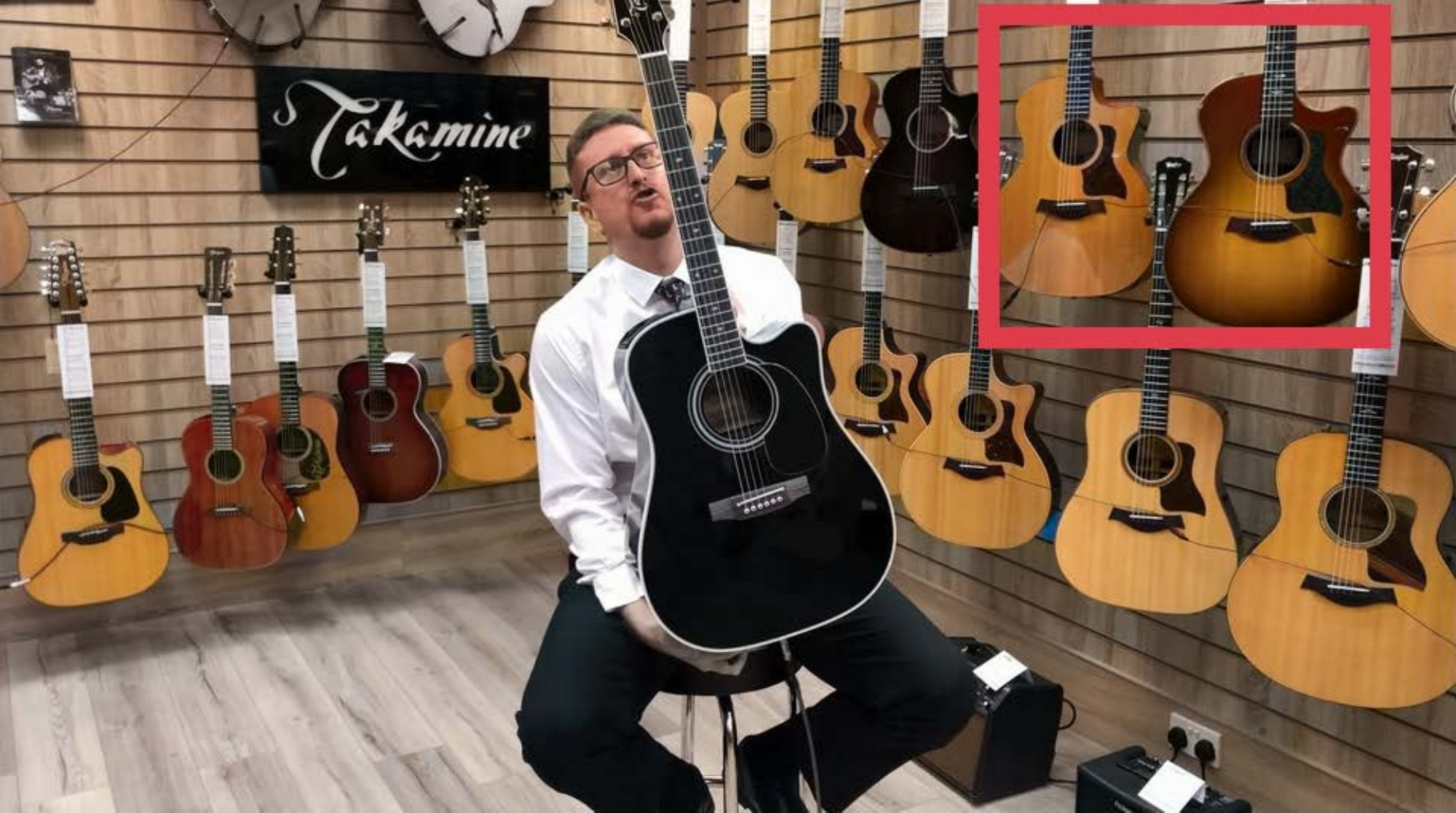} 
        \\
        \makebox[0.45\textwidth]{ (g) DeepExemplar~\cite{zhang2019deep}} &
        \makebox[0.45\textwidth]{ (h) ColorMNet (Ours)}  
         %& \vspace{-0.7em}
	\end{tabularx}
	\vspace{-0.8em}
	\caption{{Colorization results on clip \textit{014} from the NVCC2023 validation dataset~\cite{Kang_2023_CVPR}. The results shown in (b) and (c) still contain significant color-bleeding artifacts. In contrast, our proposed method generates a vivid frame in (h) that is not only more colorful compared with (d-g) but also faithful to the exemplar image in (a).}}
	\label{fig:12}
	%\vspace{-3mm}
\end{figure*}

\begin{figure*}[!htb]
	\setlength\tabcolsep{1.0pt}
	\centering
	\small
	\begin{tabularx}{1\textwidth}{cccc}
        \includegraphics[width=0.246\textwidth, height = 0.28\textwidth] {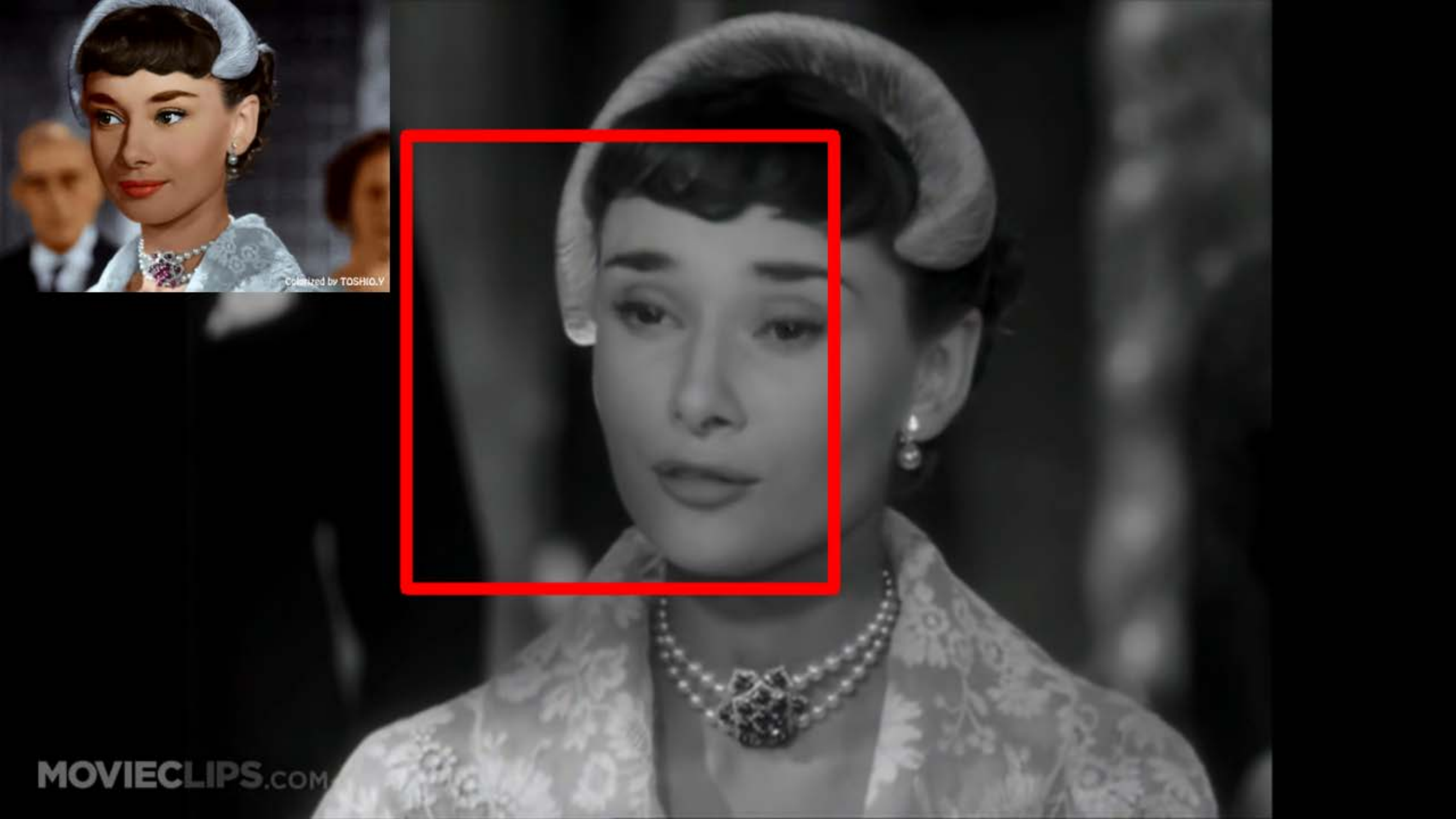} &
        \includegraphics[width=0.246\textwidth, height = 0.28\textwidth] {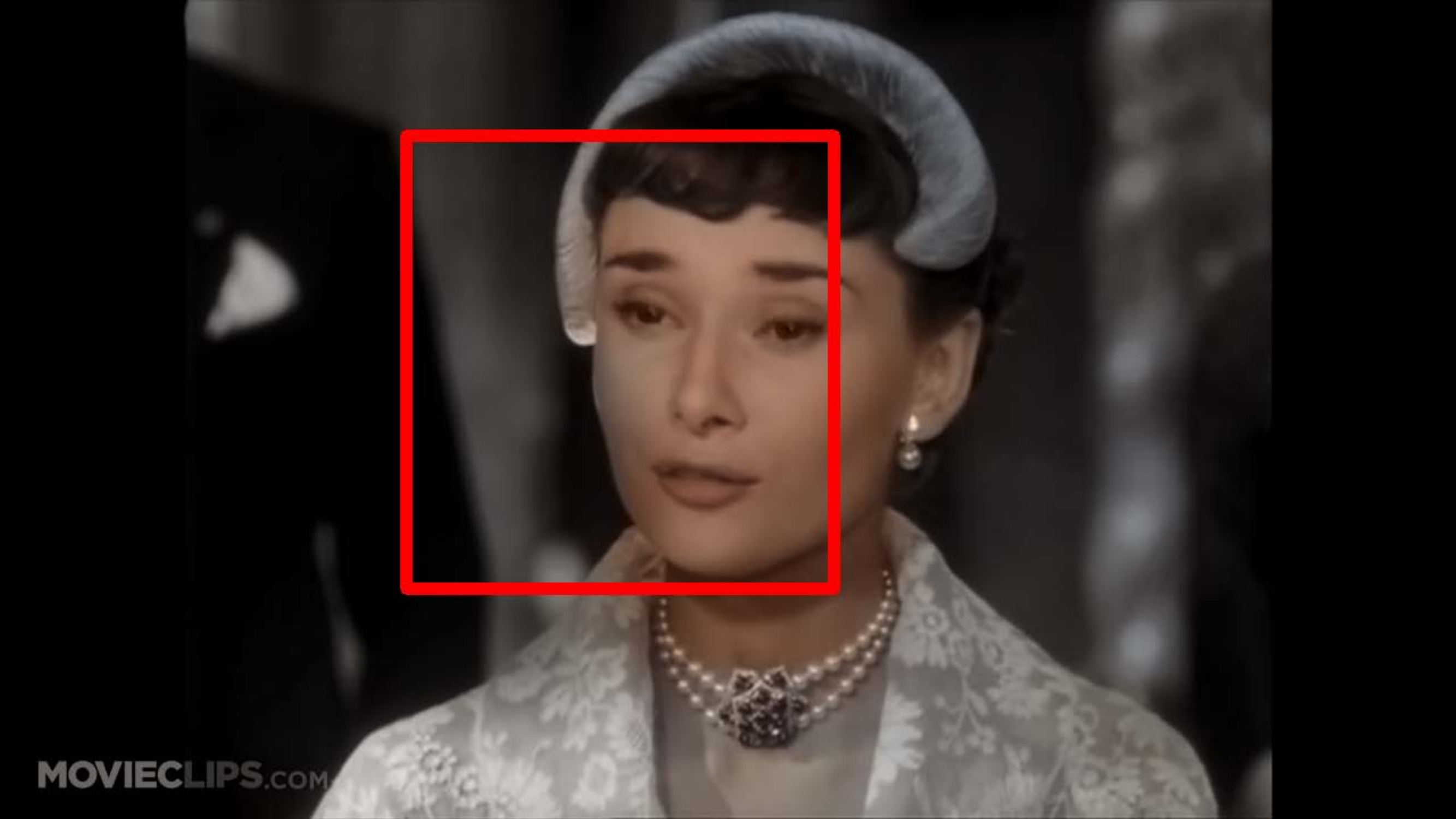} &
       \includegraphics[width=0.246\textwidth, height = 0.28\textwidth] {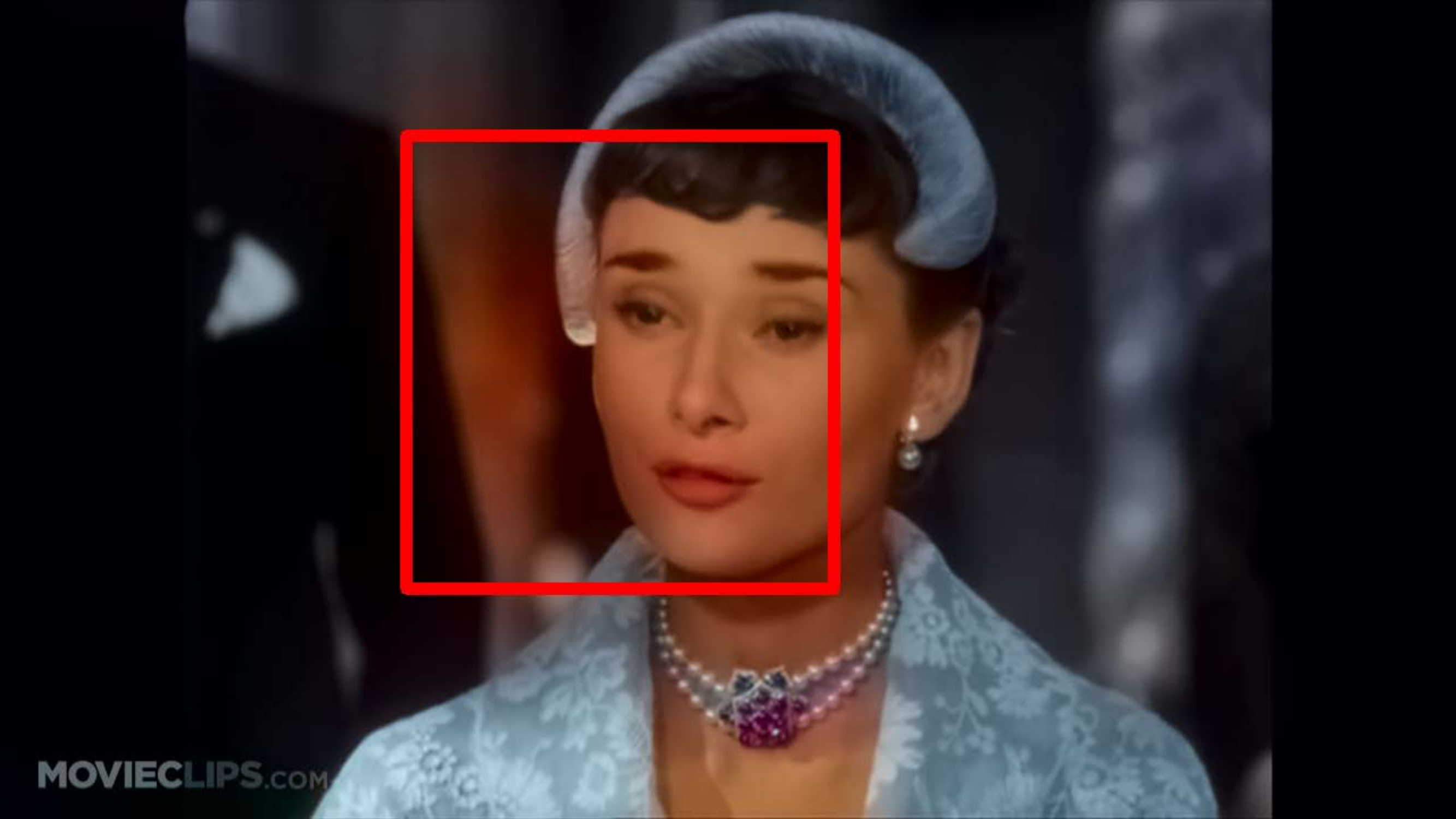} &
        \includegraphics[width=0.246\textwidth, height = 0.28\textwidth] {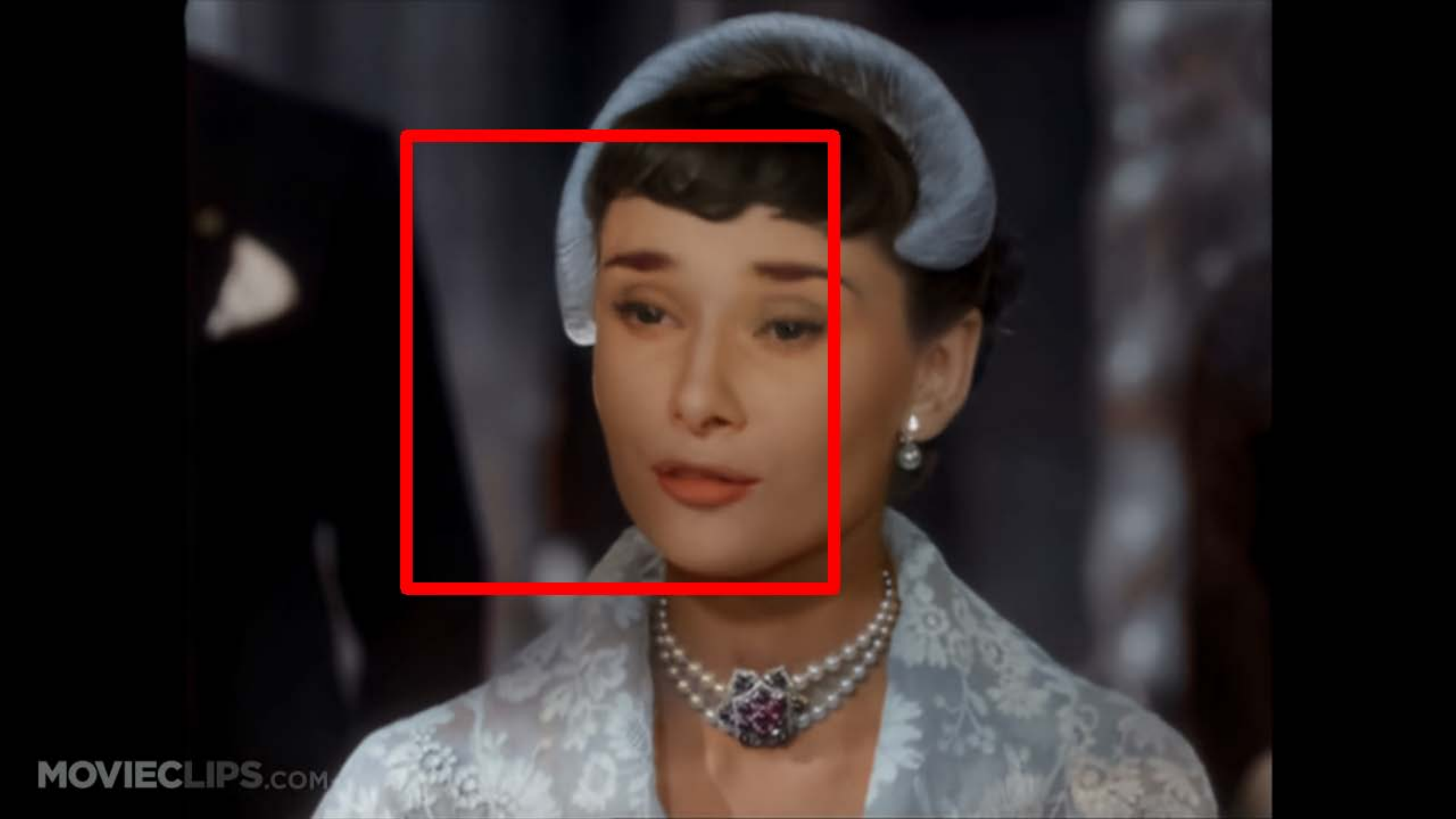} 
        \\ 
        \includegraphics[width=0.246\textwidth, height = 0.28\textwidth] {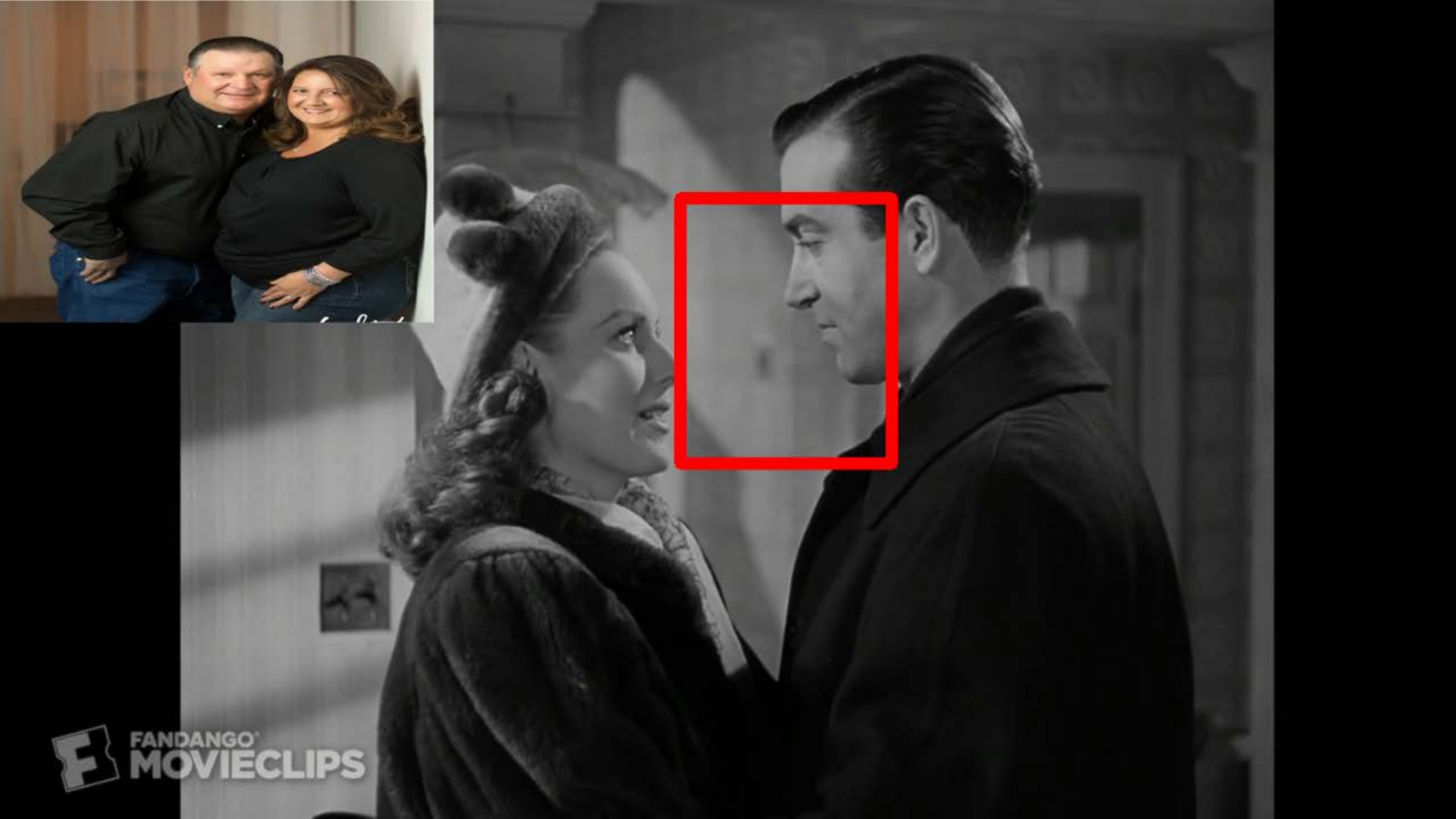} &
        \includegraphics[width=0.246\textwidth, height = 0.28\textwidth] {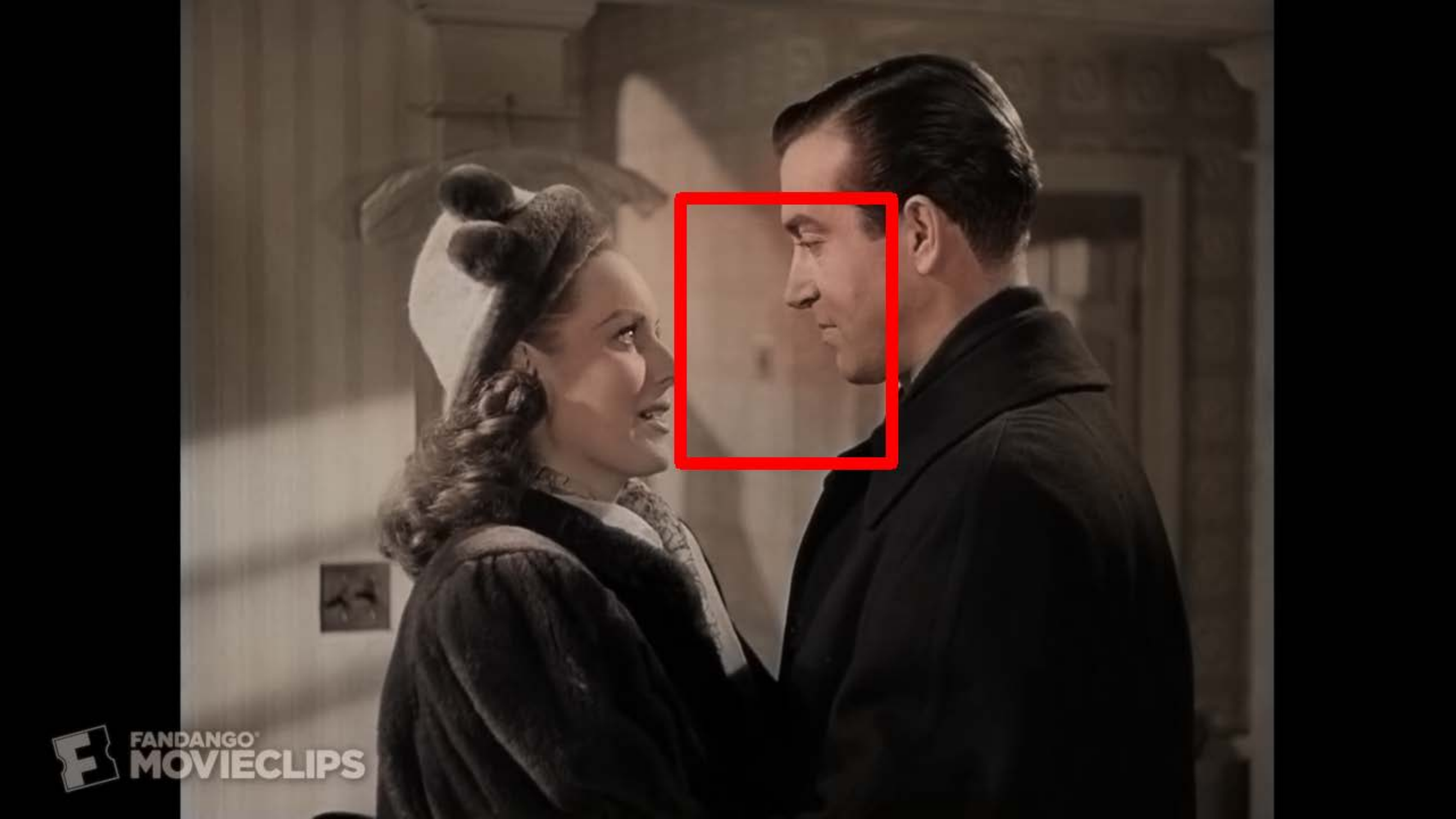} &
       \includegraphics[width=0.246\textwidth, height = 0.28\textwidth] {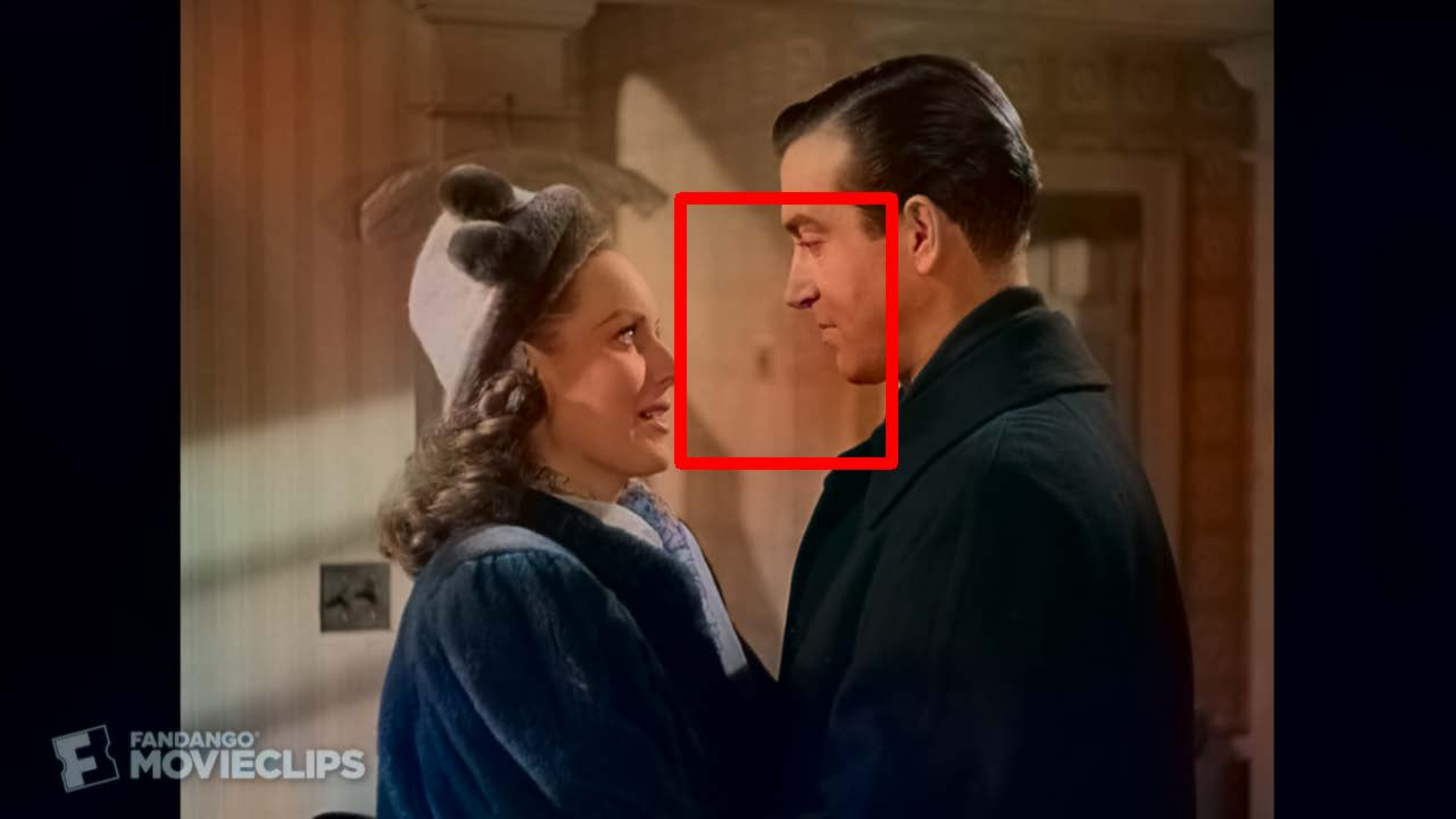} &
        \includegraphics[width=0.246\textwidth, height = 0.28\textwidth] {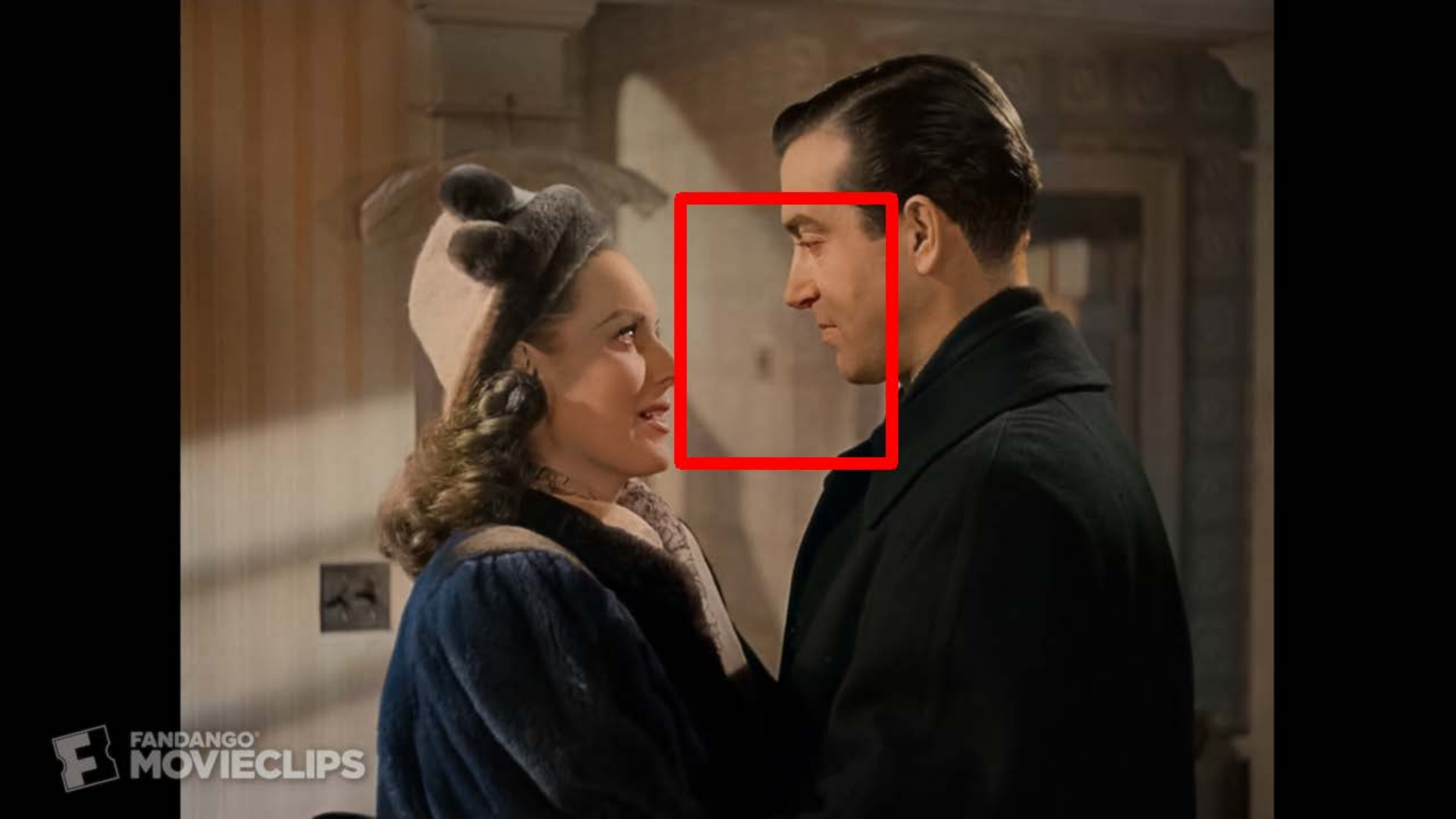} 
        \\ 
        \includegraphics[width=0.246\textwidth, height = 0.28\textwidth] {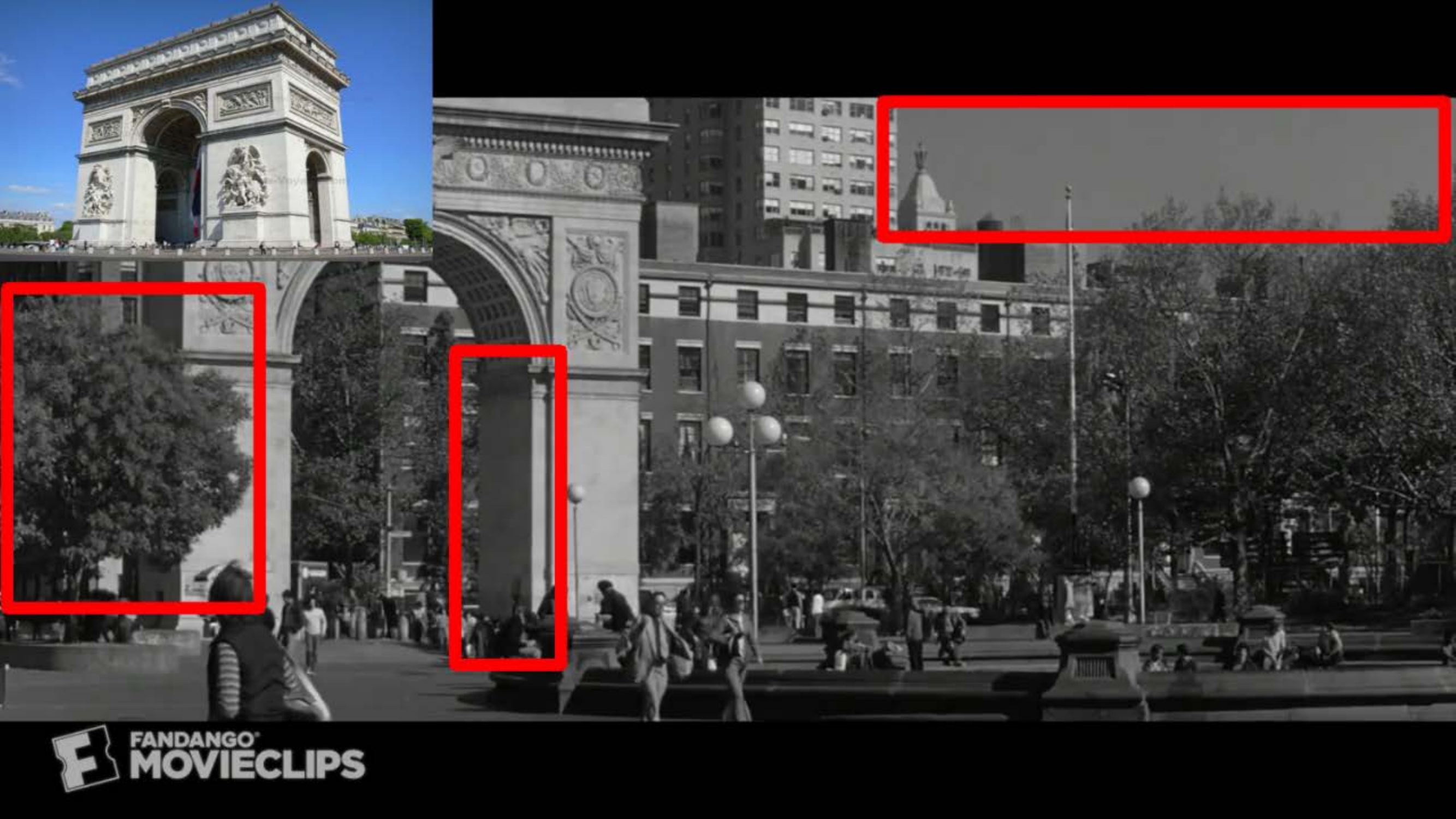} &
        \includegraphics[width=0.246\textwidth, height = 0.28\textwidth] {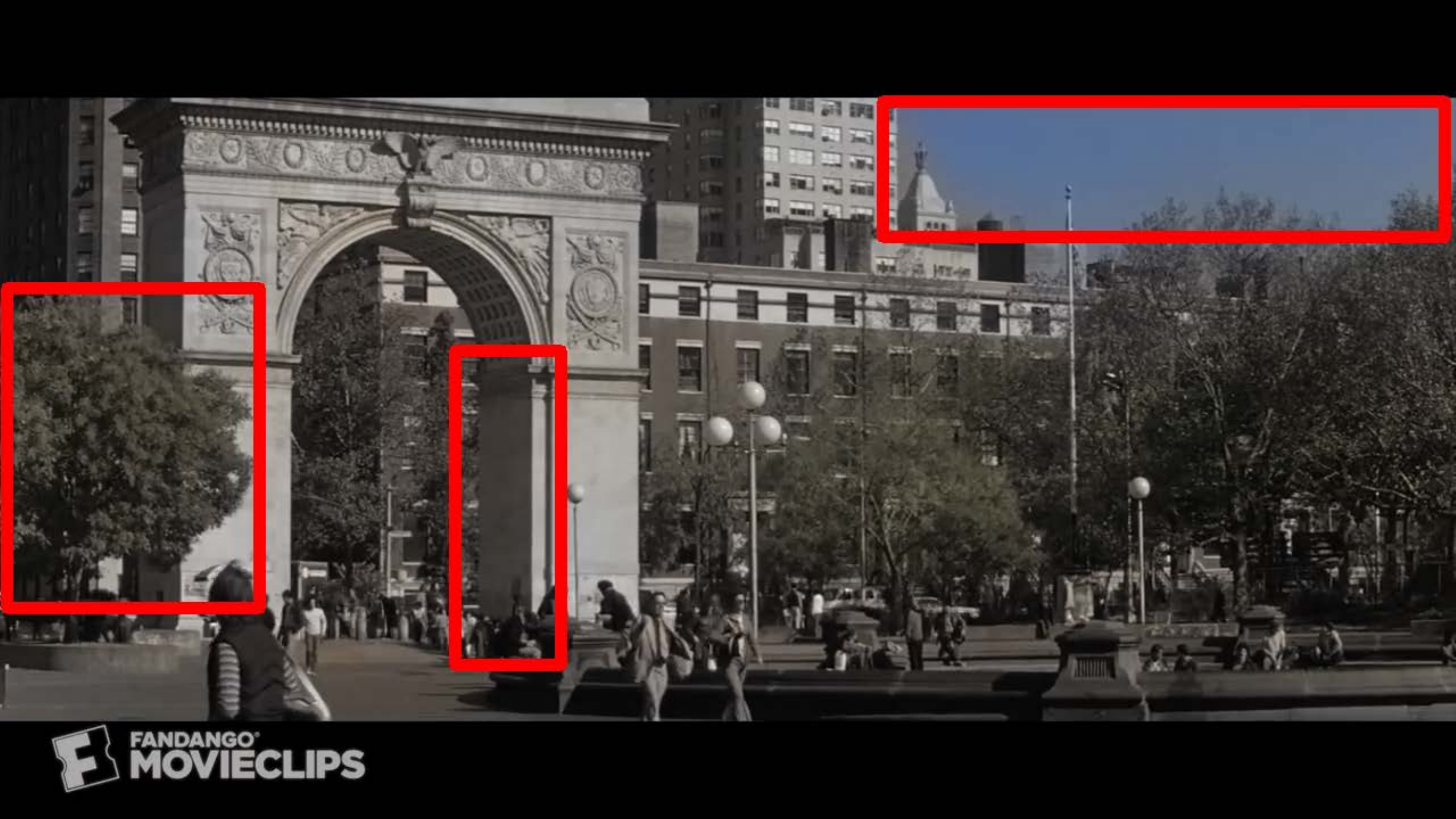} &
       \includegraphics[width=0.246\textwidth, height = 0.28\textwidth] {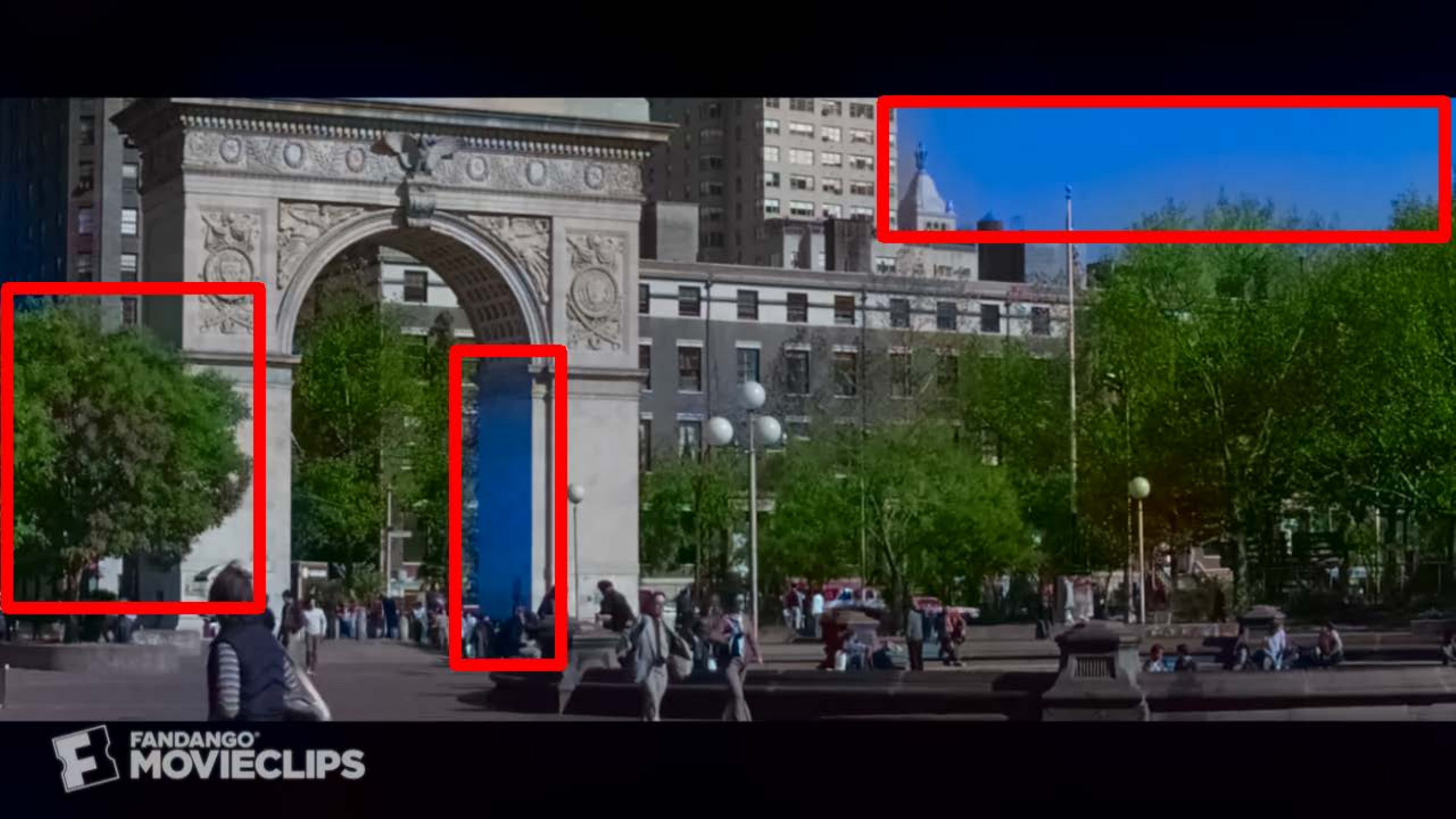} &
        \includegraphics[width=0.246\textwidth, height = 0.28\textwidth] {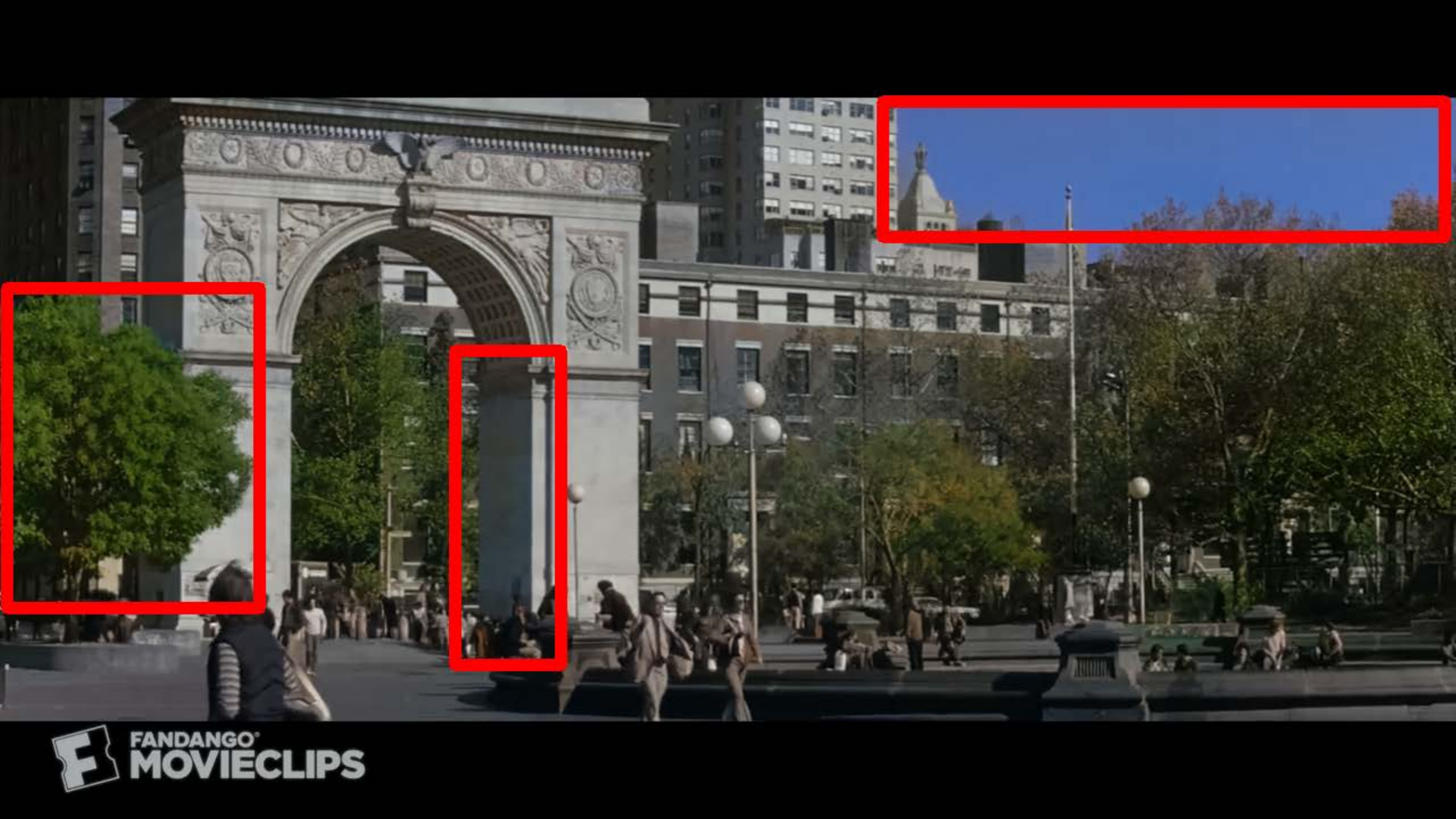} 
        \\
        \includegraphics[width=0.246\textwidth, height = 0.28\textwidth] {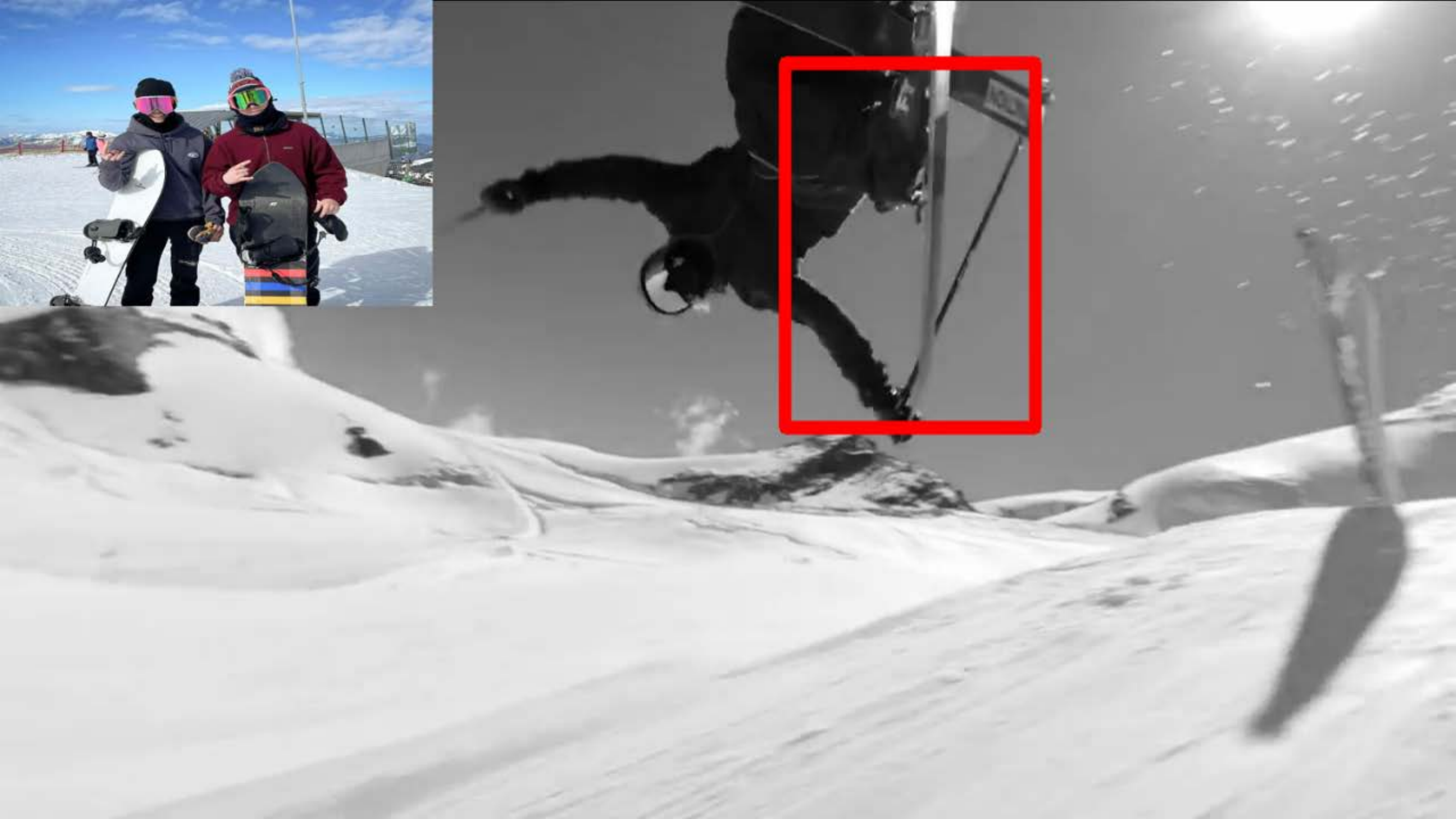} &
        \includegraphics[width=0.246\textwidth, height = 0.28\textwidth] {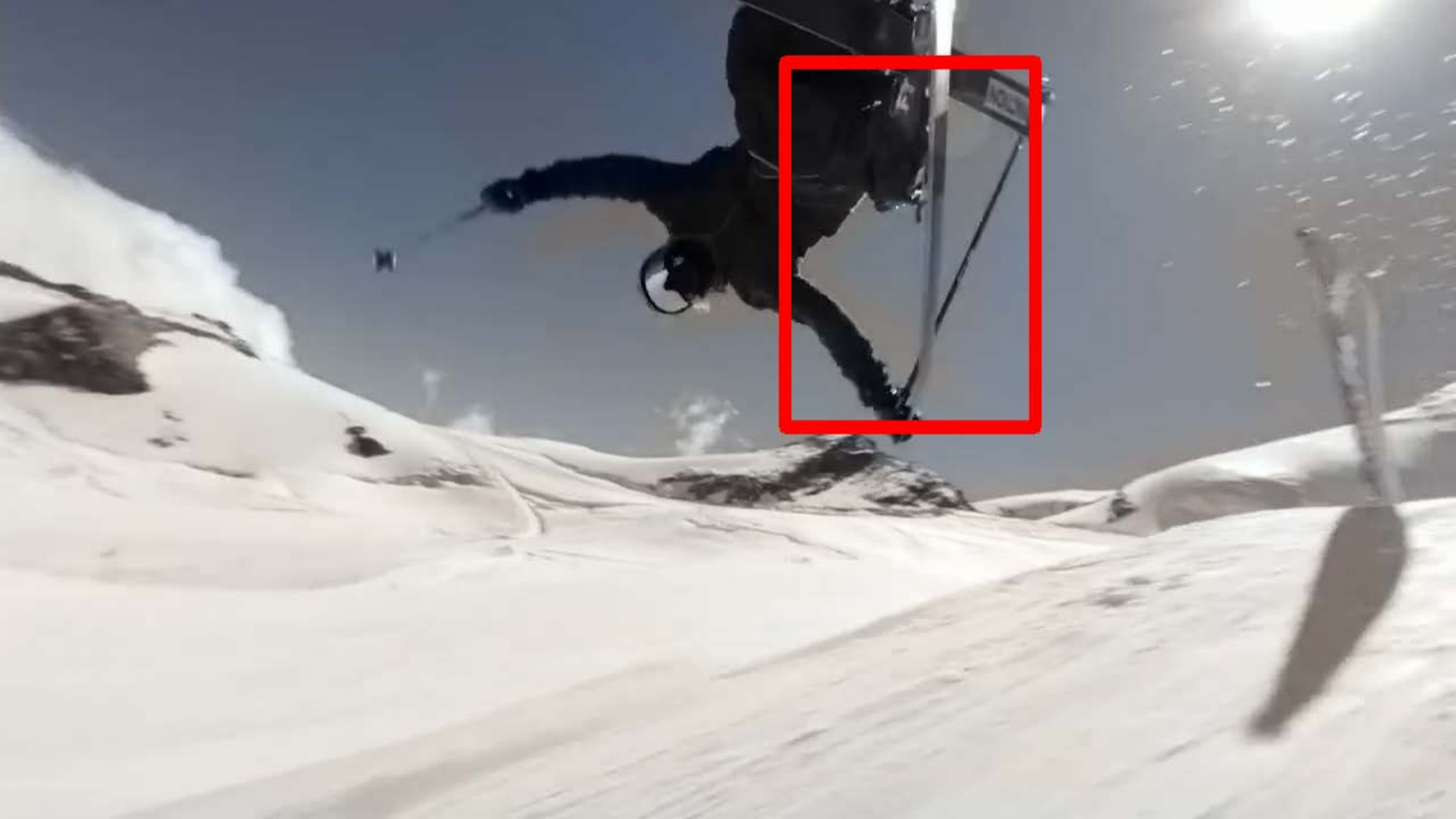} &
       \includegraphics[width=0.246\textwidth, height = 0.28\textwidth] {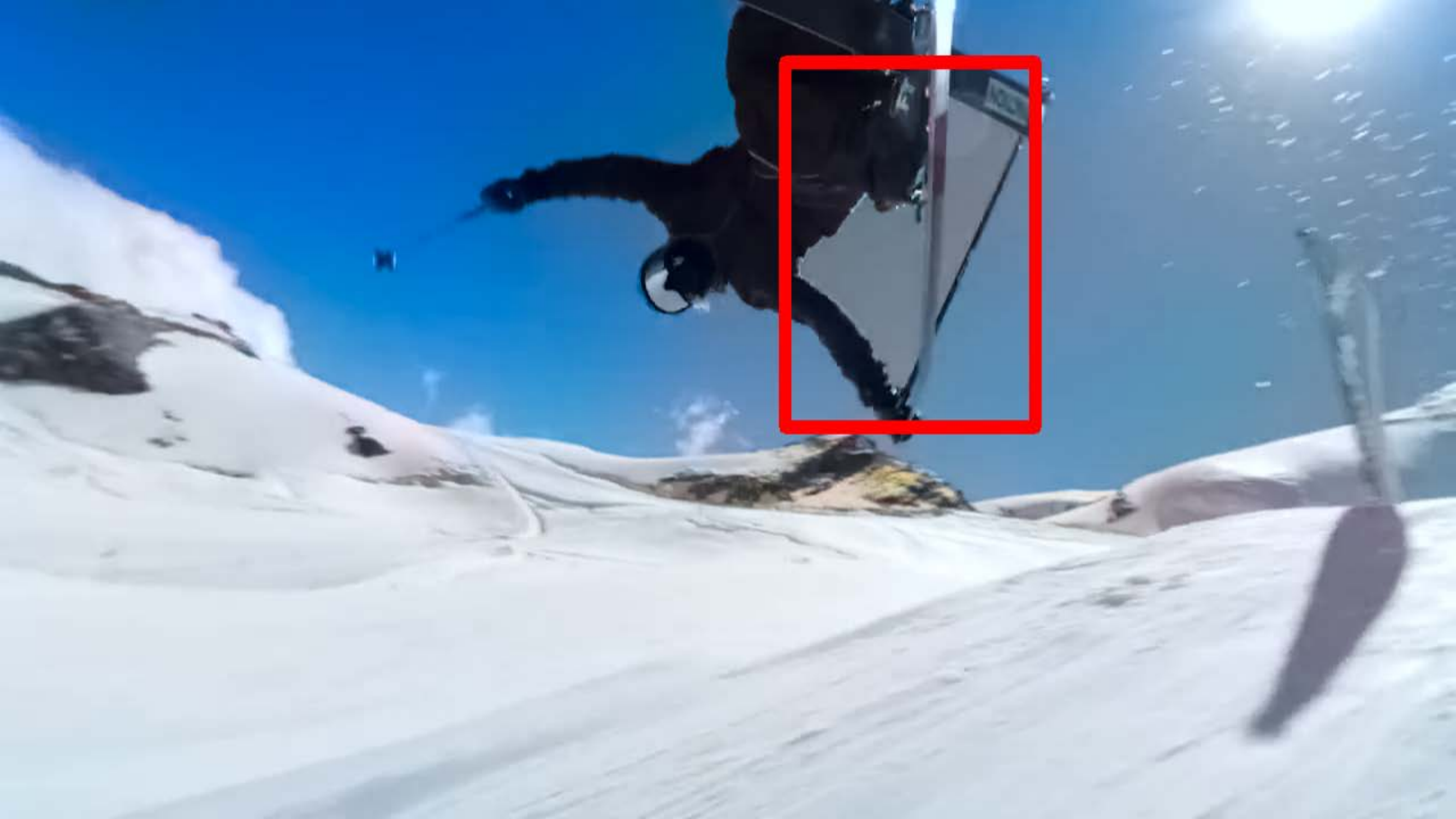} &
        \includegraphics[width=0.246\textwidth, height = 0.28\textwidth] {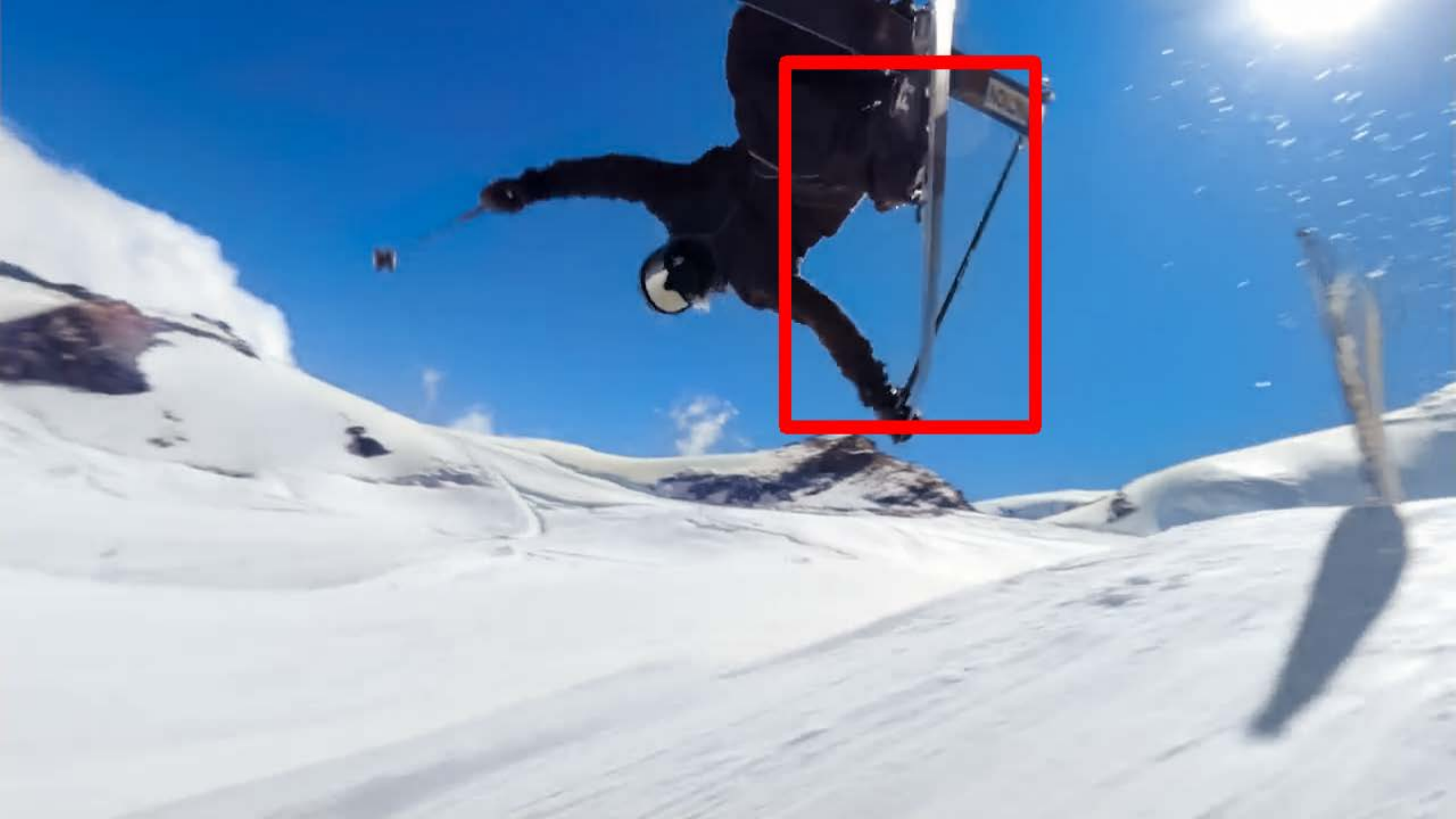} 
        \\
        \makebox[0.246\textwidth]{\tiny (a) Inputs} &
        \makebox[0.246\textwidth]{\tiny (b) DeepRemaster~\cite{IizukaSIGGRAPHASIA2019}}  &
        \makebox[0.246\textwidth]{\tiny (c) DeepExemplar~\cite{zhang2019deep}} &
        \makebox[0.246\textwidth]{\tiny (d) ColorMNet (Ours)}  
         %& \vspace{-0.7em}
	\end{tabularx}
	\vspace{-0.8em}
	\caption{{Colorization results on real-world videos. From top to bottom are respectively the film clip from \textit{Roman Holiday (1953)}, the film clip from \textit{Miracle on 34th Street (1947)}, the film clip from \textit{Manhattan (1979)} and a real-world video collected from the internet. We obtain the exemplars by searching the internet to find the most visually similar images to the input video frames. DeepRemaster~\cite{IizukaSIGGRAPHASIA2019} cannot gerenate vivid frames. The results shown in (c) generated by DeepExemplar~\cite{zhang2019deep} still contain significant color-bleeding artifacts (the wall of the building and the skiing man) and cannot maintain faithfulness to the given exemplar images (over-saturated colors on the both the face of the man and the face of the woman). In contrast, our proposed method generates vivid and realistic frames.}}
	\label{fig:13}
	%\vspace{-3mm}
\end{figure*}

In this section, we provide more visual comparisons with state-of-the-art methods on both synthetic and real-world videos. 
Figures~\ref{fig:1}-\ref{fig:13} show the comparisons, where our method generates better colorized frames.
In \textit{`ComparisonsWithSOTA.mp4'}, we show that the proposed method generates vivid and realistic videos.

{\small
\bibliographystyle{splncs04}
\bibliography{main}
}

% 2

% 3

% 4

% 5

% 6

% 7

% 8

% 9

% 11

% 12

% 13
